\documentclass[10pt,twocolumn,letterpaper]{article}

\usepackage[pagenumbers]{iccv} 

\usepackage{booktabs}
\usepackage{array}
\usepackage{multirow}
\usepackage{color}
\usepackage{colortbl}
\usepackage{bm}
\usepackage{bbm}
\usepackage{xspace}
\usepackage{xparse}
\usepackage{xcolor}
\usepackage{algorithm}
\usepackage{url}
\usepackage{pifont}
\usepackage{stfloats}
\usepackage{graphicx}

\definecolor{Gray}{gray}{0.93}
\definecolor{LightCyan}{rgb}{0.88,0.95,1}
\definecolor{blond}{rgb}{0.98, 0.94, 0.75}
\definecolor{OurColor}{rgb}{0.87, 0.94, 0.84}

\definecolor{pastelblue}{RGB}{102,153,204}
\definecolor{pastelgreen}{RGB}{119,221,119}
\definecolor{pastelpurple}{RGB}{200,160,220}

\def \ie {\emph{i.e.}}
\def \eg {\emph{e.g.}}

\newcommand{\cmark}{\ding{51}}%
\newcommand{\xmark}{\ding{55}}%

\newcommand{\tit}[1]{\smallbreak\noindent\textbf{#1.}}
\newcommand{\tinytit}[1]{\noindent\textbf{#1.}}

\newcommand{\ours}{\texttt{\textbf{ScanDiff}}\xspace}
\newcommand{\scandiff}{\texttt{ScanDiff}\xspace}

\definecolor{iccvblue}{rgb}{0.21,0.49,0.74}
\usepackage[pagebackref,breaklinks,colorlinks,allcolors=iccvblue]{hyperref}

\def\paperID{9855} 
\def\confName{ICCV}
\def\confYear{2025}

\title{Modeling Human Gaze Behavior with Diffusion Models\\for Unified Scanpath Prediction}

\author{$^1$Giuseppe Cartella, $^1$Vittorio Cuculo, $^2$Alessandro D'Amelio, \\$^1$Marcella Cornia, $^2$Giuseppe Boccignone, $^1$Rita Cucchiara
\\
$^1$University of Modena and Reggio Emilia, Italy \quad $^2$University of Milan, Italy \\
{\tt\small $^1$\{name.surname\}@unimore.it, $^2$\{name.surname\}@unimi.it}
}

\begin{document}
\maketitle
\begin{abstract}
Predicting human gaze scanpaths is crucial for understanding visual attention, with applications in human-computer interaction, autonomous systems, and cognitive robotics. While deep learning models have advanced scanpath prediction, most existing approaches generate averaged behaviors, failing to capture the variability of human visual exploration. In this work, we present \scandiff, a novel architecture that combines diffusion models with Vision Transformers to generate diverse and realistic scanpaths. Our method explicitly models scanpath variability by leveraging the stochastic nature of diffusion models, producing a wide range of plausible gaze trajectories. Additionally, we introduce textual conditioning to enable task-driven scanpath generation, allowing the model to adapt to different visual search objectives. Experiments on benchmark datasets show that \scandiff surpasses state-of-the-art methods in both free-viewing and task-driven scenarios, producing more diverse and accurate scanpaths. These results highlight its ability to better capture the complexity of human visual behavior, pushing forward gaze prediction research. Source code and models are publicly available at \href{https://aimagelab.github.io/ScanDiff}{\texttt{https://aimagelab.github.io/ScanDiff}}.
\end{abstract}    
\section{Introduction}
\label{sec:intro}

Understanding and predicting human visual attention remains a central problem in computer vision~\cite{cartella2024trends,chen2024gazexplain,mondal2023gazeformer,kummerer2022deepgaze,cartella2024unveiling}, with broad relevance to fields such as human-computer interaction~\cite{jiang2023ueyes}, autonomous driving~\cite{pal2020looking}, and cognitive robotics~\cite{qian2023gvgnet}. Visual attention deployment is a dynamic and selective mechanism that allows humans to efficiently process the vast amount of information in complex visual stimuli. A critical aspect of computational modeling in this domain involves the prediction of human gaze scanpaths -- the sequences of fixations and saccades that represent the dynamic process of visual exploration.

\begin{figure}[t]
\vspace{-0.05cm}
    \centering
    \includegraphics[width=0.98\linewidth]{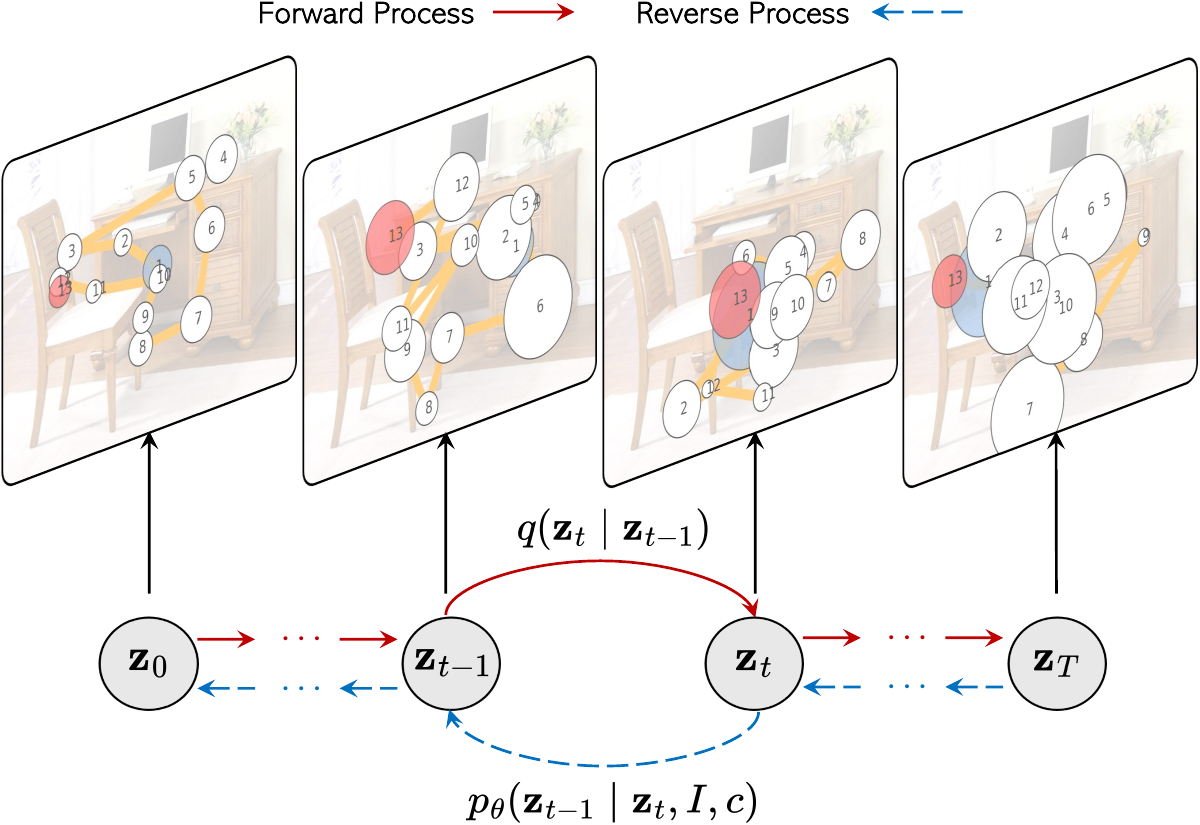}
    \vspace{-0.2cm}
    \caption{The diffusion process of \scandiff that generates realistic scanpaths through learned transitions conditioned on image $I$ and viewing task $c$.}
    \label{fig:first_page}
    \vspace{-0.5cm}
\end{figure}

Models based on deep convolutional~\cite{kummerer2014deep} and recurrent architectures~\cite{chen2018scanpath}, as well as more recent Transformer-based methods~\cite{mondal2023gazeformer,yang2024unifying}, have significantly improved the ability to predict eye movements. These models are effective in both free-viewing scenarios, where observers explore without an explicit task, and in visual search, where exploration follows predefined goals. However, most of these approaches generate scanpaths that reflect an averaged behavior, failing to capture the rich variability observed in individual visual exploration
~\cite{risko2012curious,yamashita2024personality,le2024way}. As noted in~\cite{canosa2009real}, the decision of where to look next at any given moment is neither entirely deterministic nor completely random.
Modeling the effects of randomness allows us to efficiently address the influence of complex factors, such as oculomotor biases, traits, and motor response variability, at both the individual and group levels.
Indeed,  variability -- and the resulting stochasticity of gaze allocation -- goes beyond merely revealing individual idiosyncrasies, which are significant in clinical and psychological studies. It also enables the observer to remain responsive to new signals and promotes a flexible shift of attention. This flexibility, in turn, facilitates efficient learning and exploration of the environment, an essential capability for autonomous systems in computer vision and robotics~\cite{bajcsy2018revisiting,tang2022perception}.

Recent advances in generative modeling, particularly diffusion probabilistic models~\cite{ho2020denoising,dhariwal2021diffusion}, offer a promising alternative by using stochastic sampling to learn and generate diverse sequential outputs. Early applications to scanpath prediction have shown promise in generating human-like gaze behaviors that capture the inherent variability of individual scanpaths on 360° images~\cite{jiao2024diffgaze,wang2024scantd} and text~\cite{bolliger2023scandl}.
When combined with the sequence modeling strengths of Transformer-based architectures~\cite{vaswani2017attention,dosovitskiy2021image}, these approaches move beyond deterministic predictions to simulate a broader range of plausible scanpaths aligned with cognitive theories of attention~\cite{zhao2024cognition}.

Building on these insights, we propose \scandiff, a unified architecture that integrates diffusion models with Vision Transformers~\cite{dosovitskiy2021image,oquab2023dinov2} to generate diverse and realistic gaze scanpaths. Unlike existing approaches, \scandiff explicitly models scanpath variability by leveraging the stochastic nature of diffusion models, enabling the generation of diverse yet plausible gaze trajectories. Furthermore, our method incorporates textual conditioning and a length prediction module, allowing the model to flexibly adapt to diverse visual search objectives within a unified framework.

Through extensive experiments on COCO-FreeView \cite{yang2023predicting}, MIT1003~\cite{judd2009learning}, and COCO-Search18~\cite{Yang_2020_CVPR,chen2021coco}, we demonstrate that \scandiff sets a new state of the art in scanpath prediction across both free-viewing and task-driven scenarios.
Additionally, we present a novel analysis of the variability of predicted scanpaths, highlighting that our approach generates highly diverse eye trajectories, better capturing human gaze behaviors than competitors. This is achieved by leveraging existing scanpath prediction metrics and incorporating a new measure that penalizes excessive similarity among generated scanpaths.
\noindent In summary, our key contributions are as follows:

\begin{itemize}[noitemsep,topsep=0pt]
\item A novel diffusion-based architecture that models the inherent stochasticity of human gaze, enabling the generation of diverse and realistic gaze trajectories.
\item A unified framework that integrates textual conditioning and a length prediction module, allowing the model to adapt to both free-viewing and task-driven scenarios.
\item A comprehensive evaluation that includes a novel analysis of scanpath variability, demonstrating that \scandiff outperforms existing methods in capturing the diversity of human gaze behavior, along with achieving state-of-the-art results in traditional scanpath prediction metrics.
\end{itemize}
\section{Related Work}
\label{sec:related}

\tinytit{Scanpath Prediction}  The study of visual attention in computer vision has seen significant progress since the seminal works in~\cite{aloimonos1987active,ballard1991animate,bajcsy1992active,IttiKoch98}.
In particular, research on modeling scanpaths -- \ie, the sequence of gaze fixations and subsequent shifts (saccades) -- has surged in recent years, with applications expanding across multiple domains~\cite{cartella2024trends}.

A considerable amount of research has been dedicated to predicting scanpaths under free-viewing conditions, where the observer has no predefined task~\cite{IttiKoch98,bfpha04,chen2018scanpath,kummerer2022deepgaze,chen2024beyond,damelio2025tpp,assens2018pathgan}. Yet, echoing the foundational work in this field, some studies have shifted focus toward goal-driven attention modeling, in which an observer purposively engages in a specific task, such as locating an object within a scene~\cite{yang2020predicting,chen2021predicting,mondal2023gazeformer,damelio2025tpp} or searching for targets not present in the image~\cite{yang2022target,yang2024unifying}. Other works aimed at simulating human-like attention in visual question answering~\cite{chen2021predicting} and image captioning tasks~\cite{he2019human}.

Recent approaches have explored predicting attention dynamically as a person views an image while hearing a referring expression specifying the target object~\cite{mondal2024look,liu2025eyear}. Others have used vision-and-language models to jointly predict scanpaths and generate language-based explanations~\cite{chen2024gazexplain}, or to predict subjective feedback like satisfaction and aesthetic quality alongside human attention patterns~\cite{li2023uniar}. Notably, some efforts have focused on using diffusion models to generate scanpaths, though these have been limited to specific settings, such as reading~\cite{bolliger2023scandl} or viewing 360° images~\cite{wang2024scantd,jiao2024diffgaze}. To the best of our knowledge, we are the first to explore the potential of diffusion-based architectures for free-viewing and visual search tasks in natural scenes.

\tit{Diffusion Models for Sequence Modeling} Recently, diffusion models have emerged as one of the most successful probabilistic generative architectures across various fields, particularly in computer vision~\cite{croitoru2023diffusion}. They have also gained popularity as a non-autoregressive alternative for modeling sequences~\cite{yang2024survey}, demonstrating success in generating various types of sequences, including continuous time-series~\cite{coletta2023constrained,lim2023regular,crabbe2024frequency}, text~\cite{gong2022diffuseq,li2022diffusion}, and audio~\cite{kong2021diffwave}. Notably, these models have recently been adopted for generating spatio-temporal data, such as GPS trajectories~\cite{zhu2024controltraj,zhu2023difftraj,wei2024diff} and human motion data~\cite{song2024controllable,zhang2024motiondiffuse,chu2024simulating}, including eye movement patterns~\cite{bolliger2023scandl,wang2024scantd,jiao2024diffgaze}. In contrast to these methods, we focus on both standard free-viewing and goal-oriented settings, introducing a novel approach that can predict scanpaths of variable lengths, thereby enabling greater variability and more realistic gaze behavior.
\section{Proposed Method}
\label{sec:methodology}
\begin{figure*}[t]
    \centering
    \includegraphics[width=0.98\linewidth]{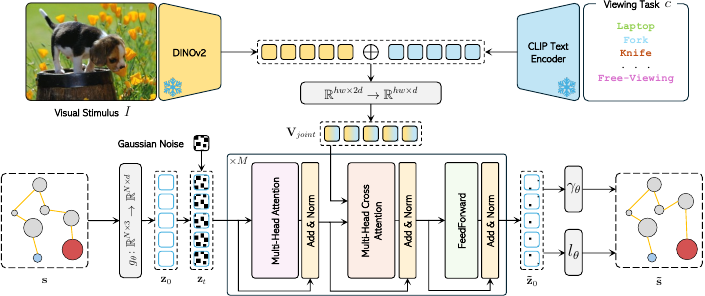 }
    \vspace{-0.15cm}
    \caption{Overview of \scandiff. Given a stimulus $I$ and a viewing task $c$, a scanpath $\tilde{s}$ is generated through a diffusion process.}
    \label{fig:method}
    \vspace{-0.3cm}
\end{figure*}

Scanpath generation aims to predict the spatial and temporal dynamics of human eye movements in response to a given visual stimulus. 
The generation can be performed under the free-viewing task or the goal-directed task, as in the case of object visual search~\cite{mondal2023gazeformer,yang2022target,Yang_2020_CVPR}, visual question answering~\cite{chen2021predicting} or incremental object referral~\cite{mondal2024look}.
We propose \scandiff, a novel scanpath prediction architecture based on diffusion models to generate realistic and diverse gaze patterns (see Fig.~\ref{fig:method}).
Its multimodal nature enables the unified prediction of various types of visual attention, seamlessly adapting to different viewing tasks and stimuli.

\subsection{Preliminaries}
Diffusion models are a class of generative models able to model the ground-truth distribution of a given dataset by 
reversing a diffusion process that gradually adds noise to the input data. 
They consist of a forward and a backward process. Given a sample $\mathbf{x}_0$ drawn from a real-world data distribution $\mathbf{x}_0 \sim q(\mathbf{x})$, the forward process gradually corrupts the input data by adding Gaussian noise
for a number of timesteps $T$ according to a variance schedule $\beta_1, \ldots, \beta_T$. This produces, at each timestep $t$, a latent variable $\mathbf{x}_t$ with distribution $q(\mathbf{x}_t \mid \mathbf{x}_{t-1})$, defined as: 
\begin{equation}
    q(\mathbf{x}_t \mid \mathbf{x}_{t-1}) = \mathcal{N} \left( \mathbf{x}_t; \sqrt{1-\beta_t} \mathbf{x}_{t-1}, \beta_t \boldsymbol{\mathcal{I}} \right),
    \label{eq:diff_process}
\end{equation}
with $\boldsymbol{\mathcal{I}}$ being the identity matrix. In the reverse process, the final goal is to recover $\mathbf{x}_0$ by denoising $\mathbf{x}_T$. This process is defined by a Markov chain parameterized by $\theta$:
\begin{equation}
    p_\theta(\mathbf{x}_{0:T}) := p_\theta(\mathbf{x}_T) \prod_{t=1}^{T} p_\theta(\mathbf{x}_{t-1} | \mathbf{x}_t).
    \label{eq:denoising_process}
\end{equation}

In particular, each transition $p_\theta(\mathbf{x}_{t-1} | \mathbf{x}_t) = \mathcal{N} \big(\mathbf{x}_{t-1}; \mu_\theta(\mathbf{x}_t, t), \Sigma_\theta(\mathbf{x}_t, t) \big)$ 
is parameterized by a function $\phi_\theta$, where $\mathbf{\mu}_\theta$ and $\Sigma_\theta$ represent the predicted mean and variance of the true posterior distribution, respectively.

\tit{Problem Definition}
Given an image or stimulus $I \in \mathbb{R}^{H \times W \times 3}$ and a viewing task $c$, the final objective is to predict a human-like scanpath represented as an ordered sequence of $N$ fixations $\mathbf{s} = \{f_1,f_2,\ldots,f_N\}$. Each fixation $f_i$ consists of a pair $(r_i, m_i)$, where $r_i=(x_i, y_i)\in \mathbb{R}^2$ is the 2D spatial fixation location, while $m_i\in \mathbb{R}^+$ is the fixation duration.
In this work, we propose a non-autoregressive approach to generate scanpath trajectories by learning a diffusion model $\phi_\theta$. Starting from a noisy sample drawn from a Gaussian distribution, the learned model iteratively refines it to produce the final scanpath trajectory.

\subsection{ScanDiff Model}
\subsubsection{Forward Process: Scanpath Embedding}
Let $\mathbf{s} \in \mathbb{R}^{N\times 3}$ be a sequence of $N$ ground-truth fixations. To enable a structured latent space that better captures temporal and spatial dependencies of the scanpath, we learn a linear projection $g_\theta: \mathbb{R}^{N\times 3} \to \mathbb{R}^{N\times d}$ to map the scanpath $\mathbf{s}$ into an augmented embedding space, thus obtaining the initial uncorrupted latent variable $\mathbf{z}_0 = g_\theta(\mathbf{s}) \in \mathbb{R}^{N\times d}$. 
During the forward process (see Fig.~\ref{fig:first_page}), we gradually corrupt the whole embedded sequence $\mathbf{z}_0$ by adding Gaussian noise over $T$ timesteps, following a predefined variance schedule. At each timestep $t$, the noisy latent representation $\mathbf{z}_t$ is obtained through a Markovian diffusion process, as defined in Eq.~\ref{eq:diff_process}.
By the final timestep, the representation $\mathbf{z}_T$ approaches an isotropic Gaussian distribution, effectively removing any trace of the original scanpath structure.

\subsubsection{Conditional Denoising Process}
Scanpath prediction involves the generation of an ordered sequence of $N$ fixations conditioned on a given stimulus $I$ and a viewing task $c$. Therefore, referring to Eq.~\ref{eq:denoising_process}, the conditioned denoising process can be rewritten as:
\begin{equation}
    p_\theta(\mathbf{z}_{0:T} \mid I, c) := p(\mathbf{z}_T) \prod_{t=1}^{T} p_\theta(\mathbf{z}_{t-1} \mid \mathbf{z}_t, I, c).
    \label{eq:conditioned_denoising_process}
\end{equation}

Our model $\phi_\theta$ is based on an encoder-only  Transformer~\cite{vaswani2017attention},
modified to incorporate an additional multi-head cross-attention layer between the self-attention and feed-forward layers~\cite{peebles2023scalable}. This enables the model to effectively condition on the image $I$ and the viewing task $c$.

\tit{Image Encoding}
We process each stimulus $I\in\mathbb{R}^{H\times W\times 3}$ using a Transformer-based visual backbone $v(\cdot)$ which outputs a dense feature map $v(I)\in \mathbb{R}^{h\times w\times d_v}$. Here, $h$ and $w$ denote the number of patches along the height and width of the image, respectively, while $d_v$ refers to the dimensionality of the visual embedding space.
\tit{Task Encoding}
Our model features a unified architecture that seamlessly adapts to different tasks without requiring any architectural modifications. This flexibility is achieved by using a text encoder to represent the viewing task $c$. 
Specifically, for the free-viewing task, we represent $c$ as an empty string. In contrast, for visual search, $c$ corresponds to the textual label of the target object to look for in the image (\eg~``laptop'').
To extract task representations, we employ a pre-trained text encoder $\psi(\cdot)$ which maps $c$ to a feature vector $\psi(c) \in \mathbb{R}^{d_{t}}$ in the textual embedding space, where $d_t$ denotes the dimensionality of the textual features.

\tit{Multimodal Conditioning}
To condition the denoising process on both the image $I$ and task $c$, we project the visual features $v(I)$ and the textual features $\psi(c)$ in a joint multimodal embedding space. Following previous works~\cite{mondal2023gazeformer}, we first map the visual and textual features into a common $d$-dimensional space using two independent linear transformations. The textual features are then repeated $hw$ times and concatenated with the visual features along the channel dimension, resulting in a feature map of size $hw\times 2d$.
Finally, this feature map is linearly projected into a $d$-dimensional feature space to obtain the final multimodal embedding $\mathbf{V}_{joint} \in \mathbb{R}^{hw \times d}$, which effectively combines visual and task semantic information.  
This multimodal conditioning enables a unified model to adapt to various tasks, from free-viewing to visual search.
The resulting multimodal embedding $\mathbf{V}_{joint}$ is then passed through the cross-attention layer of the Transformer. In parallel, $\mathbf{z}_t$ is augmented with a learnable positional encoding and 
a sinusoidal diffusion timestep embedding.
For the sake of simplicity, in what follows, we refer to $\mathbf{z}_t$ as the combination of the noisy scanpath embedding, the positional, and the diffusion timestep embeddings.

Unlike previous approaches~\cite{jiao2024diffgaze,wang2024scantd} that directly concatenate the noisy gaze sequence with the image embedding, we combine $\mathbf{z}_t$
and the visual-semantic features $\mathbf{V}_{joint}$ only in the cross-attention layer. This design choice allows the model to dynamically modulate the interaction between gaze dynamics and visual-semantic information, rather than enforcing a rigid concatenation.

\tit{Scanpath Reconstruction and Length Prediction}
The Transformer encoder output is defined as $\mathbf{\tilde{z}}_0=\phi_\theta(\mathbf{z}_t, \mathbf{V}_{joint})\in \mathbb{R}^{N\times d}$.
The spatial coordinates and the duration of the fixations are reconstructed starting from the predicted sample $\mathbf{\tilde{z}}_0$. Specifically, a feed-forward network $\gamma_\theta$ with three linear layers followed by a ReLU activation function is adopted to decode $\mathbf{\tilde{z}}_0$ to an approximation of the original scanpath $\mathbf{\tilde{s}} = \{\tilde{f}_1,\tilde{f}_2, \ldots, \tilde{f}_N\}$ , $\tilde{f}_i \in \mathbb{R}^3$.
The visual response to a given stimulus and the corresponding scanpath length can vary across subjects. 
Existing works that leverage diffusion models for scanpath generation~\cite{wang2024scantd,jiao2024diffgaze} typically produce fixed-length scanpaths. In contrast and crucially, we take a different approach: we introduce a length prediction module in the model architecture, which allows for greater flexibility. In particular, this module predicts the probability $\tilde{u}_i$ of each token in the reconstructed sample $\mathbf{\tilde{z}}_0$ to be valid through a linear function $l_{\theta}: \mathbb{R}^{N \times d} \to \mathbb{R}^{N}$. The final predicted length is equal to the number of consecutive valid tokens.

\subsection{Training and Inference}
The training objective $\mathcal{L}$ is defined as the combination of four different components:
\begin{equation}
     \mathcal{L} =  \mathcal{L}_{\text{VLB}} + \mathcal{L}_{rec} + \mathcal{L}_{val} + \mathcal{L}_{T}.
\end{equation}

The first component $\mathcal{L}_{\text{VLB}}$ aims to minimize the difference between the uncorrupted sample $\mathbf{z}_0$ and the model prediction. Formally, it is defined as:
\begin{equation}
    \mathcal{L}_{\text{VLB}}=\sum_{t=1}^{T} \|\mathbf{z}_0 - \phi_\theta(\mathbf{z}_t, \mathbf{V}_{joint})\|^2.
\end{equation}
This simplification can be derived from the variational lower bound~\cite{bolliger2023scandl}.
To reduce the noise in the optimization of $\mathcal{L}_{\text{VLB}}$ we adopt importance sampling~\cite{nichol2021improved}.

The second term $\mathcal{L}_{rec}$ measures the scanpath reconstruction error and is defined as the $L_1$ loss between the ground-truth scanpath $\mathbf{s}$ and the reconstructed one $\mathbf{\tilde{s}}$:
\begin{equation}
\begin{split}
\mathcal{L}_{rec} &= \| \mathbf{s} - \tilde{\mathbf{s}} \| \\ &= \frac{1}{N} \sum_{i=1}^{N} \left( |x_i - \tilde{x}_i| + |y_i - \tilde{y}_i| + |m_i - \tilde{m}_i| \right).
\end{split}
\end{equation}
Here, $\tilde{x}_i$ and $\tilde{y}_i$ denote the spatial coordinates of the reconstructed fixation $\tilde{f}_i$, and $\tilde{m}_i$ is the relative fixation duration. 
During training, the ground-truth scanpaths are padded or truncated to a maximum length of $L$. 
The $L-N$ padding fixations are masked out during the computation of $\mathcal{L}_{rec}$.

The term $\mathcal{L}_{val}$ represents binary cross-entropy loss for predicting the validity of each reconstructed fixation:
\begin{equation}
    \mathcal{L}_{val}= \frac{1}{L} \sum_{i=0}^{L} \text{BCE}(u_i, \tilde{u}_i).
\end{equation}

At the final diffusion step, the mean prediction should converge to zero under the assumption that the noise prior follows a standard Gaussian distribution. To stabilize the diffusion process, we define the loss $\mathcal{L}_{T} = \|\mu(\mathbf{z}_T)\|^2$ which penalizes any residual bias in the mean prediction at the final timestep ensuring it ends in a clean isotropic Gaussian. This also regularizes training by enforcing theoretical constraints otherwise weakened by finite-step approximations.

\tit{Sampling a Scanpath}
At inference time, our model generates a scanpath in response to a visual stimulus $I$ and a viewing task $c$ (\eg~free-viewing or visual-search).
We sample $\mathbf{z}_T\sim\mathcal{N}(\mathbf{0},\boldsymbol{\mathcal{I}}) \in \mathbb{R}^{N\times d}$ and the model $\phi_\theta$ iteratively denoises $\mathbf{z}_T$ to $\mathbf{z}_0$. 
At each sampling step $t$ the multimodal features $\mathbf{V}_{joint}$ are fused with $\mathbf{z}_t$ in the cross-attention layer to condition the reverse process. After denoising $\mathbf{z}_T$ into $\mathbf{z}_0$, this is fed through two independent inverse embedding layers to obtain the predicted scanpath and its relative length $N$.
To account for the variability in the observational patterns across different subjects, we generate multiple scanpath trajectories by sampling distinct noisy samples $\mathbf{z}_T$.
\section{Experimental Results}
\label{sec:experiments}

\subsection{Experimental Setup}
\label{subsec:experiments}
\tinytit{Datasets}
We evaluate our model on both the free-viewing and visual search tasks. For free-viewing experiments, we adopt COCO-FreeView~\cite{yang2023predicting} and MIT1003~\cite{judd2009learning}, which comprise 6,202 and 1,003 images, respectively. To train our model, we combine images from both datasets, using 70\% for training, 15\% for validation, and 15\% for testing.
To assess the performance on the visual search task, we employ COCO-Search18~\cite{Yang_2020_CVPR,chen2021coco}, which features eye gaze behavior from 10 people while searching for the presence of a specific object (among 18 diverse categories) in the scene. Images are divided into target-present and target-absent splits, with 3,101 items each. In this setting, we employ the training, validation, and test sets used in previous works~\cite{chen2024gazexplain,mondal2023gazeformer}.

\tit{Evaluation Metrics}
The similarity between generated and human scanpaths is measured through the MultiMatch (MM)~\cite{jarodzka2010vector, dewhurst2012depends}, ScanMatch (SM)~\cite{cristino2010scanmatch}, Sequence Score (SS)~\cite{Yang_2020_CVPR}, and Semantic Sequence Score (SemSS)~\cite{yang2022target} metrics. In particular, we adopt the same evaluation protocol proposed in~\cite{damelio2025tpp} where the distribution of human vs. generated metrics is compared against the distribution of the human consistency metrics, using the Kullback-Leibler divergence.
Beyond similarity, capturing the diversity of generated scanpaths is crucial to prevent the model from collapsing into a deterministic solution, thereby preserving the natural variability observed in human eye movements. To this end, we adopt two additional metrics: 
the Individual Scanpath Recall~\cite{yang2024unifying}, which we rename as Recall Sequence Score (RSS),
and a newly introduced metric that favors the diversity of generated scanpaths termed as Diversity-aware Sequence Score (DSS). 

In particular, the RSS measures the extent to which the generated scanpaths cover the variability of human scanpaths for a given stimulus. For a human scanpath in the dataset, it is considered covered if its SS with at least one generated scanpath surpasses a predefined threshold. The RSS is then computed as the ratio of covered human scanpaths to the total number of human scanpaths.
This metric evaluates whether the model can replicate the range of individual behaviors observed in humans.

The novel DSS we propose extends the standard sequence similarity measures by incorporating a term that penalizes excessive similarity among the generated scanpaths when humans do not reflect such behavior. Given a set of generated scanpaths $\mathbf{s}_g$ and corresponding human scanpaths $\mathbf{s}_h$ for a specific visual stimulus, DSS is computed as 
\begin{equation}
    \text{DSS}(\mathbf{s}_g,\mathbf{s}_h) = \frac{\text{SS}(\mathbf{s}_g, \mathbf{s}_h)}{1+|\text{SS}(\mathbf{s}_g,\mathbf{s}_g) - \text{SS}(\mathbf{s}_h, \mathbf{s}_h)|}
\end{equation}
where SS is the average sequence score calculated over the possible combinations of different scanpaths. The denominator penalizes models that produce overly uniform predictions, encouraging outputs that not only match human behavior but also reflect its natural variability.

\tit{Implementation and Training Details}
We employ the DINOv2 ViT-B/14 model~\cite{oquab2023dinov2} as the pre-trained visual backbone, considering its rich semantic understanding of the visual scene. In particular, we use the DINOv2 variant with registers~\cite{darcet2023vision}.
To align with its training resolution, we resize all images to a resolution of $518\times 518$, resulting in a feature map $v(I)$ of $37\times 37$ patches, with an embedding dimension of $d_v = 768$.
For textual encoding, we utilize the pre-trained CLIP ViT-B/32 model~\cite{radford2021learning}, which projects the viewing task $c$ into a feature space of dimension $d_{t} = 512$.
The modified architecture of the Transformer encoder consists of $M=6$ layers, each with $8$ attention heads and a hidden dimension $d=512$.
The model is trained with the AdamW optimizer, a batch size of $128$, a learning rate set to $1\times 10^{-4}$, a weight decay of $1\times 10^{-2}$, and a number of diffusion steps $T = 1000$. In addition, a squared-root noise schedule is adopted. 
Following previous works~\cite{chen2024gazexplain,chen2021predicting}, the maximum scanpath length is set to 16.
During training, the spatial coordinates of each fixation are scaled in the range $[0,1]$, and fixation durations are retained in seconds.

\begin{table*}[t]
  \centering
  \small
  \setlength{\tabcolsep}{.2em}
  \resizebox{\linewidth}{!}{
  \begin{tabular}{lc cccccc c cc c cc cc cccccc c cc c cc}
    \toprule
    & & \multicolumn{12}{c}{\textbf{COCO-FreeView}} & & \multicolumn{12}{c}{\textbf{MIT1003}} \\
    \midrule
    & & \multicolumn{6}{c}{\textbf{MM} $\downarrow$} & & \multicolumn{2}{c}{\textbf{SM} $\downarrow$} & & \multicolumn{2}{c}{\textbf{SS} $\downarrow$} & & & \multicolumn{6}{c}{\textbf{MM} $\downarrow$} & & \multicolumn{2}{c}{\textbf{SM} $\downarrow$} & & \multicolumn{2}{c}{\textbf{SS} $\downarrow$} \\
    \cmidrule{3-8} \cmidrule{10-11} \cmidrule{13-14} \cmidrule{17-22} \cmidrule{24-25} \cmidrule{27-28}
    & & Sh & Len & Dir & Pos & Dur & Avg & & w/ Dur & w/o Dur & & w/ Dur & w/o Dur & & & Sh & Len & Dir & Pos & Dur & Avg & & w/ Dur & w/o Dur & & w/ Dur & w/o Dur \\
    \midrule
    Itti-Koch~\cite{IttiKoch98} && 0.504 & 0.507 & 0.237 & 1.325 & - & 0.643 & & -	& 4.317 & & - & 1.624 & & & 1.040 & 0.702 & 0.353 & 2.900 & - & 1.249 & & - & 3.233 & & - & 6.639 \\
    CLE (Itti)~\cite{bfpha04,IttiKoch98} && 0.052 & 0.317 & 0.427 & 1.966 & - & 0.691 & & -	& 3.576 & & - & 1.747 & & & 0.061 & 0.124 & 0.414 & 1.515 & - & 0.529 & & - & 3.454 & & - & 5.397 \\
    CLE (DG)~\cite{bfpha04,kummerer2014deep} && 0.037 & 0.180 & 0.323 & 1.823 & - & 0.591 && - & 3.657 & & - & 1.750 & & & 0.099 & 0.038 & 0.458 & 1.066 & - & 0.415 & & - & 2.566 & & - & 6.234 \\
    PathGAN~\cite{assens2018pathgan} & & 0.070 & 0.406 & 1.009 & 0.073 & 0.031 & 0.318 & & 1.210 &	1.383 & & 0.718 & 1.012 & & & 0.063 & 0.234 & 1.603 & 0.514 & 0.165 & 0.516 & & 2.255 & 1.069 & & 1.087 & 1.190 \\
    G-Eymol~\cite{zanca2019gravitational} & & 0.583 & 0.741 & 1.296 & 0.550 & 0.676 & 0.769 & & 9.350 &	8.990 & & 8.622 & 4.127 & & & 0.870 & 0.523 & 0.444 & 0.431 & 0.187 & 0.491 & & 9.942 & 2.513 & & 9.799 & 3.771 \\
    IOR-ROI-LSTM~\cite{chen2018scanpath} & & 1.107 & 0.442 & \underline{0.013} & 0.444 & 0.028 & 0.407 & & 1.540 & 1.520 & & 0.546 & 1.005 & & & 0.677 & 0.446 & \underline{0.021} & 1.099 & 0.051 & 0.459 & & 0.985 & 0.875 & & 0.437 & 5.302 \\
    DeepGazeIII~\cite{kummerer2022deepgaze} & & 0.037 & \underline{0.016} & 0.019 & \underline{0.028} & - & \underline{0.025} & & - & 0.368 & & - & 0.393 & & & - & - & - & - & - & - & & - & - & & - & - \\
    ChenLSTM~\cite{chen2021predicting} & & 0.034 & 0.128 & 0.105 & 0.045 & 0.189 & 0.100 & & 0.574 & 0.373 & & 0.344 & 0.442 & & & 0.028 & 0.073 & 0.149 & 0.107 & 0.110 & 0.094 & & 0.168 & 0.161 & & 0.192 & 0.316 \\
    HAT~\cite{yang2024unifying} & & 1.099 & 0.434 & 0.042 & 0.444 & - & 0.505 & & - & 1.025 & & - & 0.331 & & & 1.196 & 0.522 & 0.381 & 2.386 & - & 1.121 & & - & 2.112 & & - & 1.305 \\
    ChenLSTM-ISP~\cite{chen2024beyond} & & 0.038 & 0.173 & 0.166 & 0.077 & 0.188 & 0.128 & & 0.683 & 0.576 & & 0.377 & 0.579 & & & 0.034 & 0.124 & 0.175 & 0.114 & 0.095 & 0.108 & & 0.264 & 0.214 & & 0.267 & 0.240 \\
    GazeXplain~\cite{chen2024gazexplain} & & 0.151 & 0.195 & 0.874 & 0.164 & 0.382 & 0.353 & & 3.915 & 3.423 & & 2.278 & 5.616 & & & \underline{0.018} & 0.065 & 0.079 & 0.058 & 0.188 & 0.082 & & \underline{0.035} & 0.094 & & 0.072 & 1.419 \\
    \midrule
    \rowcolor{Gray}
    DeepGazeIII~\cite{kummerer2022deepgaze} & & \textbf{\underline{0.009}} & \textbf{0.017} & 0.059 & 0.038 & - & \textbf{0.031} & & - & 0.348 & & - & 0.417 & & & \textbf{0.025} & 0.020 & 0.210 & 0.074 & - & 0.082 & & - & 0.210 & & - & 3.878  \\
    \rowcolor{Gray}
    ChenLSTM~\cite{chen2021predicting} & & 0.715 & 0.411 & 0.056 & 0.129 & 0.092 & 0.280 & & 0.116 & 0.110 & & 0.022 & 0.093 & & & 0.251 & 0.153 & 0.181 & 0.136 & 0.059 & 0.156 & & 0.373 & 0.251 & & 0.284 & 0.236 \\
    \rowcolor{Gray}
    GazeXplain~\cite{chen2024gazexplain} & & 0.346 & 0.226 & 0.032 & \textbf{0.033} & 0.068 & 0.141 & & 0.049 & 0.038 & & 0.017 & \underline{\textbf{0.007}} & & & 0.060 & 0.046 & 0.065 & 0.025 & 0.044 & 0.048 & & 0.158 & \textbf{\underline{0.051}} & & 0.128 & \textbf{\underline{0.043}} \\  
    \rowcolor{Gray}
    TPP-Gaze~\cite{damelio2025tpp} & & 0.063 & \textbf{0.017} & 0.061 & 0.038 & \textbf{\underline{0.010}} & 0.038 & & 0.125 & 0.226 & & 0.033 & 0.130 & & & 0.039 & 0.036 & 0.139 & 0.068 & \textbf{\underline{0.027}} & 0.062 & & 0.244 & 0.257 & & 0.144 & 0.280 \\  
    \rowcolor{OurColor} 
    \textbf{\ours (Ours)} & & 0.131 & 0.048 & \textbf{0.021} & 0.037 & 0.151 & 0.078 & & \underline{\textbf{0.015}} & \underline{\textbf{0.027}} & & \underline{\textbf{0.013}} & 0.038 & & & 0.050 & \textbf{\underline{0.015}} & \textbf{0.042} & \textbf{\underline{0.019}} & 0.072 & \textbf{\underline{0.040}} & & \textbf{0.041} & 0.065 & & \textbf{\underline{0.026}} & 0.047 \\
    \bottomrule
  \end{tabular}
  }
  \vspace{-0.15cm}
  \caption{Performance comparison of different models on the COCO-FreeView~\cite{yang2023predicting} and MIT1003~\cite{judd2009learning} datasets. Models trained using identical settings and datasets to \scandiff are highlighted in \colorbox{Gray}{\textbf{gray}}. Among these, the highest performance for each metric is marked in \textbf{bold}. \underline{Underlined} values denote the top overall performance across all models and metrics.}
  \label{tab:results_cocofreeview_mit}
  \vspace{-0.3cm}
\end{table*}

\subsection{Comparison with the State of the Art} 
We evaluate \scandiff by comparing it with existing scanpath prediction models in both free-viewing and visual search tasks. Our evaluation includes a diverse range of approaches and architectures, covering both traditional model-based methods (\eg~Itti-Koch~\cite{IttiKoch98}, CLE~\cite{bfpha04}, and G-Eymol~\cite{zanca2019gravitational}) and deep learning-based models (\eg~PathGAN~\cite{assens2018pathgan}, IOR-ROI-LSTM~\cite{chen2018scanpath}, DeepGazeIII~\cite{kummerer2022deepgaze}, ChenLSTM~\cite{chen2021predicting}, Gazeformer~\cite{mondal2023gazeformer}, HAT~\cite{yang2024unifying}, ChenLSTM-ISP~\cite{chen2024beyond}, GazeXplain~\cite{chen2024gazexplain}, and TPP-Gaze~\cite{damelio2025tpp}).

\tit{Free-Viewing Results}
Table~\ref{tab:results_cocofreeview_mit} presents a comprehensive evaluation across the considered free-viewing datasets. For a fair comparison, the most recent models were re-trained using identical settings and datasets to \scandiff. These results are reported in gray color at the bottom of the table.

For the COCO-FreeView dataset, \scandiff demonstrates competitive performance across multiple metrics. Our approach achieves the best results in the MM-direction metric among models trained with identical settings. This indicates the superior ability of our model to predict saccade directions that match human scanpaths. Furthermore, our model shows strong performance in the MM-position metric and achieves the best overall SM and SS metrics when considering fixation duration, demonstrating effective modeling of temporal dynamics. The MIT1003 dataset results further validate the effectiveness of \scandiff. Our model achieves the highest MM average score among all the competitors, demonstrating its overall strong performance. The superior results in SM and SS with duration is confirmed also on this dataset, outperforming all competing methods. This strong performance on the duration-aware metrics highlights the ability of our model to effectively represent the temporal aspects of visual attention. The unified diffusion-based architecture allows capturing both the spatial patterns of eye movements and their temporal characteristics, enabling more realistic scanpath generation.

\tit{Visual Search Results}
The results summarized in Table~\ref{tab:results_search} demonstrate the superior performance of our proposed model on the COCO-Search18 dataset across multiple evaluation metrics and search scenarios. We evaluate performance in both target-present and target-absent conditions, which represent fundamentally different search behaviors in human visual attention.
In the target-present scenario, \scandiff achieves state-of-the-art performance across all metrics. Most notably, our model exhibits a significant improvement in MM distributions, with an average KL divergence of $0.048$, which represents a $71.3\%$ reduction compared to the second best model (\ie, GazeXplain at $0.167$). For SemSS, \scandiff attains KL divergence values of $0.072$ and $0.078$ with and without duration information, respectively, demonstrating consistent performance across temporal aspects of gaze behavior.
The target-absent condition presents a particularly challenging scenario, as human attention patterns become more exploratory when the target object cannot be found. Even in this setting, \scandiff outperforms all baseline methods by substantial margins.

It is worth noting that while some competing methods like GazeXplain perform reasonably well in specific metrics, they lack the consistent performance across all evaluation dimensions that \scandiff demonstrates. This consistency across metrics and conditions indicates that our model better captures the underlying mechanisms of human visual search behavior in both goal-directed (target-present) and exploratory (target-absent) scenarios. 

\tit{Qualitative Results} Fig.~\ref{fig:qualitatives} shows some qualitative results comparing \scandiff with other competitors in both free-viewing and visual search settings. These results confirm the effectiveness of our model also from a qualitative point-of-view, highlighting its ability in generating human-like eye movement trajectories across diverse scenarios.

\begin{table*}[t]
    \centering
  \small
  \setlength{\tabcolsep}{.3em}
  \resizebox{0.95\linewidth}{!}{
  \begin{tabular}{lc c c cc c cc c cc cc c c cc c cc c cc}
    \toprule
    & & \multicolumn{10}{c}{\textbf{Target-Present}} & & \multicolumn{10}{c}{\textbf{Target-Absent}} \\
    \midrule
    & & \multicolumn{1}{c}{\textbf{MM} $\downarrow$} & & \multicolumn{2}{c}{\textbf{SM} $\downarrow$} & & \multicolumn{2}{c}{\textbf{SS} $\downarrow$} & & \multicolumn{2}{c}{\textbf{SemSS} $\downarrow$} & & & \multicolumn{1}{c}{\textbf{MM} $\downarrow$} & & \multicolumn{2}{c}{\textbf{SM} $\downarrow$} & & \multicolumn{2}{c}{\textbf{SS} $\downarrow$} & & \multicolumn{2}{c}{\textbf{SemSS} $\downarrow$} \\
    \cmidrule{3-3} \cmidrule{5-6} \cmidrule{8-9} \cmidrule{11-12} \cmidrule{15-15} \cmidrule{17-18} \cmidrule{20-21} \cmidrule{23-24}
    & & Avg & & w/ Dur & w/o Dur & & w/ Dur & w/o Dur & & w/ Dur & w/o Dur & & & Avg & & w/ Dur & w/o Dur & & w/ Dur & w/o Dur & & w/ Dur & w/o Dur \\
    \midrule
    PathGAN~\cite{assens2018pathgan} & & 0.513 & & 1.891 & 2.451 & & 0.808 & 0.939 & & 0.313 & 0.468 & & & 0.125 & & 0.792 & 0.357 & & 0.498 & 0.944 & & 0.212 & 0.465 \\    
    ChenLSTM~\cite{chen2021predicting} & & 0.197 & & 0.011 & 0.236 & & 0.040 & 0.262 & & 0.084 & 0.162 & & & 0.075 & & 0.010 & 0.012 & & 0.036 & 0.775 & & 0.044 & 0.677 \\
    Gazeformer~\cite{mondal2023gazeformer} & & 0.281 & & 0.027 & 0.340 & & 0.119 & 0.268 & & 0.131 & 0.208 & & & 0.089 & & 0.061 & 0.075 & & 0.102 & 0.245 & & 0.085 & 0.139 \\
    HAT~\cite{yang2024unifying} & & 0.118 & & - & 0.263 & & - & 0.148 & & - & 3.837 & & & 0.052 & & - & 0.063 & & - & 0.097 & & - & 3.472 \\
    ChenLSTM-ISP~\cite{chen2024beyond} & & 0.174 & & 0.013 & 0.306 & & \underline{0.015} & 0.257 & & \underline{0.043} & 0.097 & & & 0.082 & & 0.028 & 0.146 & & 0.063 & 0.727 & & 0.052 & 0.561 \\
    GazeXplain~\cite{chen2024gazexplain} & & 0.166 & & 0.023 & 0.237 & & 0.070 & 0.232 & & 0.140 & 0.188 & & & 0.046 & & 0.062 & 0.048 & & 0.038 & 0.191 & & 0.043 & 0.206 \\
    TPP-Gaze~\cite{damelio2025tpp} & & 0.524 & & 1.618 & 3.218 & & 0.579 & 1.590 & & 0.554 & 1.147 & & & 0.098 & & 0.511 & 0.529 & & 0.242 & 0.325 & & 0.093 & 0.164 \\    
    \midrule
    \rowcolor{Gray}
    Gazeformer~\cite{mondal2023gazeformer} & & 0.251 & & 0.045 & 0.508 & & 0.048 & 0.349 & & 0.095 & 0.262 & & & 0.526 & & 1.184 & 1.688 & & 0.319 & 0.520 & & 0.671 & 1.043 \\
    \rowcolor{Gray} 
    GazeXplain~\cite{chen2024gazexplain} & & 0.167 & & \underline{\textbf{0.010}} & 0.238 & & 0.050 & 0.217 & & 0.092 & 0.224 & & & 0.037 & & 0.030 & 0.026 & & 0.028 & 0.146 & & 0.038 & 0.143 \\
    \rowcolor{Gray}
    TPP-Gaze~\cite{damelio2025tpp} & & 0.507 & & 2.317 & 3.995 & & 0.893 & 2.381 & & 0.736 & 1.605 & & & 0.135 & & 0.775 & 0.887 & & 0.427 & 0.537 & & 0.231 & 0.300 \\    
    \rowcolor{OurColor} 
    \textbf{\ours (Ours)} & & \underline{\textbf{0.048}} & &	0.037 & \underline{\textbf{0.079}} & & \textbf{0.019} & \underline{\textbf{0.074}} & & \textbf{0.072} & \underline{\textbf{0.078}} & & & \underline{\textbf{0.020}} & & \underline{\textbf{0.005}} & \underline{\textbf{0.008}} & & \underline{\textbf{0.008}} & \underline{\textbf{0.031}} & & \underline{\textbf{0.007}} & \underline{\textbf{0.024}} \\
    \bottomrule
  \end{tabular}
  }
  \vspace{-0.2cm}
  \caption{Performance comparison of different models on the COCO-Search18 dataset~\cite{Yang_2020_CVPR,chen2021coco} for both target-present and target-absent settings. Models trained using identical settings and training splits to \scandiff are highlighted in \colorbox{Gray}{\textbf{gray}}. Among these, the highest performance for each metric is marked in \textbf{bold}. \underline{Underlined} values denote the top overall performance across all models and metrics.}
  \label{tab:results_search}
  \vspace{-0.35cm}
\end{table*}

\begin{table}[t]
    \centering
  \small
  \setlength{\tabcolsep}{.18em}
  \resizebox{\linewidth}{!}{
  \begin{tabular}{lc cc c ccc c cccc}
    \toprule
    & & \multicolumn{2}{c}{\textbf{Backbones}} & & \multicolumn{3}{c}{\textbf{COCO-FreeView}} & & \multicolumn{4}{c}{\textbf{COCO-Search18}} \\
    \cmidrule{3-4} \cmidrule{6-8} \cmidrule{10-13}
    & & Textual & Visual & & \textbf{MM} $\downarrow$ & \textbf{SM} $\downarrow$ & \textbf{SS} $\downarrow$ & & \textbf{MM} $\downarrow$ & \textbf{SM} $\downarrow$ & \textbf{SS} $\downarrow$ & \textbf{SemSS} $\downarrow$ \\
    \midrule
    \rowcolor{Gray} 
    \multicolumn{13}{l}{\textit{Effect of Textual Backbone}} \\
    & & RoBERTa & DINOv2 & & 0.110 & 0.181 & 0.111 & & 0.076 & 0.143 & 0.072 & 0.132\\
    \rowcolor{OurColor} 
    \textbf{\ours} & & CLIP & DINOv2 & & \textbf{0.078} & \textbf{0.015} & \textbf{0.013} & & \textbf{0.048} & \textbf{0.037} & \textbf{0.019} & \textbf{0.072}\\
    \midrule
    \rowcolor{Gray} 
    \multicolumn{13}{l}{\textit{Effect of Visual Backbone}} \\
    & & CLIP & RN50 & & \textbf{0.049} & 0.019 & 0.029 & & 0.070 & 0.052 & 0.031 & 0.084\\
    & & CLIP & CLIP & & 0.058 & 0.199 & 0.112 & & 0.090	& 0.079 & 0.033 & \textbf{0.069}  \\
    \rowcolor{OurColor} 
    \textbf{\ours} & & CLIP & DINOv2 & & 0.078 & \textbf{0.015} & \textbf{0.013} & & \textbf{0.048} & \textbf{0.037} & \textbf{0.019} & 0.072\\
    \bottomrule
  \end{tabular}
  }
  \vspace{-0.2cm}
  \caption{Performance comparison of different textual and visual backbones on COCO-FreeView~\cite{yang2020predicting} and COCO-Search18~\cite{chen2021coco} (TP) datasets. Best results are highlighted in \textbf{bold}.}
  \label{tab:ablation1}
    \vspace{-0.35cm}
\end{table}

\subsection{Ablation Studies}

To provide insights into the design choices of our model, we conduct a series of ablation studies examining the impact of different components on scanpath prediction performance. Tables~\ref{tab:ablation1} and~\ref{tab:ablation} summarize these results across both free-viewing and visual search tasks.

\tit{Effect of Textual and Visual Backbones}
We first investigate the influence of textual and visual backbones on model performance. As shown in Table~\ref{tab:ablation1}, replacing the CLIP text encoder with RoBERTa~\cite{liu2019roberta} leads to a degradation in performance across all metrics on both datasets. This highlights the importance of vision-language pre-training for scanpath prediction, as CLIP's joint embedding space better captures the semantic relationships between textual queries and visual features that guide human attention.

For the visual backbone comparison, we test our model with ResNet-50-FPN~\cite{he2017mask}, CLIP visual encoder (always using the ViT-B version), and DINOv2. The results indicate that while ResNet-50 achieves the best MM average score on COCO-FreeView, DINOv2 consistently outperforms other visual backbones across most metrics, particularly on the more challenging COCO-Search18 dataset. Interestingly, the CLIP visual encoder performs best on the SemSS metric, suggesting its strength in capturing semantic relationships between fixations and image regions, likely due to its vision-language pre-training.

\tit{Effect of $\mathcal{L}_T$ Loss Function}
We evaluate the contribution of the diffusion prior alignment loss $\mathcal{L}_T$.
As shown in Table~\ref{tab:ablation}, including $\mathcal{L}_T$ improves overall performance, but the benefits are more pronounced on the COCO-Search18 dataset, where it improves MM, SM and SemSS. This suggests that the convergence loss is particularly valuable for modeling the sequential nature of fixations in goal-directed visual search tasks, where the temporal order of fixations is more structured compared to free-viewing scenarios.

\begin{table}[t]
\centering
  \small
  \setlength{\tabcolsep}{.25em}
  \resizebox{\linewidth}{!}{
  \begin{tabular}{lc ccc c ccc c cccc}
    \toprule
    & & & & & & \multicolumn{3}{c}{\textbf{COCO-FreeView}} & & \multicolumn{4}{c}{\textbf{COCO-Search18}} \\
     \cmidrule{7-9} \cmidrule{11-14}
    & & $\mathcal{L}_T$ & & $T$ & & \textbf{MM} $\downarrow$ & \textbf{SM} $\downarrow$ & \textbf{SS} $\downarrow$ & & \textbf{MM} $\downarrow$ & \textbf{SM} $\downarrow$ & \textbf{SS} $\downarrow$ & \textbf{SemSS} $\downarrow$ \\
    \midrule
    \rowcolor{Gray} 
    \multicolumn{14}{l}{\textit{Effect of $\mathcal{L}_T$ Loss Function}} \\
    & & \xmark & & 1000 & & 0.088 & \textbf{0.011} & \textbf{0.009} & & 0.058 & 0.040 & \textbf{0.018} & 0.076 \\
    \rowcolor{OurColor} 
    \textbf{\ours} & & \cmark & & 1000 & & \textbf{0.078} & 0.015 & 0.013 & & \textbf{0.048} & \textbf{0.037} & 0.019 & \textbf{0.072} \\
    \midrule
    \rowcolor{Gray} 
    \multicolumn{14}{l}{\textit{Varying Diffusion Timesteps}} \\
    & & \cmark & & 200 & & \textbf{0.043} & 0.072 & 0.064 & & 0.069 & 0.040 & 0.023 & 0.085 \\
    & & \cmark & & 500 & & 0.049 & 0.145 & 0.085 & & \textbf{0.046} & 0.050 & 0.030 & 0.098 \\
    & & \cmark & & 1500 & & 0.045 & 0,131 & 0.122 & & 0.057 & 0.086 & 0.061 & 0.130 \\
    \rowcolor{OurColor} 
    \textbf{\ours} & & \cmark & & 1000 & & 0.078 & \textbf{0.015} & \textbf{0.013} & & 0.048 &\textbf{} \textbf{0.037} & \textbf{0.019} & \textbf{0.072} \\
    \bottomrule
  \end{tabular}
  }
  \vspace{-0.2cm}
  \caption{Ablation study on the effect of alignment loss ($\mathcal{L}_T$) and diffusion timesteps ($T$) on COCO-FreeView~\cite{yang2020predicting} and COCO-Search18~\cite{chen2021coco} (TP) datasets. Best results are highlighted in \textbf{bold}.}
  \label{tab:ablation}
\vspace{-0.35cm}
\end{table}

\tit{Validating Diffusion Timesteps}
Finally, we investigate the impact of diffusion timesteps ($T$) on model performance by varying $T$ from $200$ to $1500$. As shown in Table~\ref{tab:ablation}, $T=1000$ achieves the best overall balance across metrics and datasets. While smaller timesteps (\eg, $T=200$) can achieve better MM scores on COCO-FreeView, they underperform on the other spatial metrics. Similarly, $T=500$ achieves a slightly better MM score on COCO-Search18, but at the cost of poorer performance on other metrics.
Increasing $T$ beyond $1000$ (\ie, $T=1500$) leads to deteriorated performance across most metrics, likely due to overfitting to noise patterns. These results highlight the critical role of properly calibrating the diffusion process: sufficient timesteps are needed to learn complex distributions, but excessive noise can degrade the ability of the model to capture meaningful patterns in scanpath data.

\begin{figure*}[t]
    \centering
    \footnotesize
    \setlength{\tabcolsep}{.2em}
    \resizebox{0.99\linewidth}{!}{
    \begin{tabular}{ccccccc}
         & IOR-ROI-LSTM~\cite{chen2018scanpath} & ChenLSTM~\cite{chen2021predicting} & GazeXplain~\cite{chen2024gazexplain} & TPP-Gaze~\cite{damelio2025tpp} & \textbf{\ours (Ours)} & Humans \\
         \addlinespace[0.08cm]
         \raisebox{2.4\normalbaselineskip}[0pt][0pt]{\rotatebox[origin=c]{90}{\texttt{Free View}}} & \includegraphics[width=0.16\linewidth]{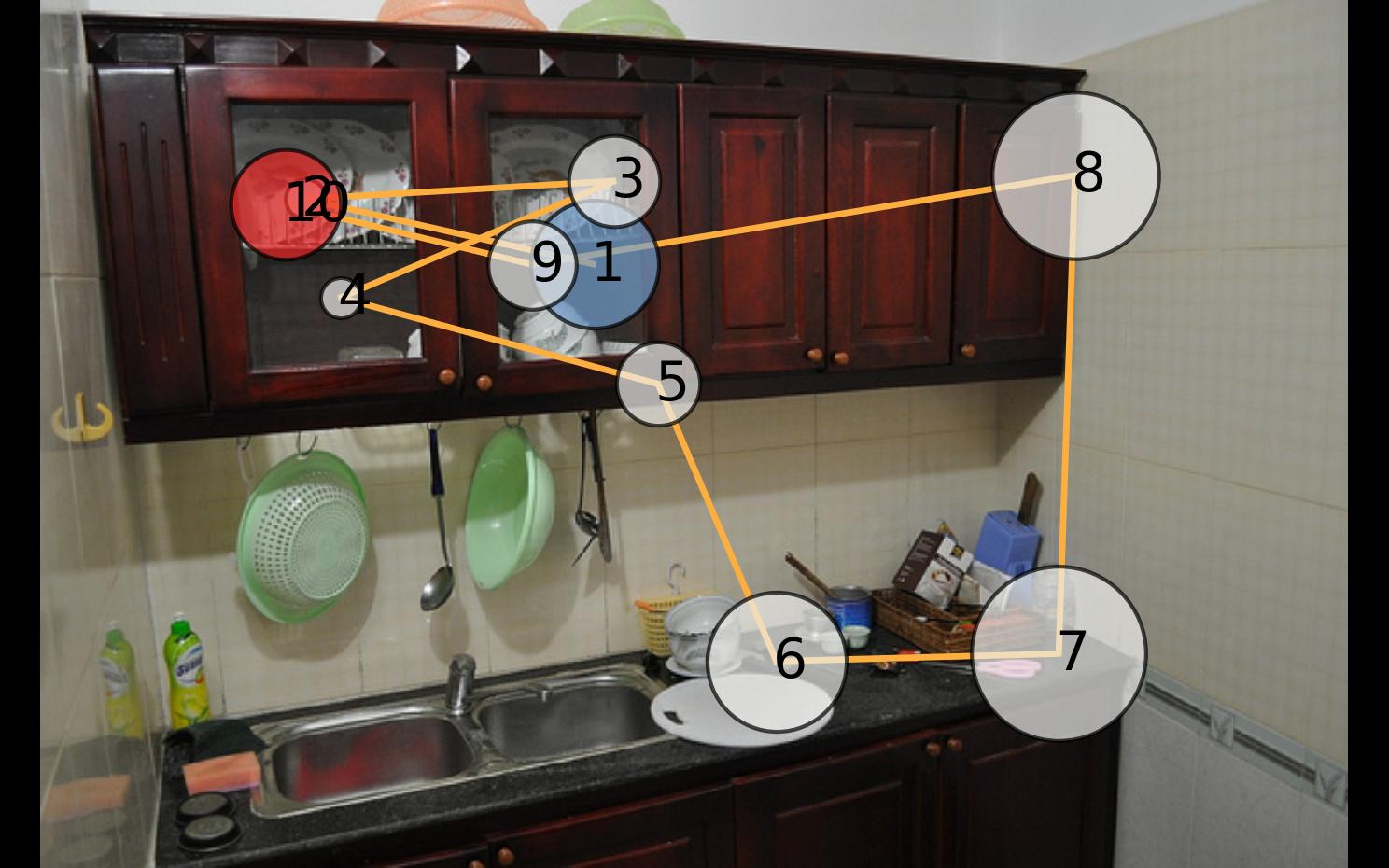} &
         \includegraphics[width=0.16\linewidth]{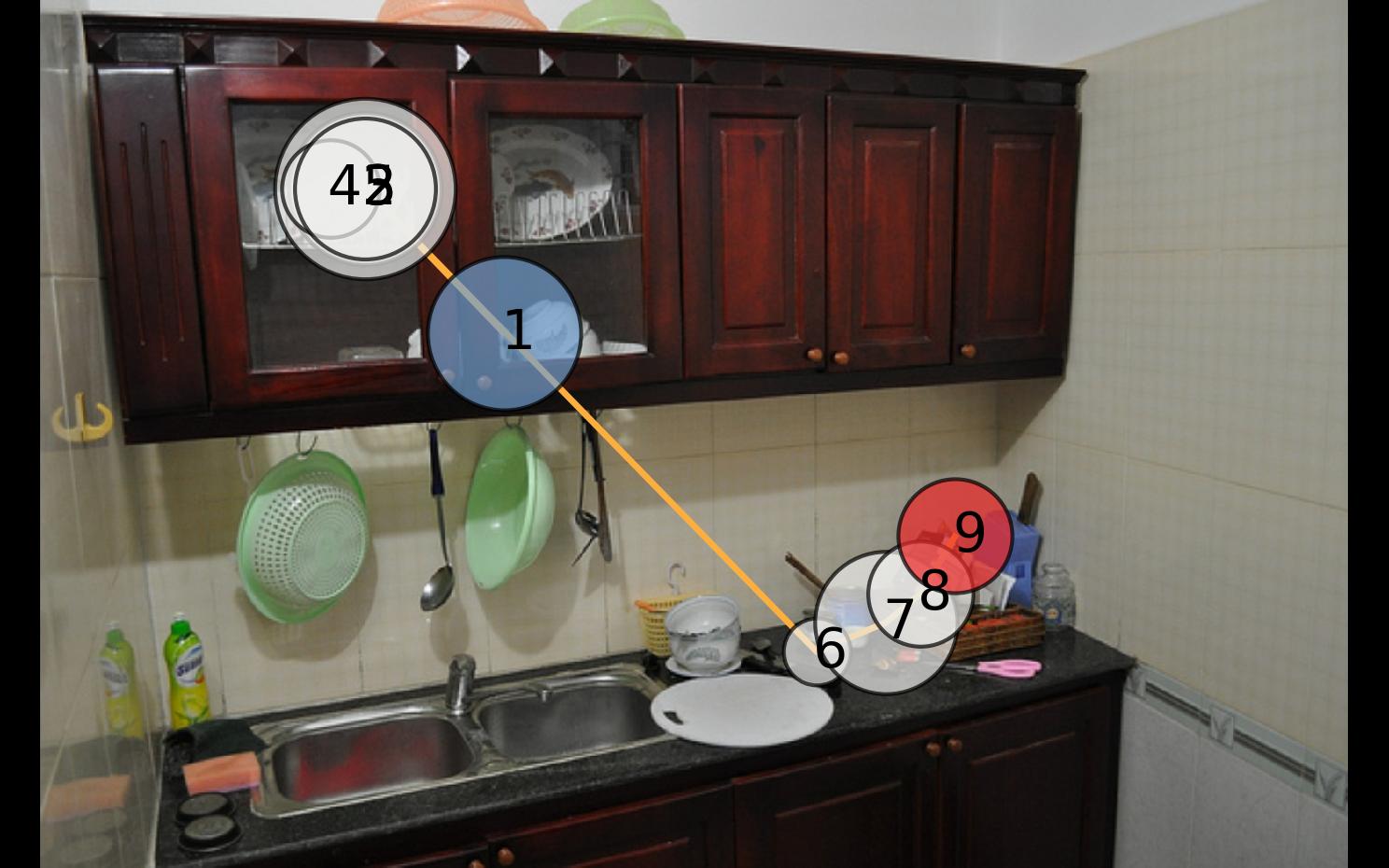} &
         \includegraphics[width=0.16\linewidth]{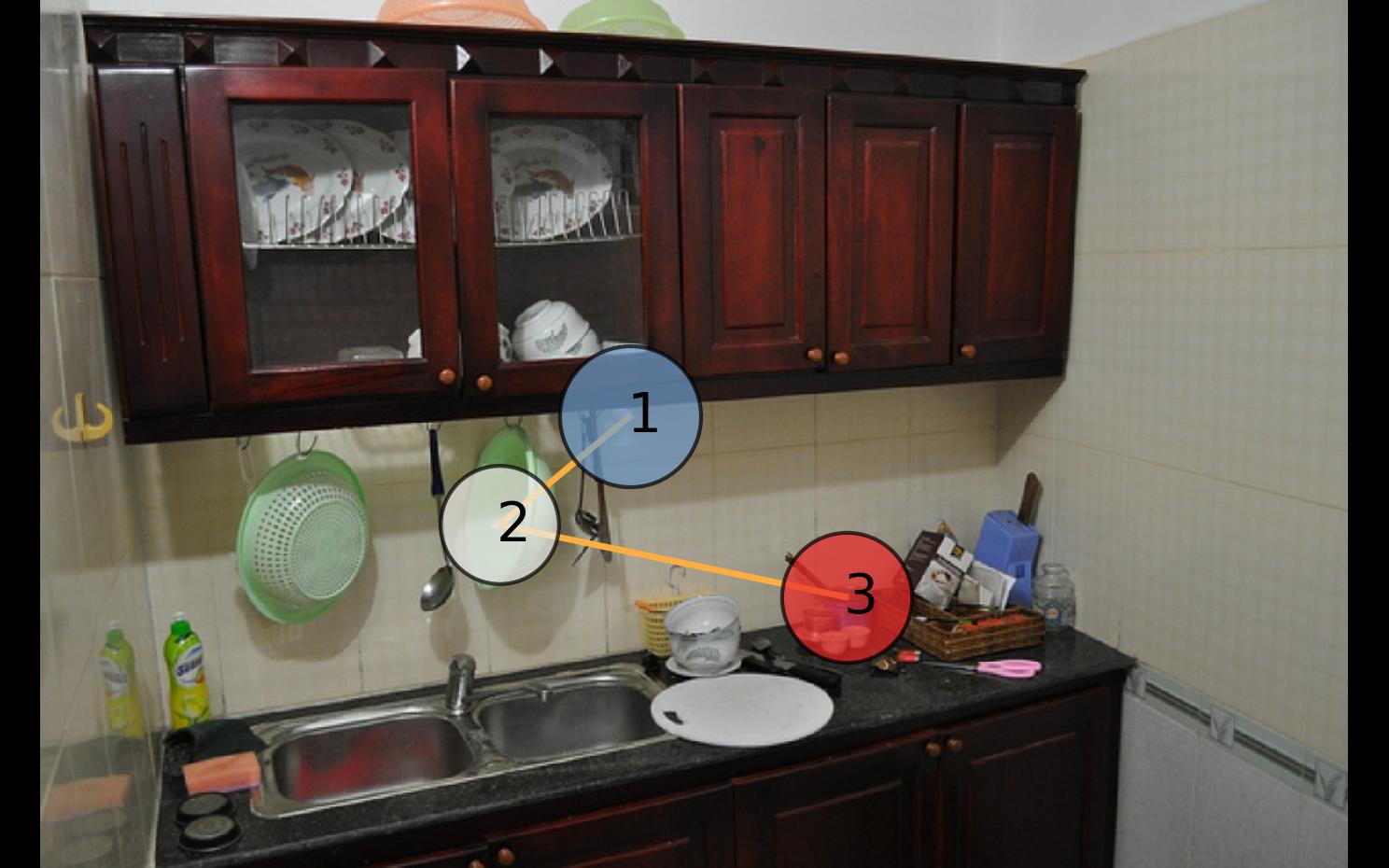} & 
         \includegraphics[width=0.16\linewidth]{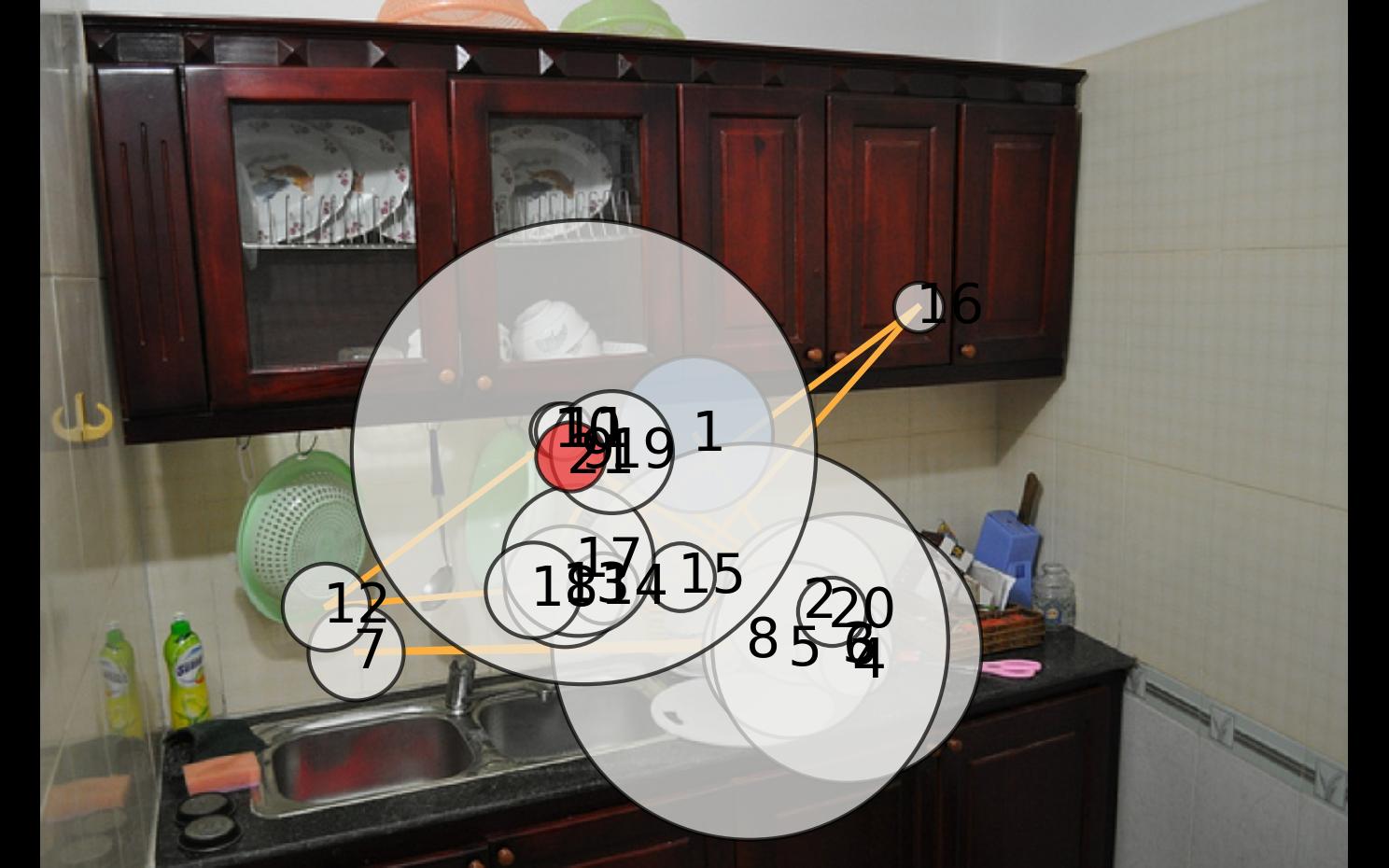} & 
         \includegraphics[width=0.16\linewidth]{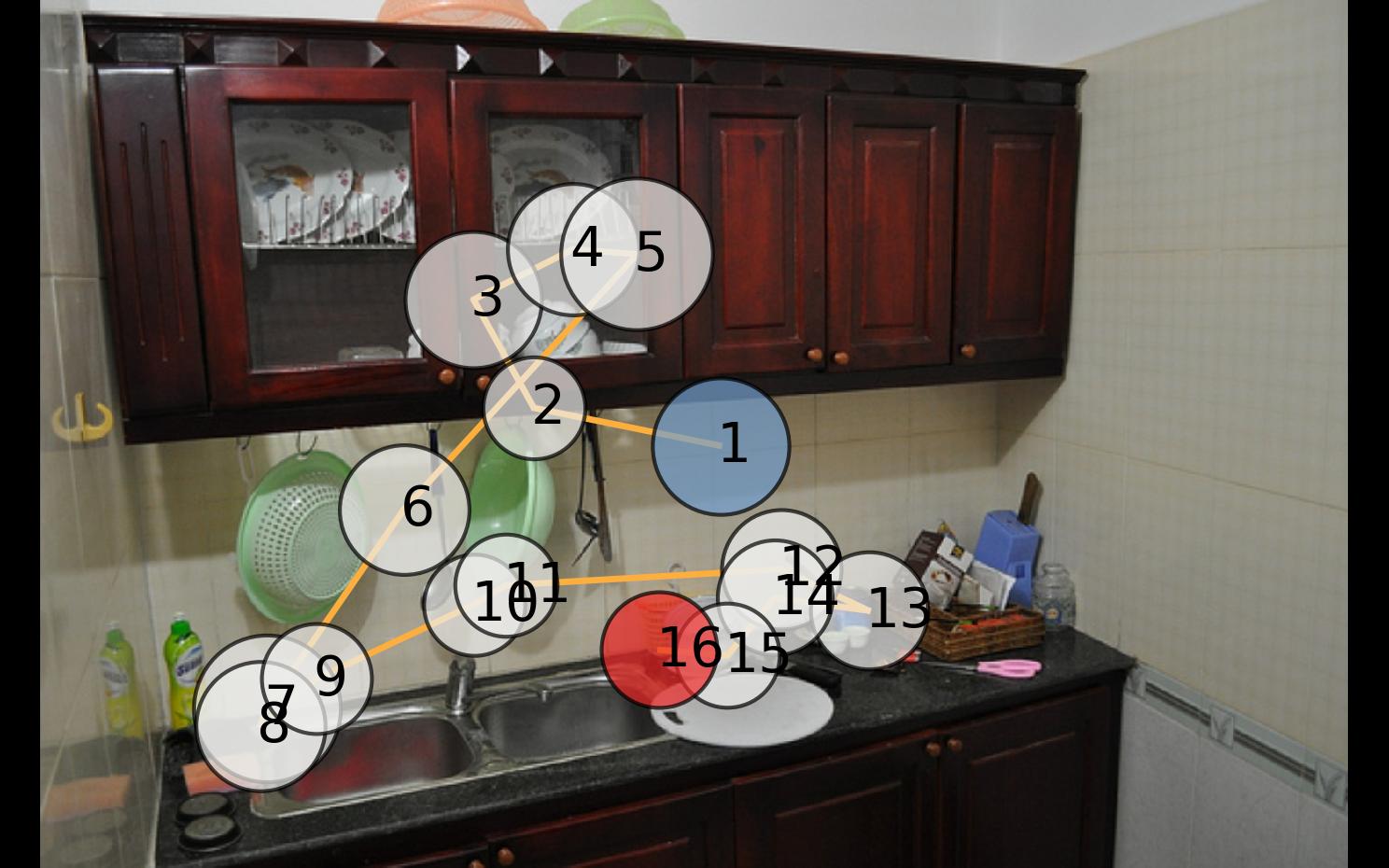} & 
         \includegraphics[width=0.16\linewidth]{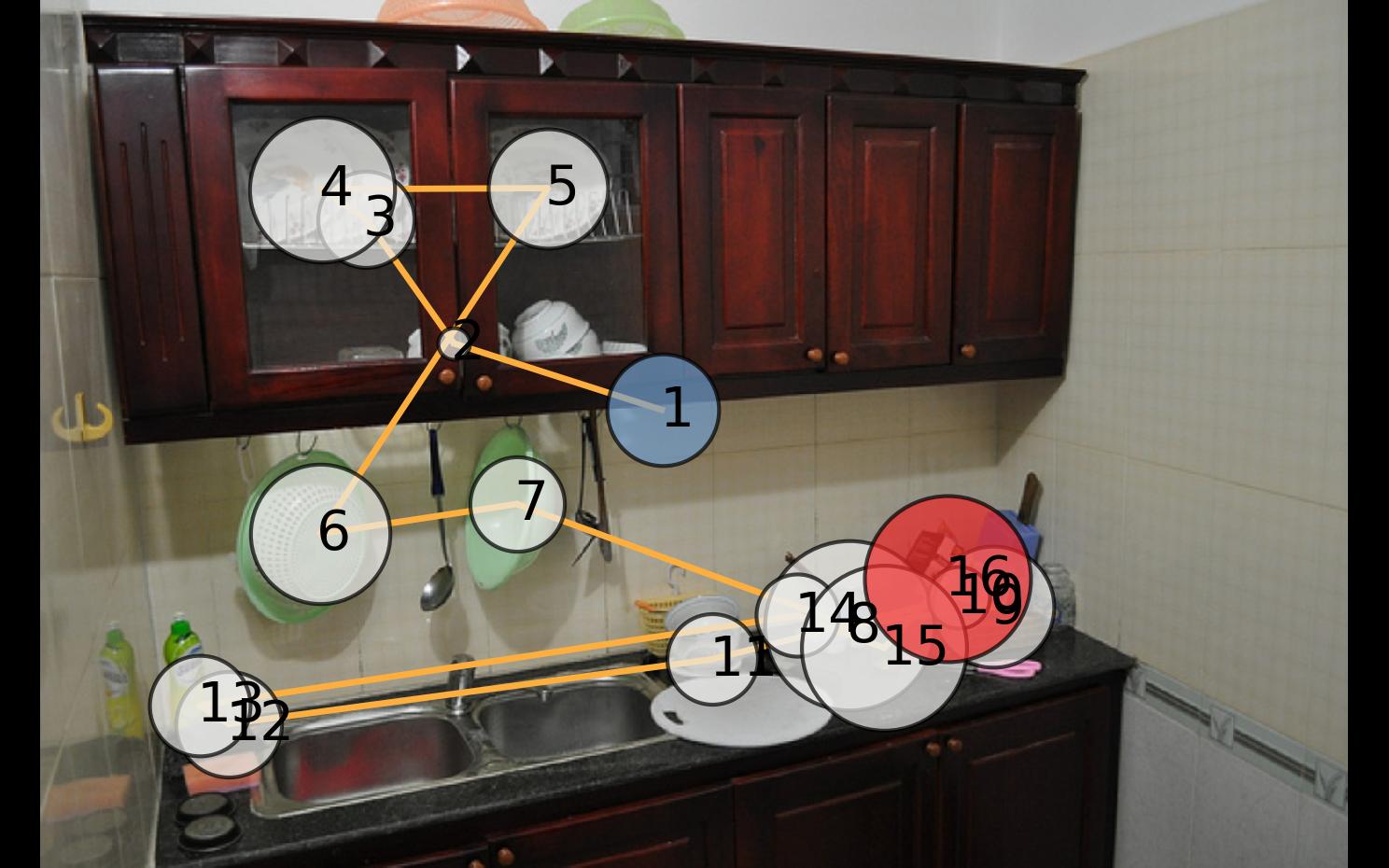} \\
         \raisebox{2.8\normalbaselineskip}[0pt][0pt]{\rotatebox[origin=c]{90}{\texttt{Free View}}} & \includegraphics[width=0.16\linewidth]{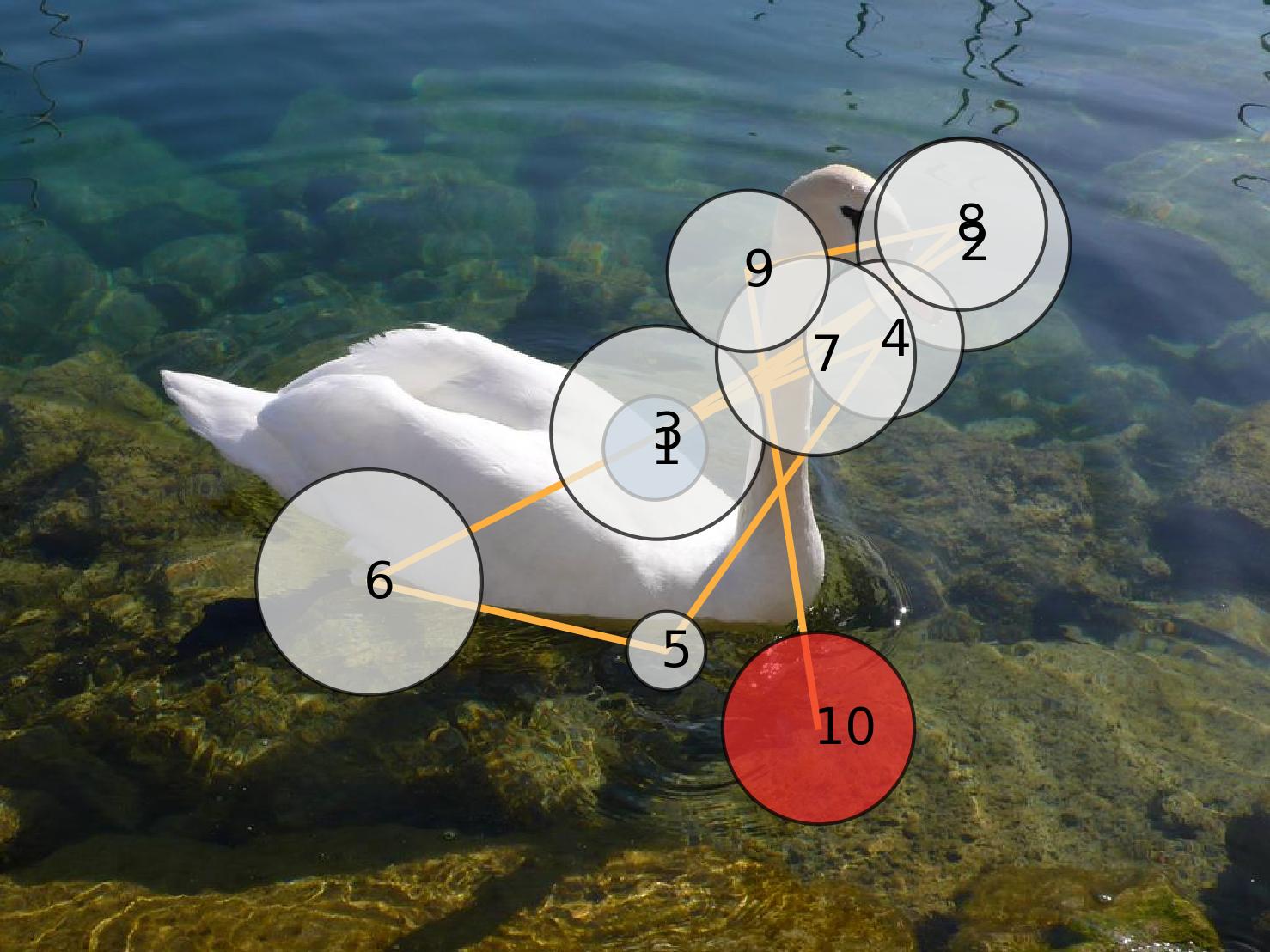} &
         \includegraphics[width=0.16\linewidth]{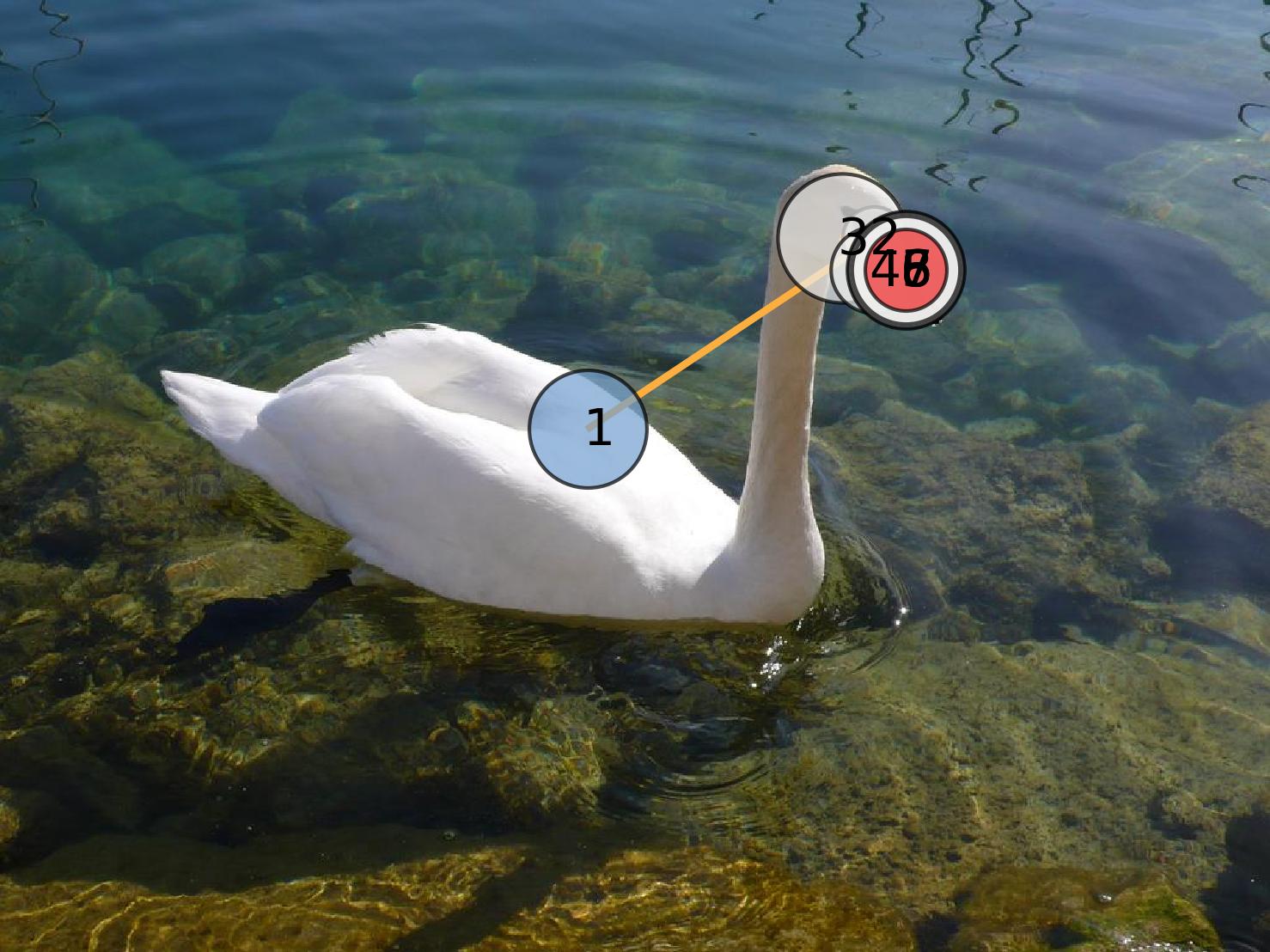} &
         \includegraphics[width=0.16\linewidth]{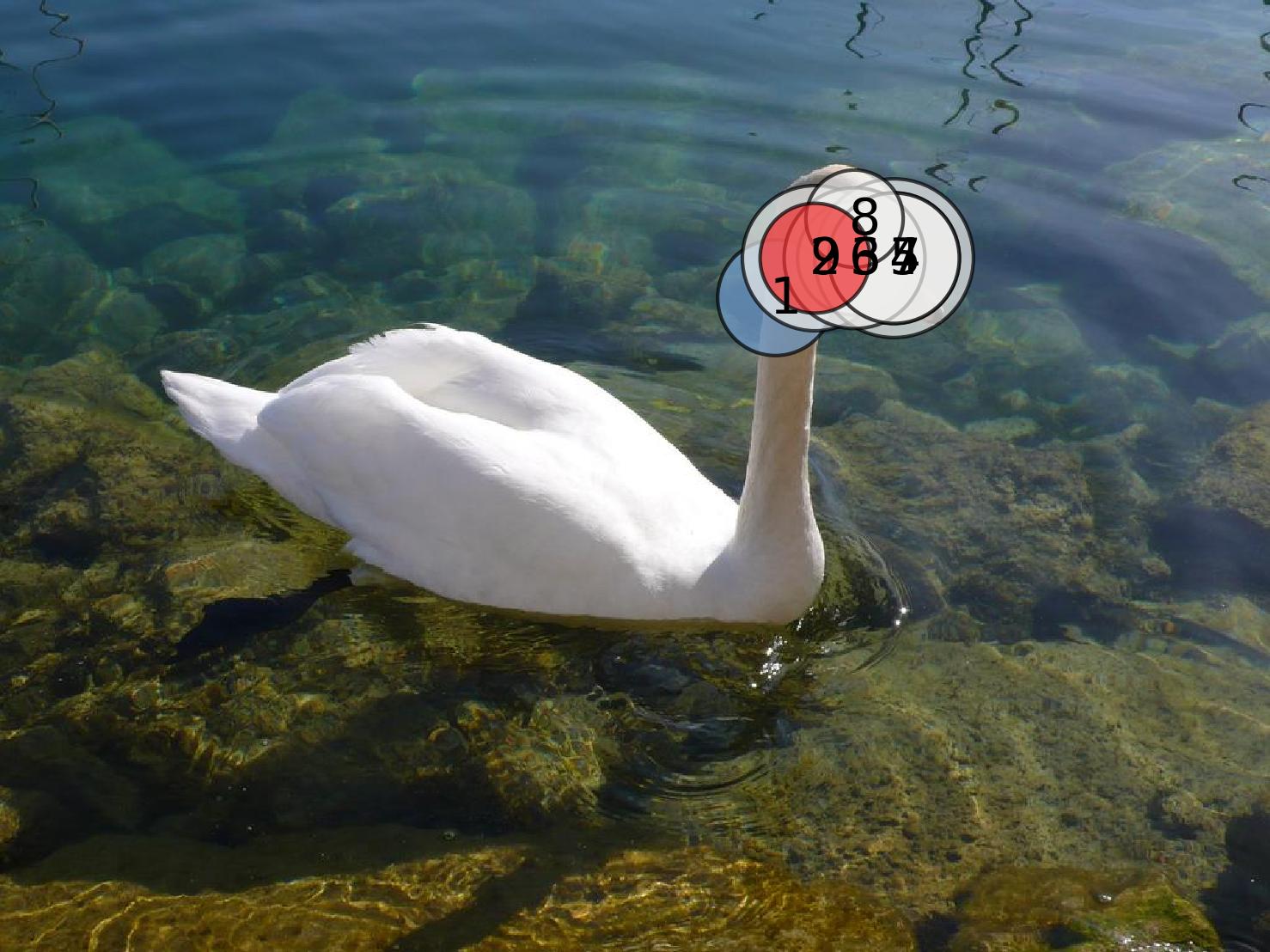} & 
         \includegraphics[width=0.16\linewidth]{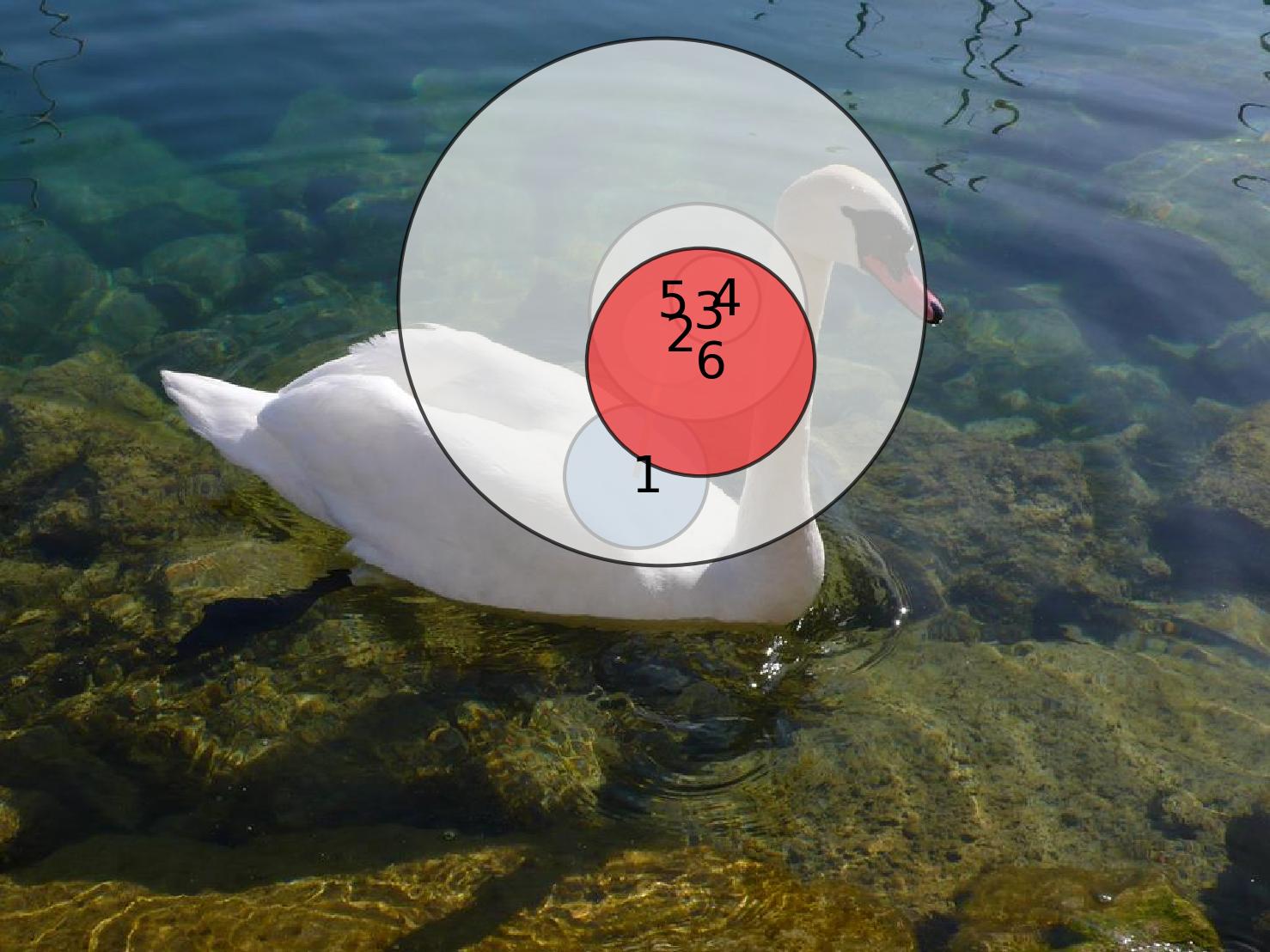} & 
         \includegraphics[width=0.16\linewidth]{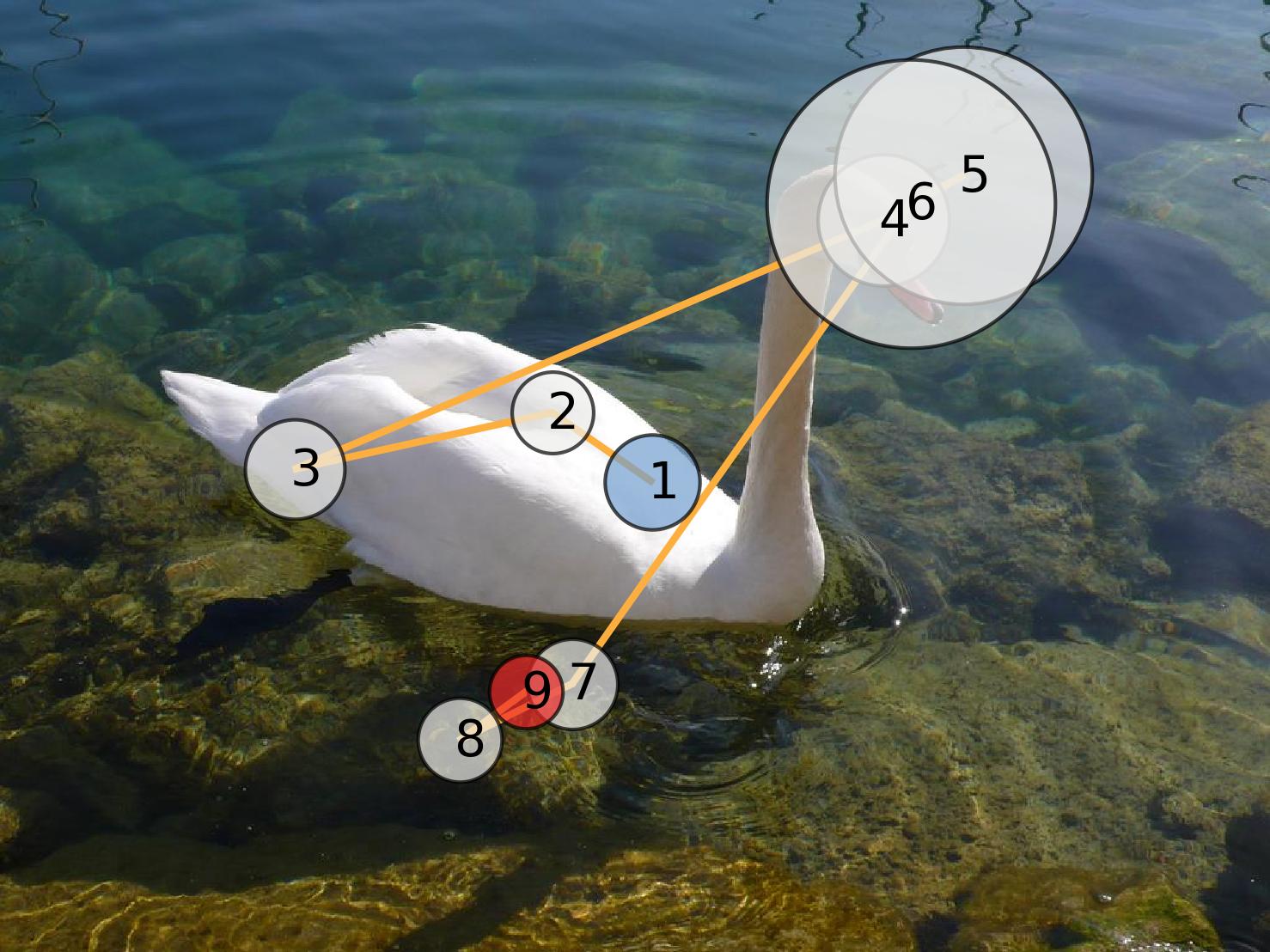} & 
         \includegraphics[width=0.16\linewidth]{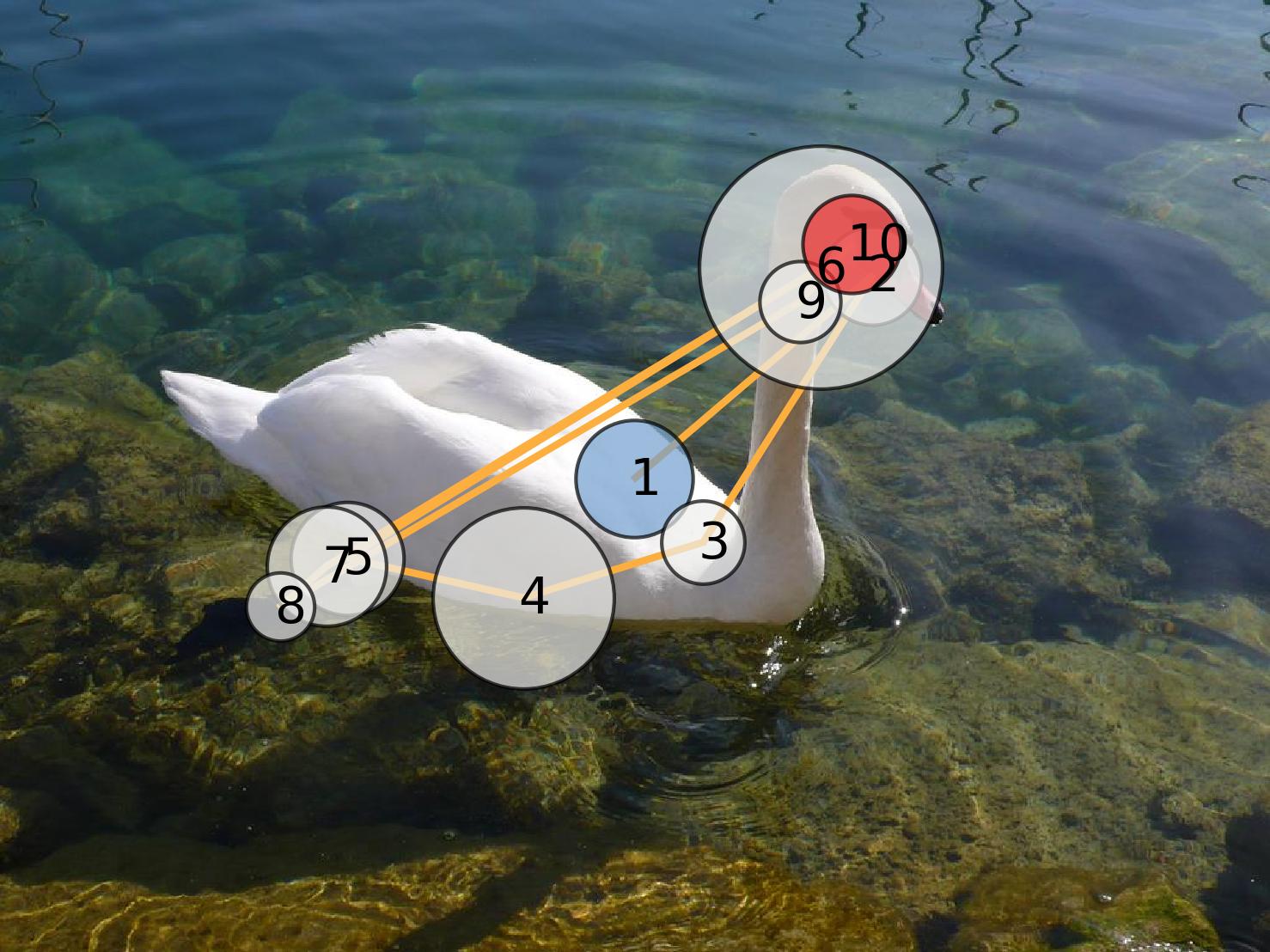} \\
        \addlinespace[0.1cm]
         & ChenLSTM~\cite{chen2021predicting} & Gazeformer~\cite{mondal2023gazeformer} & GazeXplain~\cite{chen2024gazexplain} & TPP-Gaze~\cite{damelio2025tpp} & \textbf{\ours (Ours)} & Humans \\
        \addlinespace[0.08cm]
         \raisebox{2.4\normalbaselineskip}[0pt][0pt]{\rotatebox[origin=c]{90}{\texttt{Plant}}} &
         \includegraphics[width=0.16\linewidth]{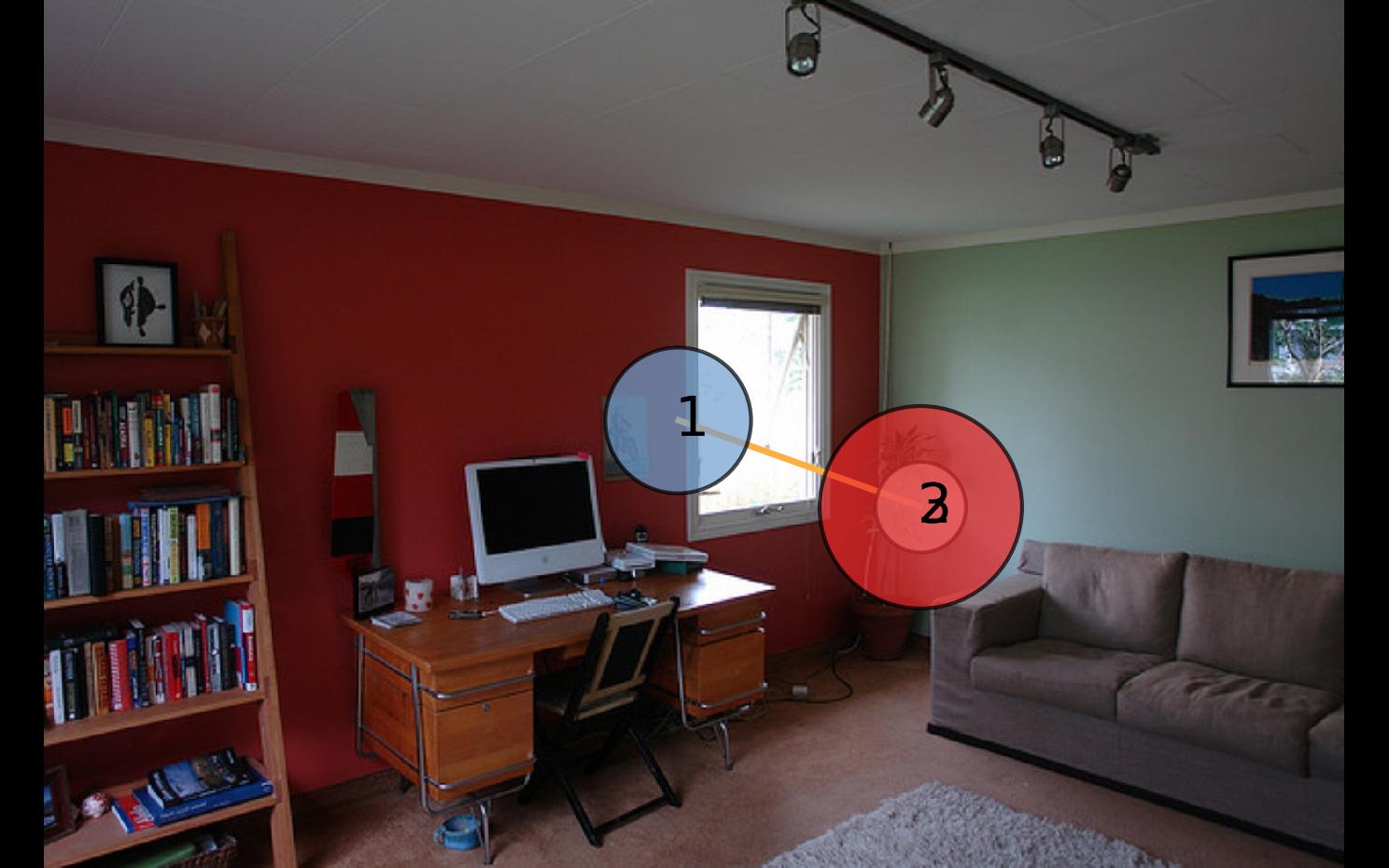} &
         \includegraphics[width=0.16\linewidth]{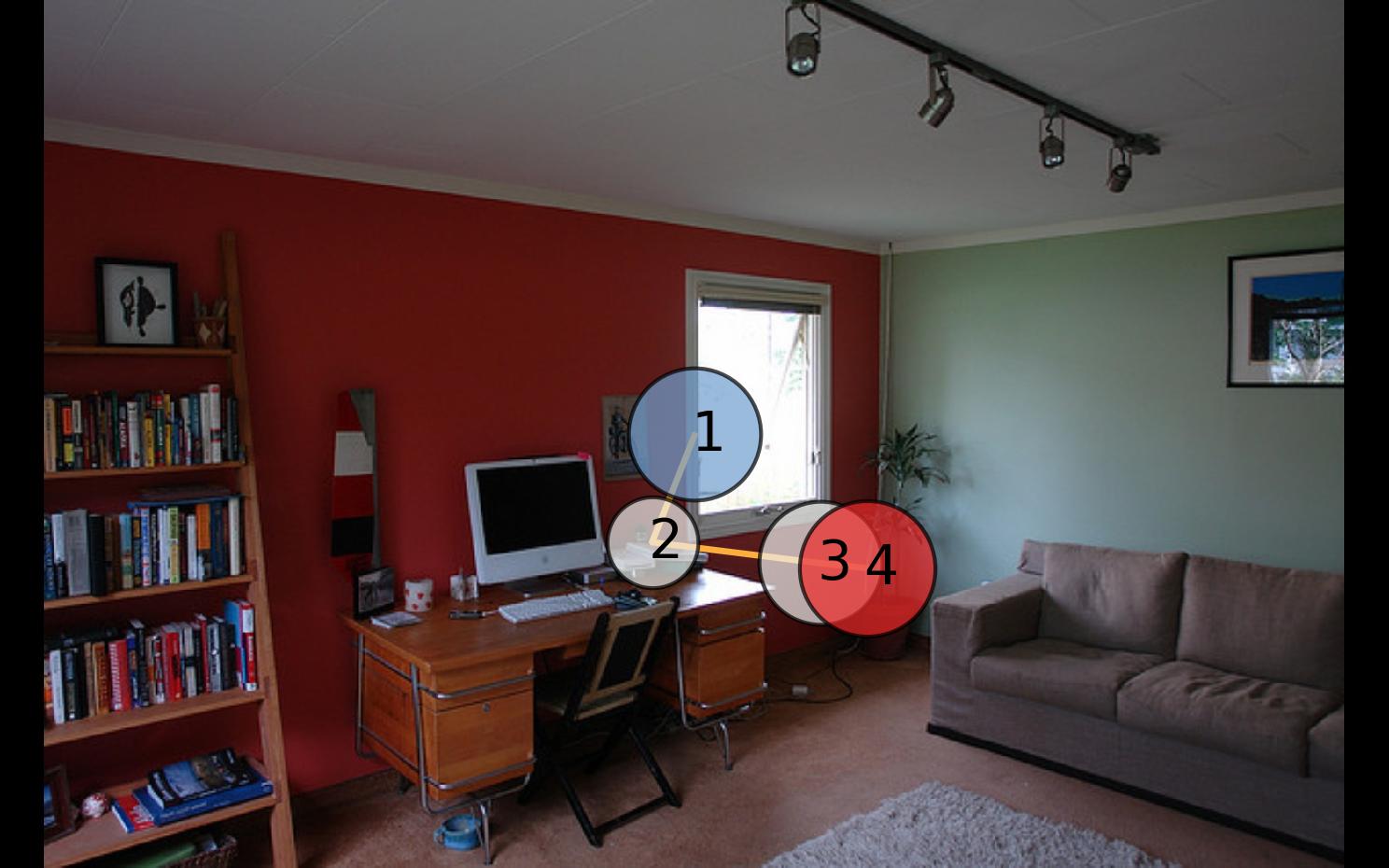} & 
         \includegraphics[width=0.16\linewidth]{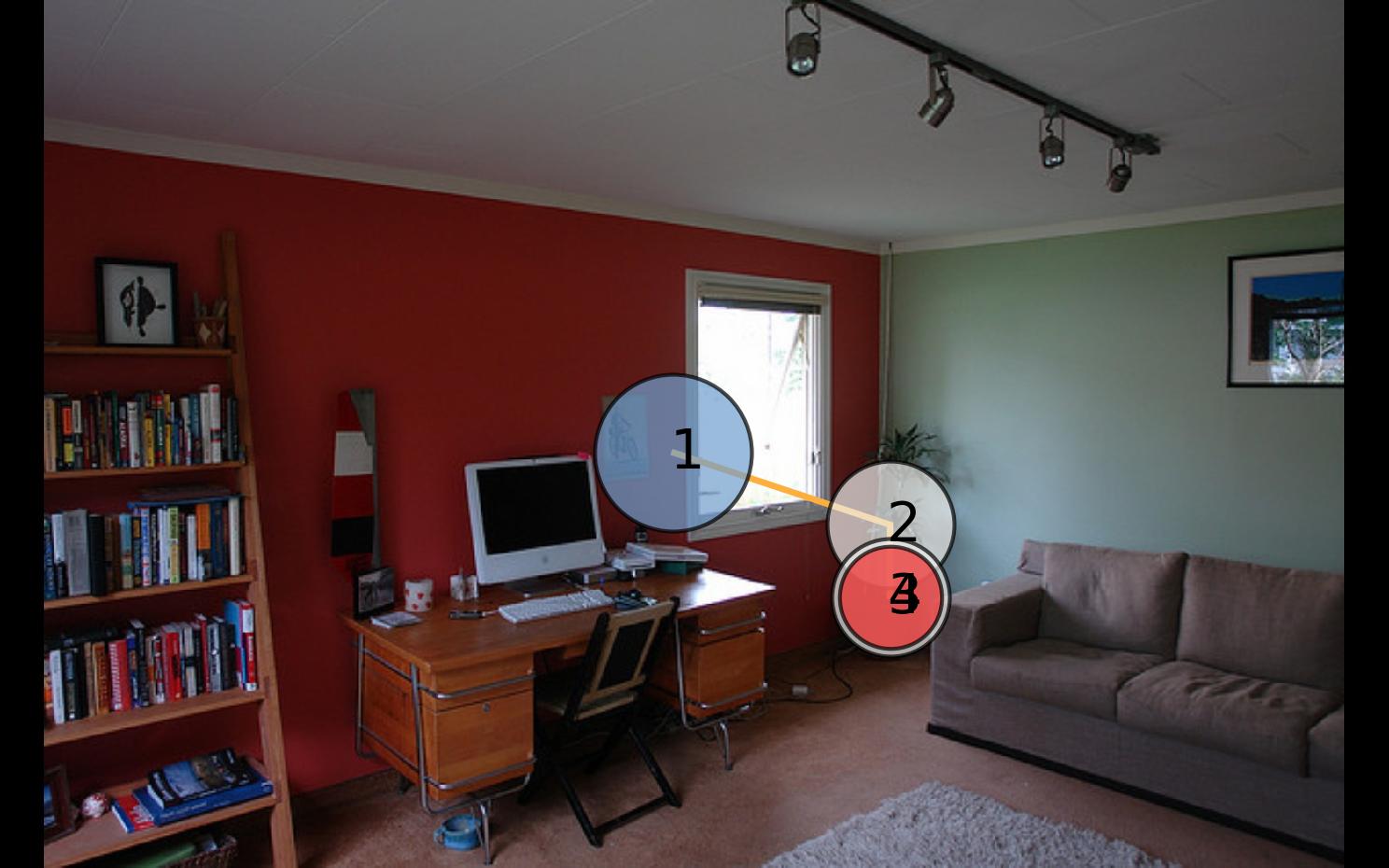} &
         \includegraphics[width=0.16\linewidth]{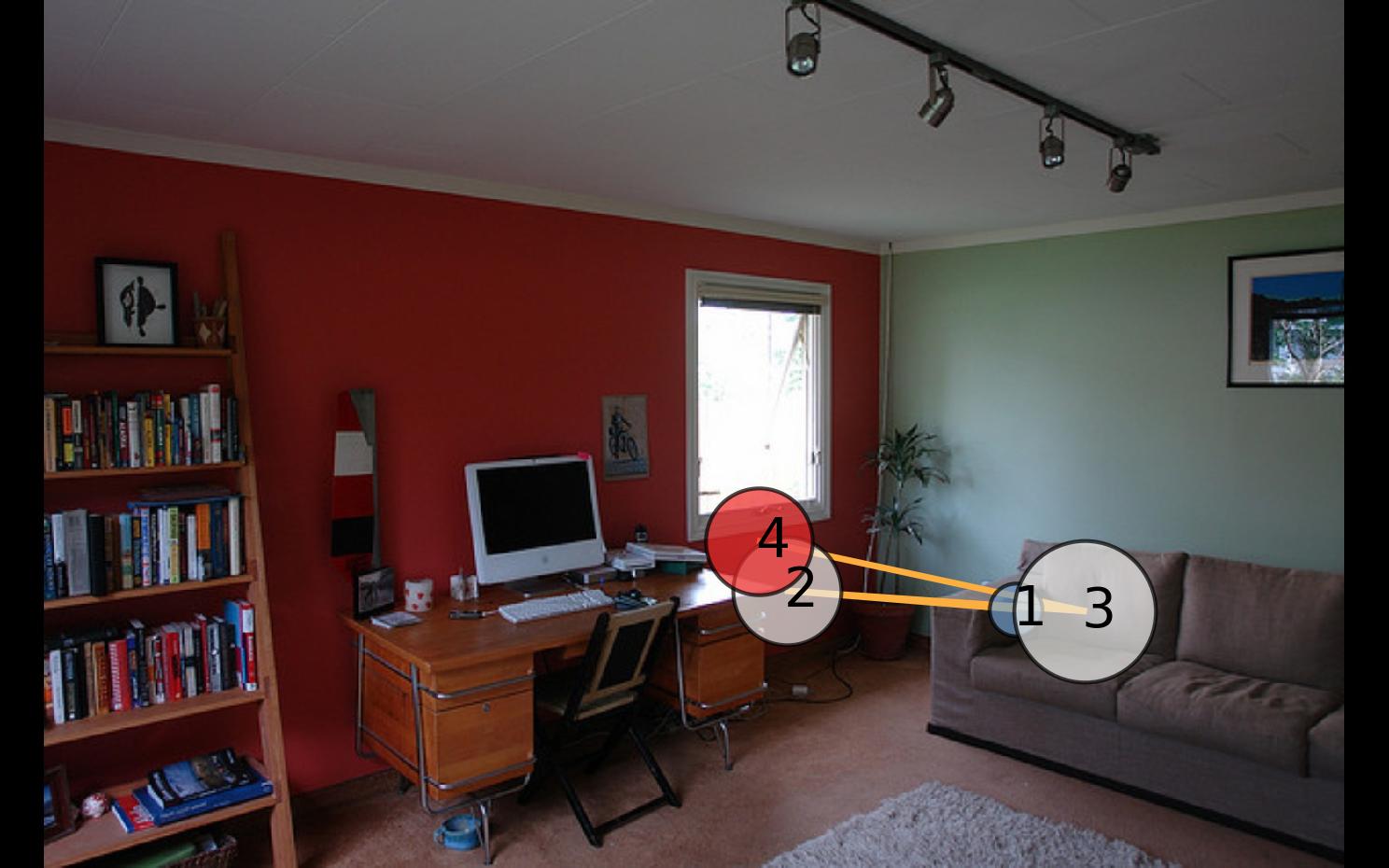} &
         \includegraphics[width=0.16\linewidth]{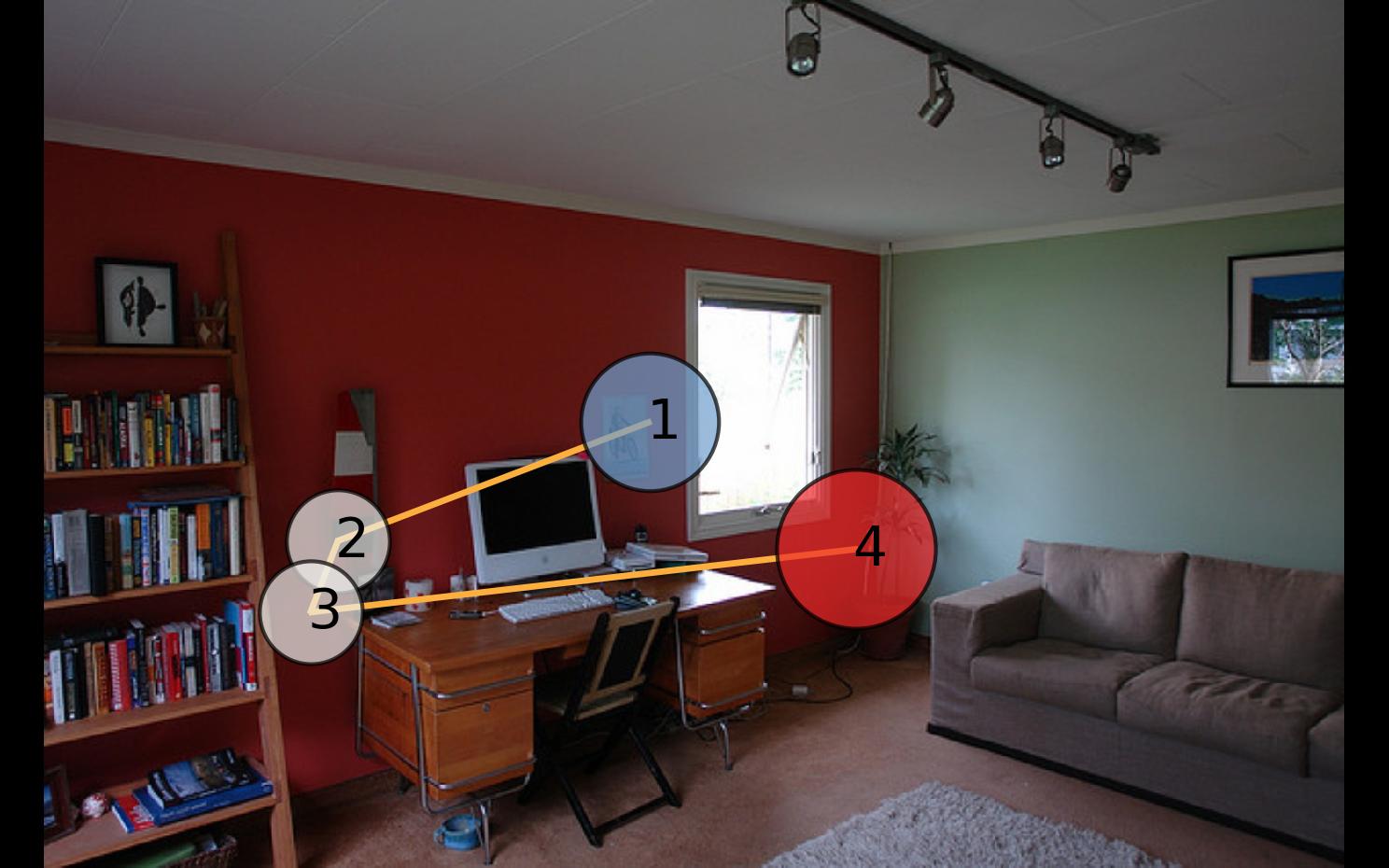} &
         \includegraphics[width=0.16\linewidth]{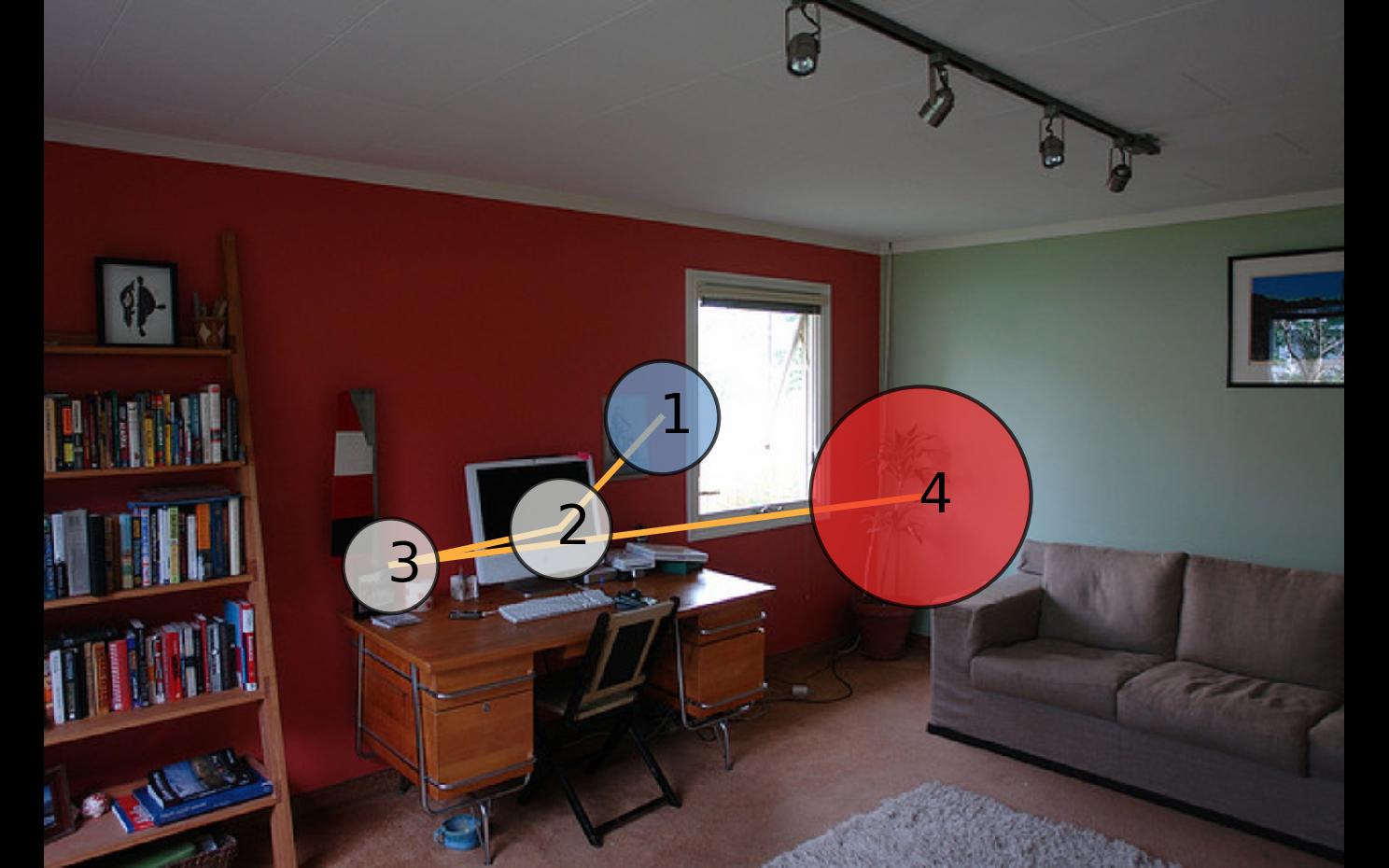} \\ 
         \raisebox{2.5\normalbaselineskip}[0pt][0pt]{\rotatebox[origin=c]{90}{\texttt{Fork}}} & \includegraphics[width=0.16\linewidth]{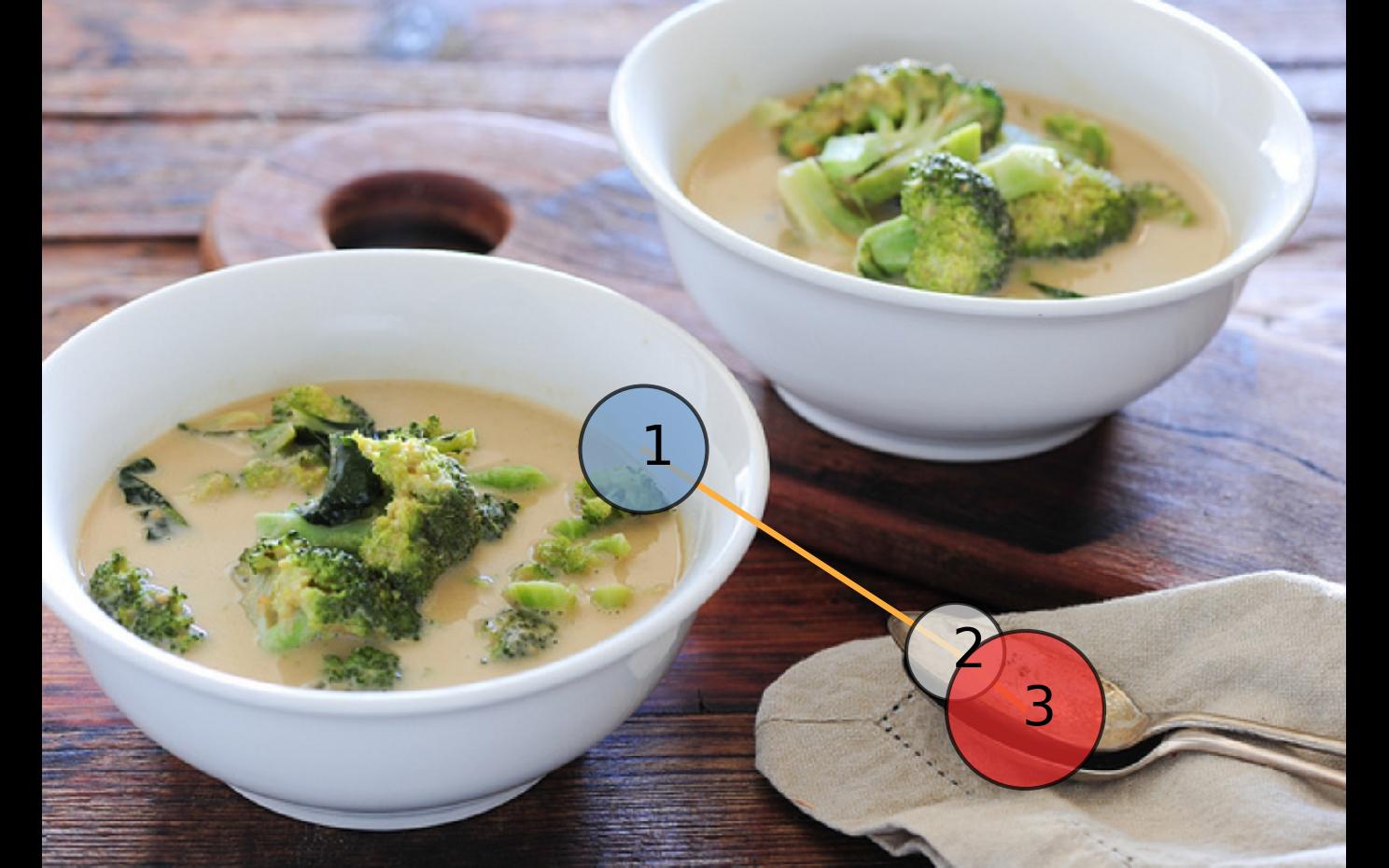} & 
         \includegraphics[width=0.16\linewidth]{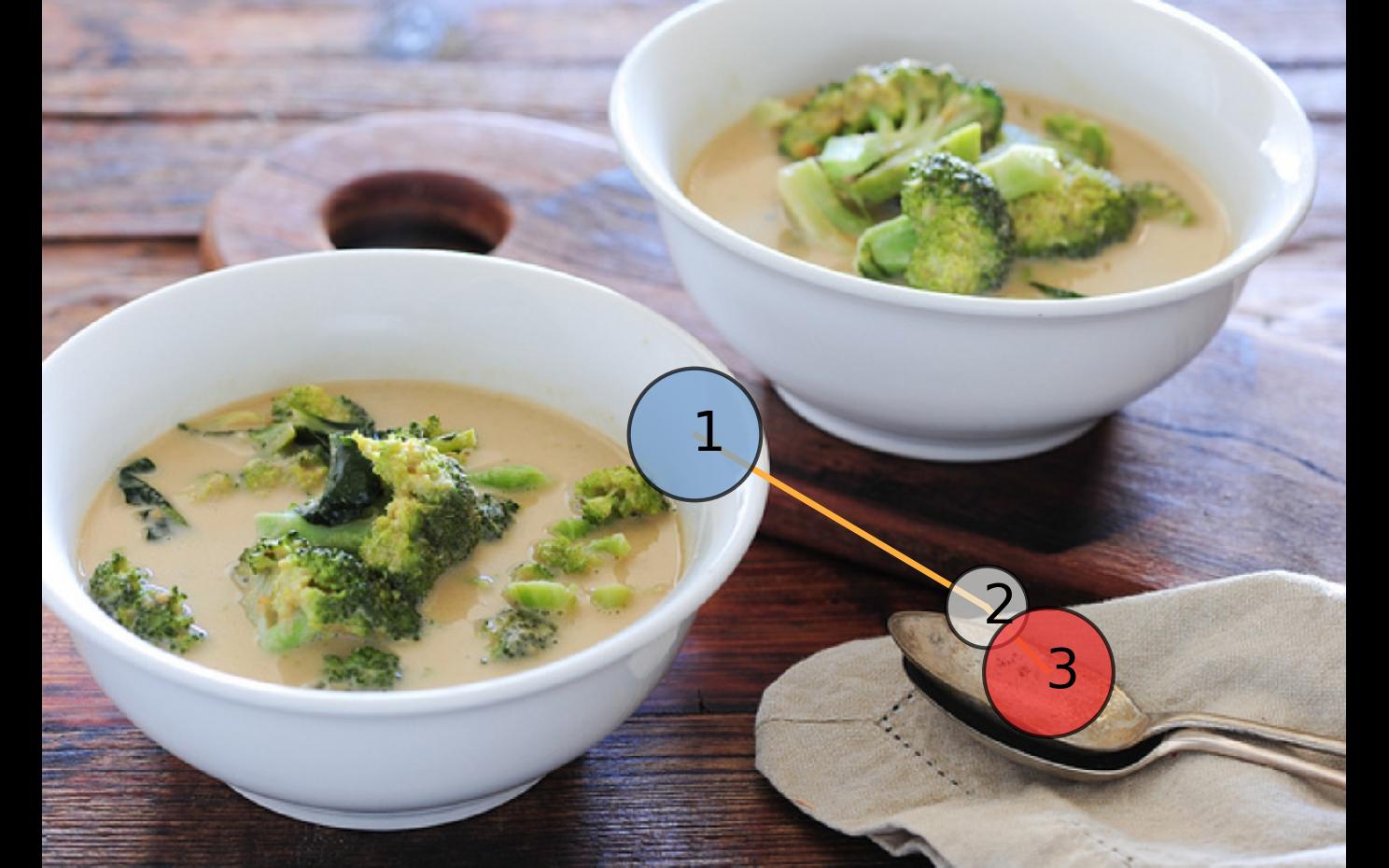} &
         \includegraphics[width=0.16\linewidth]{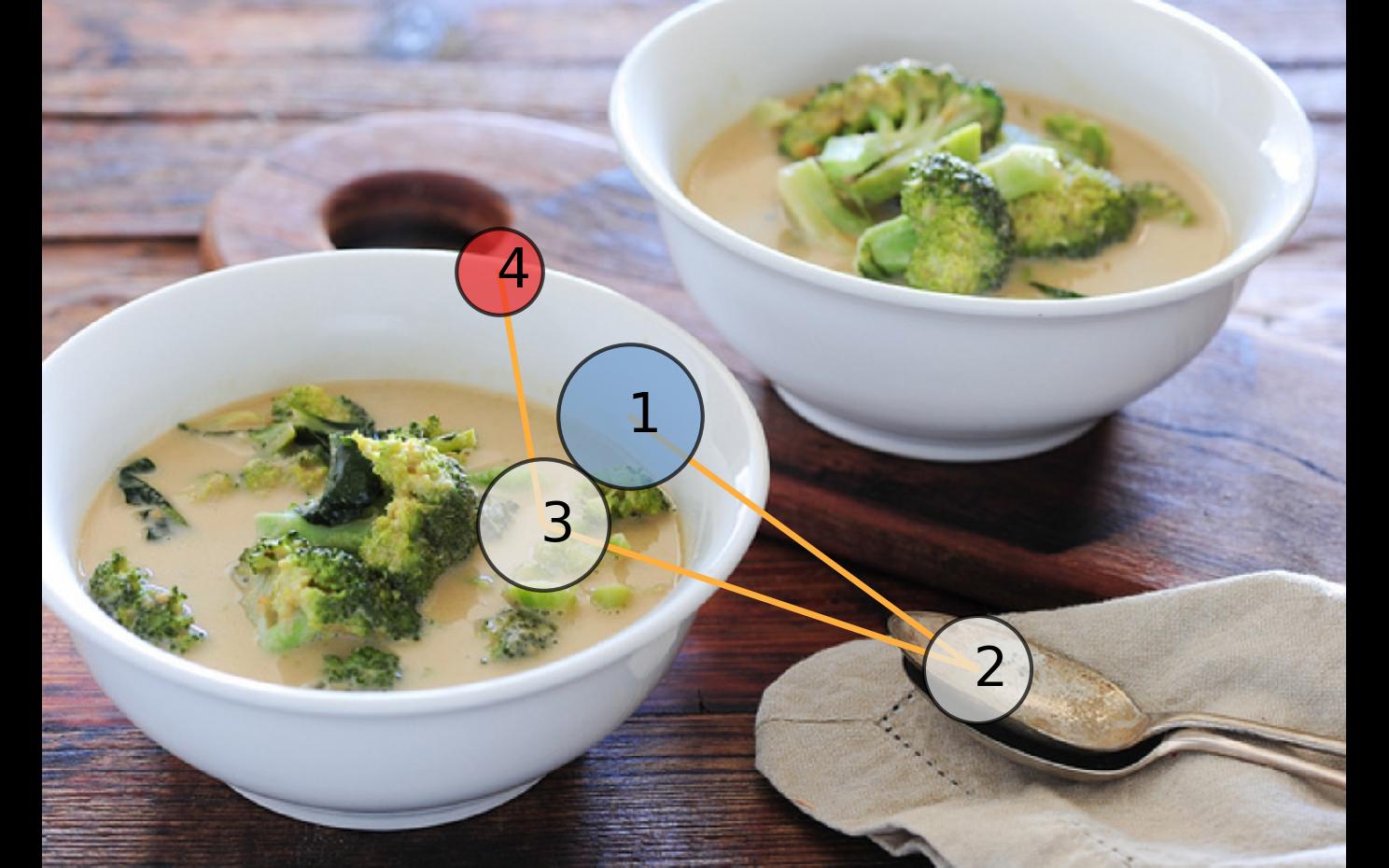} & 
         \includegraphics[width=0.16\linewidth]{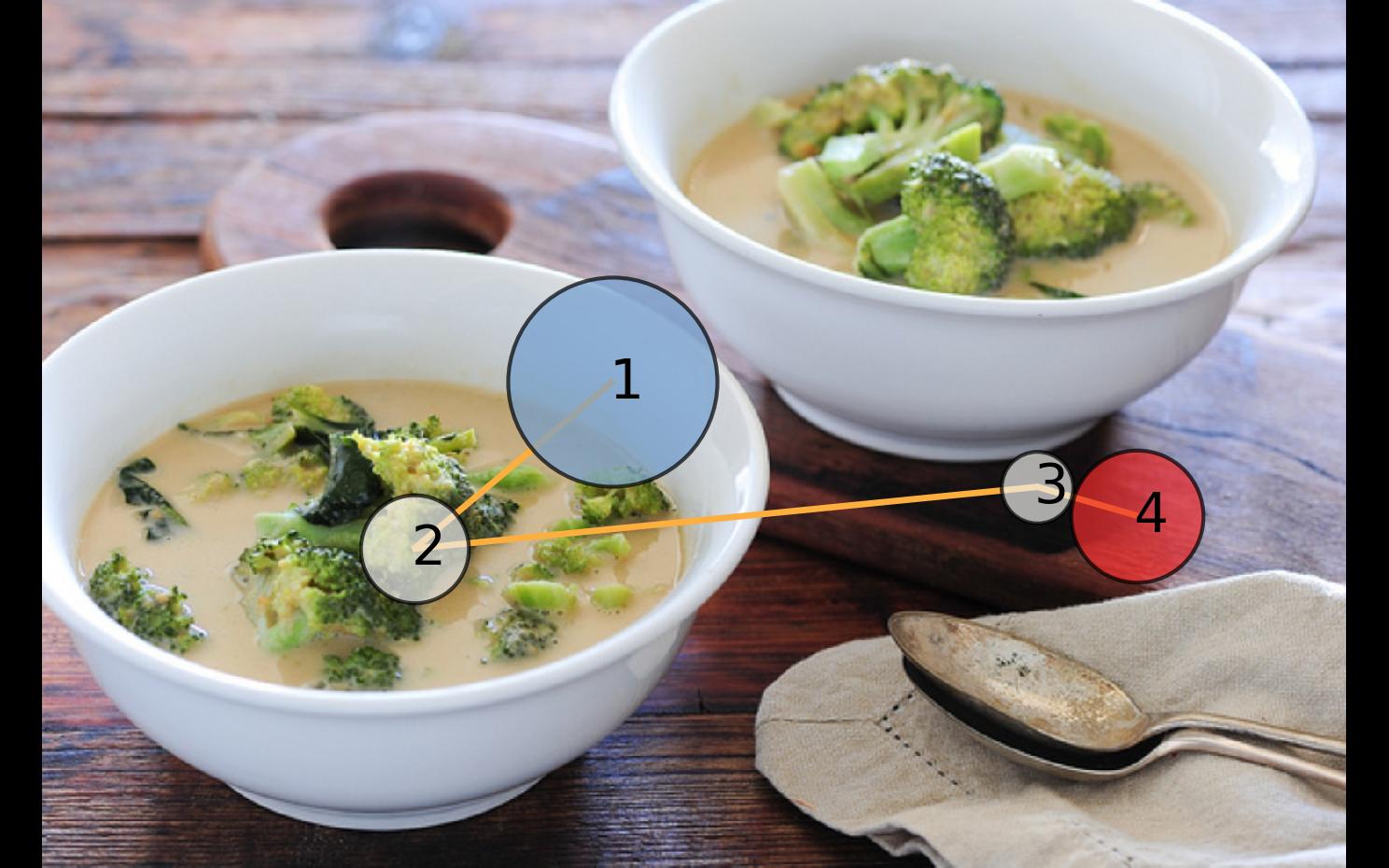} &
         \includegraphics[width=0.16\linewidth]{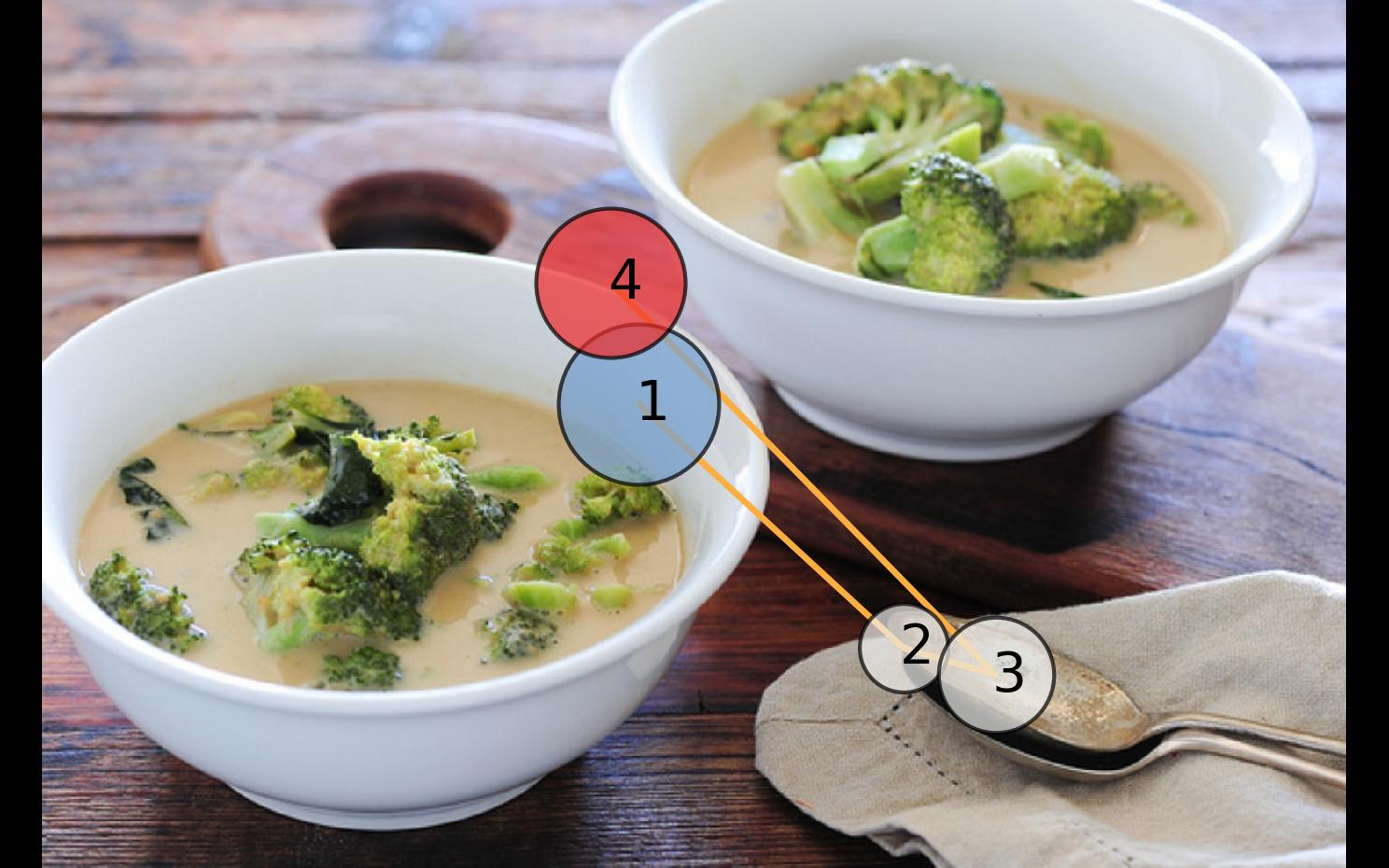} &
         \includegraphics[width=0.16\linewidth]{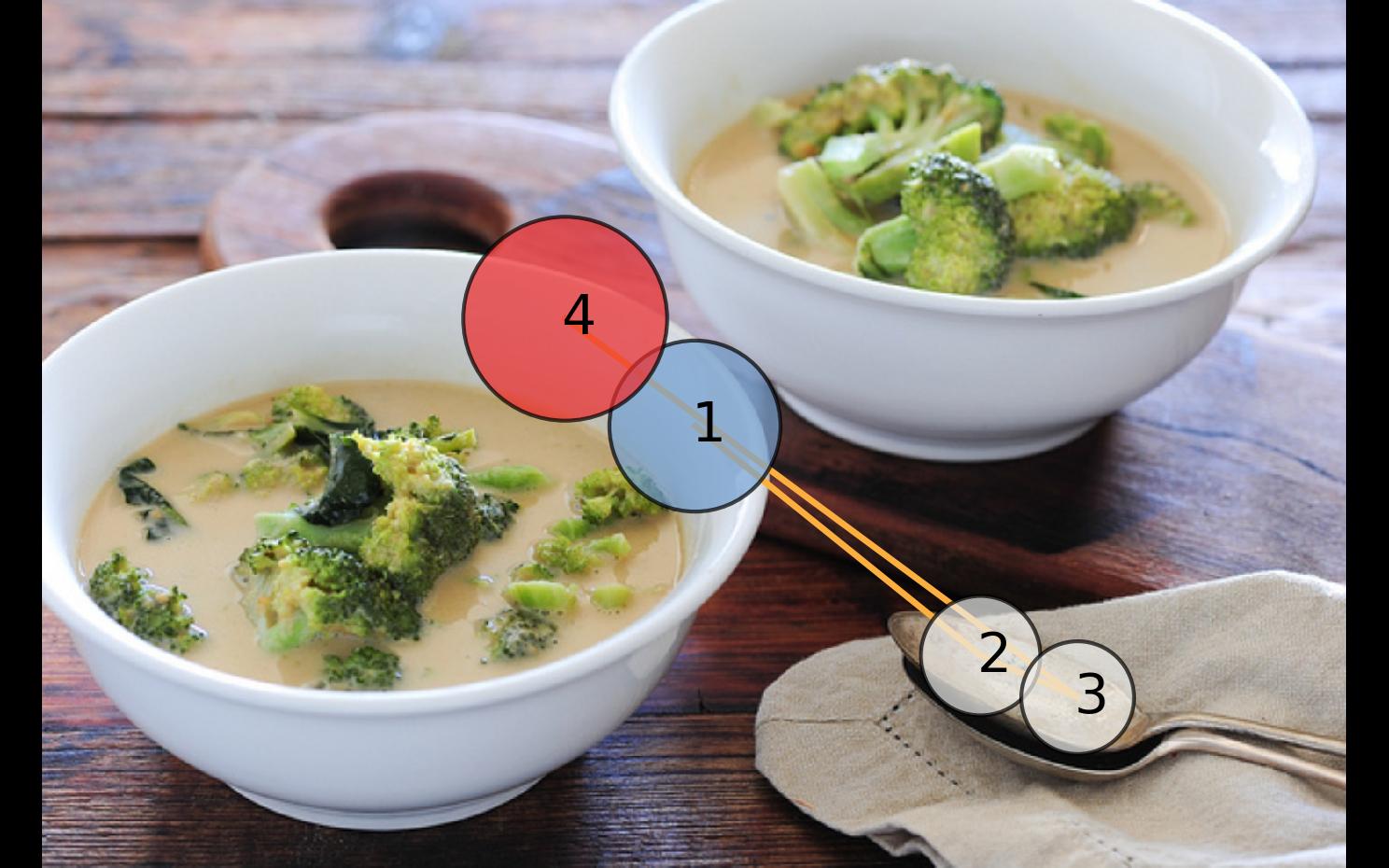} \\
    \end{tabular}
    }
    \vspace{-0.25cm}
    \caption{Comparison of simulated and human scanpaths across different datasets for both free-viewing and visual search tasks. From top to bottom: results on COCO-FreeView~\cite{yang2020predicting}, MIT1003~\cite{judd2009learning}, COCO-Search18 TP~\cite{chen2021coco}, and COCO-Search18 TA~\cite{yang2022target} datasets. }
    \label{fig:qualitatives}
    \vspace{-0.35cm}
\end{figure*}

\subsection{Scanpath Variability Analysis}
Human visual exploration is inherently variable. Individuals perceive the same stimulus in different manners depending on factors such as attention, context, and cognitive processes~\cite{henderson2003human,bisley2011neural}.
Capturing such variability is essential for developing models that accurately reflect the diverse range of human traits.
However, existing scanpath prediction models tend to align closely with the statistical mean of human gaze behavior. While this approach may improve performance on traditional evaluation metrics, it fails to reflect the natural variability in human visual attention. Commonly used metrics such as MM, SM, and SS tend to reward predictions that closely match an aggregated ground truth, thus favoring models that generate a single representative scanpath~\cite{kummerer2021state}. This is clear in several works~\cite{mondal2023gazeformer,chen2021predicting,chen2024gazexplain} where scanpath models surpass human consistency. Indeed, the average similarity between ground-truth scanpaths can be smaller than the average similarity between generated scanpaths if these well reflect an average behavior. 

\begin{table}[t]
    \centering
  \small
  \setlength{\tabcolsep}{.2em}
  \resizebox{\linewidth}{!}{
  \begin{tabular}{lc cc c cc c cc c cc}
    \toprule
    & & \multicolumn{2}{c}{\textbf{COCO-FV}} & & \multicolumn{2}{c}{\textbf{MIT1003}} & & \multicolumn{2}{c}{\textbf{TP}} & & \multicolumn{2}{c}{\textbf{TA}} \\
    \cmidrule{3-4} \cmidrule{6-7} \cmidrule{9-10} \cmidrule{12-13}
    & & \textbf{DSS} $\uparrow$ & \textbf{RSS} $\uparrow$ & & \textbf{DSS} $\uparrow$ & \textbf{RSS} $\uparrow$ & & \textbf{DSS} $\uparrow$ & \textbf{RSS} $\uparrow$ & & \textbf{DSS} $\uparrow$ & \textbf{RSS} $\uparrow$\\
    \midrule
    IOR-ROI-LSTM~\cite{chen2018scanpath} & & 0.185 & 0.428 & & 0.264 & 0.579 & & - & - & & - & -\\
    ChenLSTM~\cite{chen2021predicting} & & 0.174 & 0.420 & & 0.257 & 0.534 & & 0.386 & 0.635 & & 0.247 & 0.591\\
    Gazeformer~\cite{mondal2023gazeformer} & & - & - & & - & - & & 0.377 & 0.578 & & 0.206 & 0.417\\
    HAT~\cite{yang2024unifying} & & - & 0.645 & & - & 0.615 & & - & 0.861 & & - & 0.748\\
    ChenLSTM-ISP~\cite{chen2024beyond} & & 0.190 & 0.501 & & 0.264 & 0.619 & & 0.423 & 0.735 & & 0.268 & 0.670\\
    GazeXplain~\cite{chen2024gazexplain} & & 0.099 & 0.032 & & 0.302 & 0.674 & & 0.406 & 0.689 & & 0.283 & 0.716\\
    TPP-Gaze~\cite{damelio2025tpp} & & 0.271 & 0.732 & & 0.313 & 0.758 & & 0.284 & 0.516 & & 0.221 & 0.663\\    
     \rowcolor{OurColor} 
    \textbf{\ours (Ours)} & & \textbf{0.277} & \textbf{0.736} & & \textbf{0.354} & \textbf{0.815} & & \textbf{0.425} & \textbf{0.747} & & \textbf{0.312} & \textbf{0.800}\\
    \bottomrule
  \end{tabular}
  }
  \vspace{-0.2cm}
  \caption{Analysis of scanpath variability on free-viewing (COCO-FreeView~\cite{yang2020predicting} and MIT1003~\cite{judd2009learning}) and task oriented datasets (COCO-Search18 target-present~\cite{chen2021coco} and target-absent~\cite{yang2022target}).}
  \label{tab:diversity1}
  \vspace{-0.4cm}
\end{table}

Building upon these considerations, we present a first attempt to quantitatively assess the ability of a model to generate diverse, yet human-like, gaze trajectories. Specifically, for this study, we adopt the DSS and RSS metrics introduced in Sec.~\ref{subsec:experiments}.
Table~\ref{tab:diversity1} reports the results on both free-viewing and visual search. Notably, \scandiff achieves the best overall performance on all settings and datasets, highlighting its effectiveness in predicting accurate eye movement trajectories well aligned with the human scanpath variability. Interestingly, the performance gap between \scandiff and state-of-the-art methods becomes even more evident in the visual search task and further supported by qualitative results in the Supplementary Material. Goal-oriented scanpaths tend to be more deterministic~\cite{castelhano2009viewing,zelinsky2008theory}, particularly in the target-present setting, and are generally shorter than those in free-viewing scenarios. Nevertheless, our model effectively captures even the more subtle variability present in human gaze behavior.
\section{Conclusion}
\label{sec:conclusion}

In this paper, we introduced \scandiff, a novel diffusion-based architecture for scanpath prediction that significantly advances the state-of-the-art by modeling the inherent stochastic nature of human visual attention. Experimental results on multiple benchmark datasets demonstrate that \scandiff not only achieves state-of-the-art performance in traditional scanpath prediction metrics but also generates more diverse scanpaths that better capture the variability inherent in human visual exploration. This diversity is crucial for applications requiring realistic simulation of human visual behavior, such as human-computer interaction, autonomous systems, and cognitive robotics. 
These results highlight the importance of modeling stochasticity in visual attention deployment, suggesting that future research in gaze prediction should consider the probabilistic nature of human gaze beyond deterministic approaches.

\section*{Acknowledgments}
We acknowledge the CINECA award under the ISCRA initiative, for the availability of high-performance computing resources and support. This work was supported by the PNRR project ``Italian Strengthening of Esfri RI Resilience (ITSERR)'' funded by the European Union - NextGenerationEU (CUP B53C22001770006).

{
    \small
    \bibliographystyle{ieeenat_fullname}
    \bibliography{bibliography}
}

\clearpage
\maketitlesupplementary

\appendix

In the following, we provide additional results on the analysis of scanpath variability, demonstrating that \scandiff outperforms existing methods in capturing the diversity of human gazes, along with achieving state-of-the-art results in traditional scanpath prediction metrics.

\section{Additional Quantitative Results}
\tinytit{Additional Comparison Details}
ChenLSTM-ISP~\cite{chen2024beyond} is originally designed to generate user-specific scanpaths. In contrast, our model accounts for population diversity. To adapt ChenLSTM-ISP to our setting, we make it predict a single scanpath conditioned on an image, a task, and a user identifier. By simulating this process across different user identifiers, we obtain a population of scanpaths representing diverse subjects. For TPP-Gaze~\cite{damelio2025tpp}, we re-train it on our splits, as training data used in the original model may include samples from our test sets.

\tit{Zero-shot Evaluation}
As a complement of Table~\ref{tab:results_cocofreeview_mit} of the main paper, we report in Table~\ref{tab:results_osie} the results obtained on the OSIE dataset~\cite{xu2014predicting}, which is used exclusively for zero-shot evaluation and not included in the training process. This analysis allows us to assess the generalization capability of \scandiff when applied to unseen data. Our model achieves the best overall results when compared with models that does not use OSIE as training set. This highlights the ability of the model to generate plausible and diverse scanpaths without requiring dataset-specific fine-tuning, demonstrating its robustness also in out-of-distribution scenarios.

\tit{Additional Metrics on COCO-Search18}
Table~\ref{tab:results_search_all} presents a detailed breakdown of \scandiff's performance on the COCO-Search18 dataset across individual MM metrics for both target-present and target-absent scenarios. Extending the results reported in Table~\ref{tab:results_search} of the main paper, \scandiff consistently outperforms existing methods in modeling spatial characteristics of scanpaths with substantial margins. In the target-present condition, our approach achieves KL-divergence reductions of $68-78\%$ for shape, $79-81\%$ for length, $77-88\%$ for direction, and $50-99\%$ for position compared to models trained under identical settings (highlighted in gray). Similar improvements are observed in the target-absent condition. While TPP-Gaze shows slightly better performance in duration modeling for target-absent cases, \scandiff maintains competitive performance in duration metrics for target-present scenarios (\ie, $0.033$). These results further validate the effectiveness of our model in capturing the complex dynamics of task-driven visual behavior across different search conditions.

\tit{Additional Ablation Studies} 
Previous diffusion-based approaches~\cite{jiao2024diffgaze,wang2024scantd} perform the conditioning by directly concatenating the noisy gaze sequence with the image embedding. On a different line, we condition the denoising process through the cross-attention layer. 
Results in Table~\ref{tab:ablation_cross_attention_and_len} demonstrate that the proposed approach allows for a better semantic alignment between the scanpath and the multimodal features compared to the rigid concatenation of the input. Additionally, in Table~\ref{tab:ablation_cross_attention_and_len}, we also report an ablation study on the maximum scanpath length. We set this value to 16, matching the highest median across all datasets used in our experiments and the value used in \cite{chen2021predicting,chen2024gazexplain}. This hyperparameter serves as an upper bound, though the model can predict variable lengths.

\begin{table}[t]
  \centering
  \small
  \setlength{\tabcolsep}{.2em}
  \resizebox{\linewidth}{!}{
  \begin{tabular}{lc cccccc c cc c cc}
    \toprule
    & & \multicolumn{12}{c}{\textbf{OSIE}}\\
    \midrule    
    & & \multicolumn{6}{c}{\textbf{MM} $\downarrow$} & & \multicolumn{2}{c}{\textbf{SM} $\downarrow$} & & \multicolumn{2}{c}{\textbf{SS} $\downarrow$} \\
    \cmidrule{3-8} \cmidrule{10-11} \cmidrule{13-14} 
    & & Sh & Len & Dir & Pos & Dur & Avg & & w/ Dur & w/o Dur & & w/ Dur & w/o Dur \\
    \midrule
    Itti-Koch~\cite{IttiKoch98} && 1.886 & 1.293 & 0.482 & 2.910 & - & 1.643 & & - & 4.467 & & - & 4.086 \\
    CLE (Itti)~\cite{bfpha04,IttiKoch98} && 0.070 & 0.049 & 0.294 & 1.007 & - & 0.355 & & - & 2.560 & & - & 3.185 \\
    CLE (DG)~\cite{bfpha04,kummerer2014deep} && 0.105 & 0.025 & 0.215 & 0.906 & - & 0.313 & & - & 2.021 & & - & 3.766 \\
    PathGAN~\cite{assens2018pathgan} & & 0.070 & 0.108 & 0.575 & 2.148 & 0.199 & 0.620 & & 3.504 & 3.155 & & 2.061 & 1.960 \\
    G-Eymol~\cite{zanca2019gravitational} & & 1.531 & 0.782 & 0.238 & 2.159 & 0.324 & 1.007 & & 14.068 & 7.125 & & 9.468 & 3.341 \\
    DeepGazeIII~\cite{kummerer2022deepgaze} & & 0.058 & 0.025 & 0.143 & 0.200 & - & 0.107 & & - & 0.333 & & - & 2.465 \\
    ChenLSTM~\cite{chen2021predicting} & & 0.723 & 0.477 & 0.122 & 0.420 & \textbf{0.026} & 0.353 & & 0.781 & 0.638 & & 0.402 & 0.350 \\
    HAT~\cite{yang2024unifying} & & 2.793 & 1.248 & 0.207 & 2.236 & - & 1.621 & & - & 3.371 & & - & 1.548 \\
    TPP-Gaze~\cite{damelio2025tpp} & & 0.070 & 0.070 & 0.085 & 0.288 & 0.067 & 0.116 & & 1.058 & 0.648 & & 0.779 & 0.365 \\
    \rowcolor{OurColor} 
    \textbf{\ours (Ours)} & & \textbf{0.036} & \textbf{0.024} & \textbf{0.040} & \textbf{0.150} & 0.035 & \textbf{0.057} & & \textbf{0.219} &\textbf{0.305} & & \textbf{0.253} & \textbf{0.226} \\
    \bottomrule
  \end{tabular}
  }
  \vspace{-0.2cm}
  \caption{Performance comparison of different models on the OSIE~\cite{xu2014predicting} dataset for zero-shot prediction. Models with the highest performance for each metric is marked in \textbf{bold}.}
  \label{tab:results_osie}
  \vspace{-0.2cm}
\end{table}

\begin{table}[t]
  \centering
  \small
  \setlength{\tabcolsep}{.2em}
  \resizebox{\linewidth}{!}{
  \begin{tabular}{lc ccccc l ccccc}
    \toprule
    & & \multicolumn{5}{c}{\textbf{Target-Present}} & & \multicolumn{5}{c}{\textbf{Target-Absent}} \\
    \midrule
    & & \multicolumn{5}{c}{\textbf{MM} $\downarrow$} & & \multicolumn{5}{c}{\textbf{MM} $\downarrow$} \\
    \cmidrule{3-7} \cmidrule{9-13}
    & & Sh & Len & Dir & Pos & Dur & & Sh & Len & Dir & Pos & Dur \\
    \midrule
    PathGAN~\cite{assens2018pathgan} && 0.594 & 0.365 & 0.937 & 0.333 & 0.336 & &  0.030 & 0.103 & 0.167 & 0.153 & 0.172\\
    ChenLSTM~\cite{chen2021predicting} && 0.253 & 0.276 & 0.337 & 0.054 & 0.066 & & 0.075 & 0.044 & 0.111 & 0.052 & 0.092 \\
    Gazeformer~\cite{mondal2023gazeformer} && 0.581 & 0.301 & 0.316 & 0.056 & 0.150 & & 0.067 & 0.047 & 0.054 & 0.044 & 0.233 \\
    HAT~\cite{yang2024unifying} && 0.161 & 0.089 & 0.115 & 0.108 & - & & 0.108 & 0.040 & 0.024 & 0.034 & - \\
    ChenLSTM-ISP~\cite{chen2024beyond} && 0.272 & 0.232 & 0.302 & 0.019 & 0.044 & & 0.114 & 0.087 & 0.132 & 0.016 & 0.060\\
    GazeXplain~\cite{chen2024gazexplain} && 0.238 & 0.255 & 0.245 & 0.038 & 0.052 & & 0.019 & 0.016 & 0.015 & 0.051 & 0.128 \\
    TPP-Gaze~\cite{damelio2025tpp} && 0.676 & 0.250 & 0.845 & 0.825 & \underline{0.025} & & 0.051 & 0.017 & 0.117 & 0.287 & \underline{0.018} \\
    \midrule
    \rowcolor{Gray}
    Gazeformer~\cite{mondal2023gazeformer} && 0.436 & 0.326 & 0.313 & 0.031 & 0.147 & & 0.081 & 0.418 & 0.269 & 1.613 & 0.250 \\
    \rowcolor{Gray}
    GazeXplain~\cite{chen2024gazexplain} && 0.244 & 0.224 & 0.295 & 0.030 & 0.045 & & 0.009 & 0.023 & 0.024 & 0.035 & 0.095 \\
    \rowcolor{Gray}
    TPP-Gaze~\cite{damelio2025tpp} && 0.445 & 0.254 & 0.581 & 1.216 & 0.037 & & 0.023 & 0.032 & 0.052 & 0.547 & \textbf{0.023} \\
    \rowcolor{OurColor} 
    \textbf{\ours (Ours)} && \underline{\textbf{0.077}} & \underline{\textbf{0.048}} & \underline{\textbf{0.067}} & \underline{\textbf{0.015}} & \textbf{0.033} & & \underline{\textbf{0.008}} & \underline{\textbf{0.010}} & \underline{\textbf{0.007}} & \underline{\textbf{0.008}} & 0.067 \\
    \bottomrule
  \end{tabular}
  }
  \vspace{-0.2cm}
  \caption{Additional results on COCO-Search18 dataset~\cite{chen2021coco} for both target-present and target-absent settings. Models trained using identical settings and training splits to \scandiff are highlighted in \colorbox{Gray}{\textbf{gray}}. Among these, the highest performance for each metric is marked in \textbf{bold}. \underline{Underlined} values denote the top overall performance across all models and metrics.}
  \label{tab:results_search_all}
  \vspace{-0.3cm}
\end{table}

\begin{table}[t]
    \centering
  \small
  \setlength{\tabcolsep}{.25em}
  \resizebox{\linewidth}{!}{
  \begin{tabular}{lc cc ccc c cccc}
    \toprule
     & & & & \multicolumn{3}{c}{\textbf{COCO-FreeView}} & & \multicolumn{4}{c}{\textbf{COCO-Search18 (TP)}} \\
     \cmidrule{5-7} \cmidrule{9-12}
    & & Len & & \textbf{MM} $\downarrow$ & \textbf{SM} $\downarrow$ & \textbf{SS} $\downarrow$ & & \textbf{MM} $\downarrow$ & \textbf{SM} $\downarrow$ & \textbf{SS} $\downarrow$ & \textbf{SemSS} $\downarrow$ \\
    \midrule
    \textbf{w/o} cross-attention & & 16 & & 0.108 & 0.173 & 0.111 & & 0.439 & 1.839 & 0.862 & 0.962 \\
    \midrule
    & & 32 & & \textbf{0.052} & 0.204 & 0.113 & & 0.084 & 0.069 & 0.030 & \textbf{0.056} \\
    & & 24 & & 0.058 & 0.046 & 0.029 & & 0.075 & 0.084 & 0.044 & 0.092 \\
    \rowcolor{OurColor}
    \textbf{\ours} & & 16 & & 0.078 & \textbf{0.015} & \textbf{0.013} & & \textbf{0.048} & \textbf{0.037} & \textbf{0.019} & 0.072\\
    \bottomrule
  \end{tabular}
  }
  \vspace{-0.2cm}
  \caption{Ablation study on the effect of cross-attention compared to input concatenation and on the maximum scanpath length.}
  \label{tab:ablation_cross_attention_and_len}
   \vspace{-0.3cm}
\end{table}

\section{Additional Qualitative Results}
\begin{figure*}[t]
    \footnotesize
    \setlength{\tabcolsep}{.1em}
    \resizebox{\linewidth}{!}{
    \begin{tabular}{ccc}
         \tiny IOR-ROI-LSTM~\cite{chen2018scanpath} & \tiny ChenLSTM~\cite{chen2021predicting} & \tiny GazeXplain~\cite{chen2024gazexplain}\\
         \addlinespace[0.08cm]
         \includegraphics[width=0.16\linewidth]{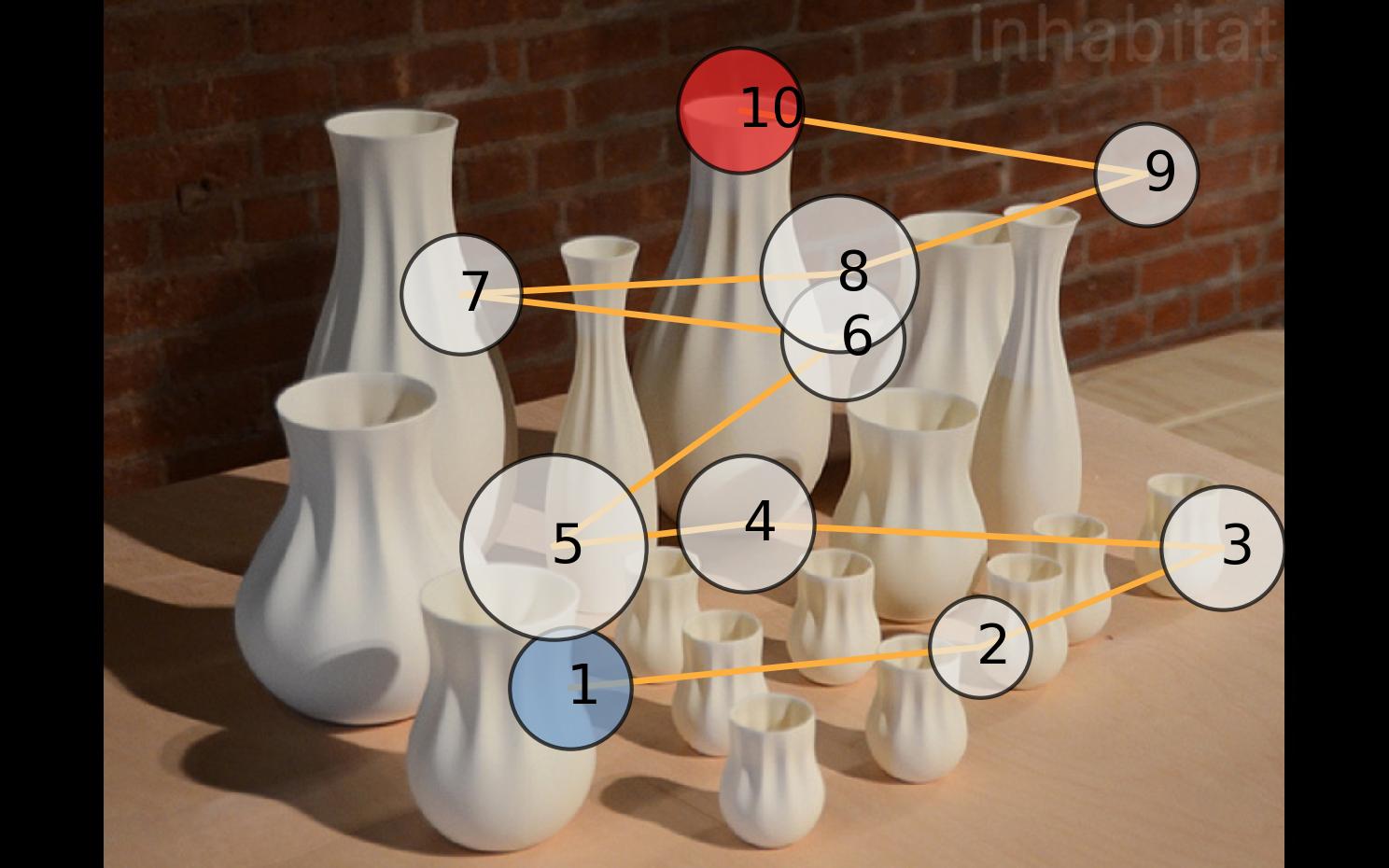} &
         \includegraphics[width=0.16\linewidth]{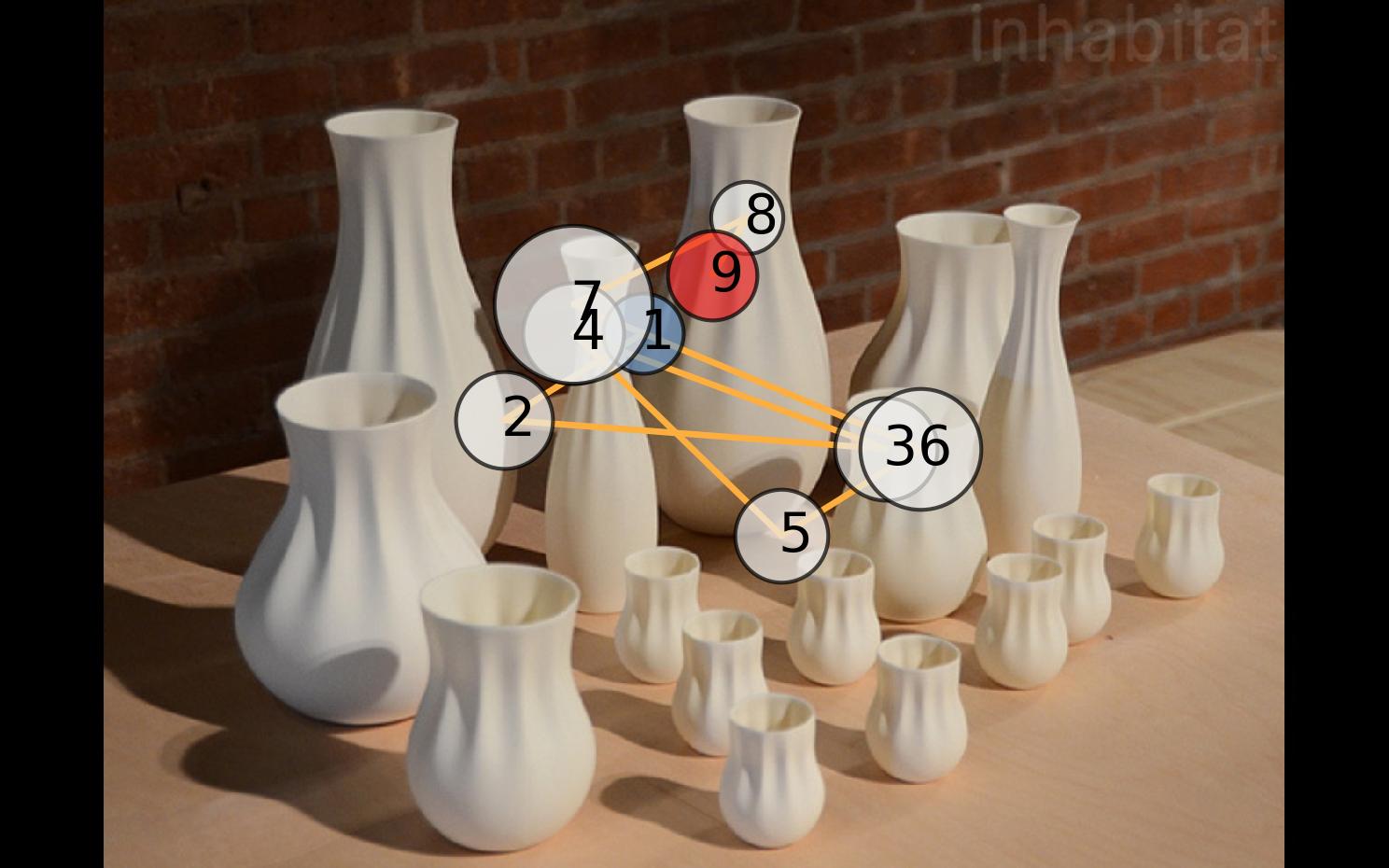} &
         \includegraphics[width=0.16\linewidth]{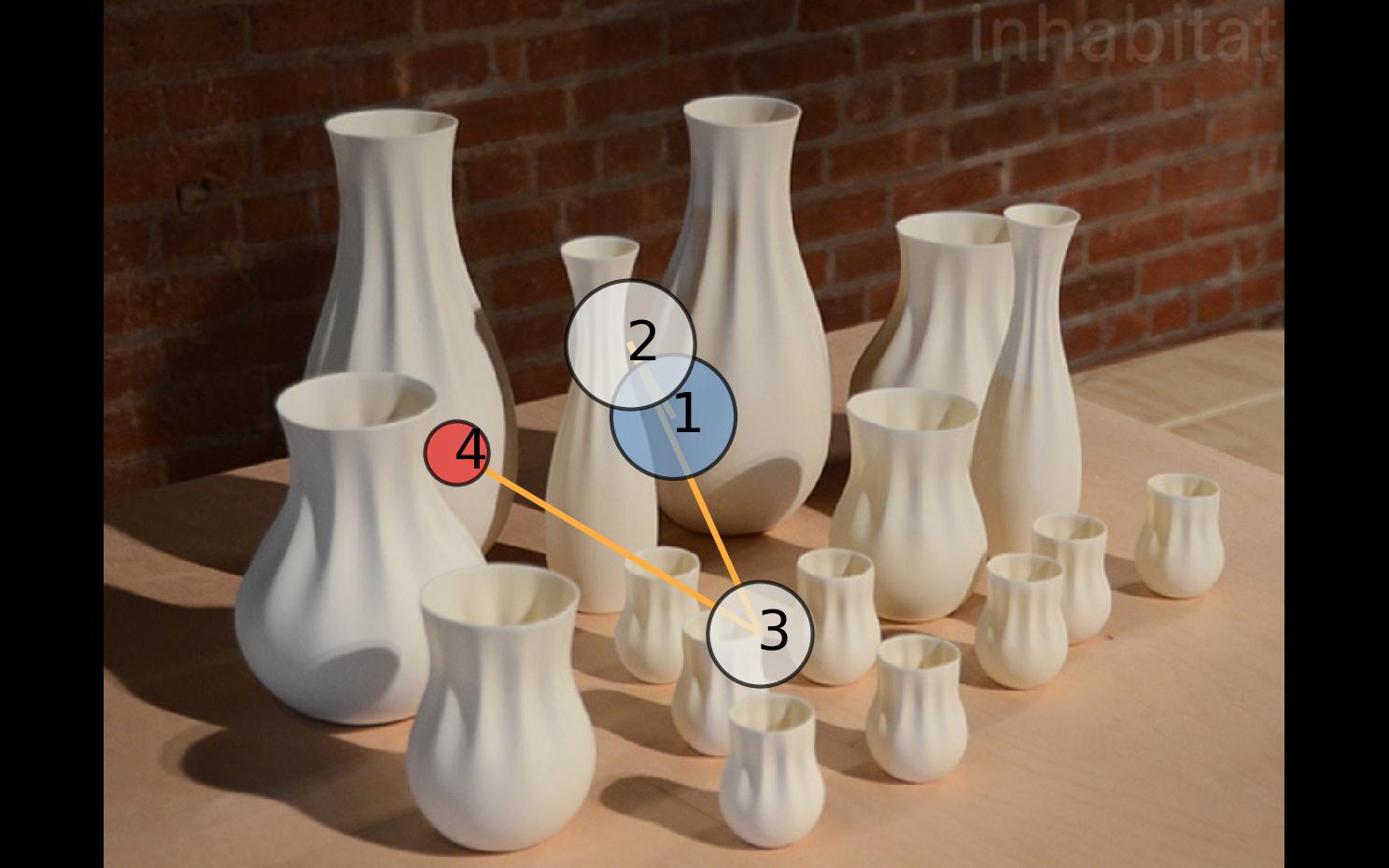} \\ 
         
         \tiny TPP-Gaze~\cite{damelio2025tpp}  & \tiny \textbf{\ours (Ours)} & \tiny Humans \\
         \includegraphics[width=0.16\linewidth]{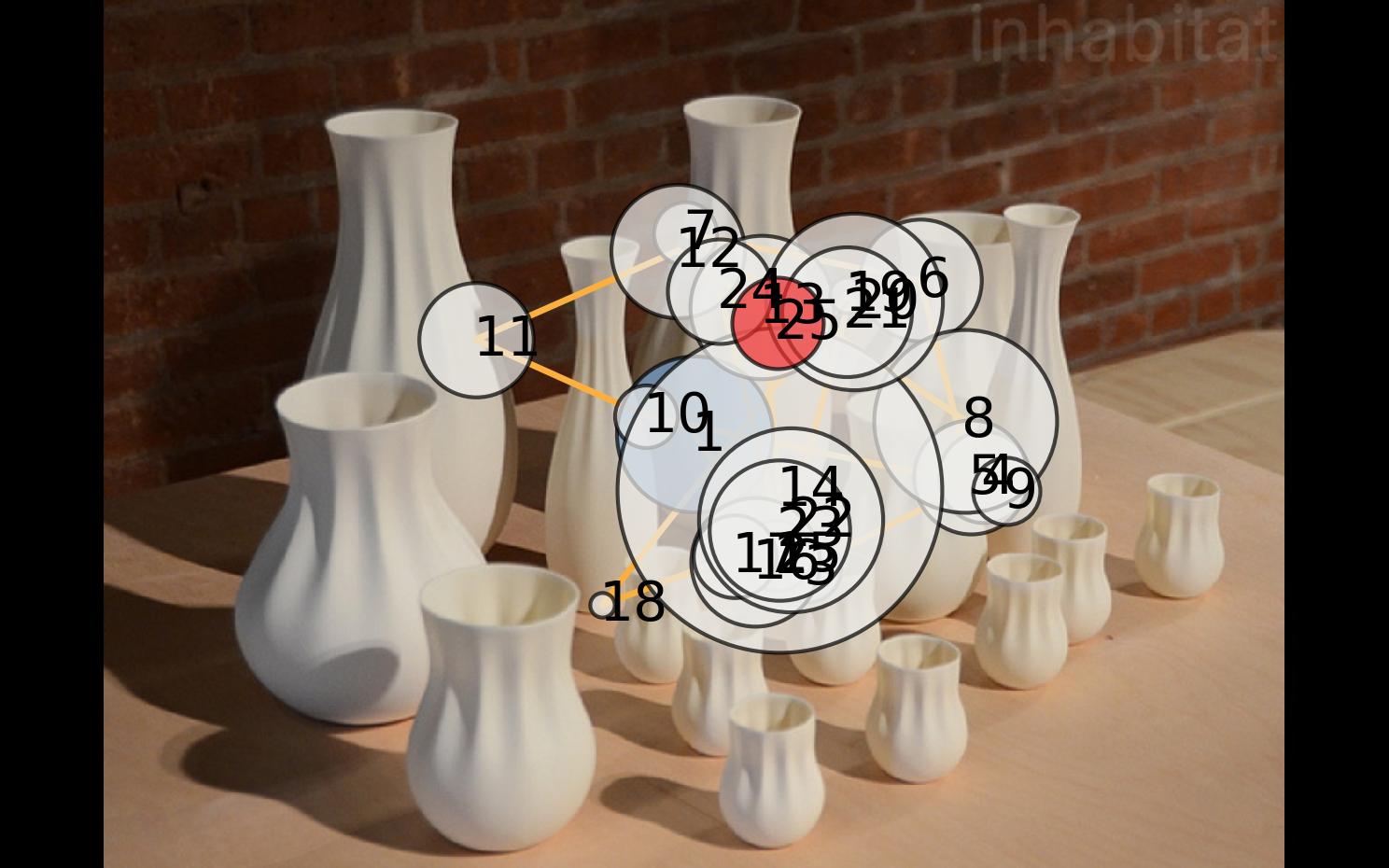} & 
         \includegraphics[width=0.16\linewidth]{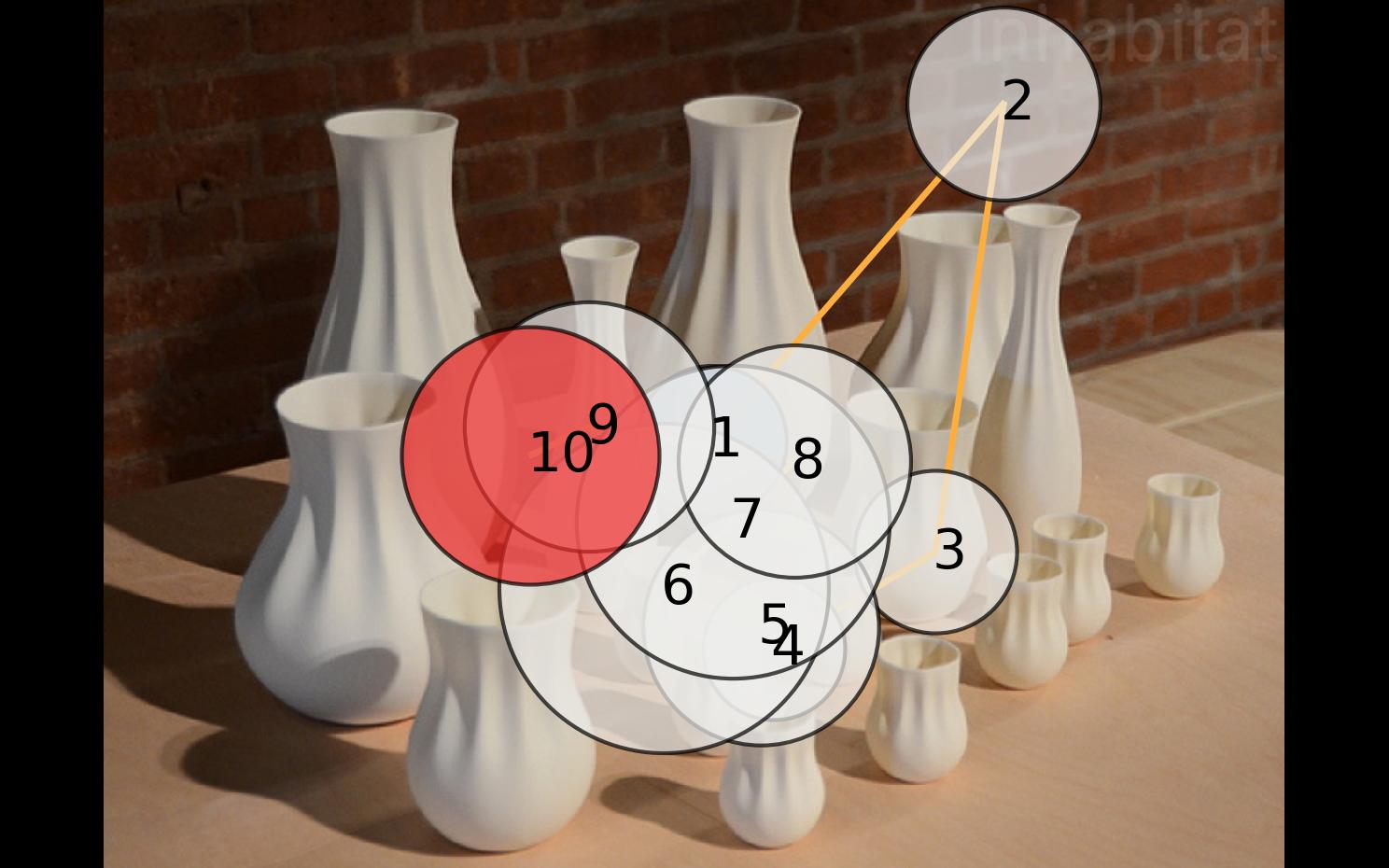} & 
         \includegraphics[width=0.16\linewidth]{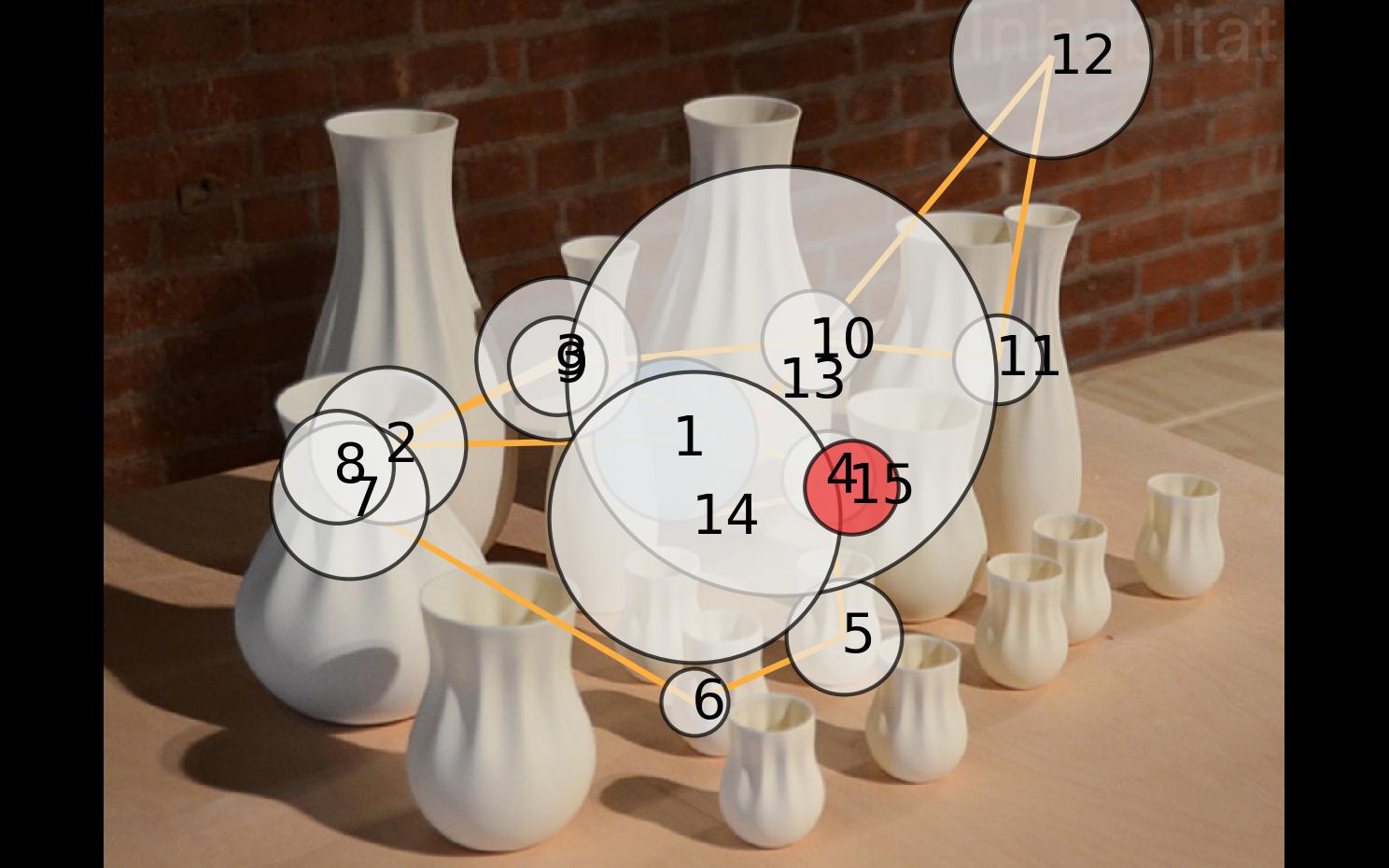} \\

         \addlinespace[0.6cm]
         
         \tiny IOR-ROI-LSTM~\cite{chen2018scanpath}~\cite{kummerer2022deepgaze} & \tiny ChenLSTM~\cite{chen2021predicting} & \tiny GazeXplain~\cite{chen2024gazexplain}\\
         \addlinespace[0.08cm]
         \includegraphics[width=0.16\linewidth]{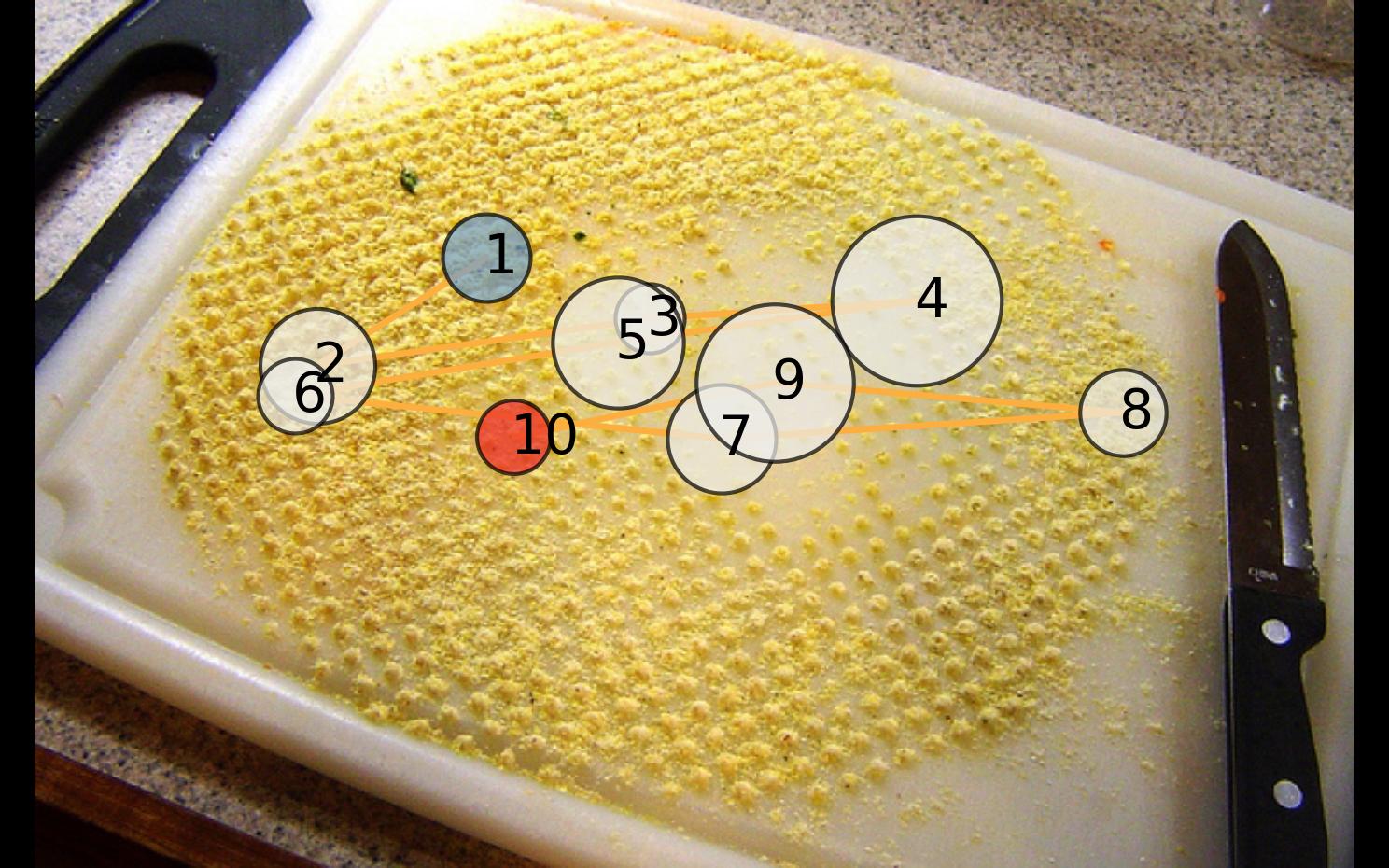} &
         \includegraphics[width=0.16\linewidth]{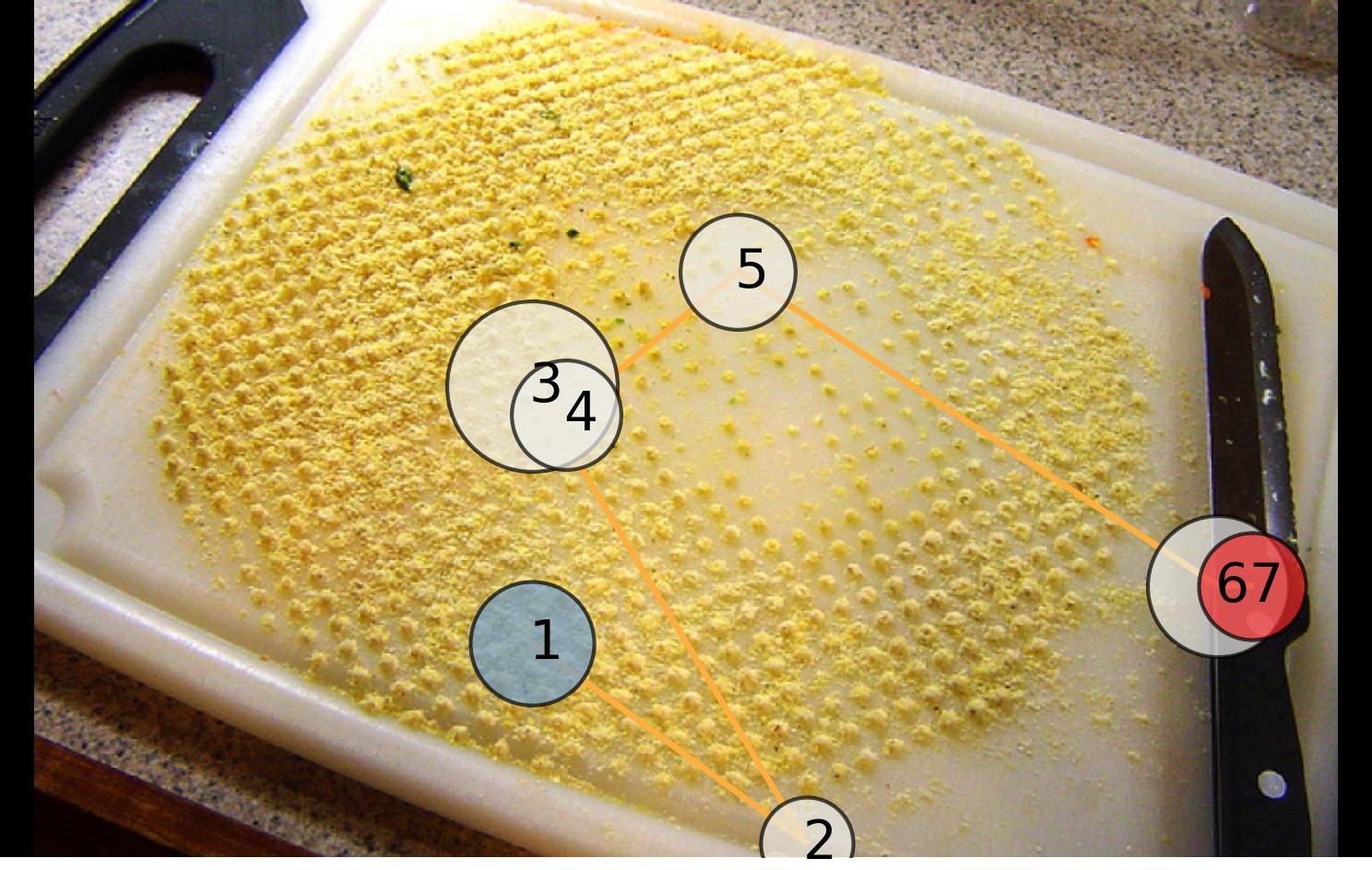} &
         \includegraphics[width=0.16\linewidth]{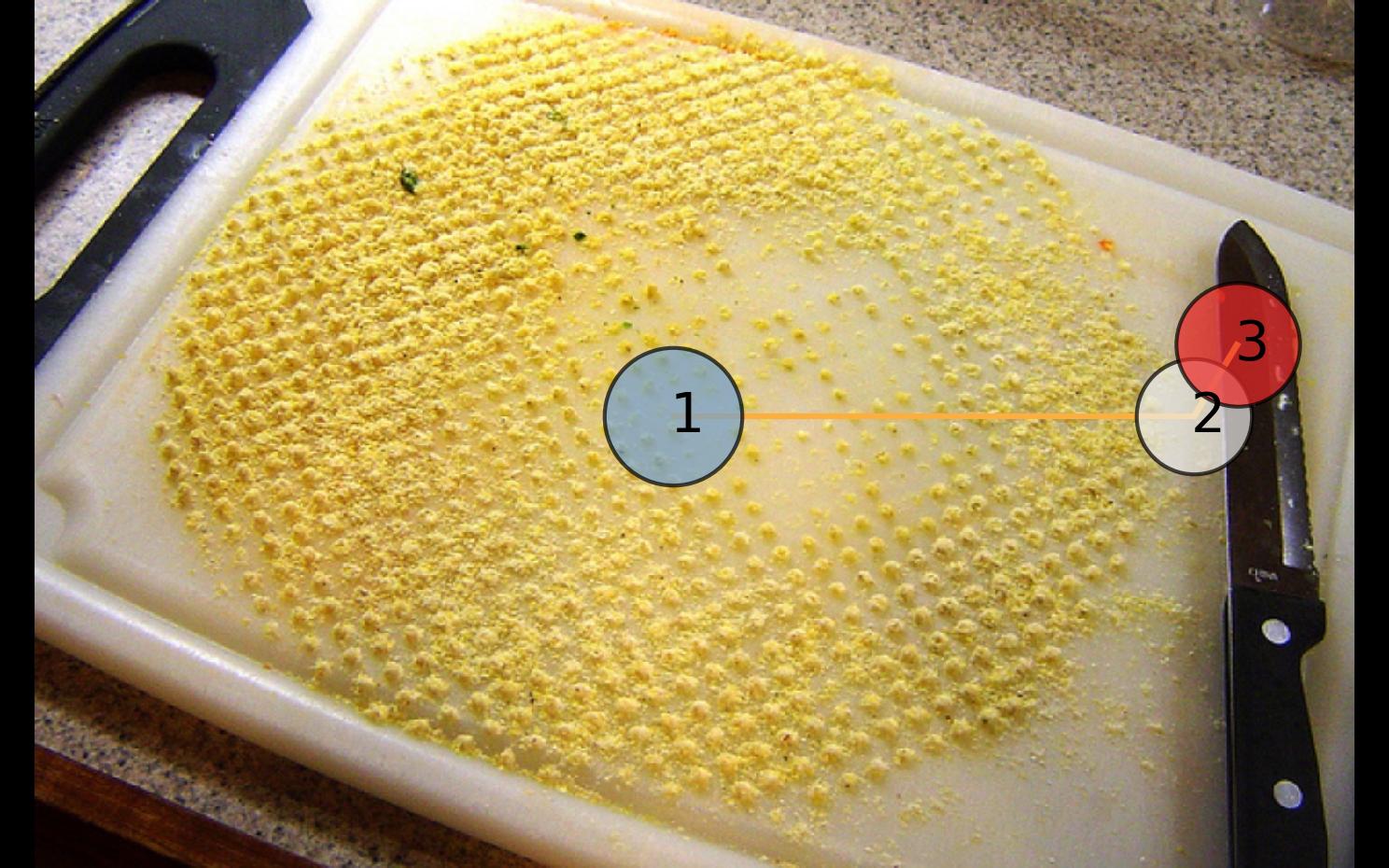} \\ 
         \tiny TPP-Gaze~\cite{damelio2025tpp} & \tiny \textbf{\ours (Ours)} & \tiny Humans \\
         \includegraphics[width=0.16\linewidth]{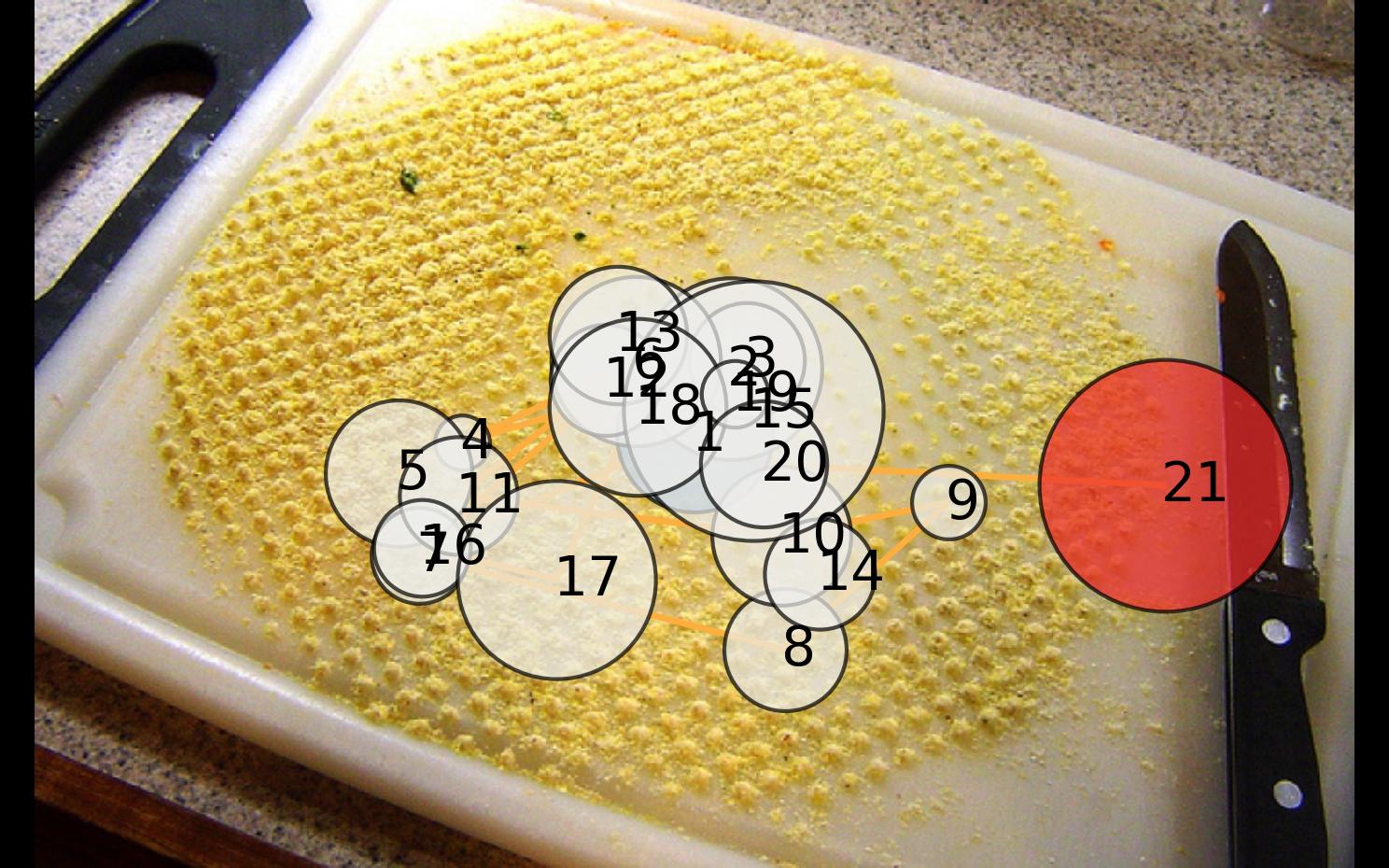} & 
         \includegraphics[width=0.16\linewidth]{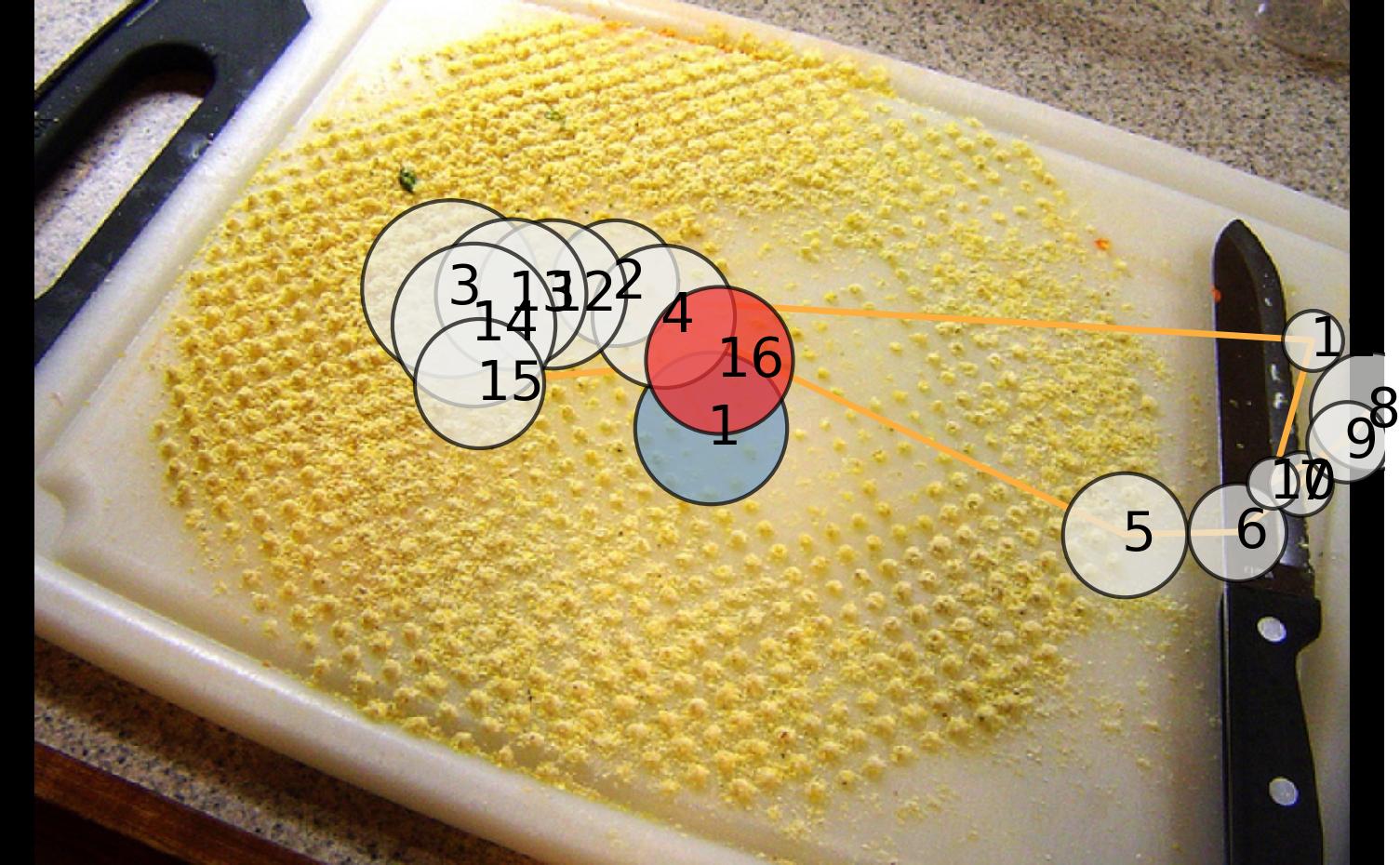} & 
         \includegraphics[width=0.16\linewidth]{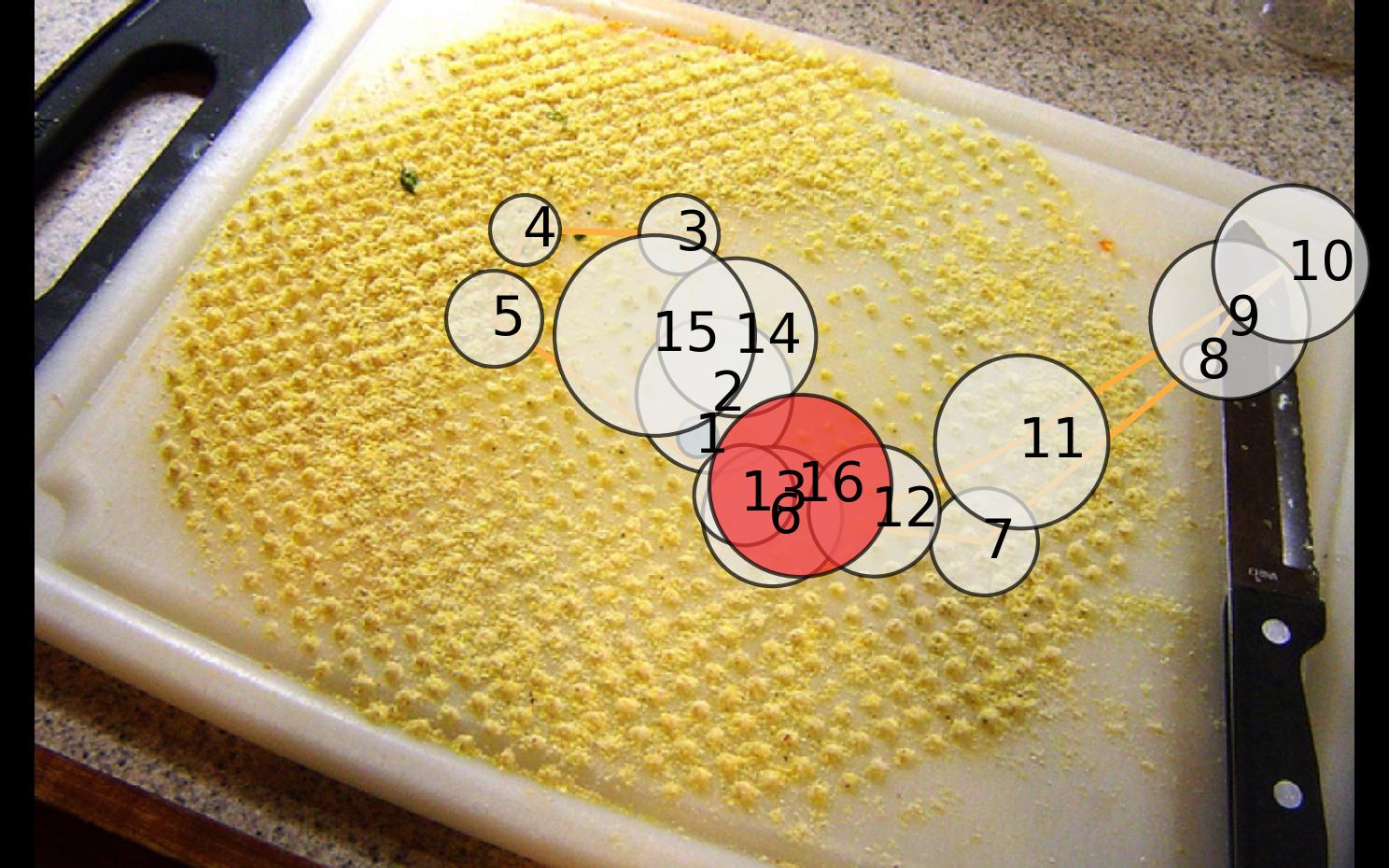}
    \end{tabular}
    }
    \vspace{-0.15cm}
    \caption{Qualitative comparison of simulated and human scanpaths on the COCO-FreeView dataset.}
    \label{fig:qualitatives_COCOFV}
    \vspace{-0.4cm}
\end{figure*}

\begin{figure*}[t]
    \footnotesize
    \setlength{\tabcolsep}{.1em}
    \resizebox{\linewidth}{!}{
    \begin{tabular}{ccc}
         \tiny IOR-ROI-LSTM~\cite{chen2018scanpath} & \tiny ChenLSTM~\cite{chen2021predicting} & \tiny GazeXplain~\cite{chen2024gazexplain}\\
         \addlinespace[0.08cm]
         \includegraphics[width=0.16\linewidth]{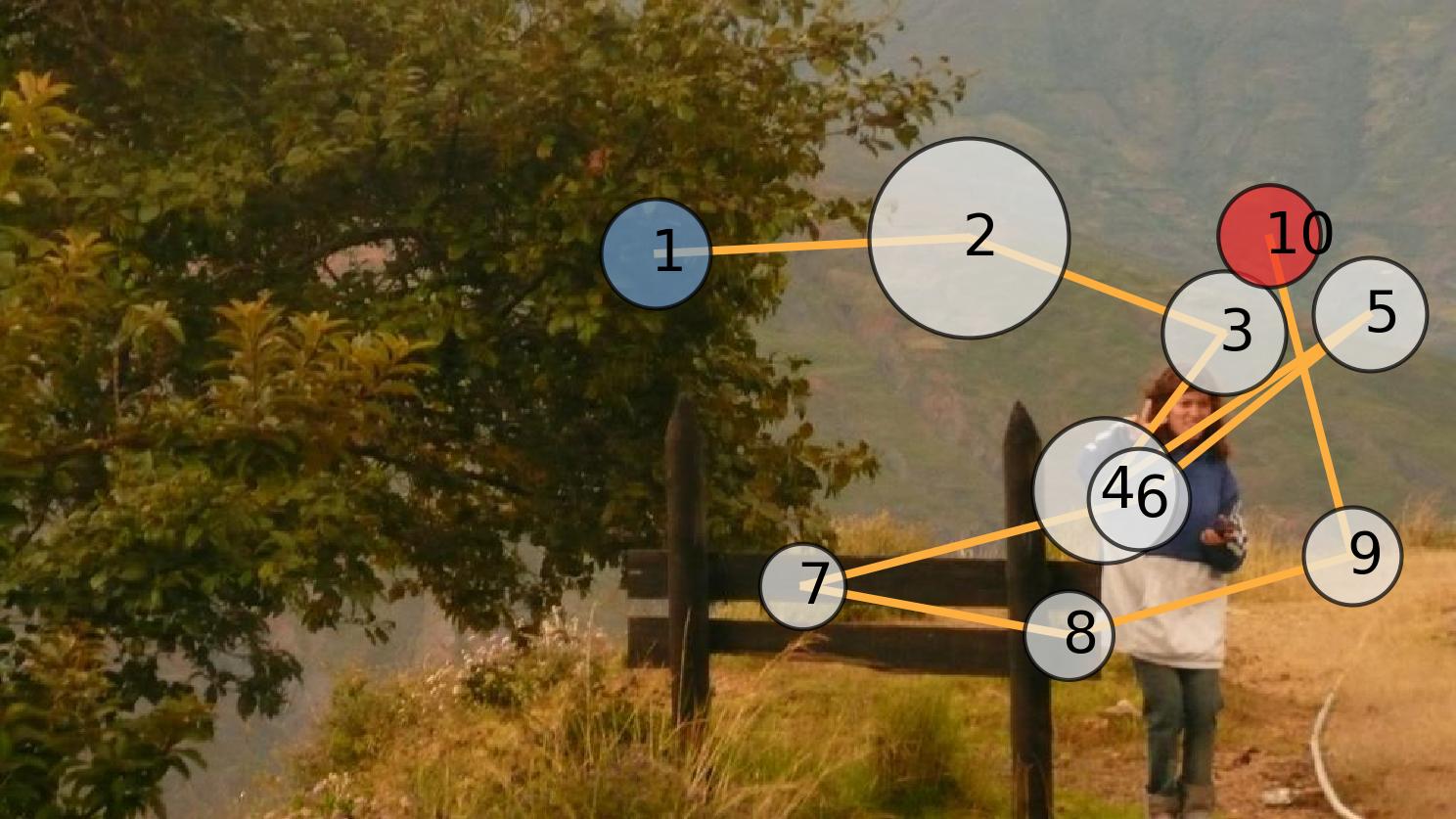} &
         \includegraphics[width=0.16\linewidth]{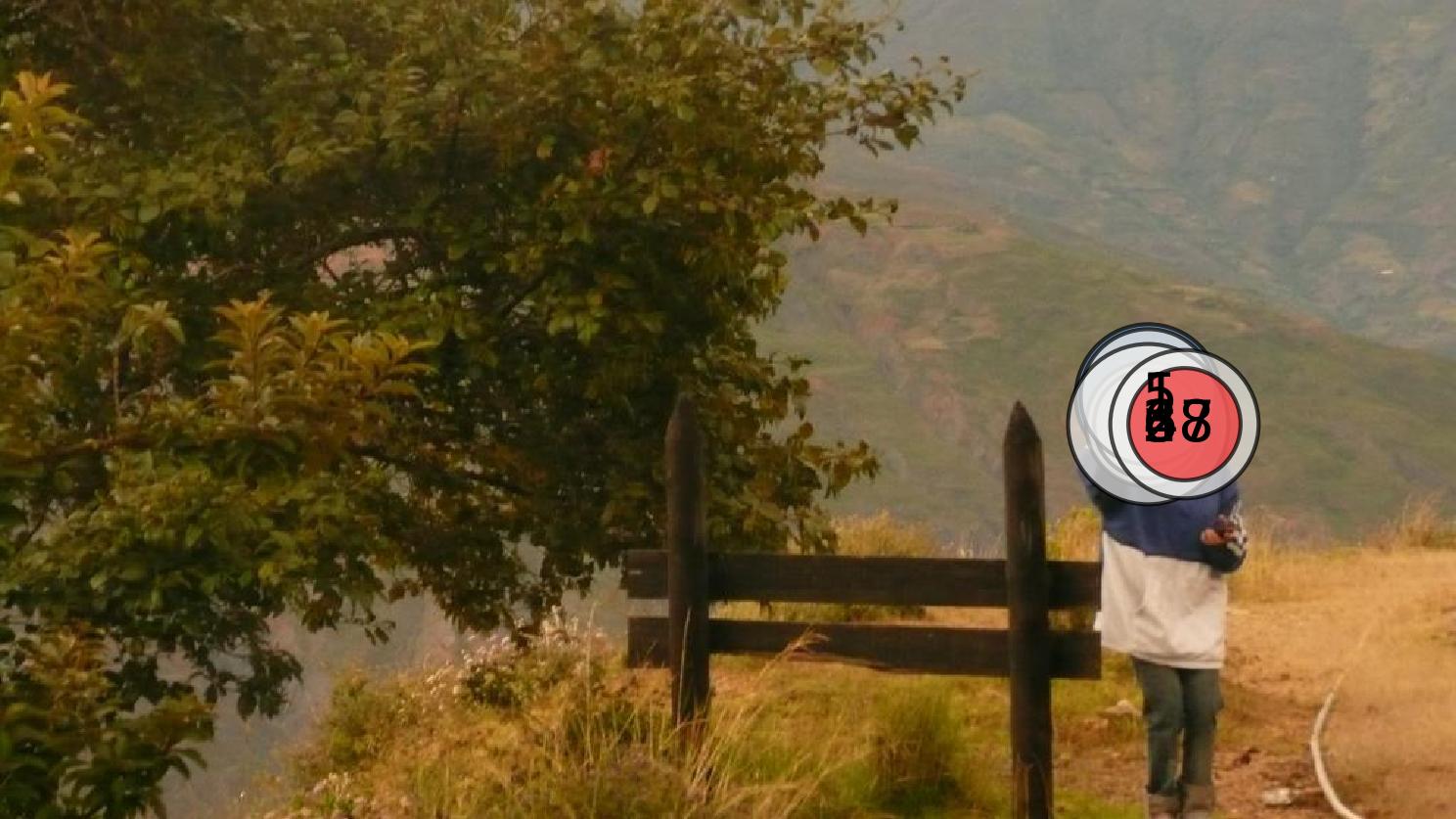} &
         \includegraphics[width=0.16\linewidth]{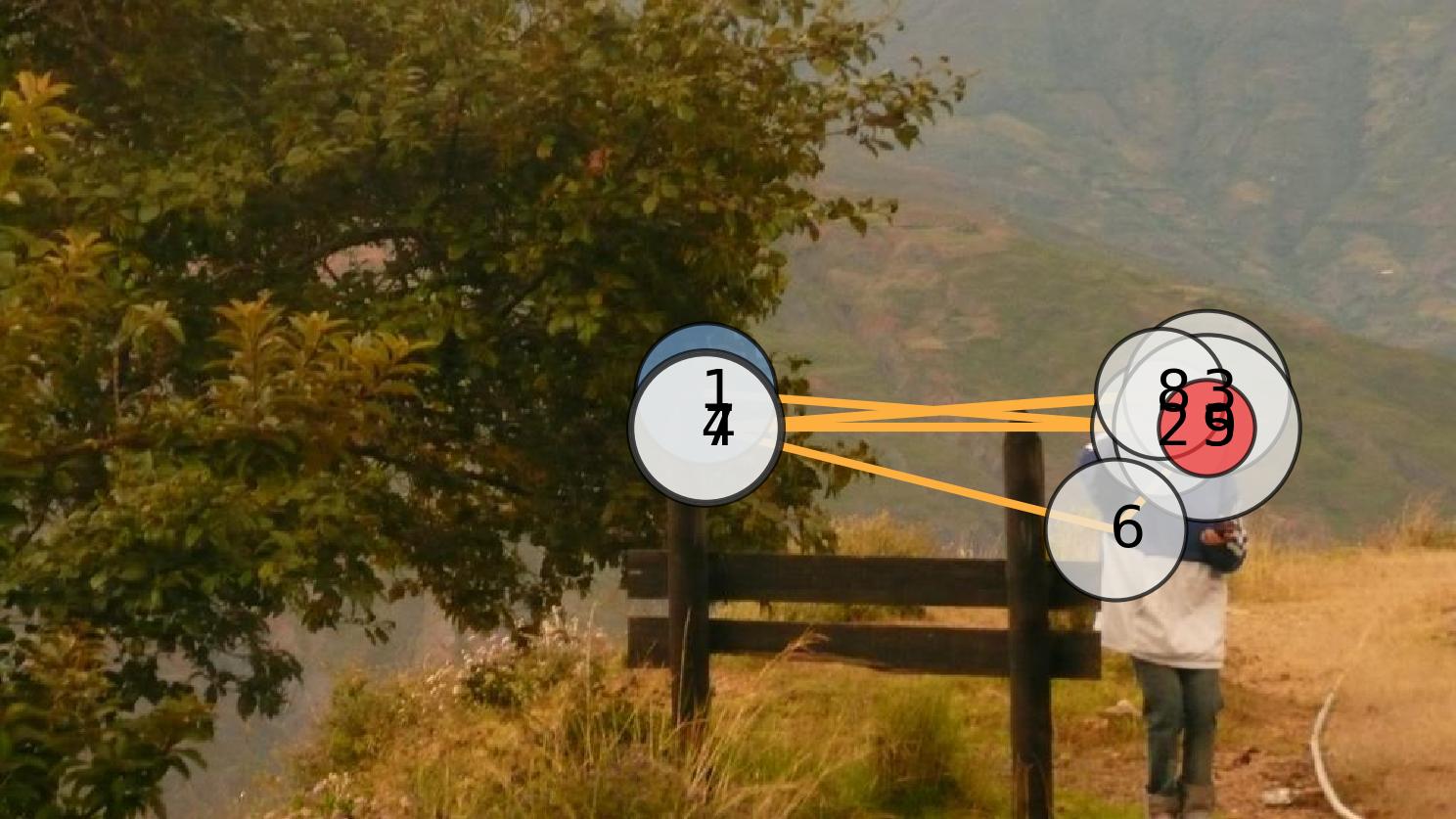} \\ 
         
         \tiny TPP-Gaze~\cite{damelio2025tpp}  & \tiny \textbf{\ours (Ours)} & \tiny Humans \\
         \includegraphics[width=0.16\linewidth]{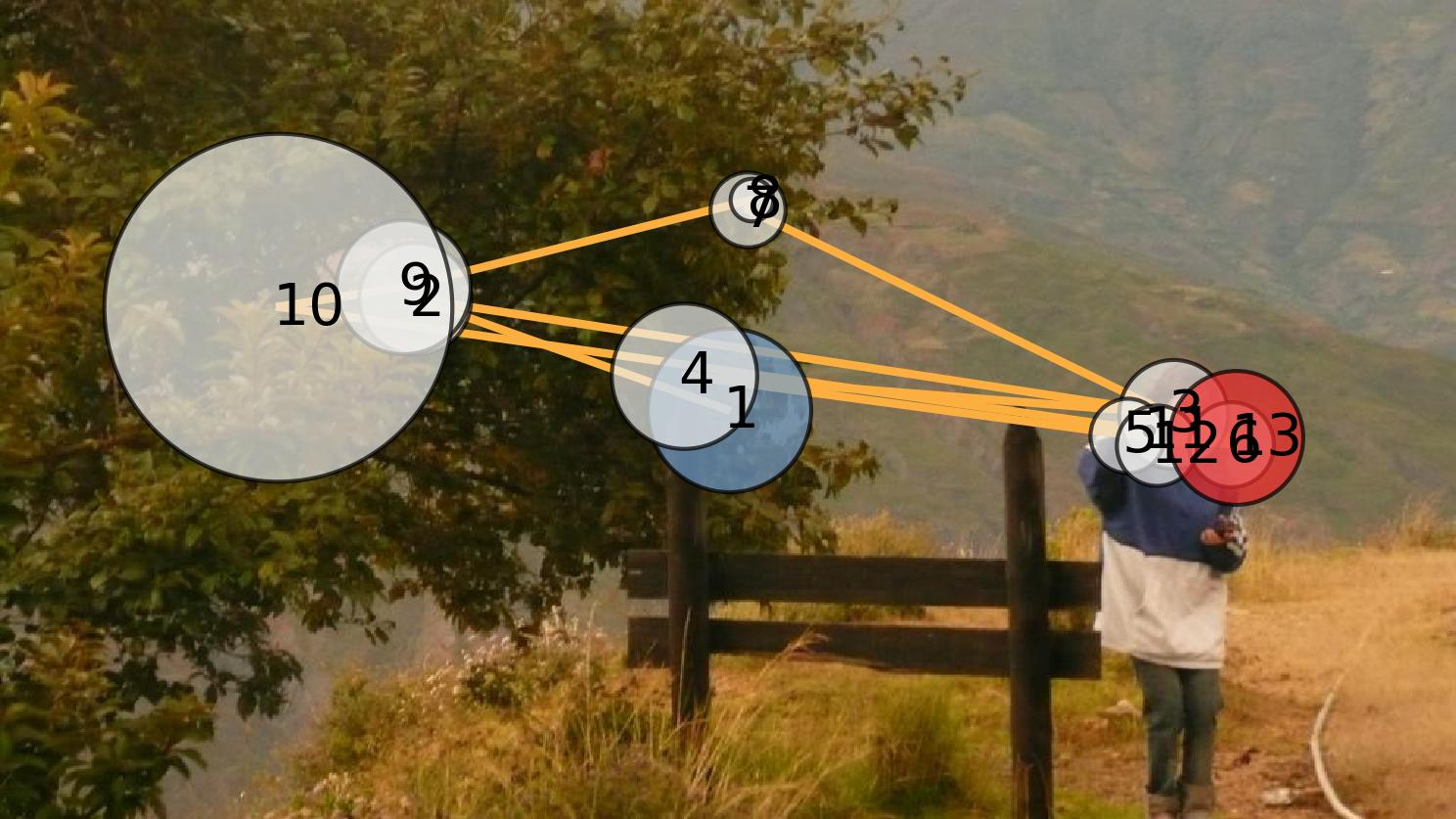} & 
         \includegraphics[width=0.16\linewidth]{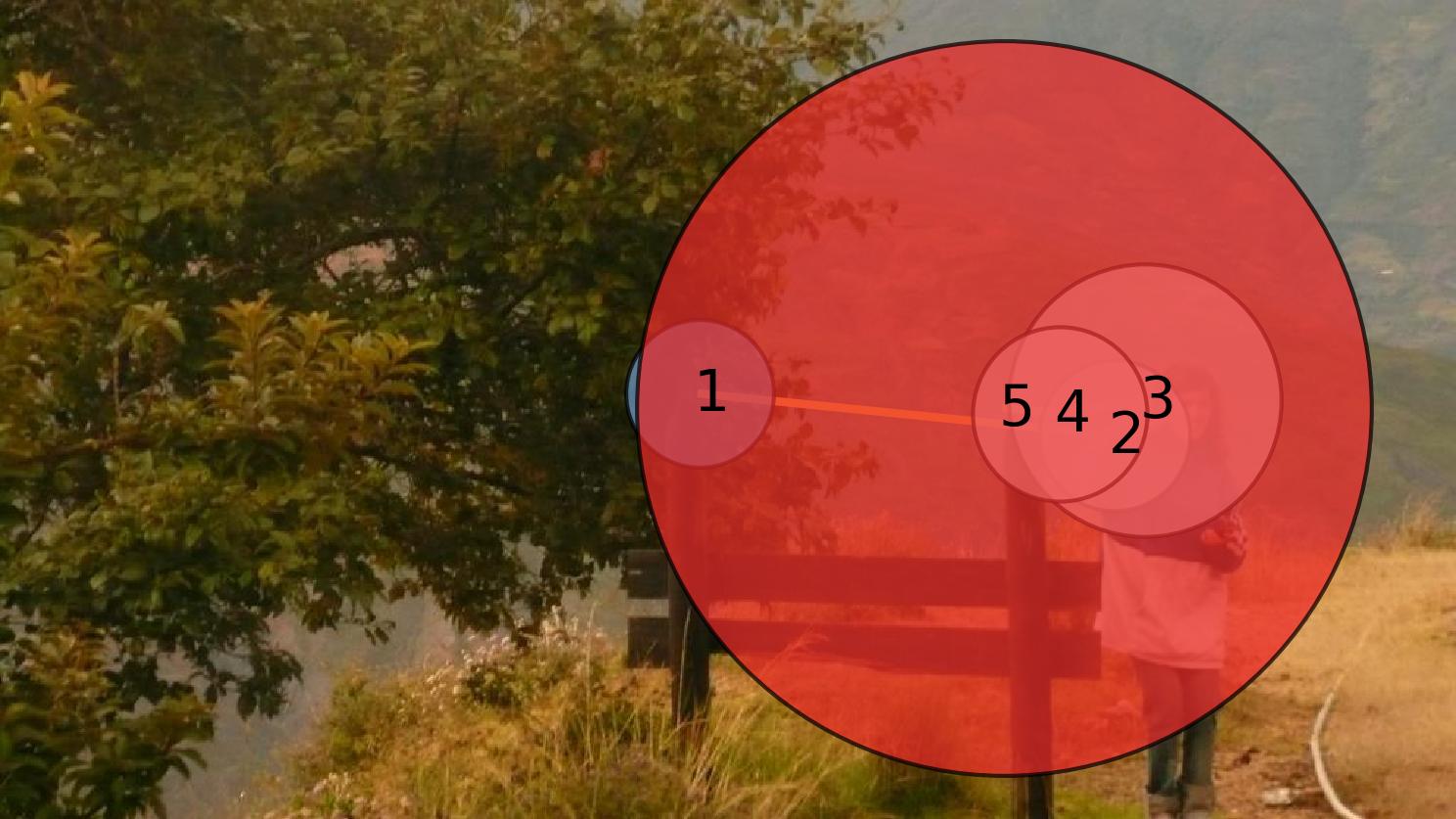} & 
         \includegraphics[width=0.16\linewidth]{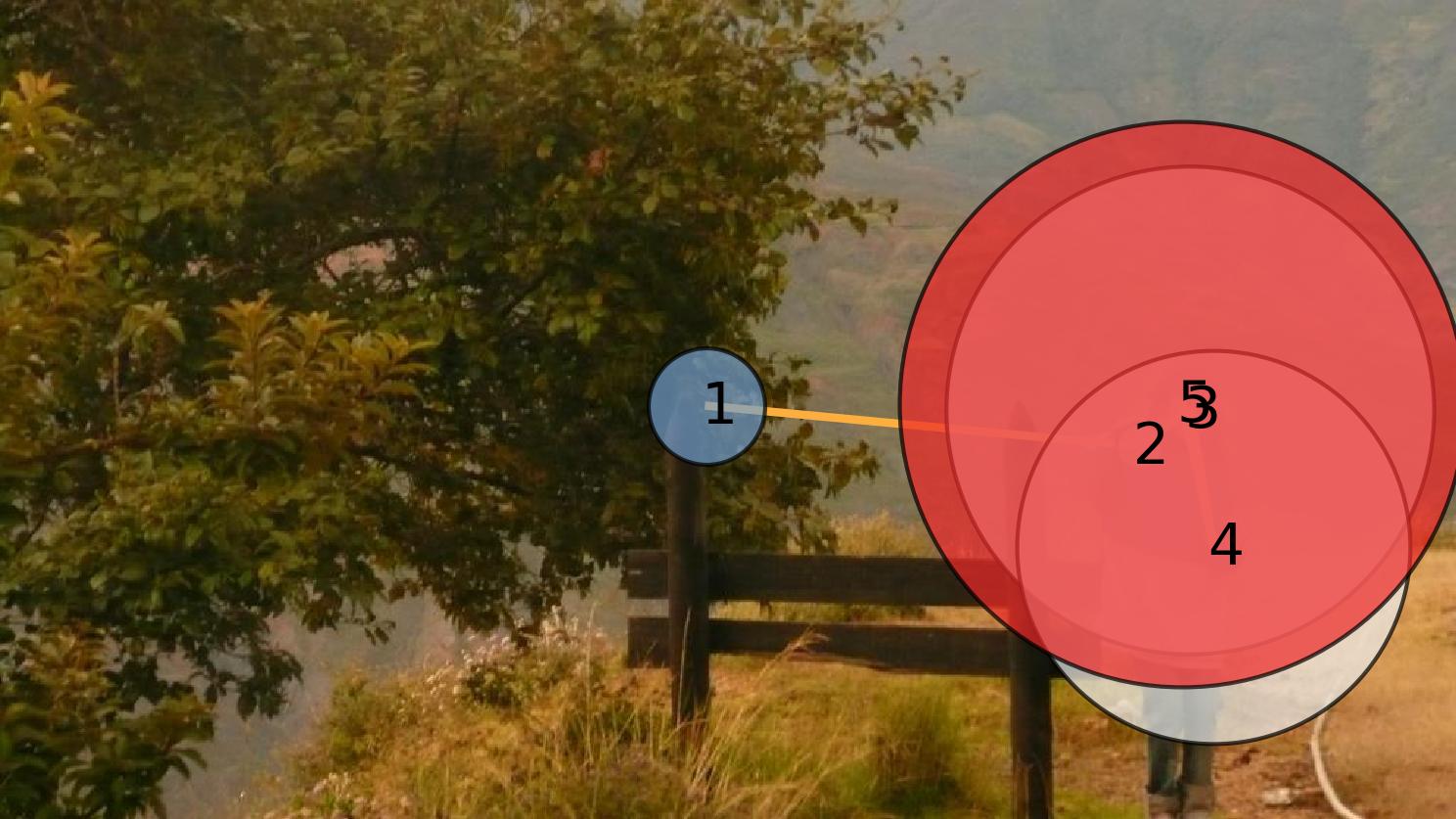} \\

         \addlinespace[0.6cm]
         
         \tiny IOR-ROI-LSTM~\cite{chen2018scanpath} & \tiny ChenLSTM~\cite{chen2021predicting} & \tiny GazeXplain~\cite{chen2024gazexplain}\\
         \addlinespace[0.08cm]
         \includegraphics[width=0.16\linewidth]{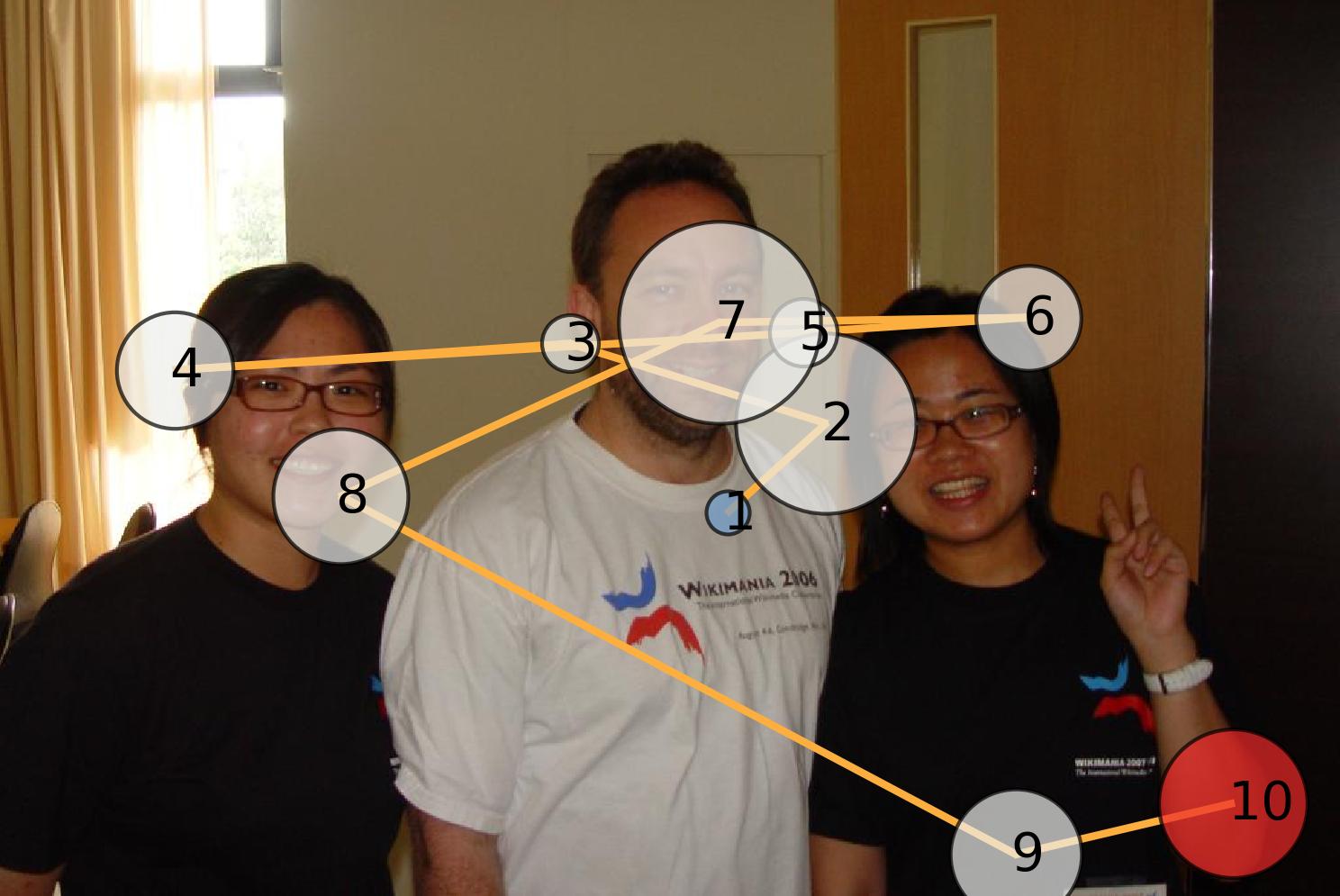} &
         \includegraphics[width=0.16\linewidth]{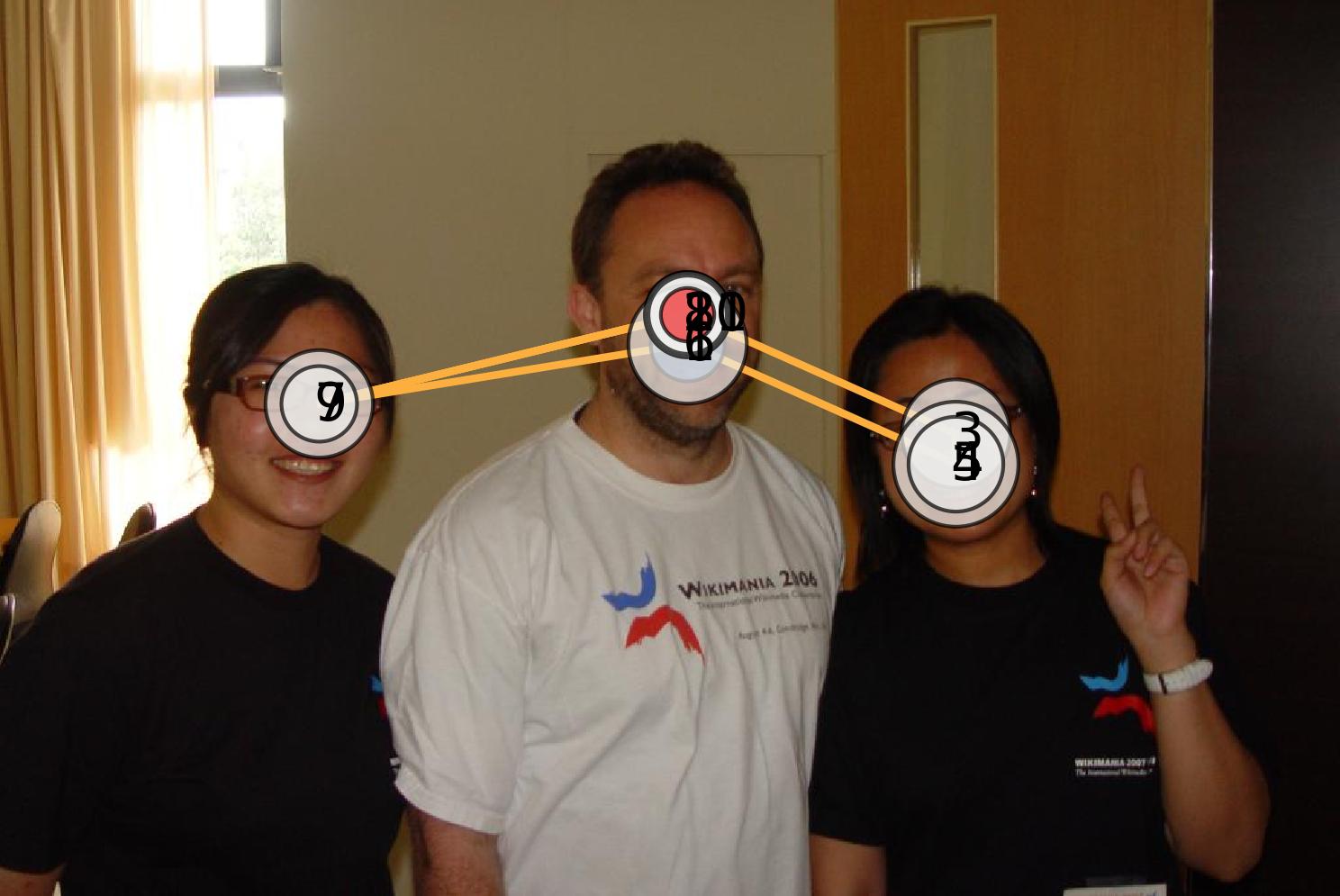} &
         \includegraphics[width=0.16\linewidth]{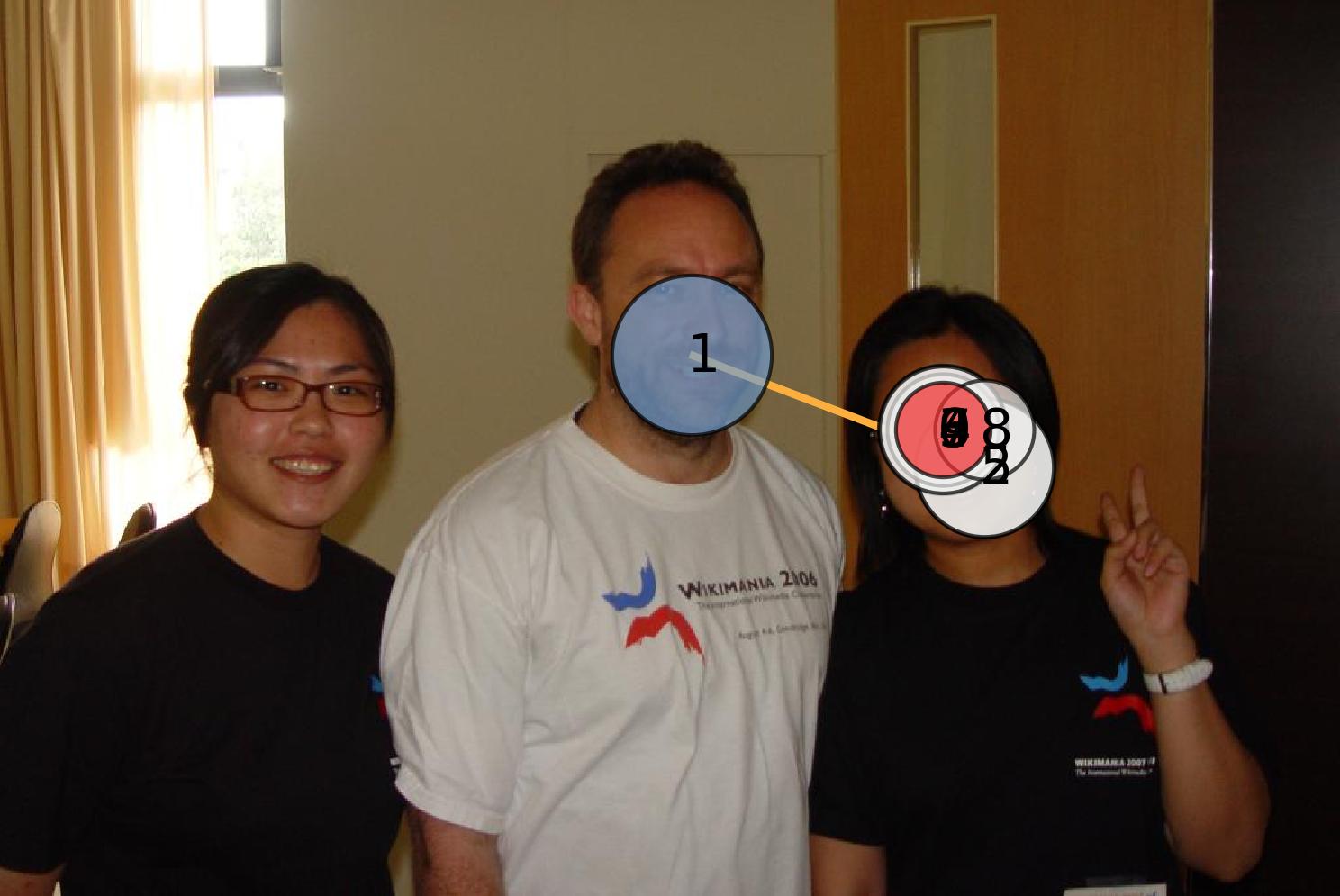} \\ 
         
         \tiny TPP-Gaze~\cite{damelio2025tpp} & \tiny \textbf{\ours (Ours)} & \tiny Humans \\
         \includegraphics[width=0.16\linewidth]{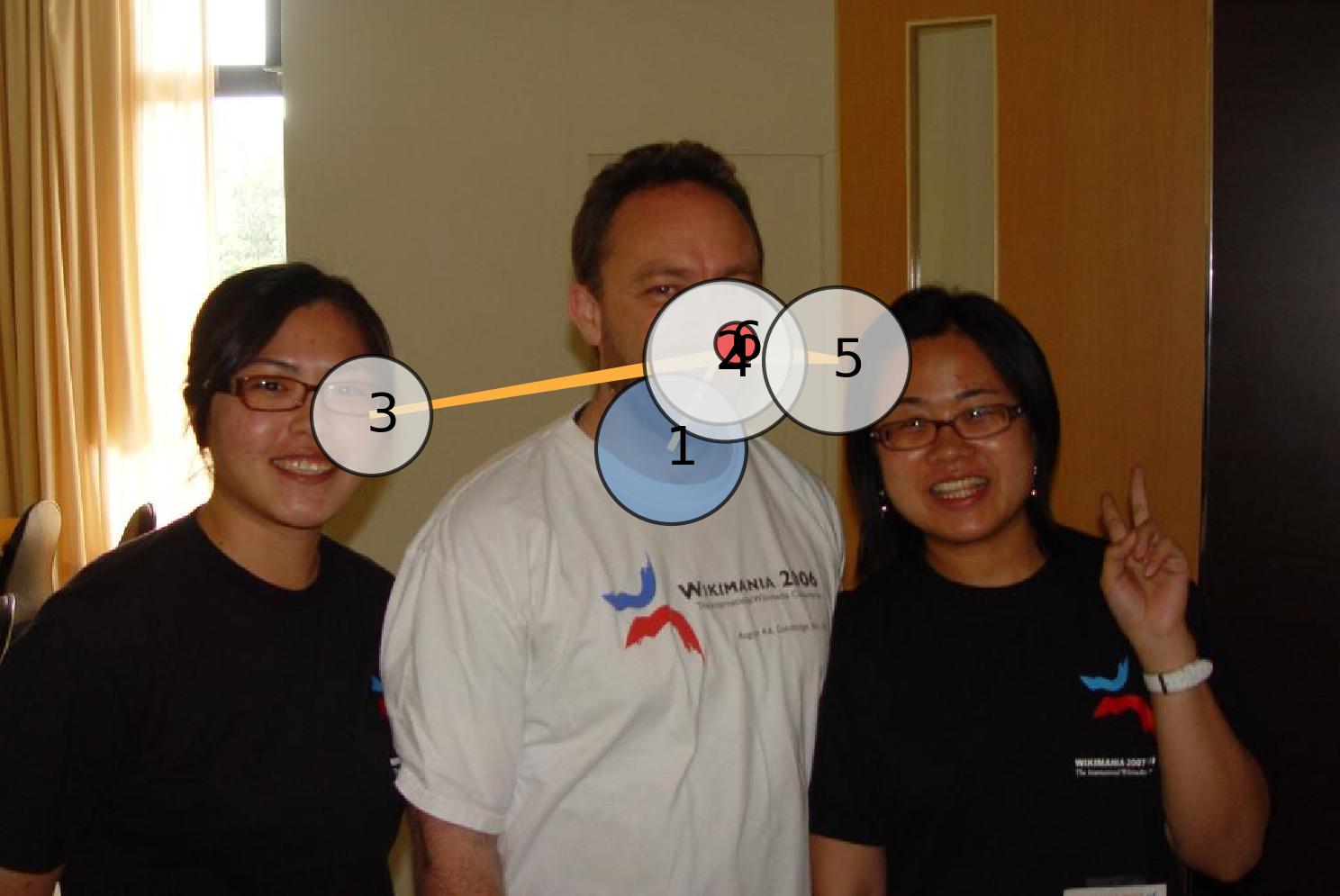} & 
         \includegraphics[width=0.16\linewidth]{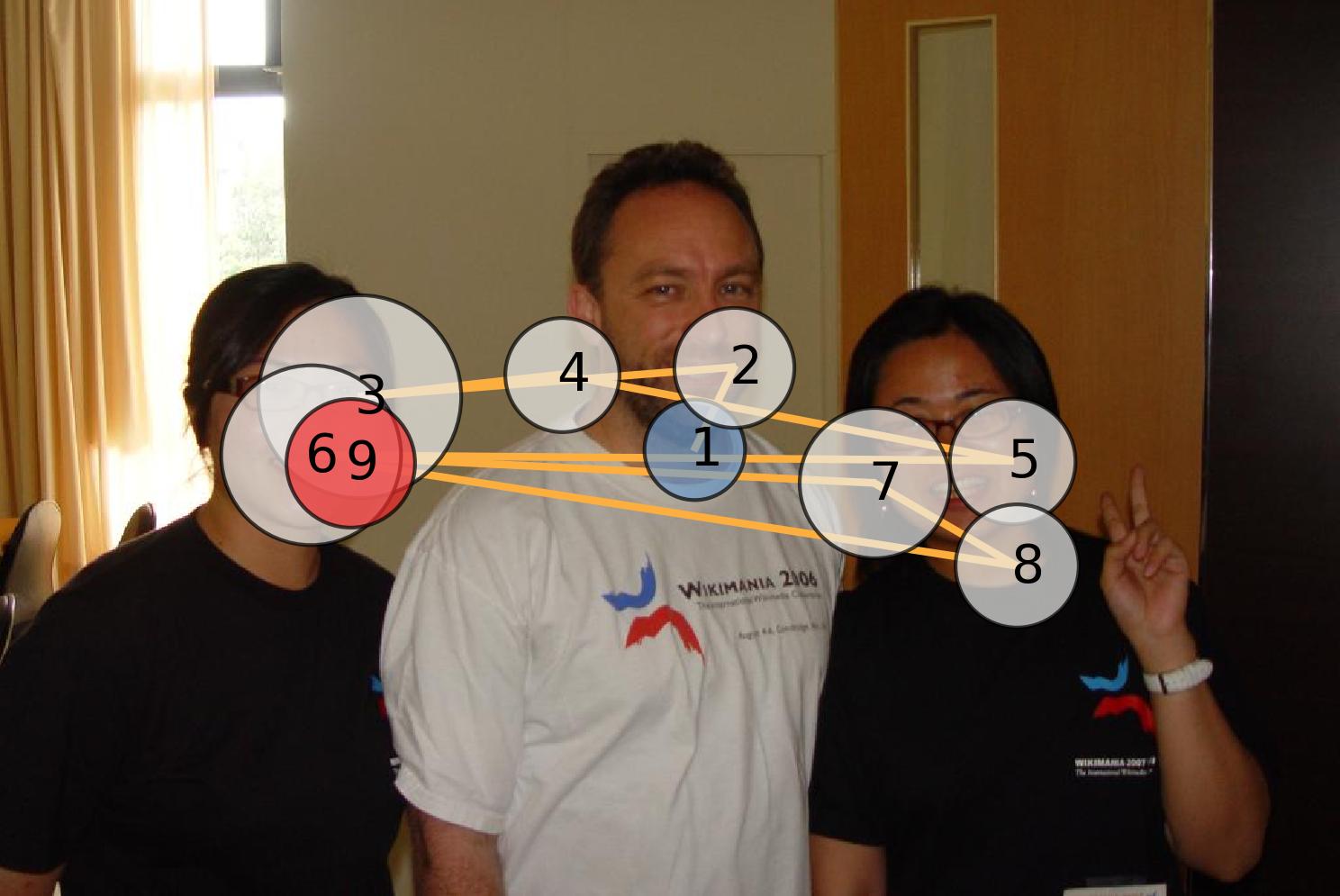} & 
         \includegraphics[width=0.16\linewidth]{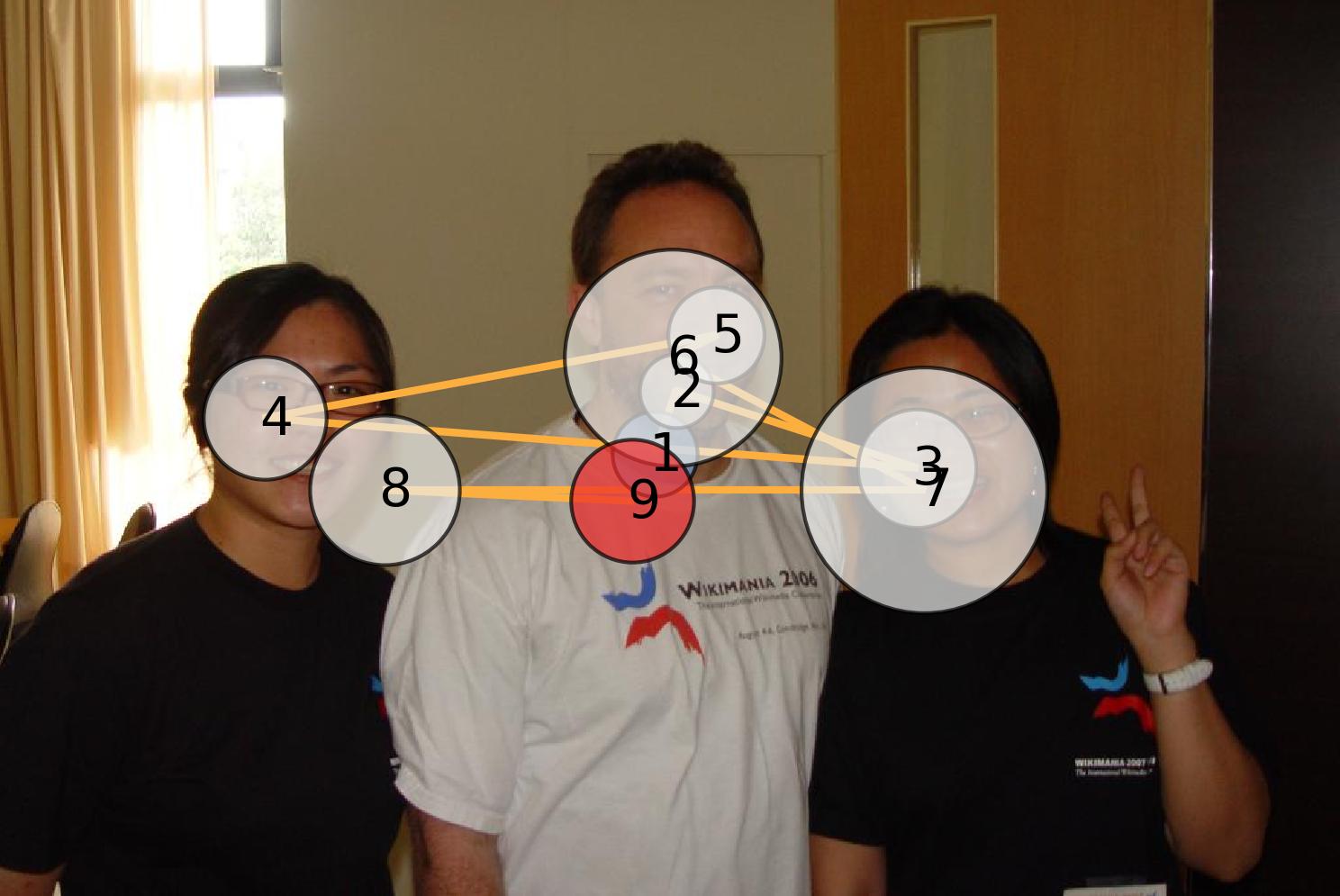}
    \end{tabular}
    }
    \vspace{-0.15cm}
    \caption{Qualitative comparison of simulated and human scanpaths on the MIT1003 dataset.}
    \label{fig:qualitatives_mit}
    \vspace{-0.4cm}
\end{figure*}

\begin{figure*}[t]
    \footnotesize
    \setlength{\tabcolsep}{.1em}
    \resizebox{\linewidth}{!}{
    \begin{tabular}{ccc}
         \tiny IOR-ROI-LSTM~\cite{chen2018scanpath} & \tiny ChenLSTM~\cite{chen2021predicting} & \tiny GazeXplain~\cite{chen2024gazexplain}\\
         \addlinespace[0.08cm]
         \includegraphics[width=0.16\linewidth]{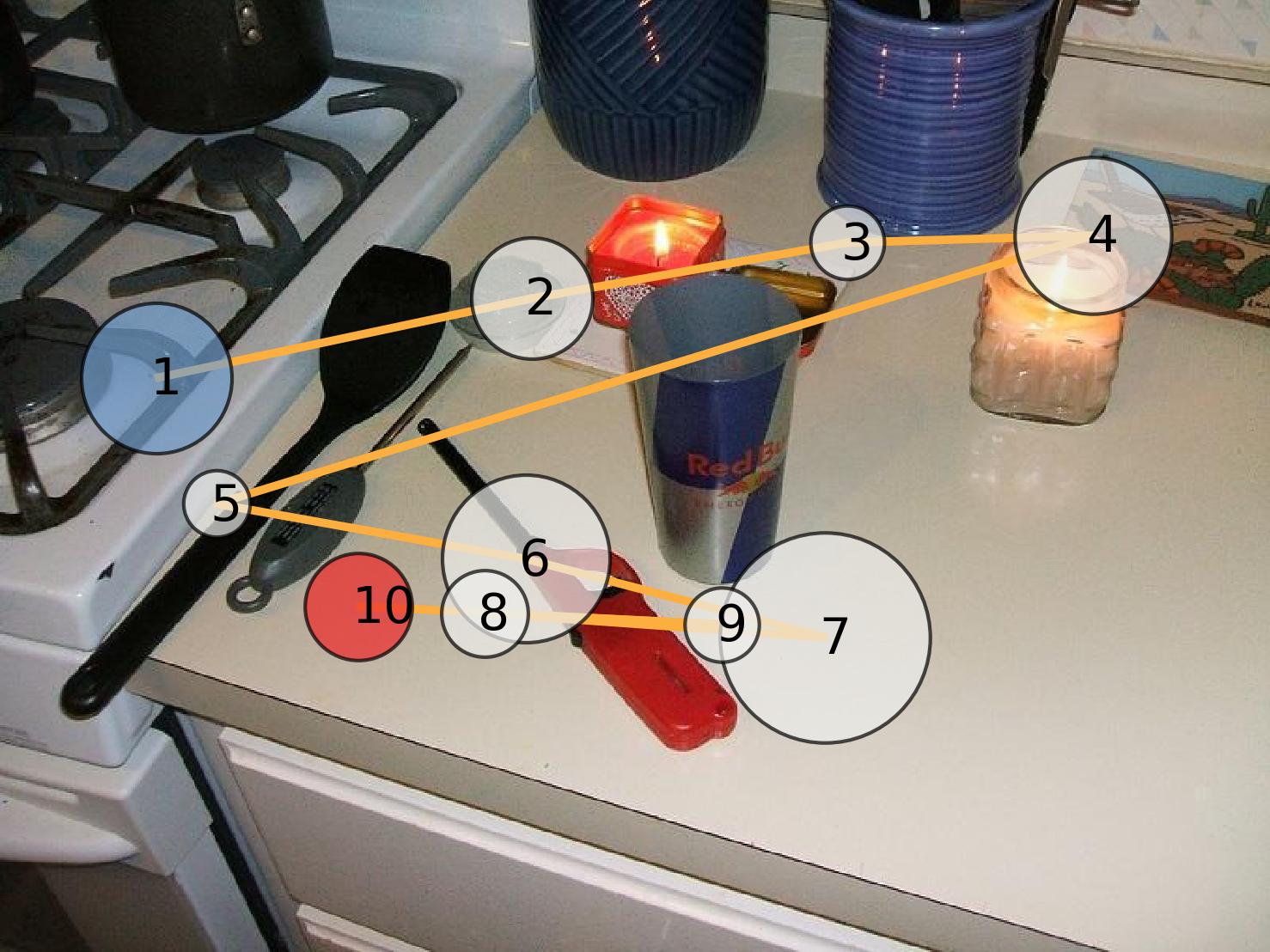} &
         \includegraphics[width=0.16\linewidth]{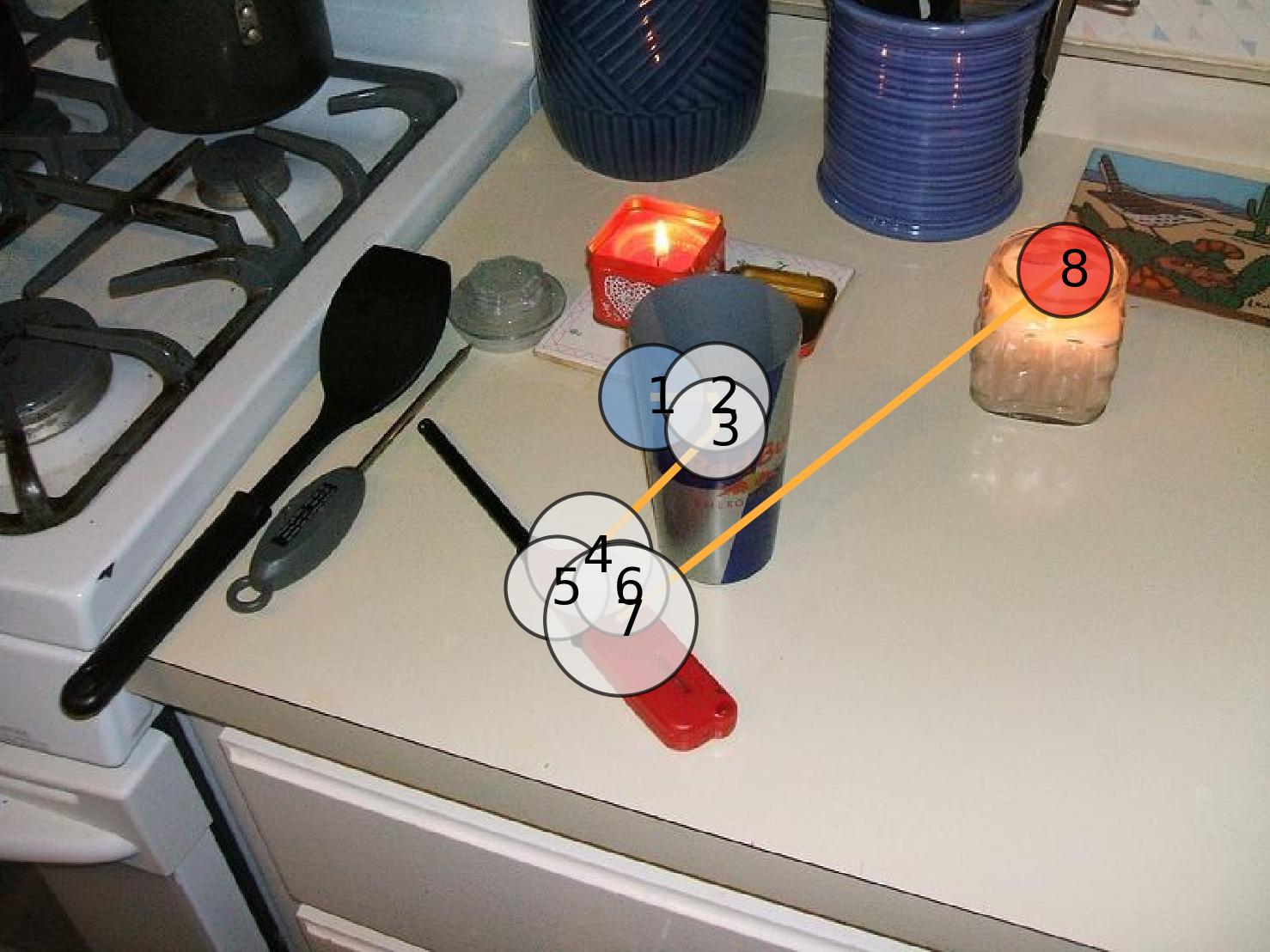} &
         \includegraphics[width=0.16\linewidth]{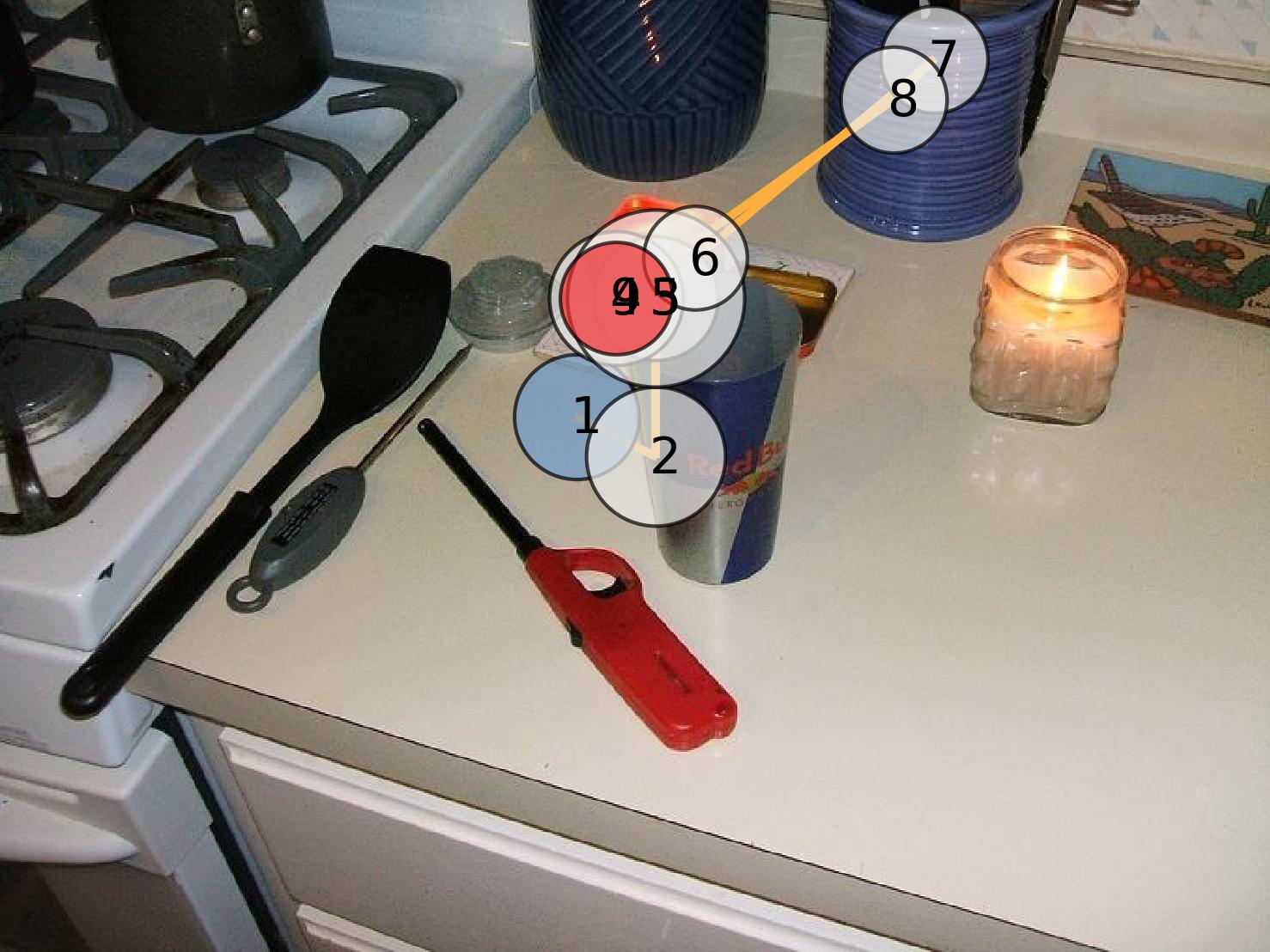} \\ 
         
         \tiny TPP-Gaze~\cite{damelio2025tpp}  & \tiny \textbf{\ours (Ours)} & \tiny Humans \\
         \includegraphics[width=0.16\linewidth]{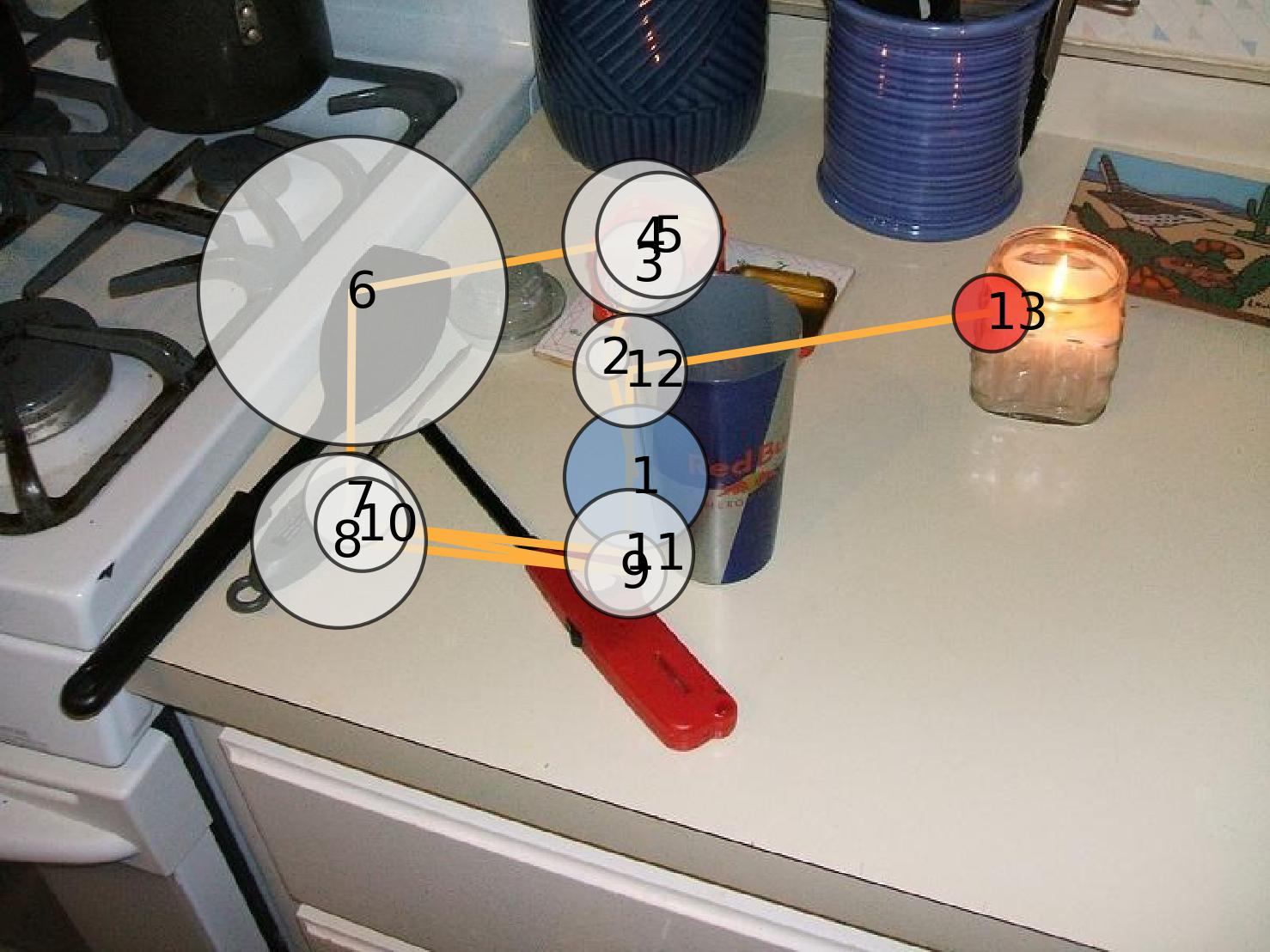} & 
         \includegraphics[width=0.16\linewidth]{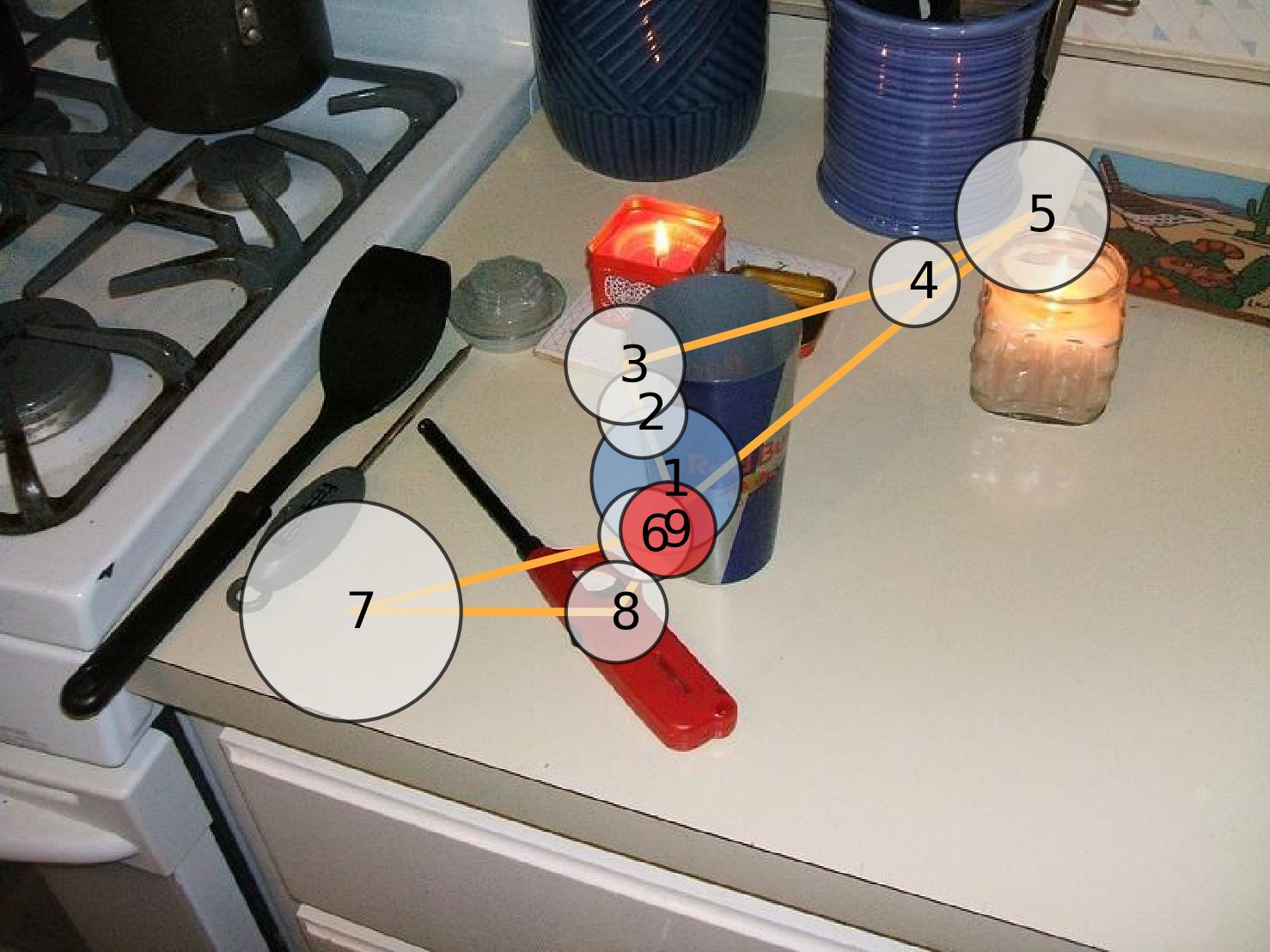} & 
         \includegraphics[width=0.16\linewidth]{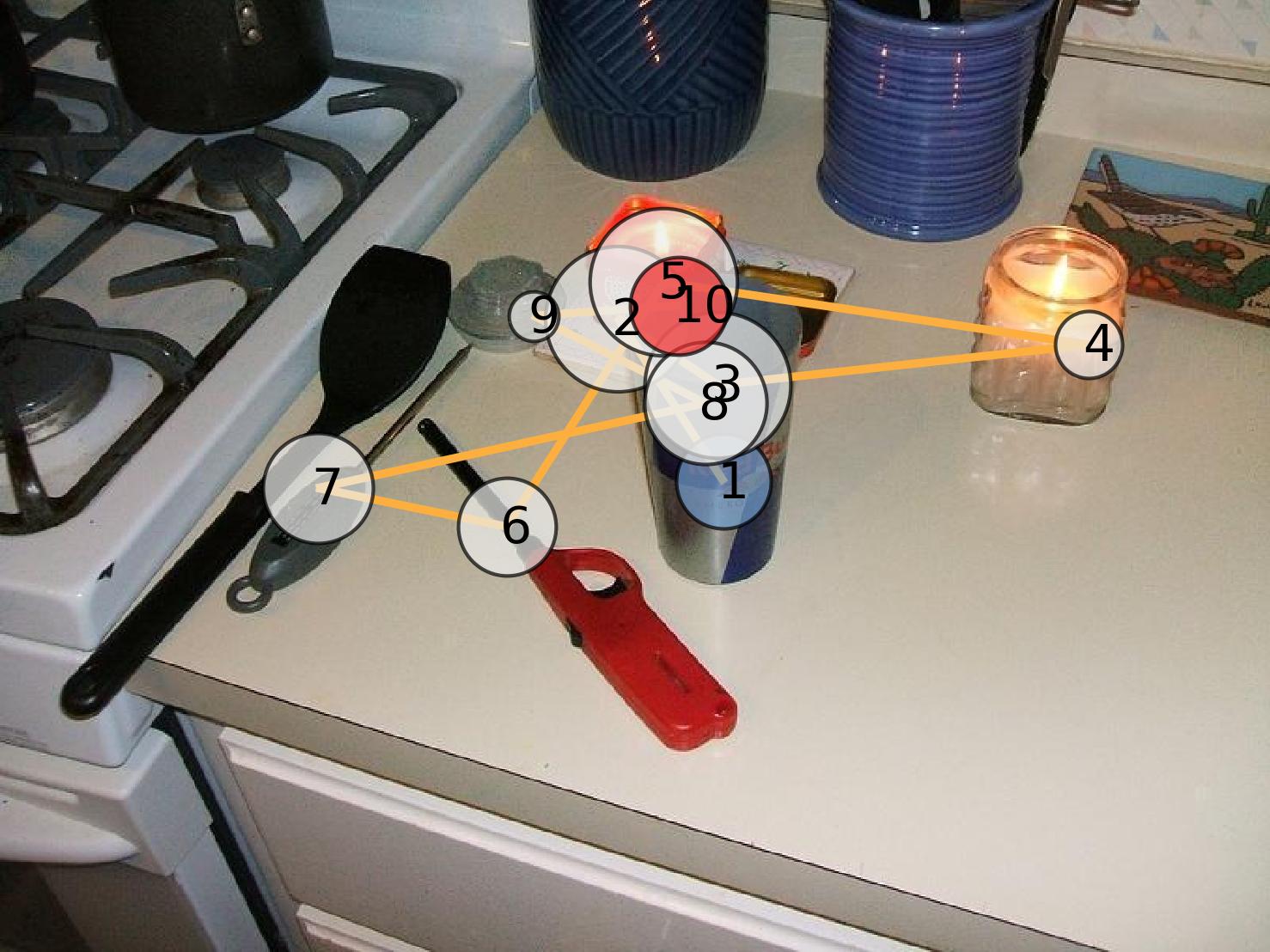} \\

         \addlinespace[0.6cm]
         
         \tiny IOR-ROI-LSTM~\cite{chen2018scanpath} & \tiny ChenLSTM~\cite{chen2021predicting} & \tiny GazeXplain~\cite{chen2024gazexplain}\\
         \addlinespace[0.08cm]
         \includegraphics[width=0.16\linewidth]{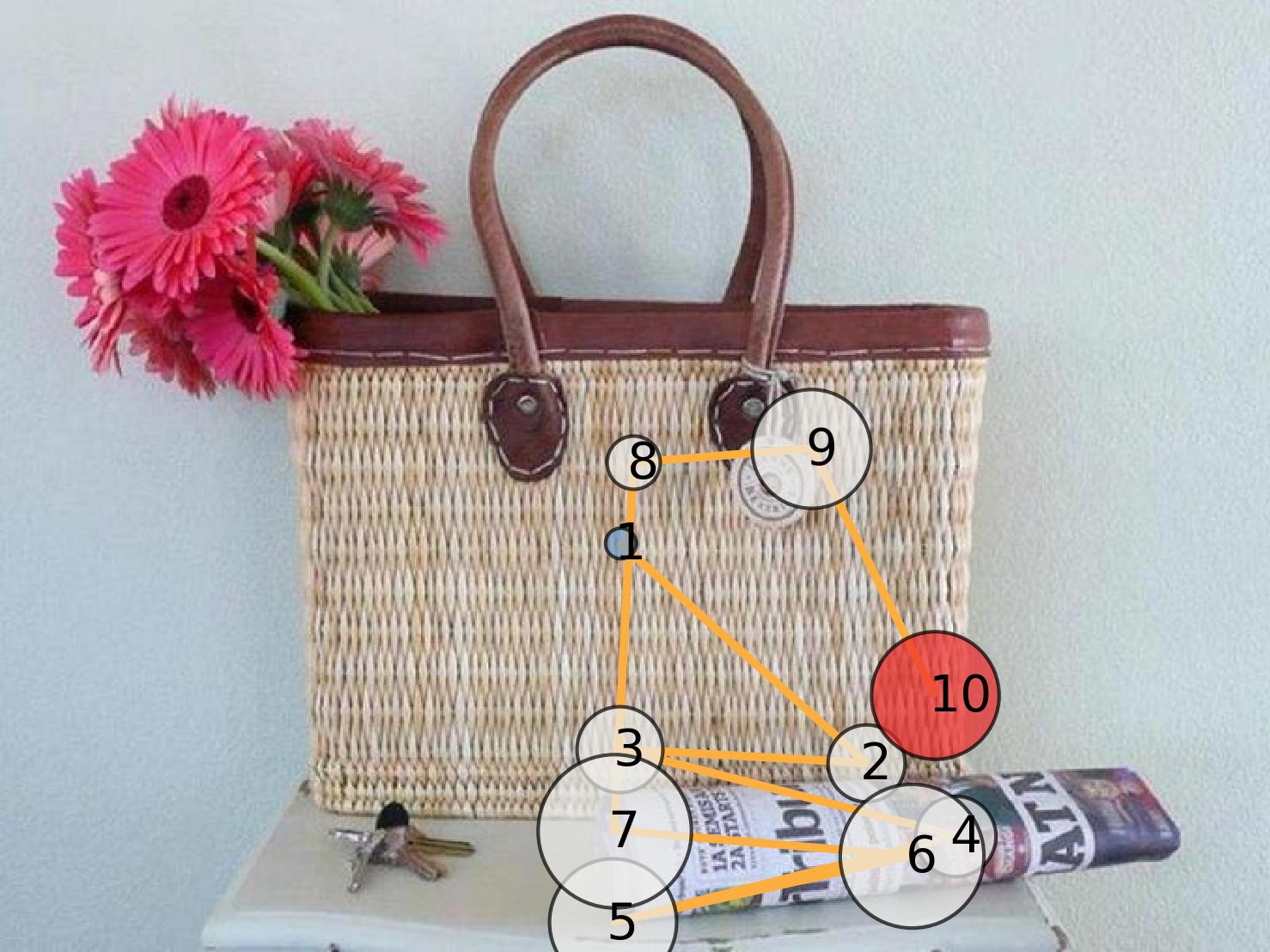} &
         \includegraphics[width=0.16\linewidth]{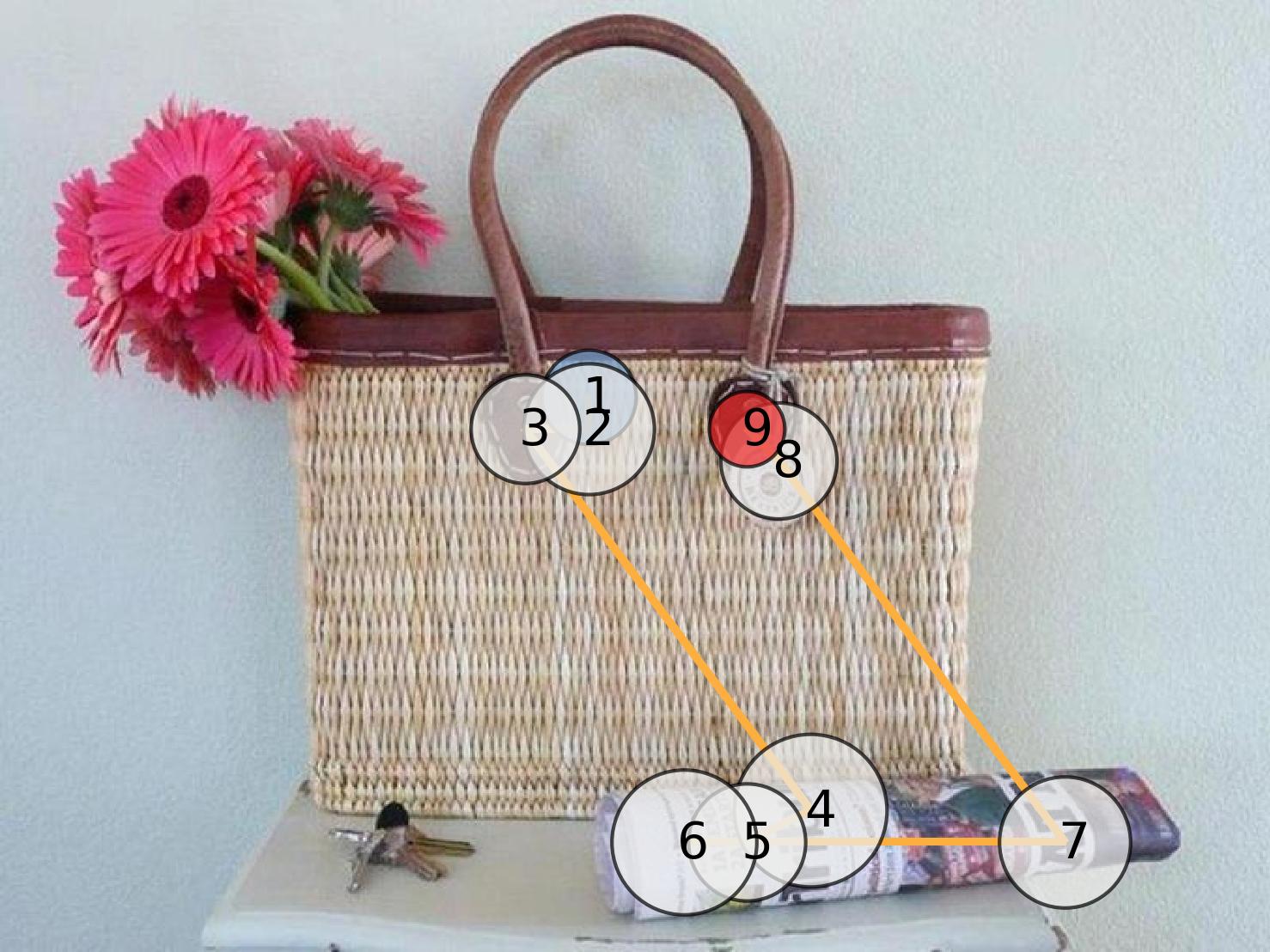} &
         \includegraphics[width=0.16\linewidth]{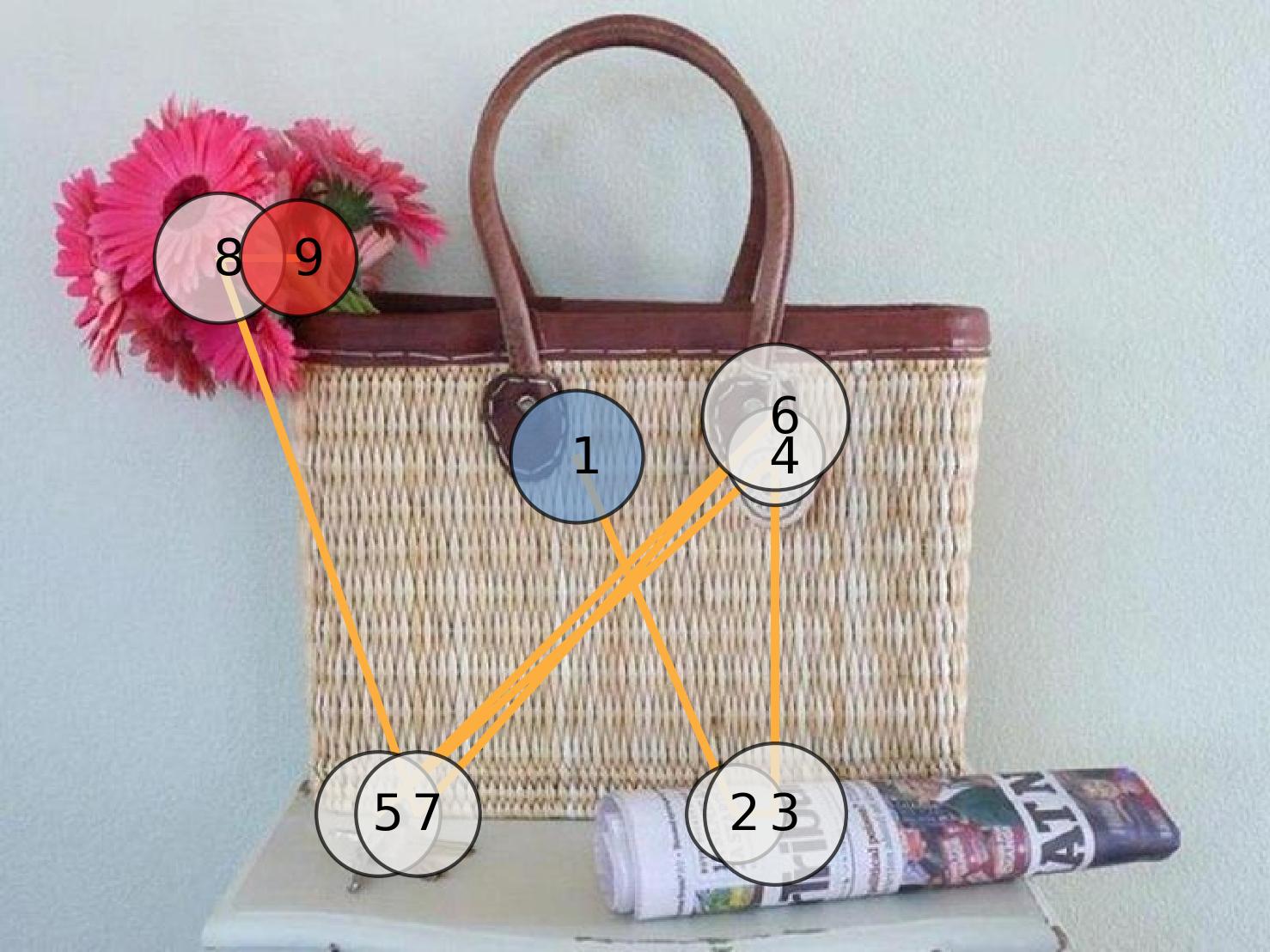} \\ 
         
         \tiny TPP-Gaze~\cite{damelio2025tpp} & \tiny \textbf{\ours (Ours)} & \tiny Humans \\
         \includegraphics[width=0.16\linewidth]{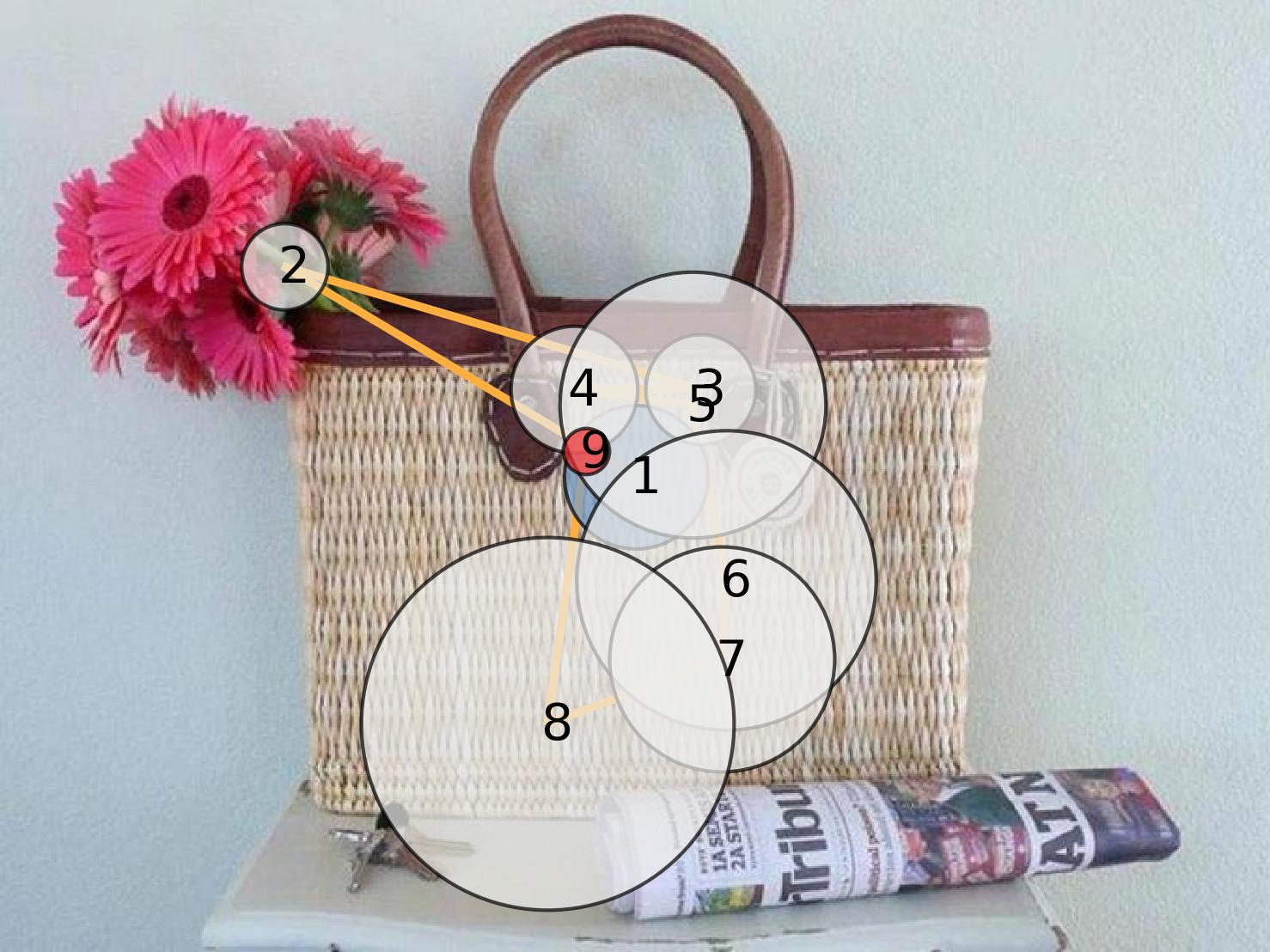} & 
         \includegraphics[width=0.16\linewidth]{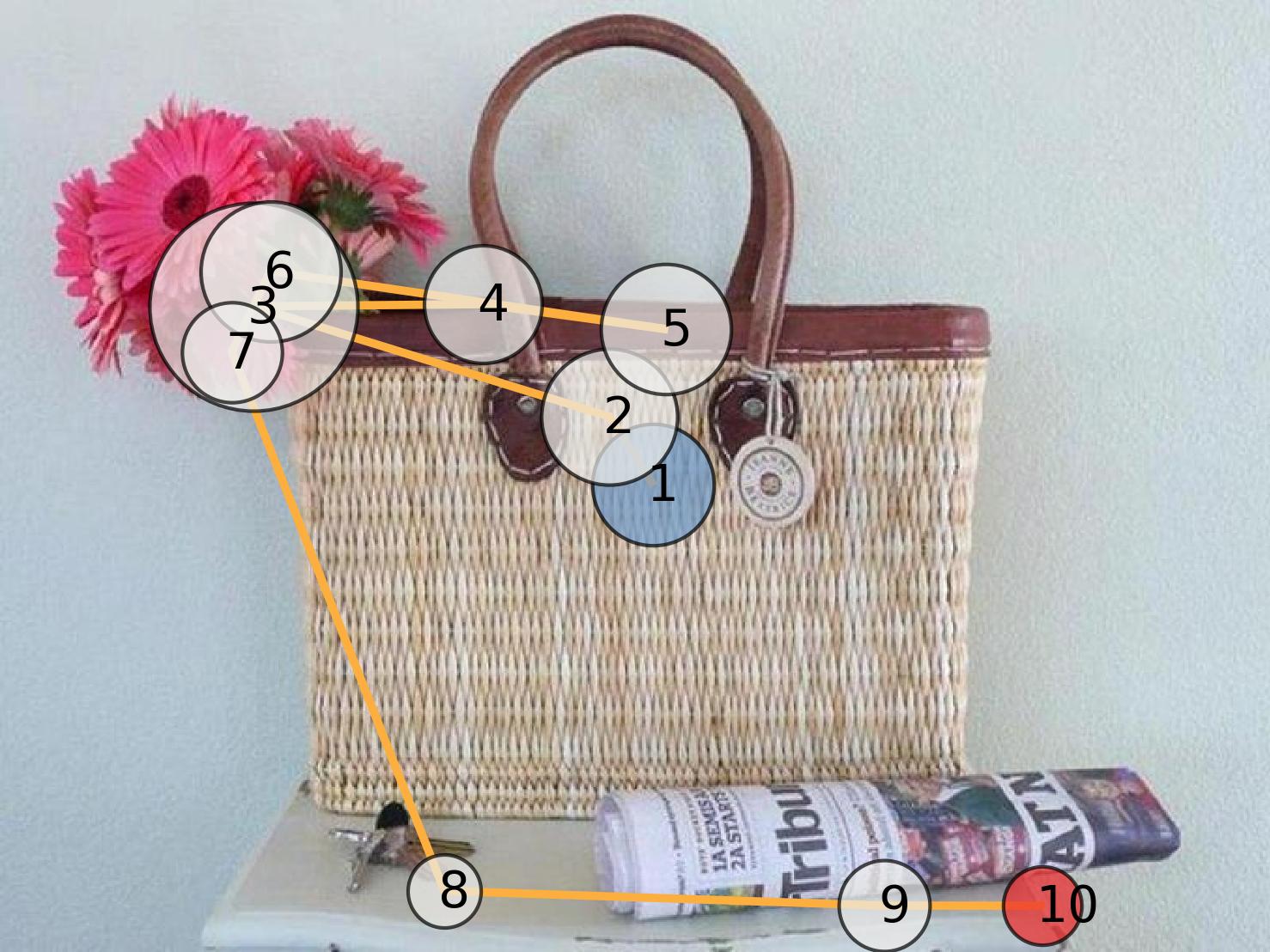} & 
         \includegraphics[width=0.16\linewidth]{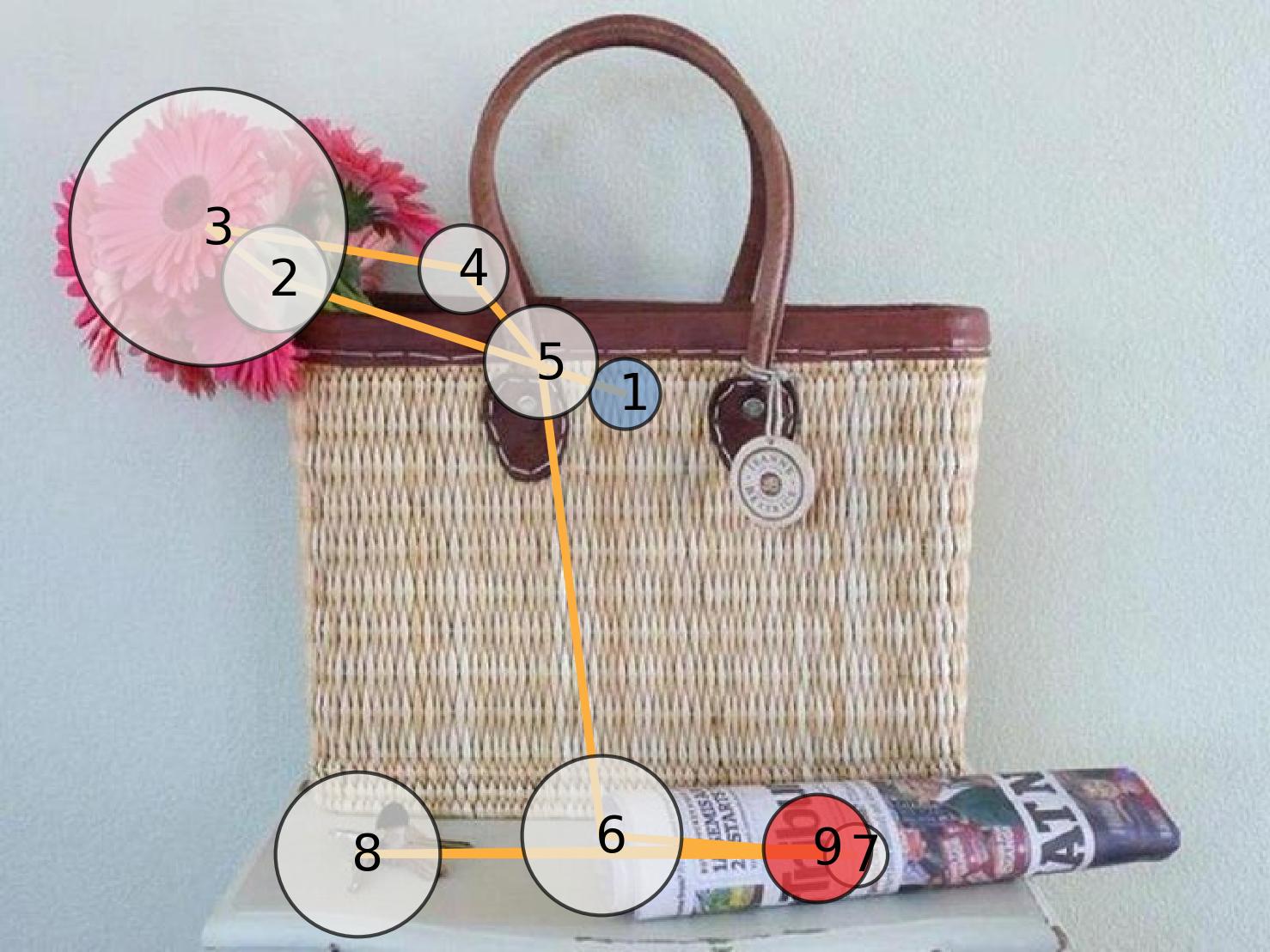}
    \end{tabular}
    }
    \vspace{-0.15cm}
    \caption{Qualitative comparison of simulated and human scanpaths on the OSIE dataset.}
    \label{fig:qualitatives_osie}
    \vspace{-0.4cm}
\end{figure*}

\begin{figure*}[t]
    \footnotesize
    \setlength{\tabcolsep}{.1em}
    \resizebox{\linewidth}{!}{
    \begin{tabular}{cc ccc}
         & & \tiny ChenLSTM~\cite{chen2021predicting} & \tiny Gazeformer~\cite{mondal2023gazeformer} & \tiny GazeXplain~\cite{chen2024gazexplain} \\
         \rotatebox{90}{\parbox[t]{0.12\linewidth}{\hspace*{\fill}\tiny \textbf{Target:} \texttt{sink}\hspace*{\fill}}} 
         & & \includegraphics[height=0.12\linewidth]{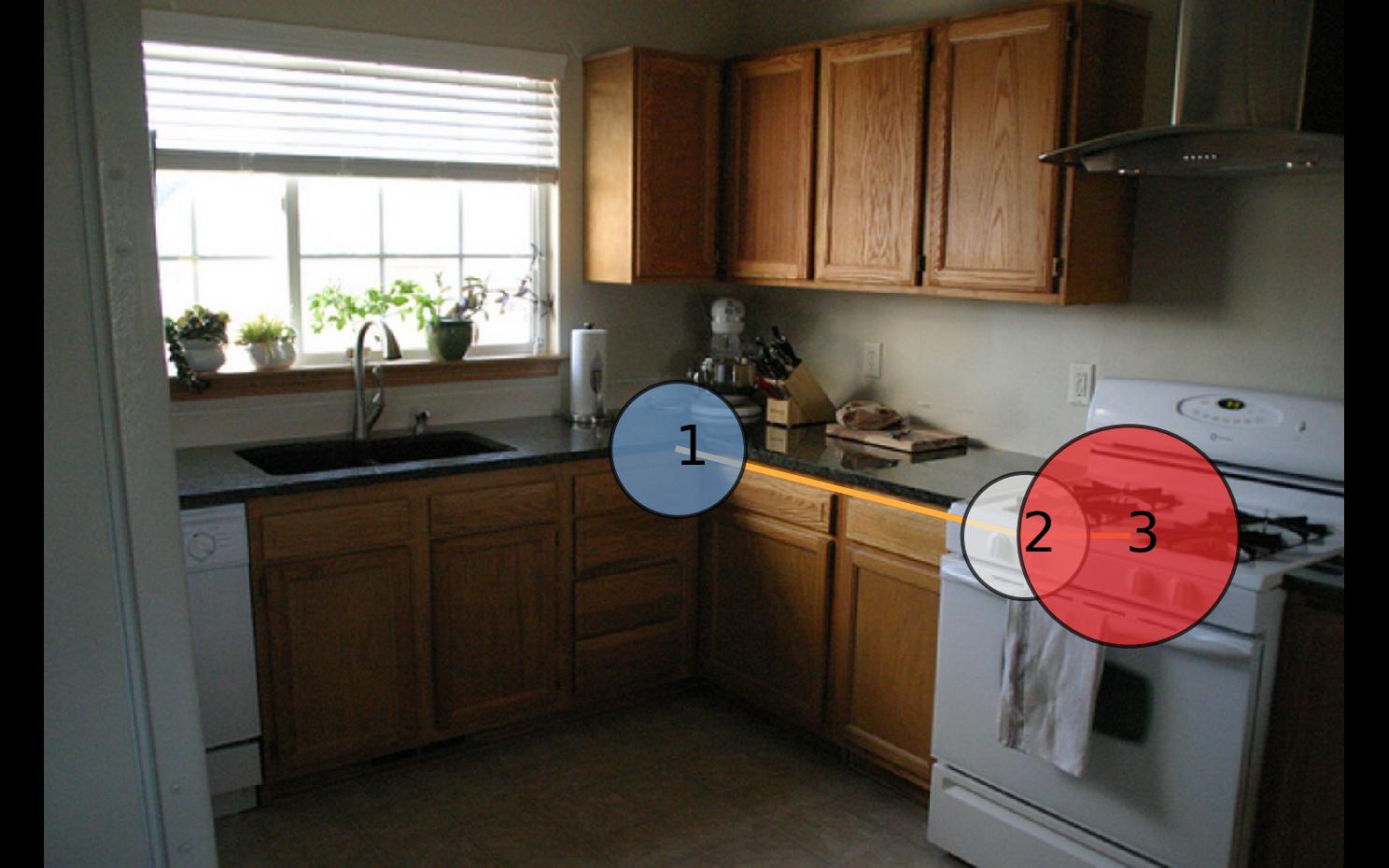} & 
         \includegraphics[height=0.12\linewidth]{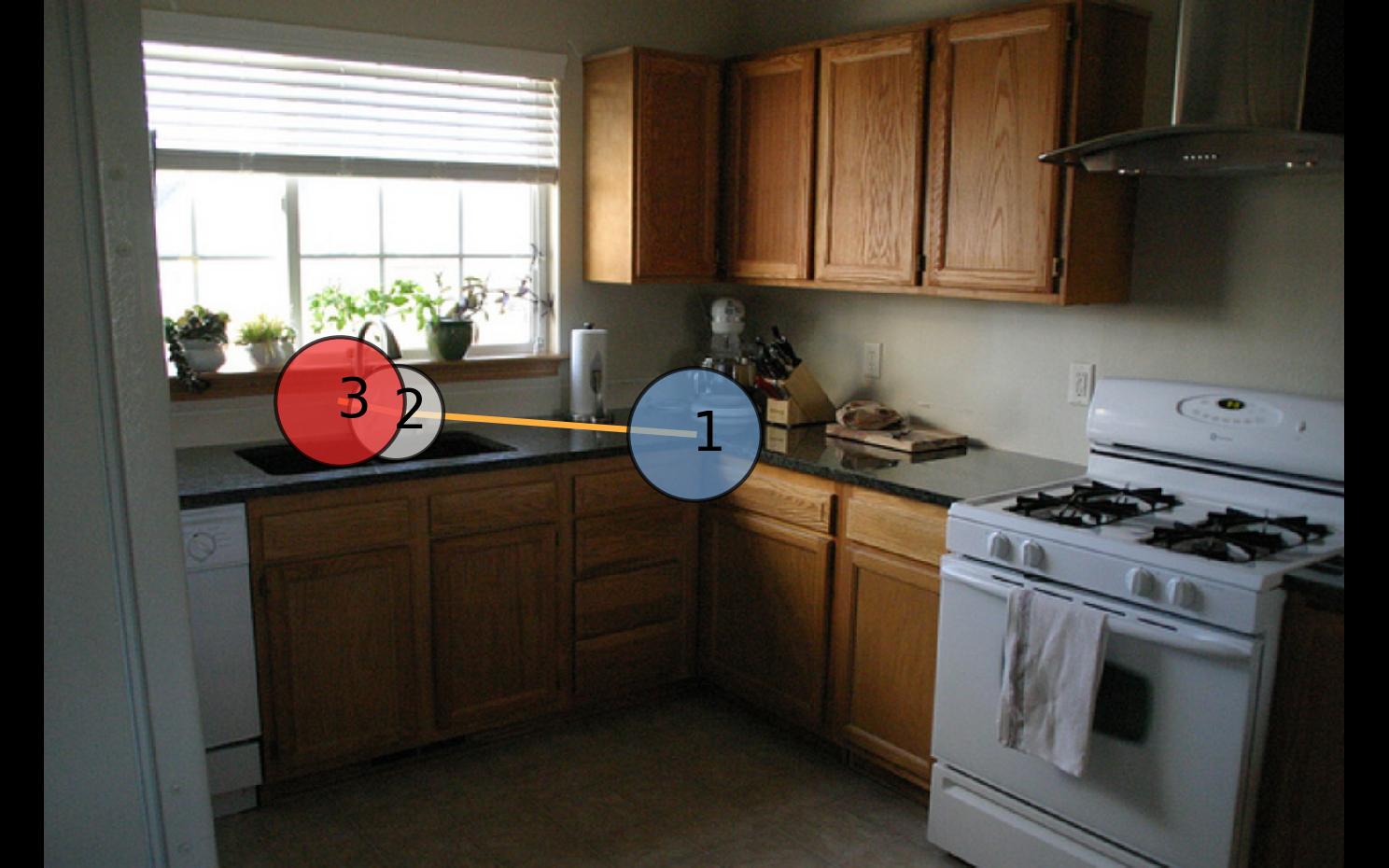} & 
         \includegraphics[height=0.12\linewidth]{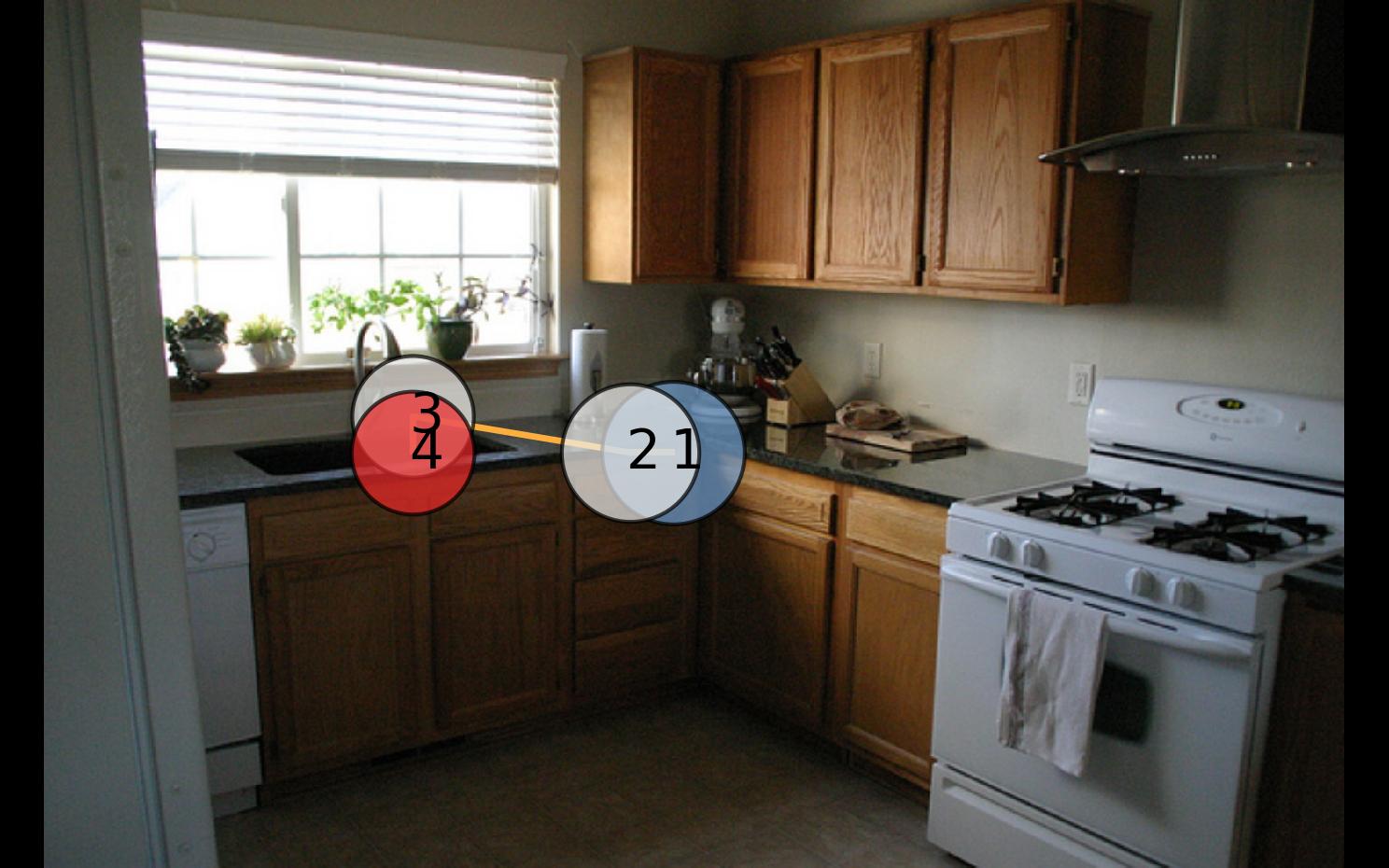} \\         
         & & \tiny TPP-Gaze~\cite{damelio2025tpp} & \tiny \textbf{\ours (Ours)} & \tiny Humans \\
         & & \includegraphics[height=0.12\linewidth]{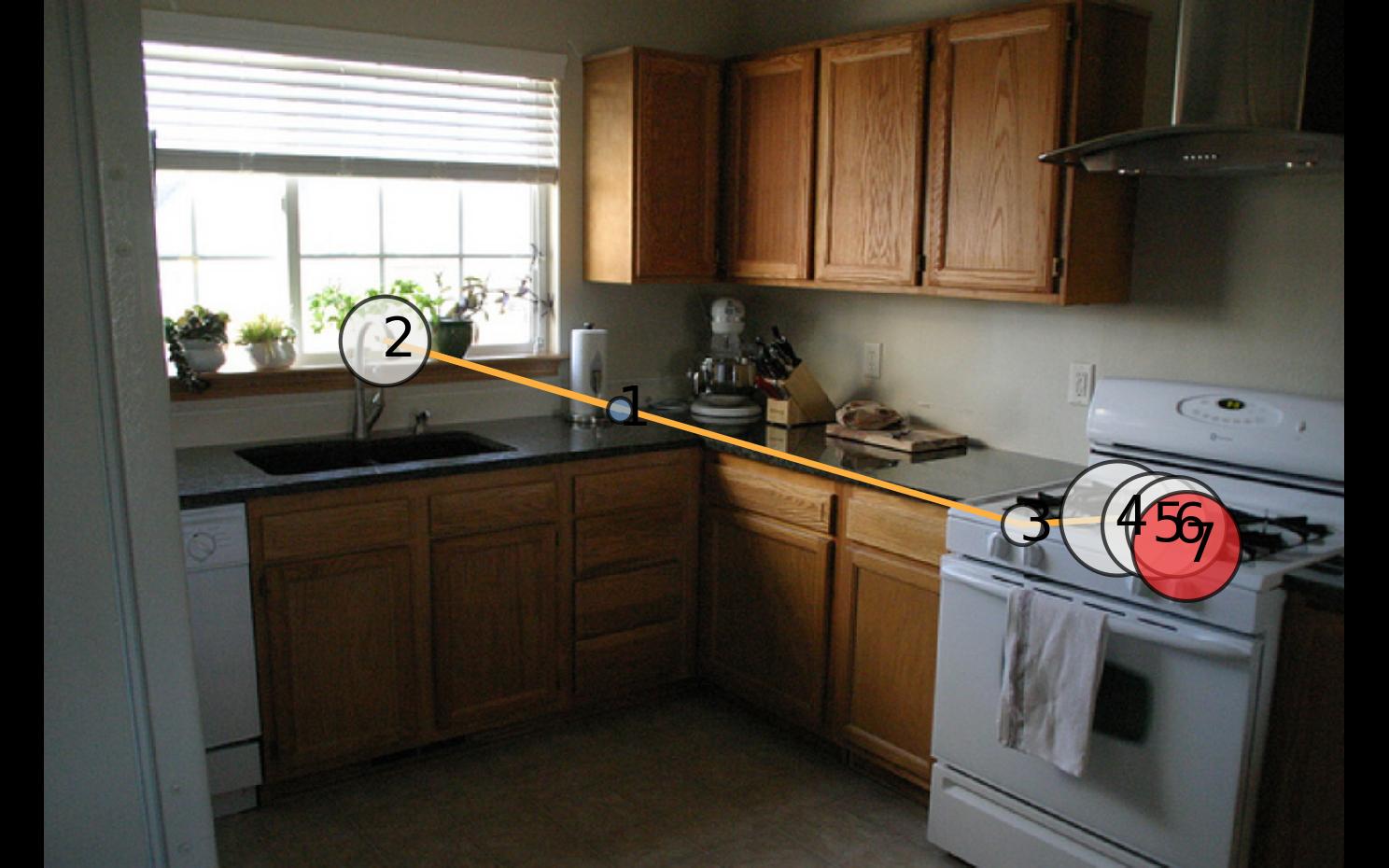} & 
         \includegraphics[height=0.12\linewidth]{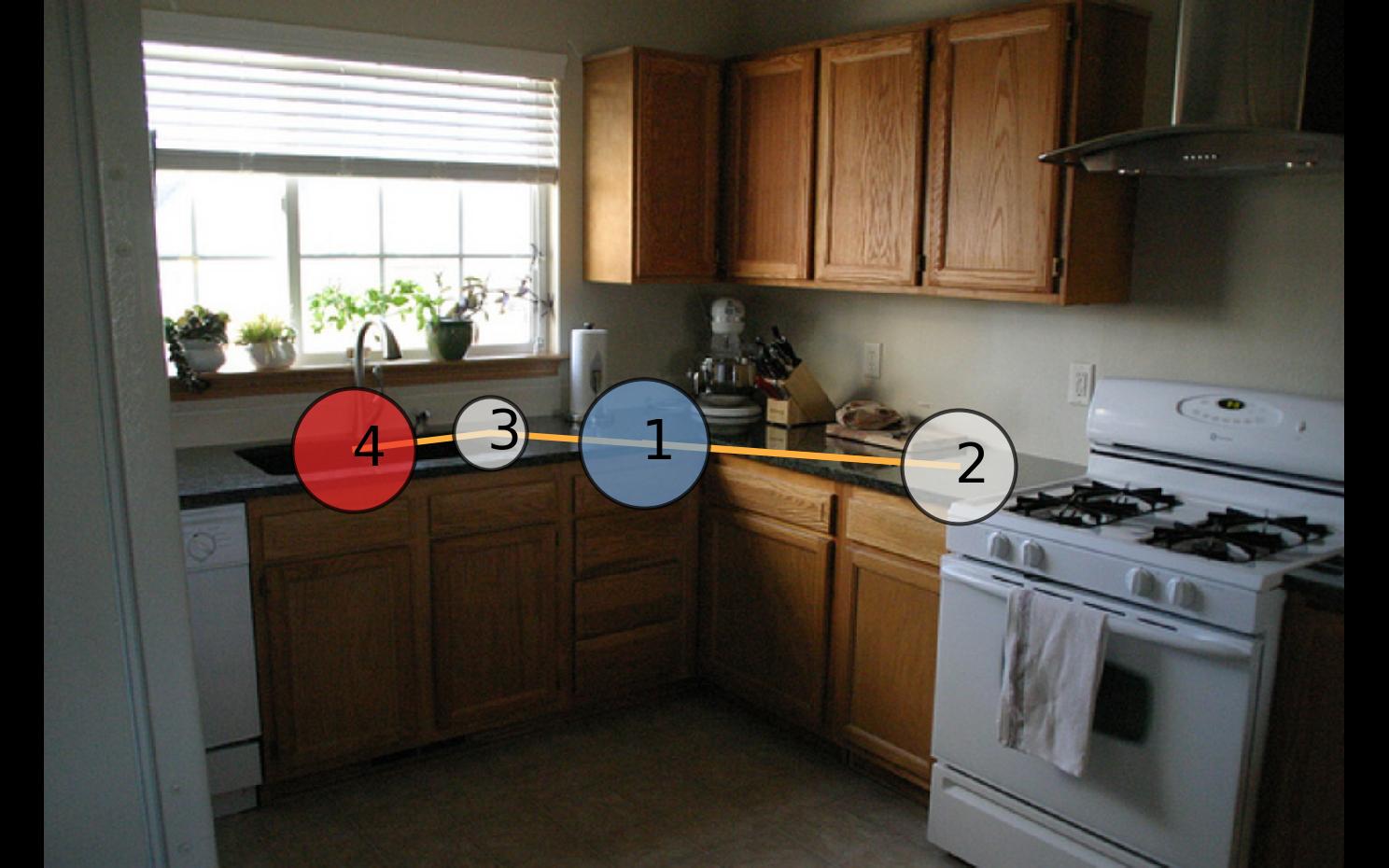} & 
         \includegraphics[height=0.12\linewidth]{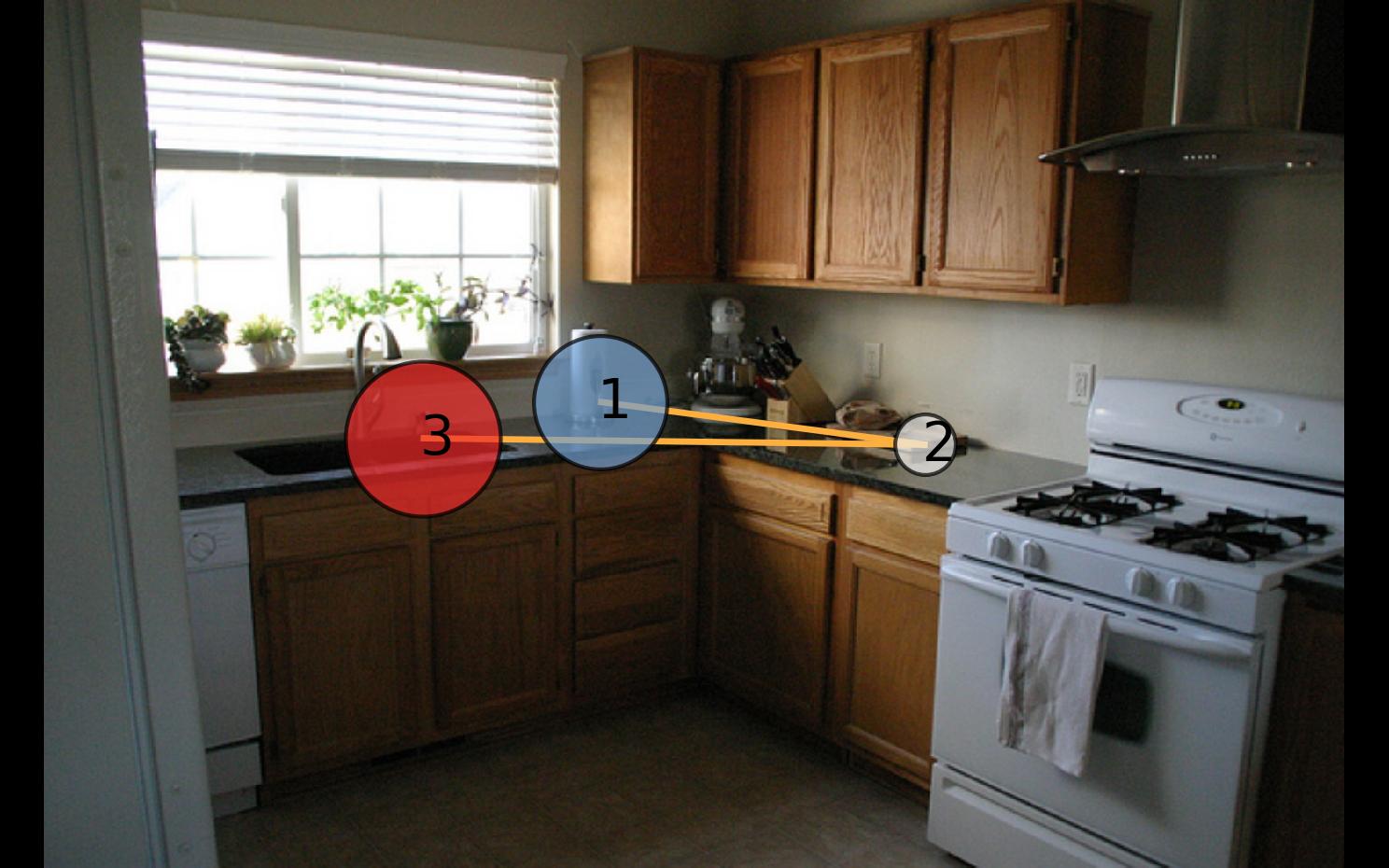} \\

         \addlinespace[0.6cm]
         
         & & \tiny ChenLSTM~\cite{chen2021predicting} & \tiny Gazeformer~\cite{mondal2023gazeformer} & \tiny GazeXplain~\cite{chen2024gazexplain} \\
         \rotatebox{90}{\parbox[t]{0.12\linewidth}{\hspace*{\fill}\tiny \textbf{Target:} \texttt{bowl}\hspace*{\fill}}} 
         & & \includegraphics[height=0.12\linewidth]{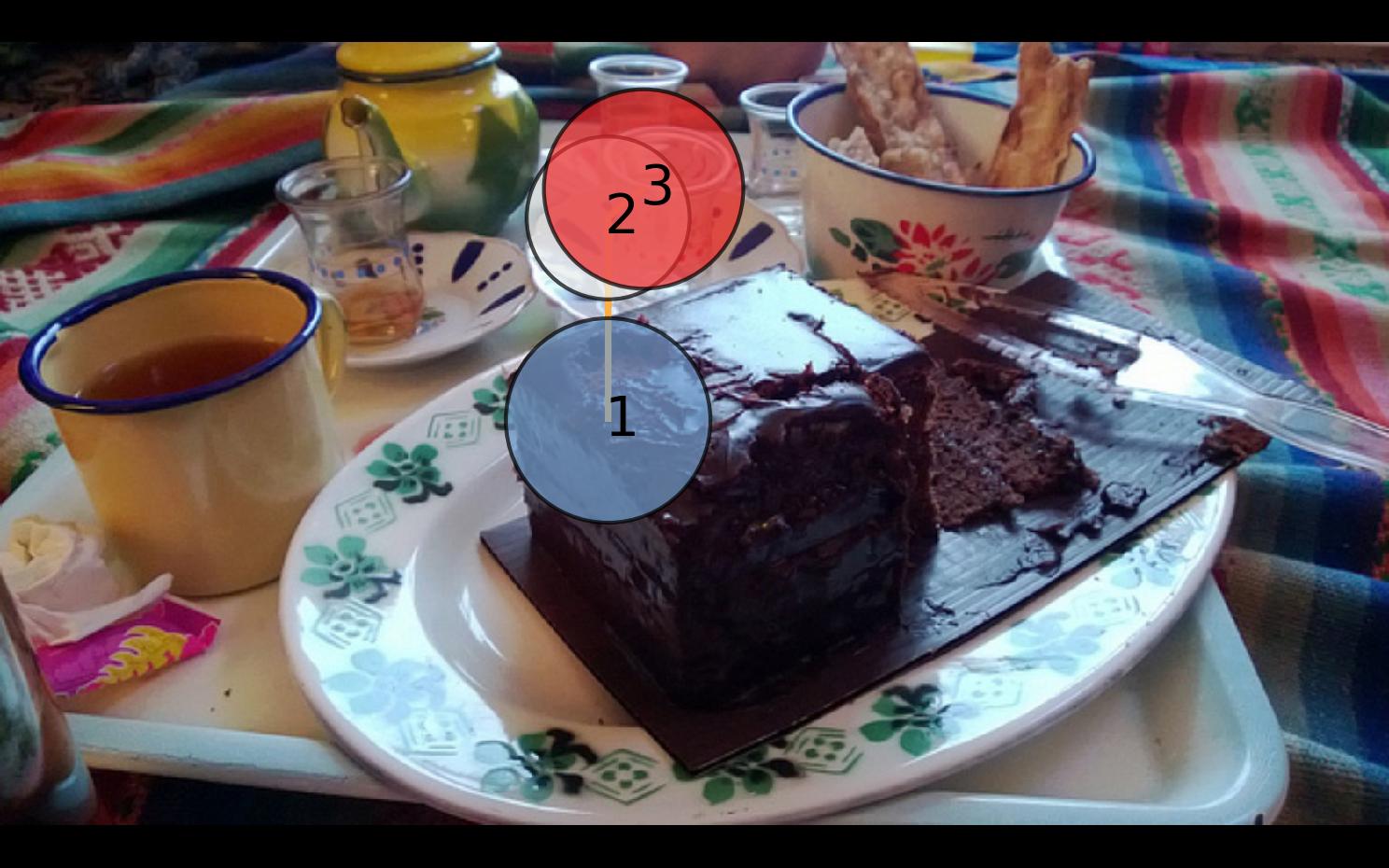} & 
         \includegraphics[height=0.12\linewidth]{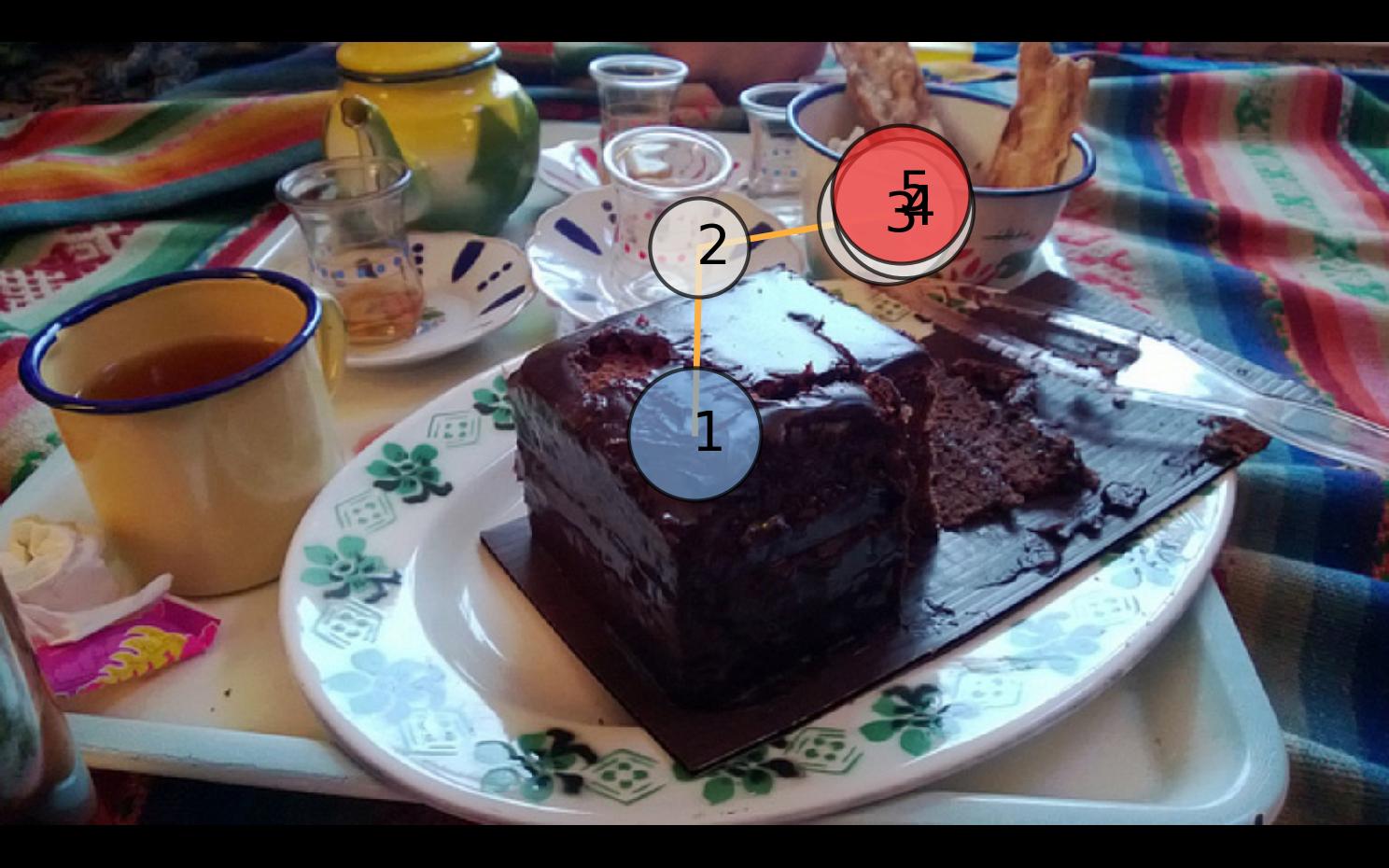} & 
         \includegraphics[height=0.12\linewidth]{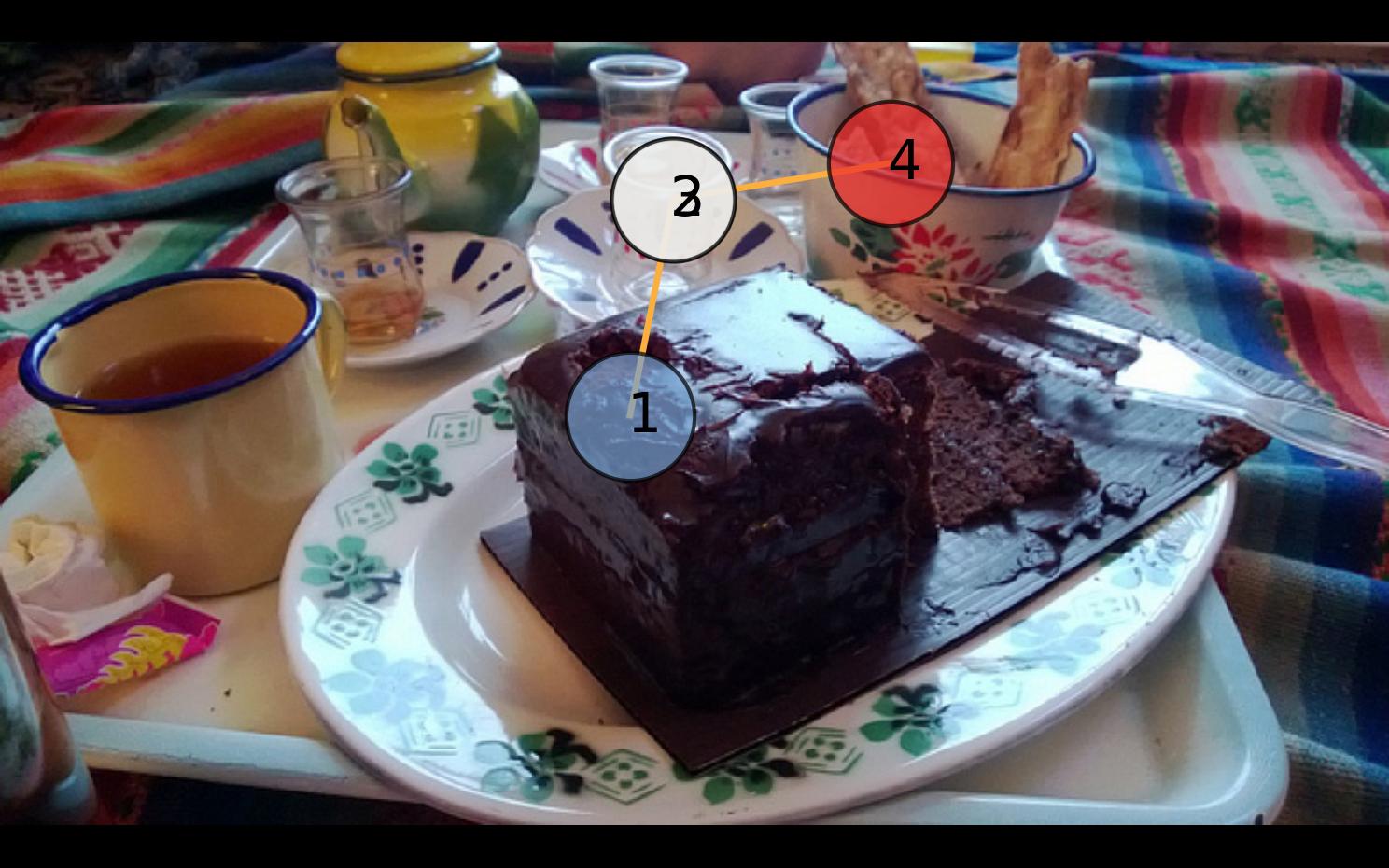} \\         
         & & \tiny TPP-Gaze~\cite{damelio2025tpp} & \tiny \textbf{\ours (Ours)} & \tiny Humans \\
         & & \includegraphics[height=0.12\linewidth]{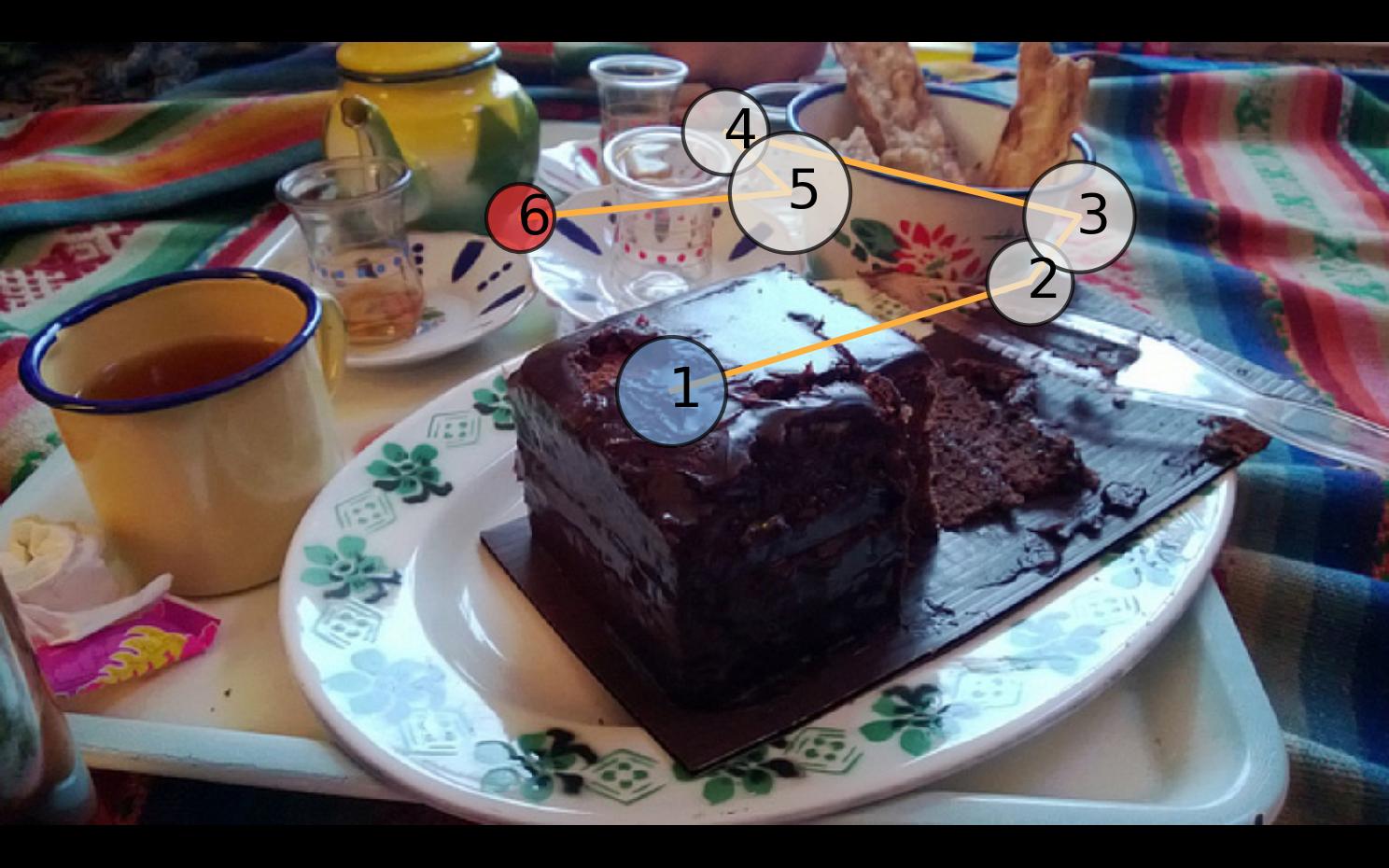} & 
         \includegraphics[height=0.12\linewidth]{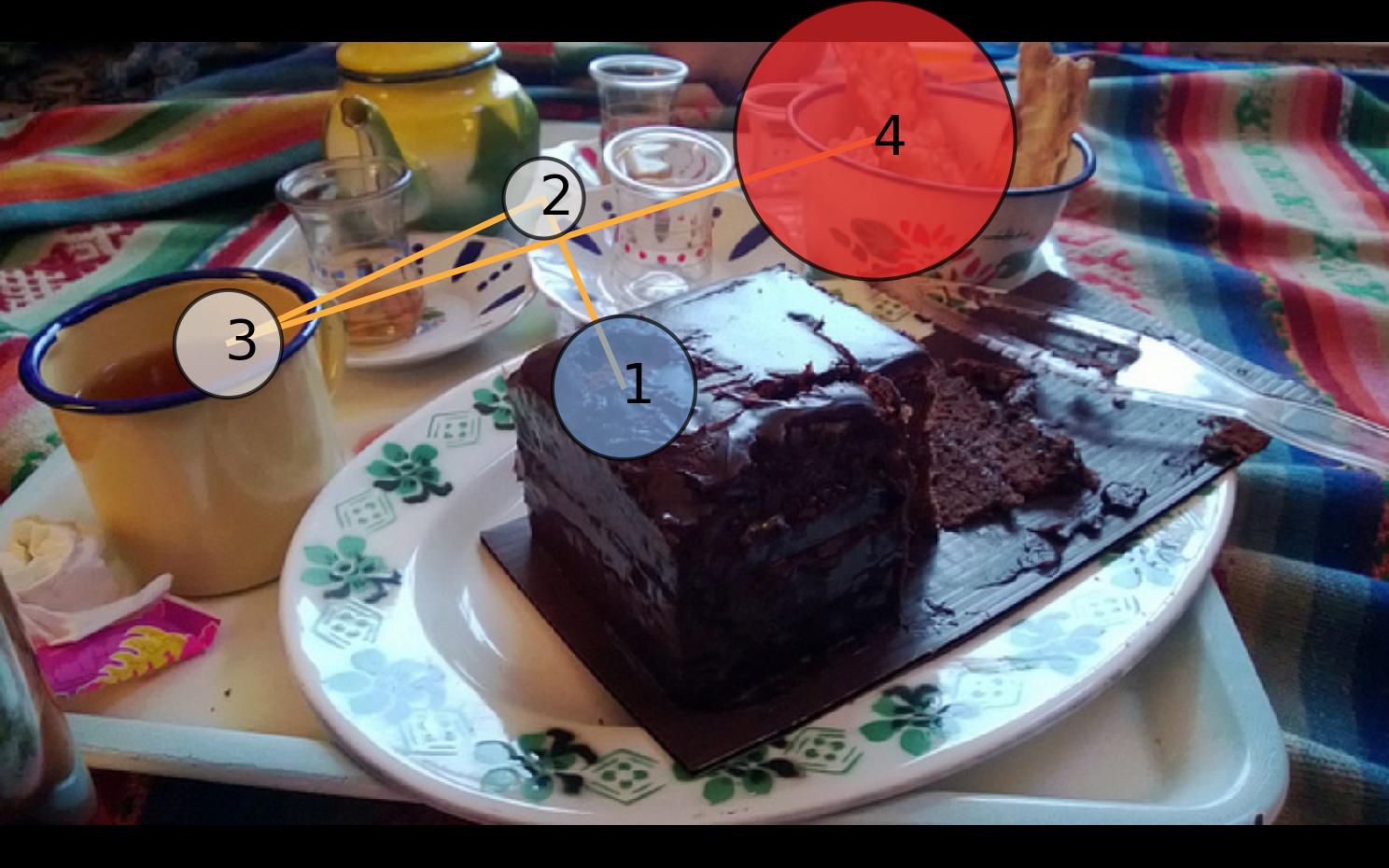} & 
         \includegraphics[height=0.12\linewidth]{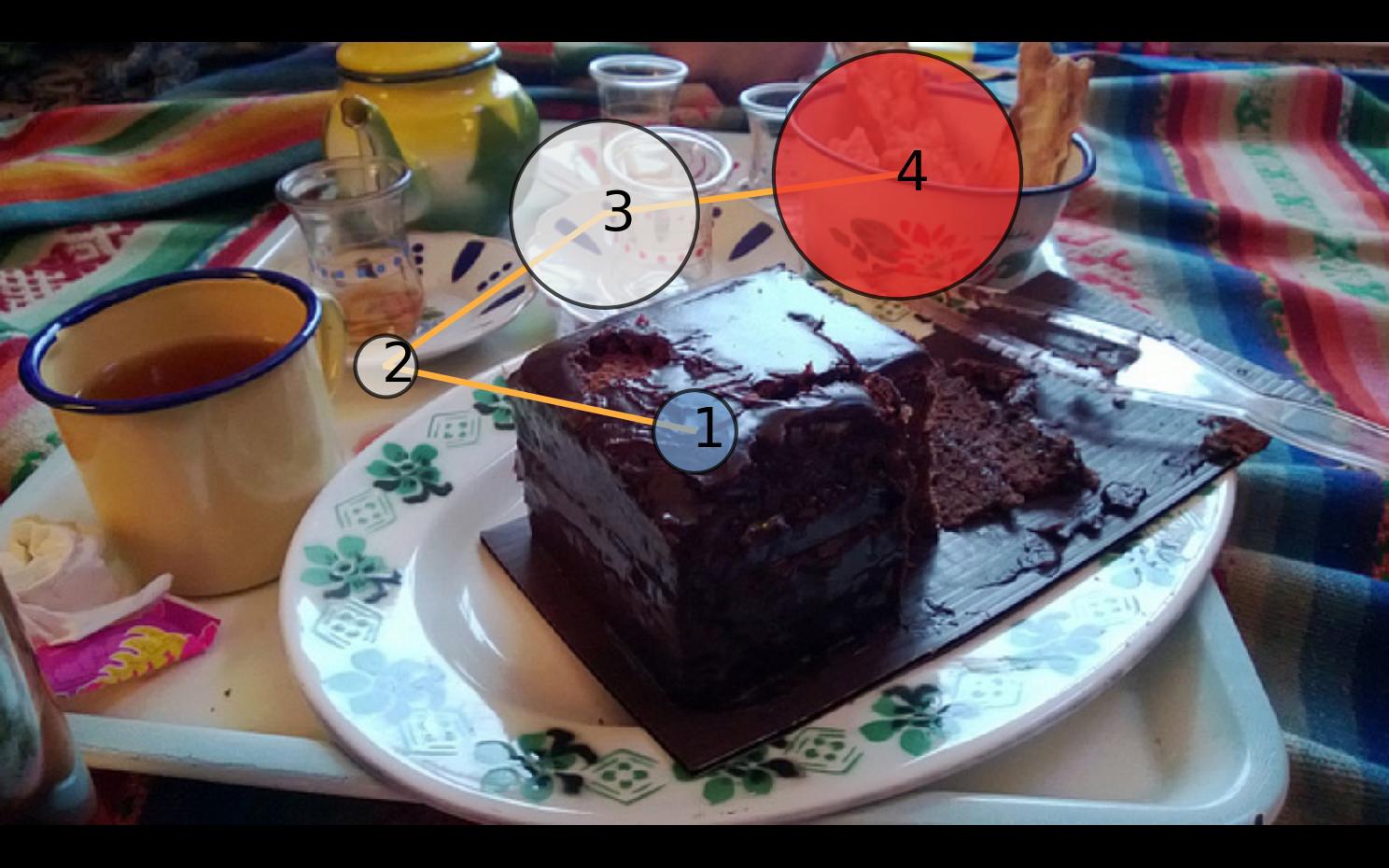} \\
         
    \end{tabular}
    }
    \vspace{-0.15cm}
    \caption{Qualitative comparison of simulated and human scanpaths on the COCO-Search18 (TP) dataset for the visual search task.}
    \label{fig:qualitatives_COCOSearch_TP}
    \vspace{-0.4cm}
\end{figure*}

\vspace{-0.1cm}
Additional qualitative results are depicted from \cref{fig:qualitatives_COCOFV} to \cref{fig:qualitatives_COCOSearch_TP} on COCO-FreeView~\cite{yang2020predicting}, MIT1003~\cite{judd2009learning}, and OSIE~\cite{xu2014predicting} for free-viewing and COCO-Search18~\cite{chen2021coco} for the visual search task, respectively.
The qualitative results support the findings of the main paper, highlighting the accuracy of \scandiff in predicting human-like scanpaths. Other methods, however, demonstrate limitations by either focusing excessively on specific elements or producing shorter scanpaths than the ones exhibited by humans. 

\begin{figure*}[t]
    \footnotesize
    \setlength{\tabcolsep}{.1em}
    \resizebox{\linewidth}{!}{
    \begin{tabular}{cccccc}
        \scriptsize IOR-ROI-LSTM~\cite{chen2018scanpath}  & \scriptsize ChenLSTM~\cite{chen2021predicting} & \scriptsize GazeXplain~\cite{chen2024gazexplain} & \scriptsize TPP-Gaze~\cite{damelio2025tpp} & \scriptsize \ours (Ours) & \scriptsize Humans \\
         \includegraphics[width=0.14\linewidth]{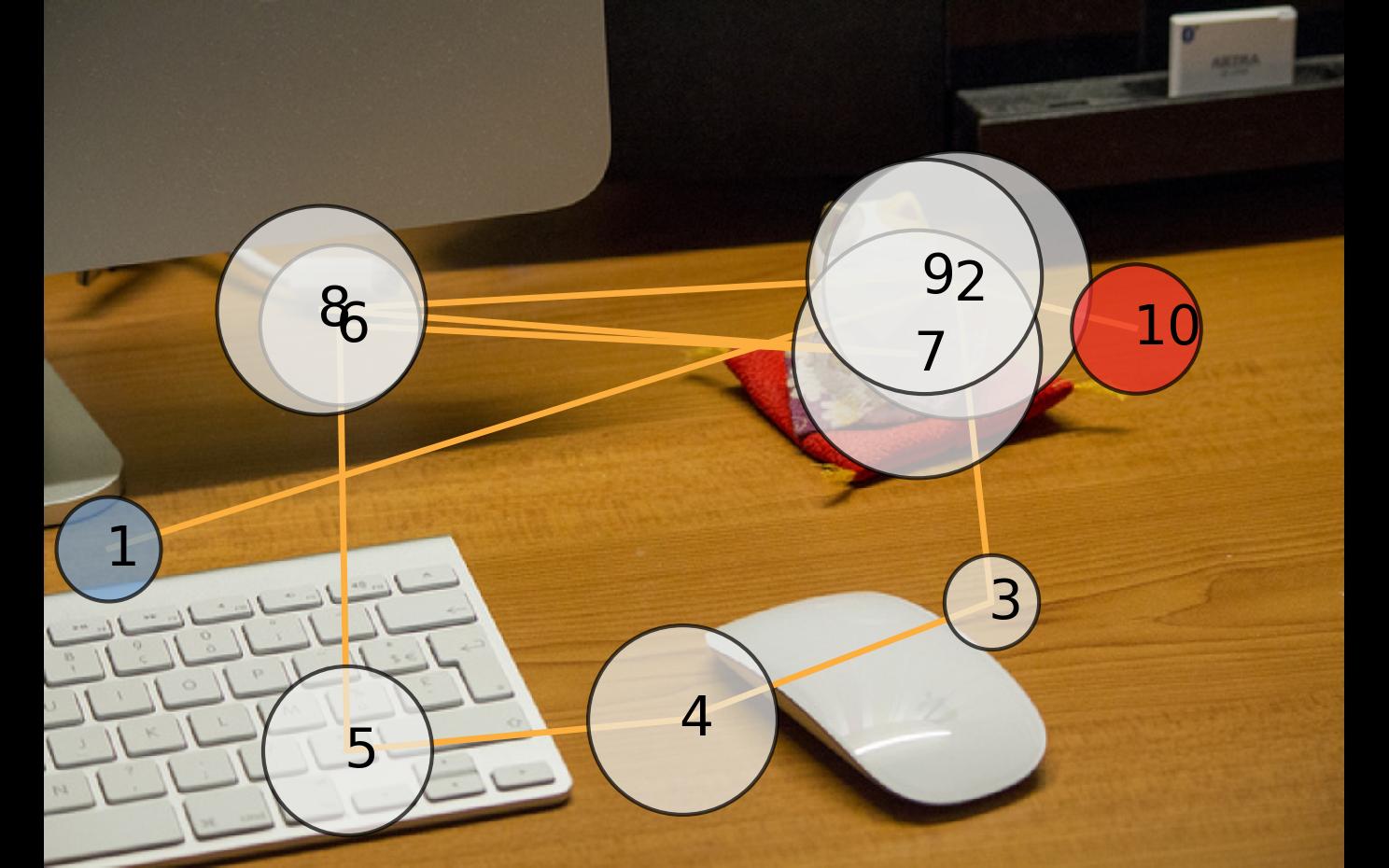} & \includegraphics[width=0.14\linewidth]{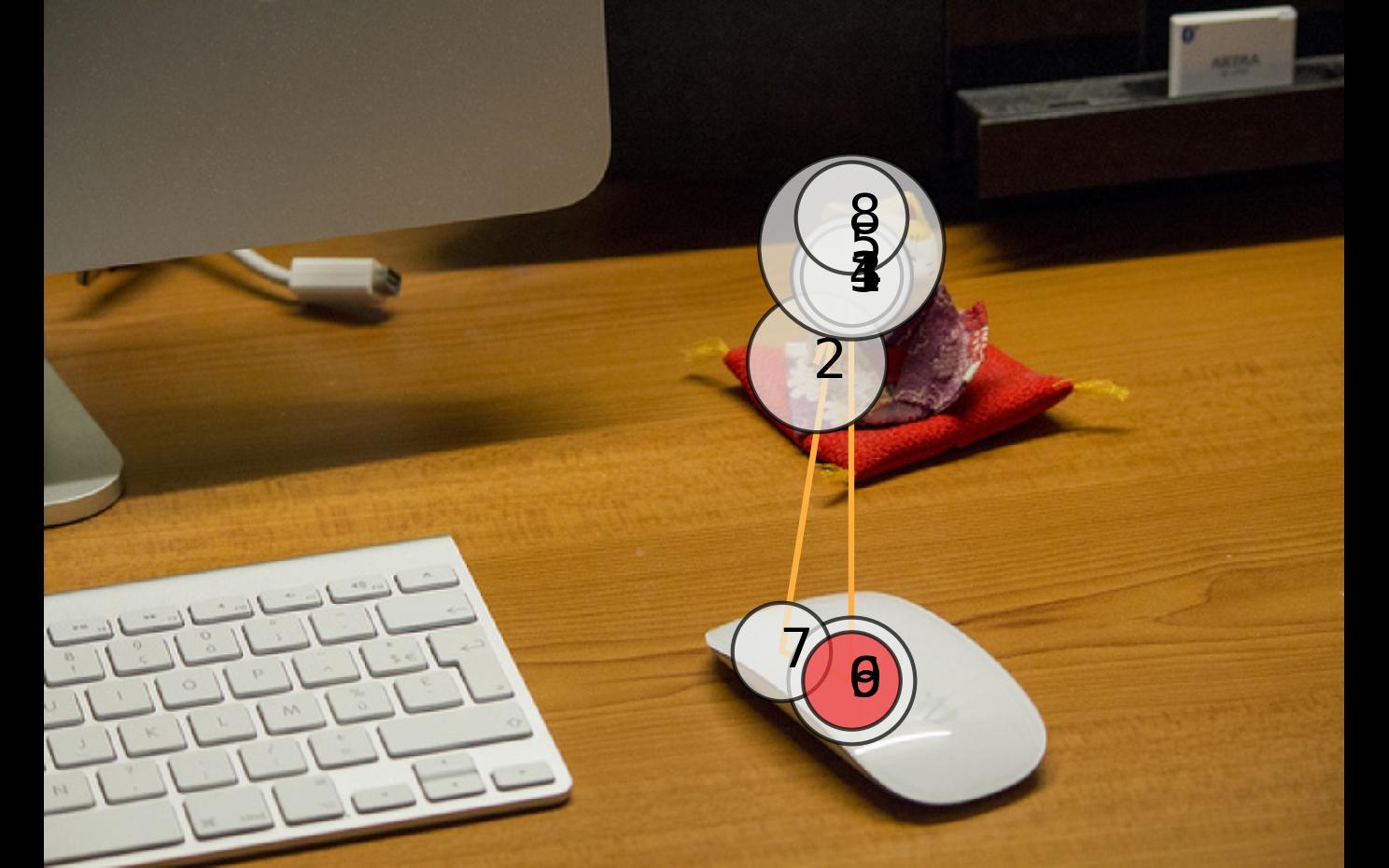} & \includegraphics[width=0.14\linewidth]{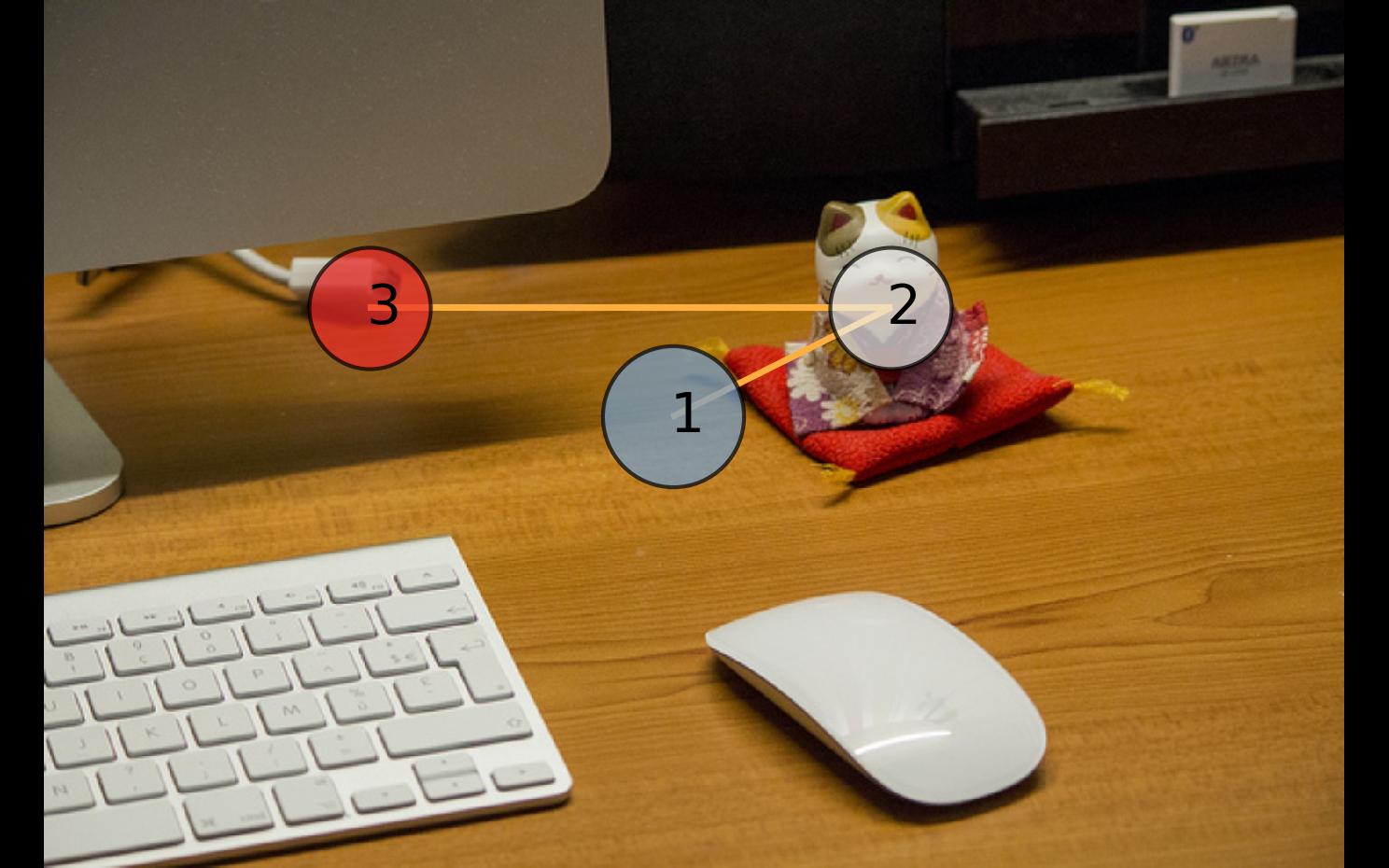} & \includegraphics[width=0.14\linewidth]{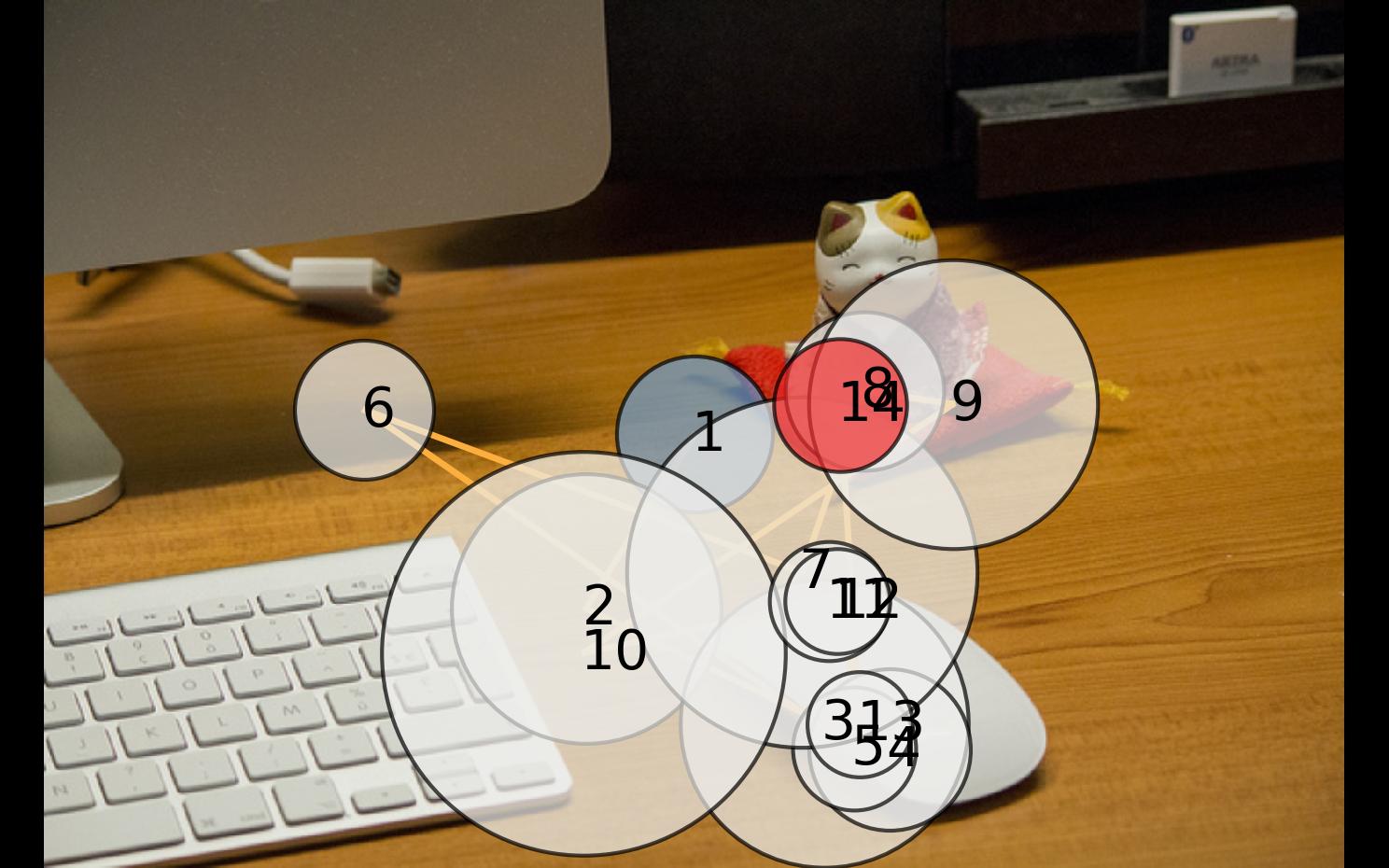} & \includegraphics[width=0.14\linewidth]{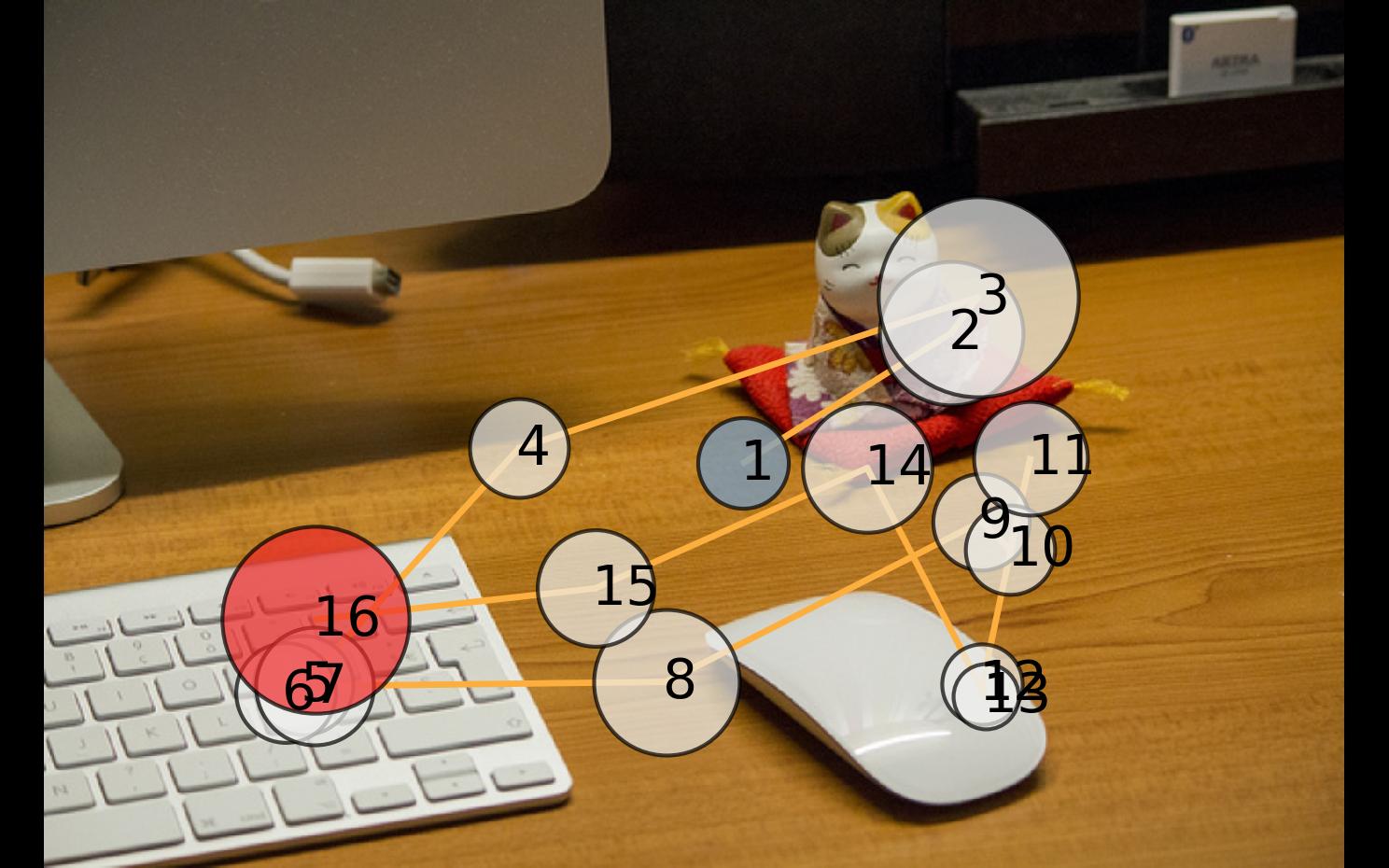} & \includegraphics[width=0.14\linewidth]{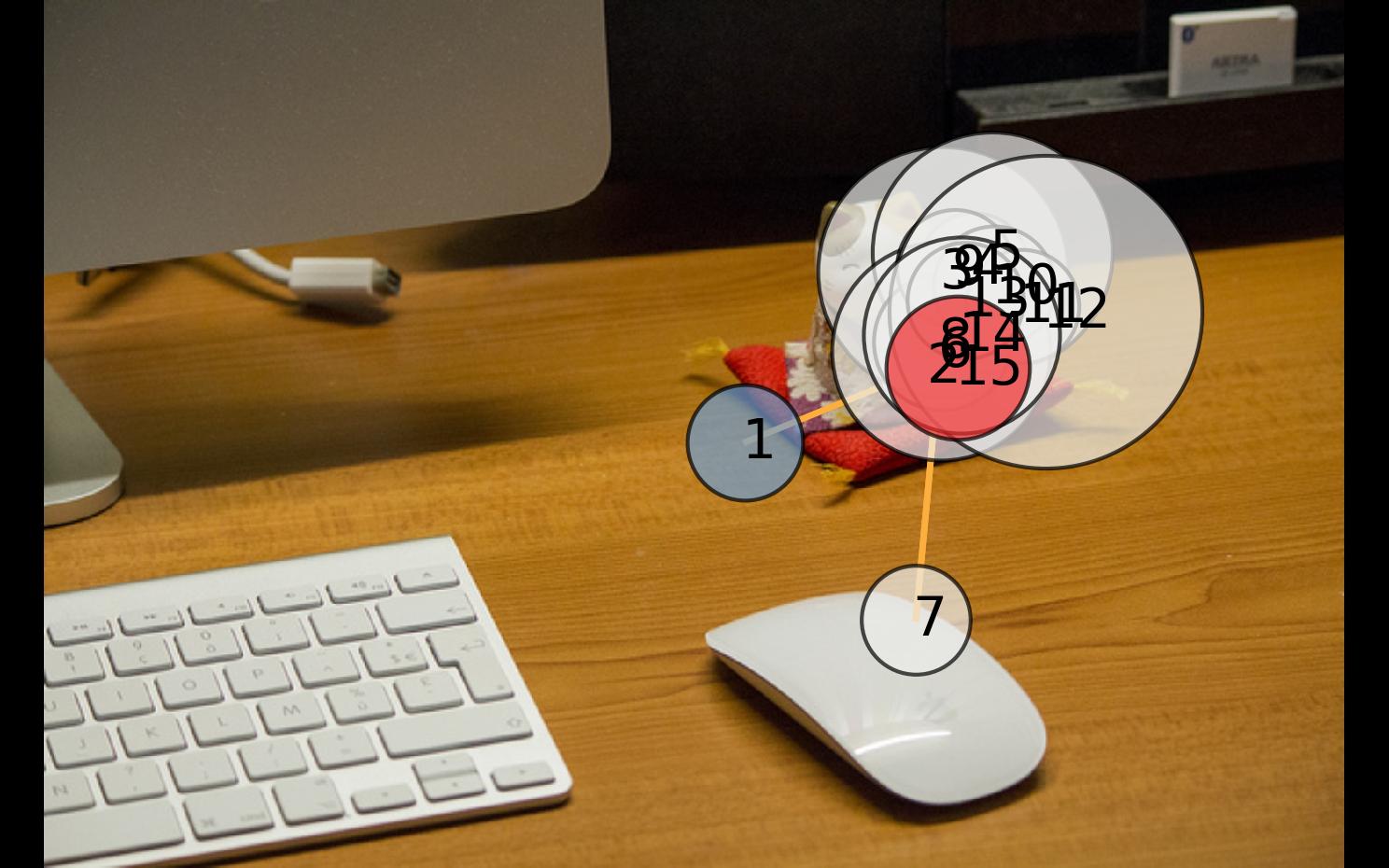} \\
         \includegraphics[width=0.14\linewidth]{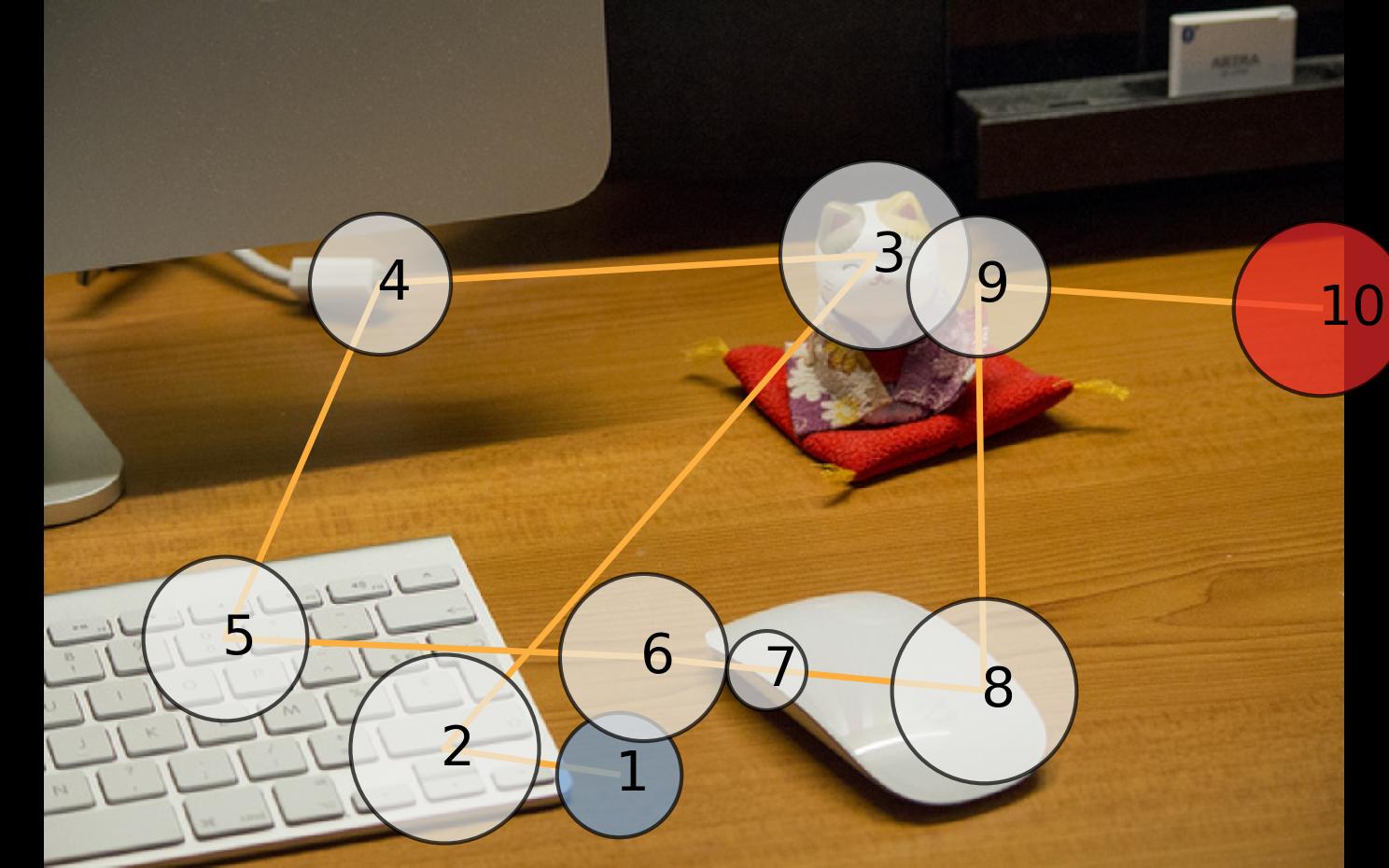} & \includegraphics[width=0.14\linewidth]{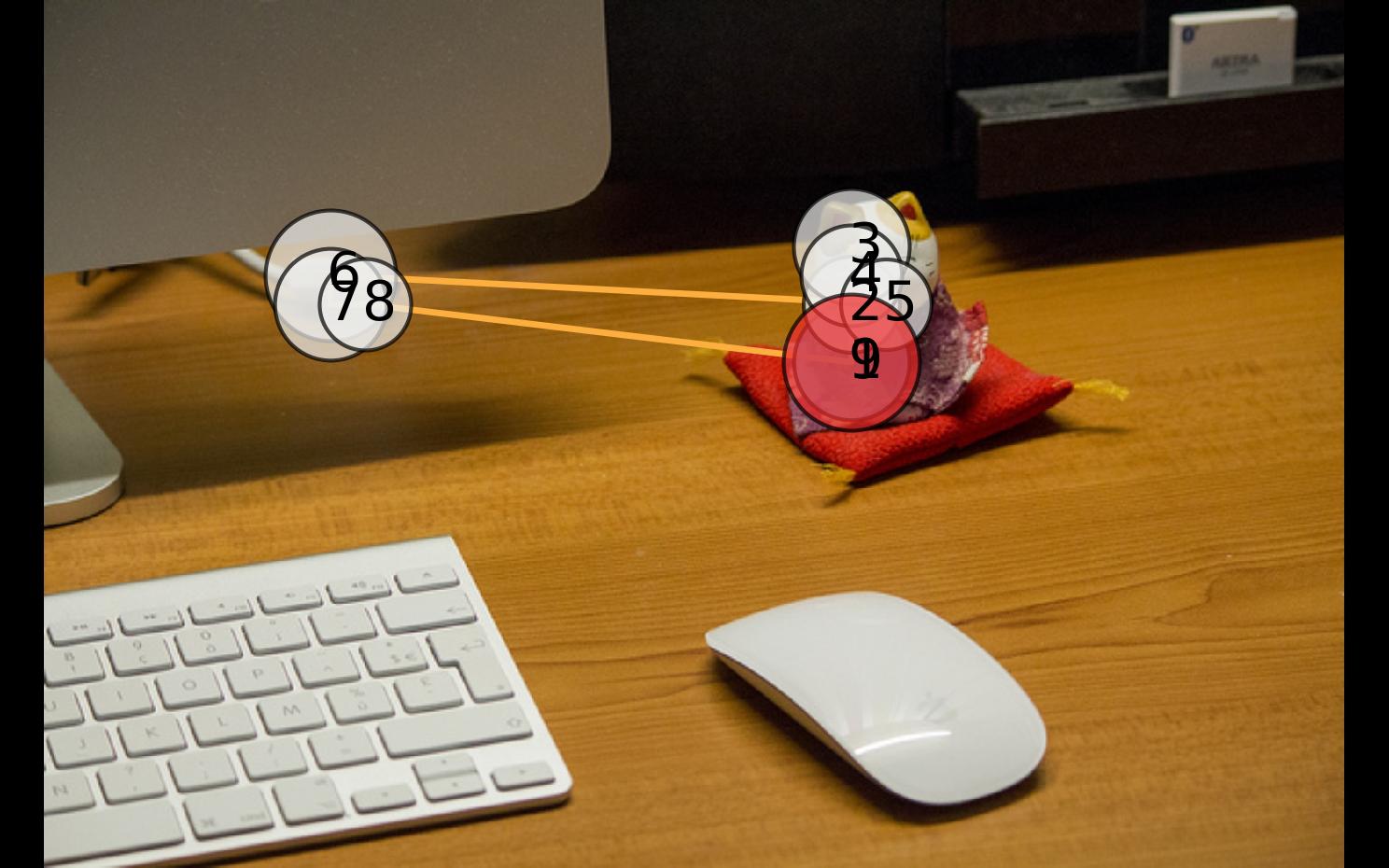} & \includegraphics[width=0.14\linewidth]{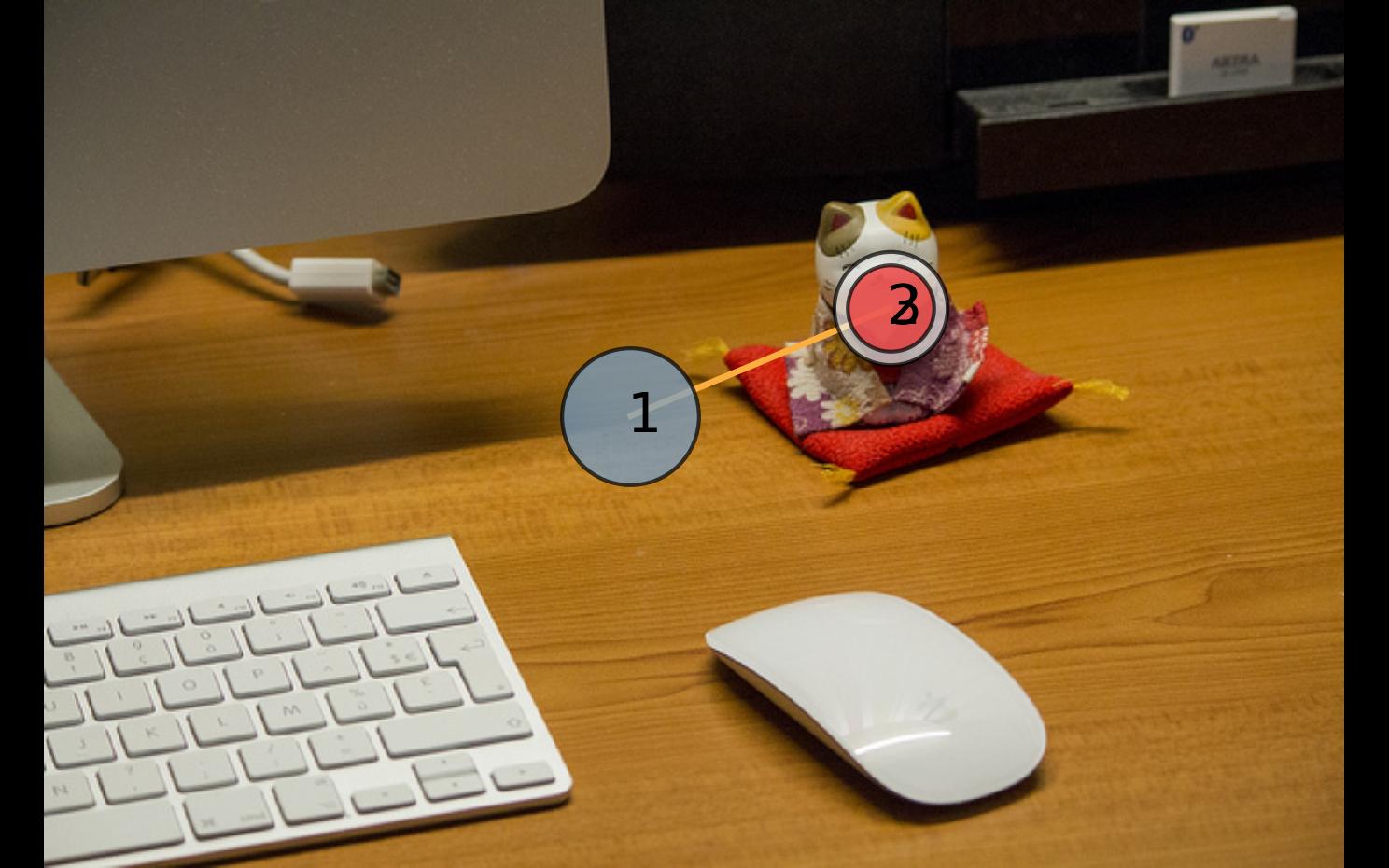} & \includegraphics[width=0.14\linewidth]{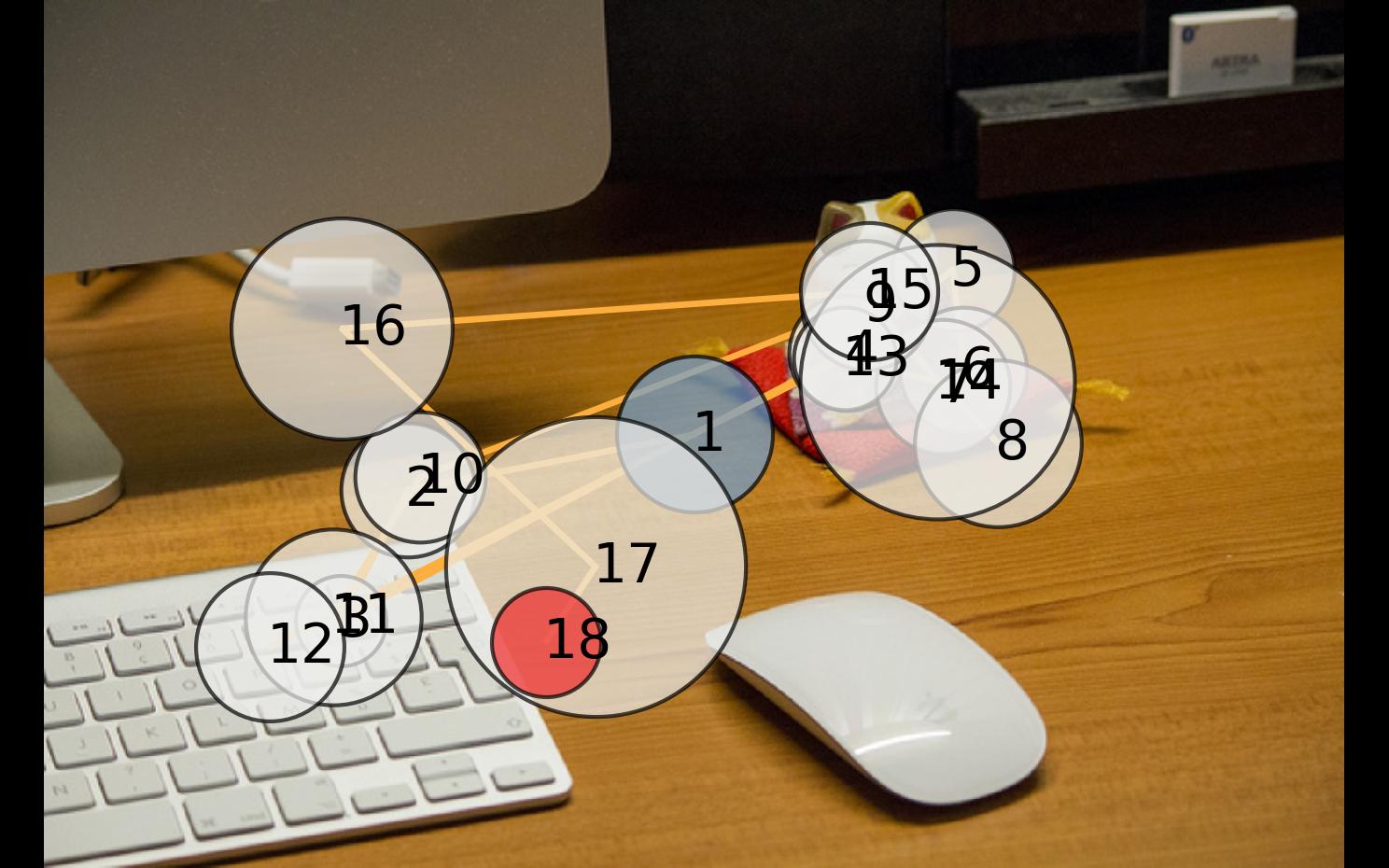} & \includegraphics[width=0.14\linewidth]{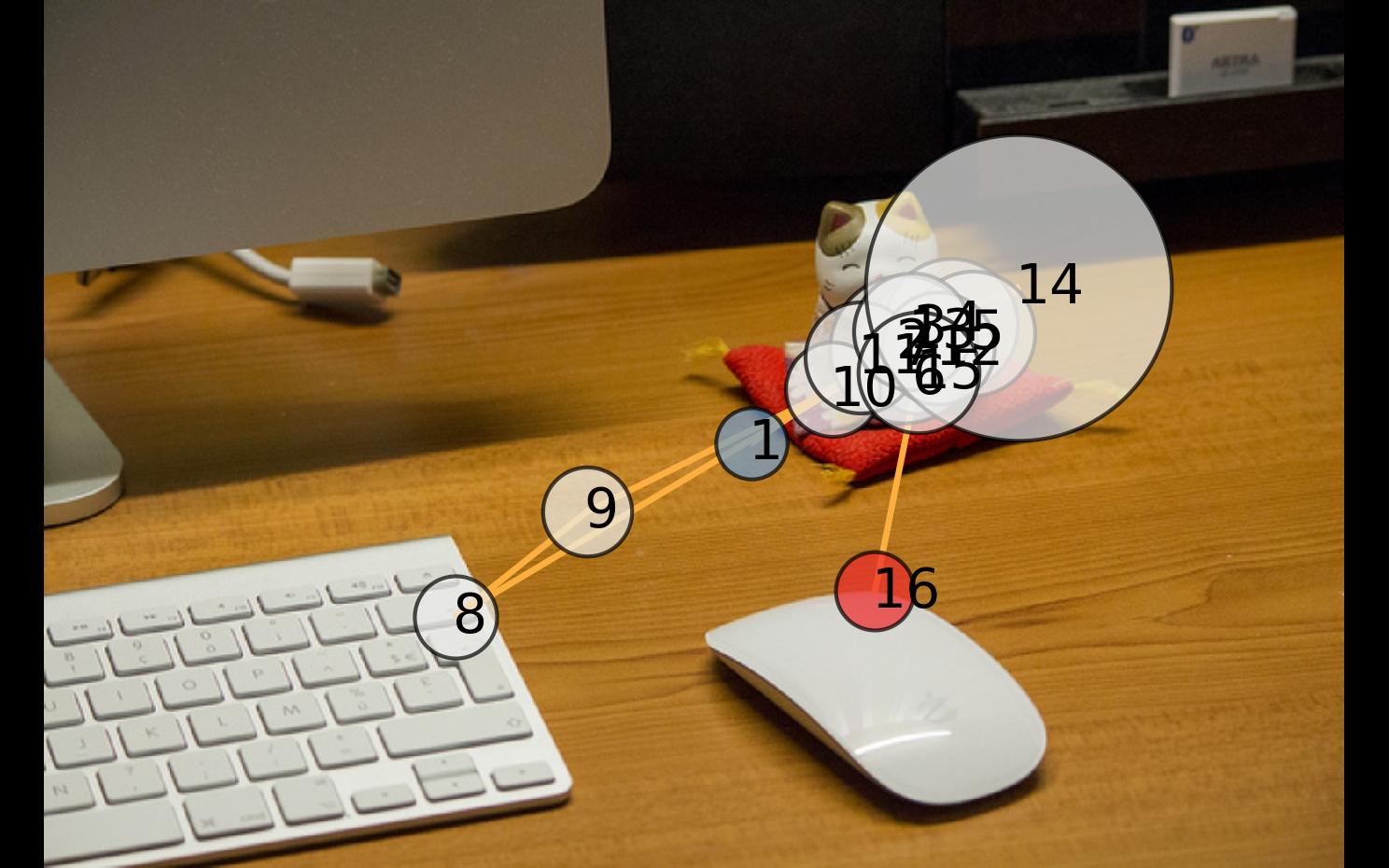} & \includegraphics[width=0.14\linewidth]{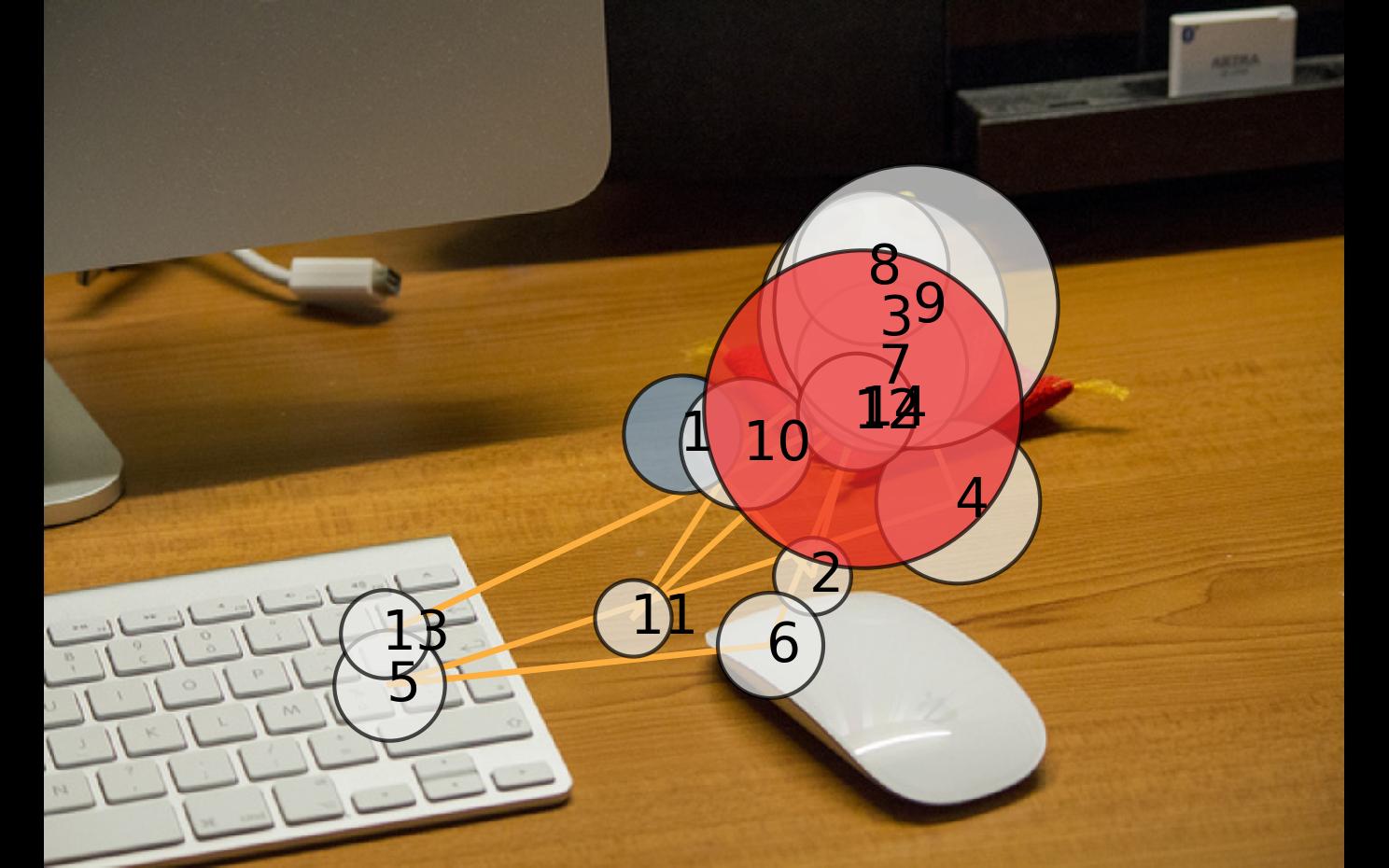} \\
         \includegraphics[width=0.14\linewidth]{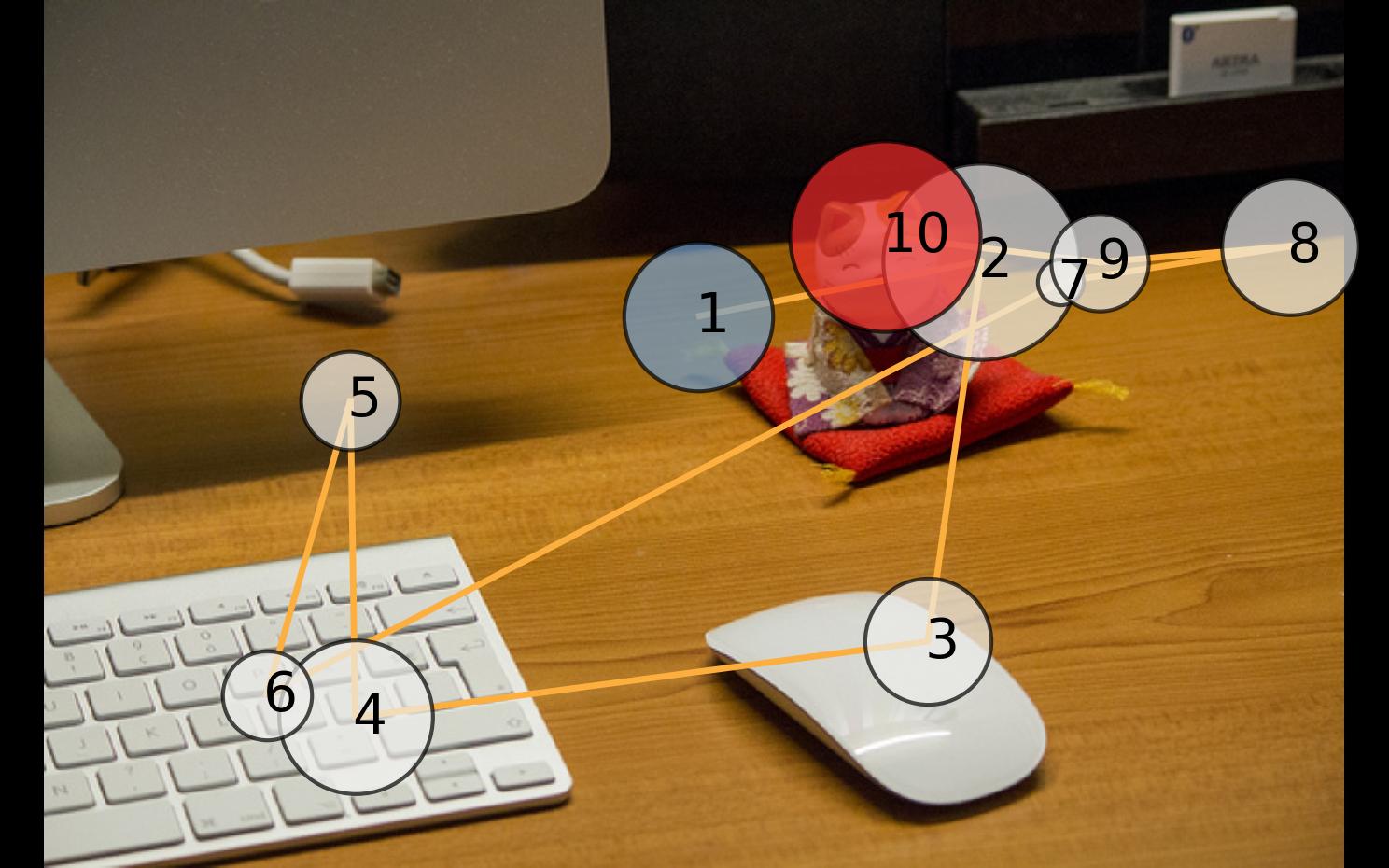} & \includegraphics[width=0.14\linewidth]{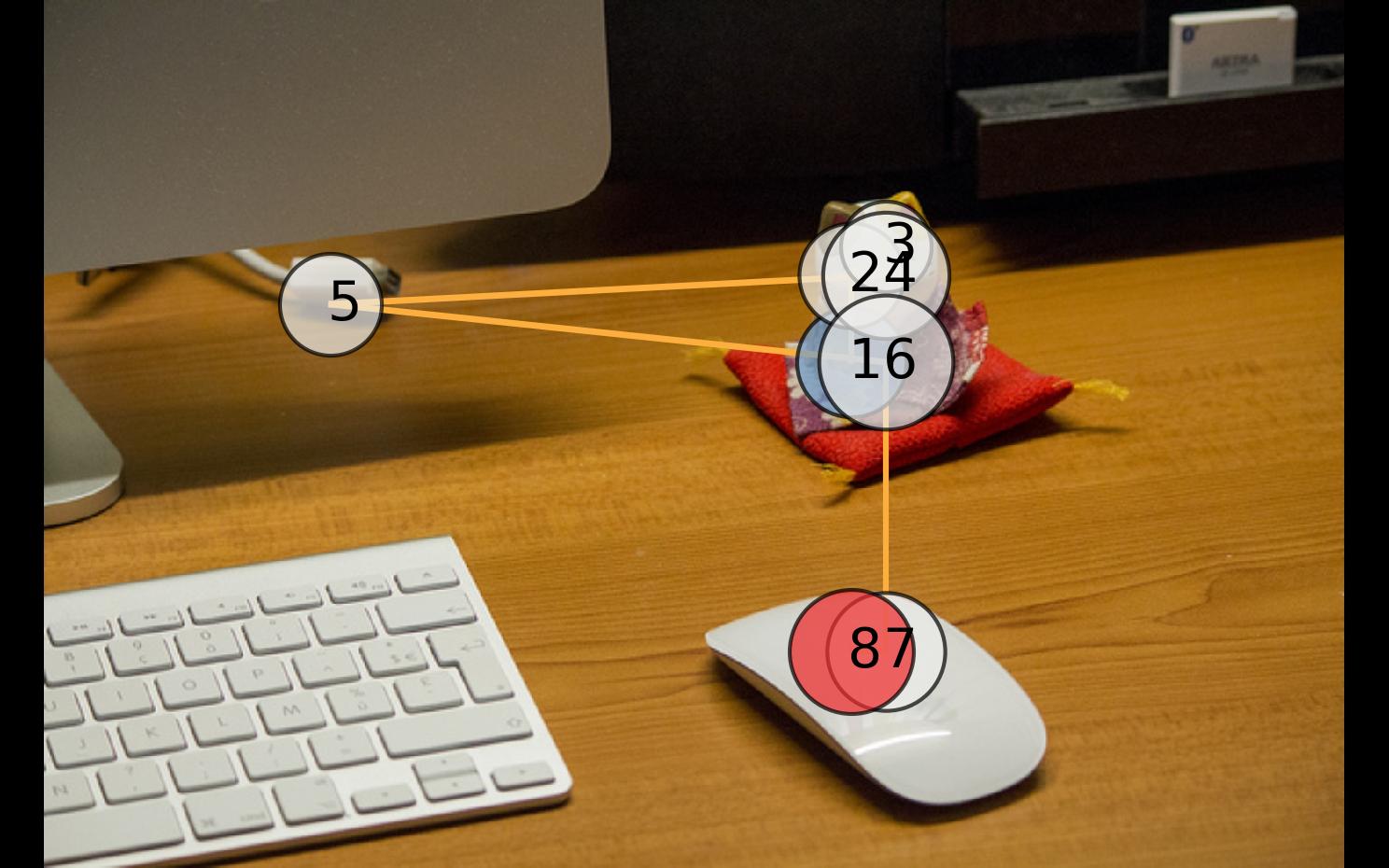} & \includegraphics[width=0.14\linewidth]{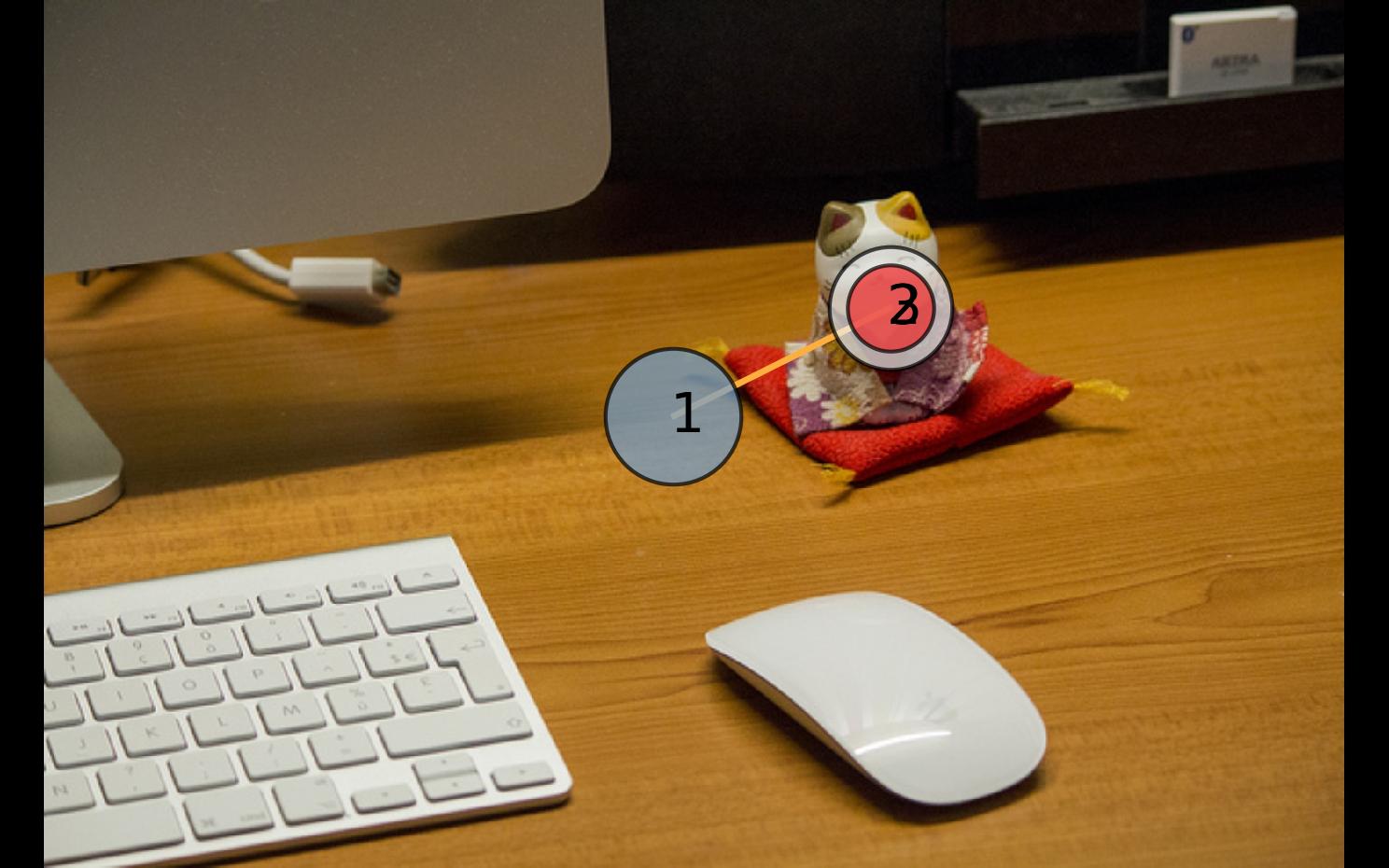} & \includegraphics[width=0.14\linewidth]{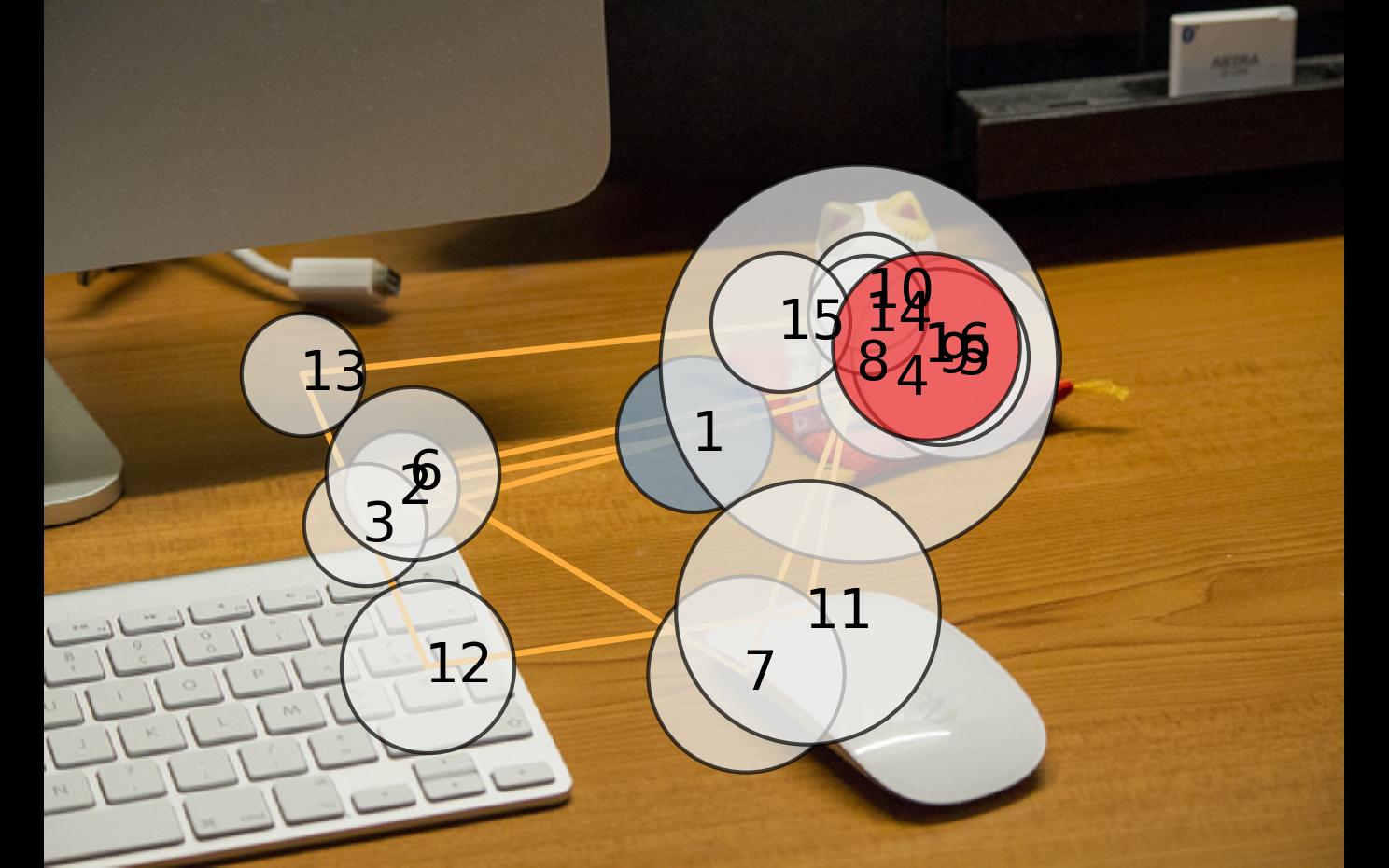} & \includegraphics[width=0.14\linewidth]{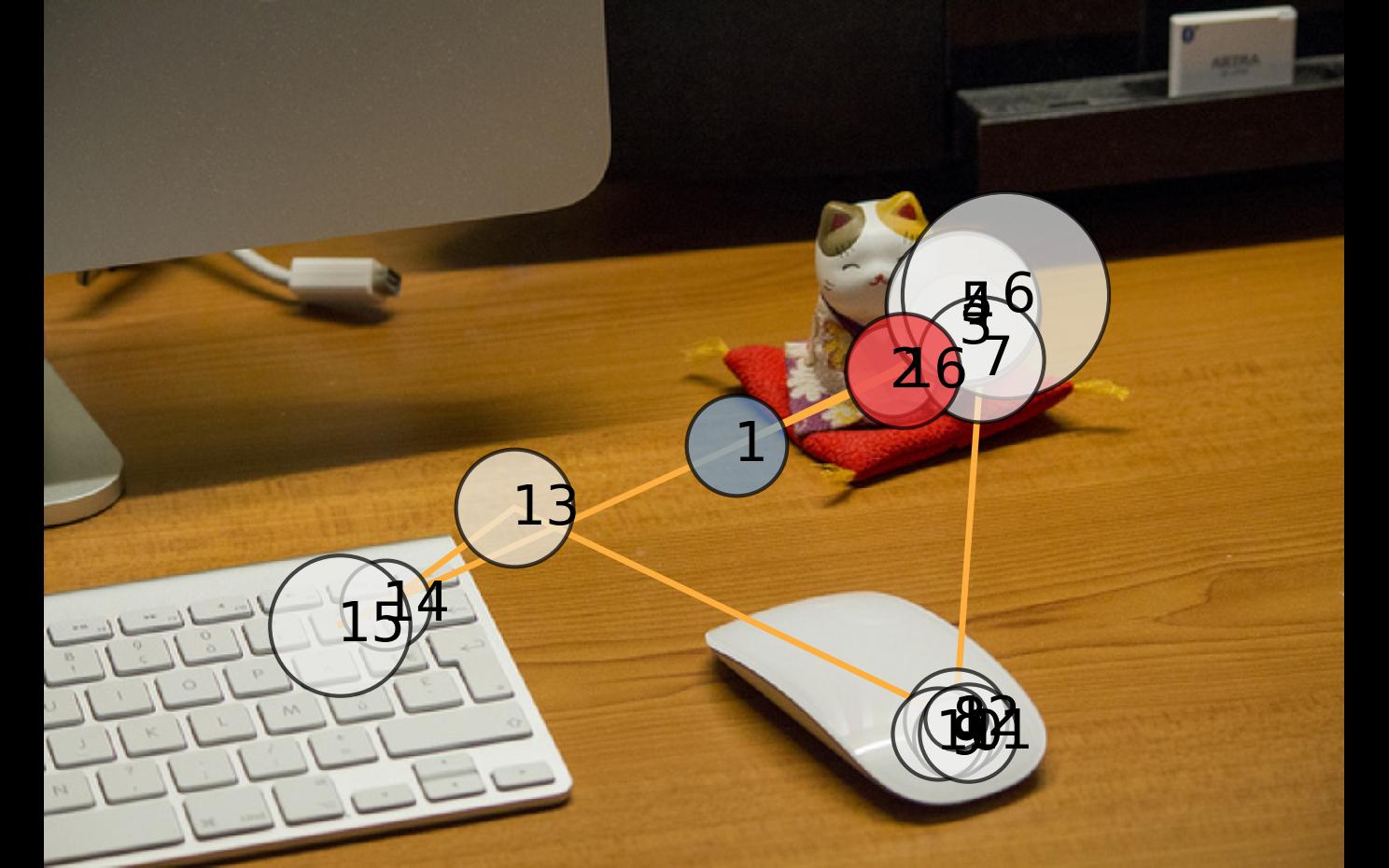} & \includegraphics[width=0.14\linewidth]{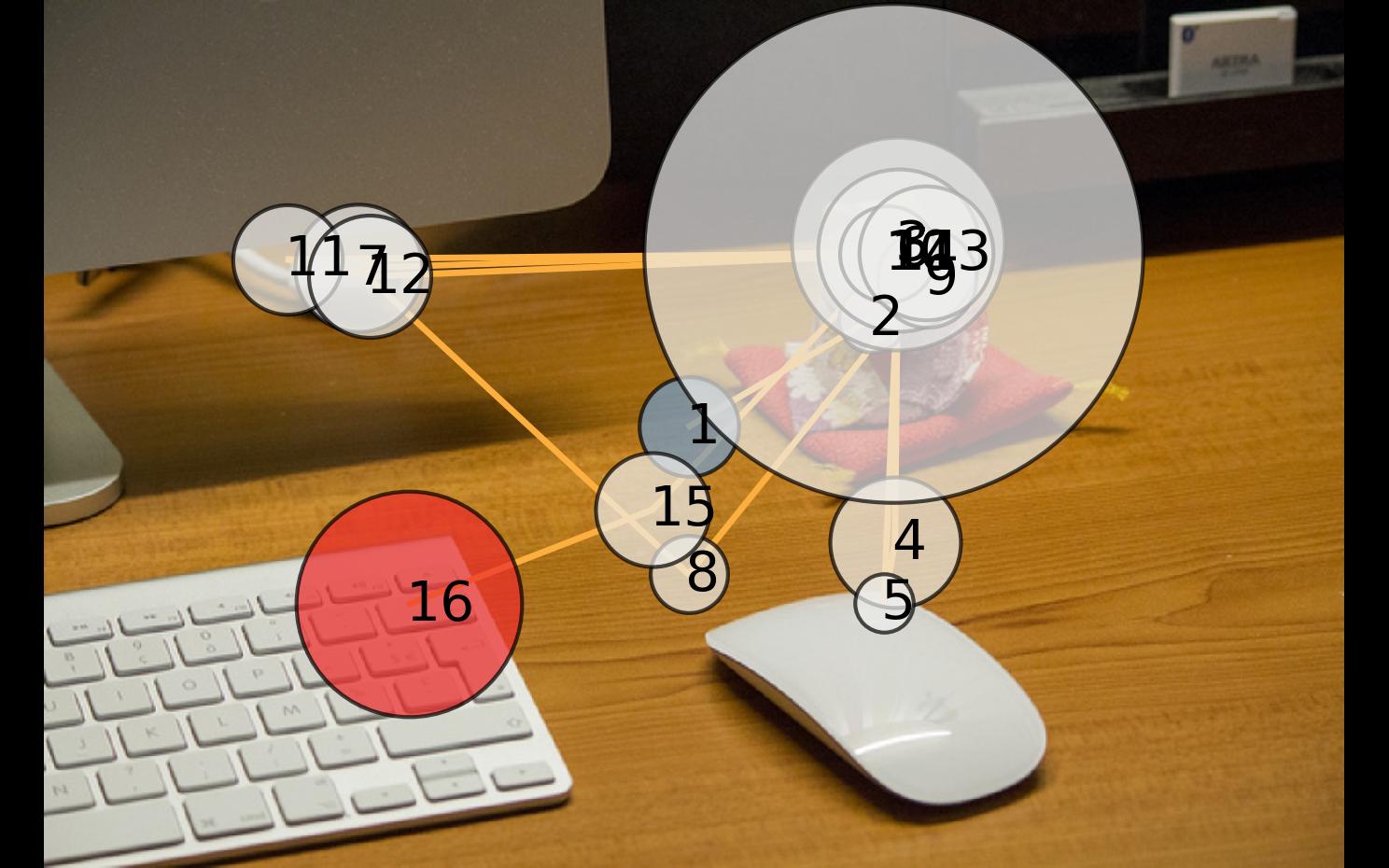} \\
        \includegraphics[width=0.14\linewidth]{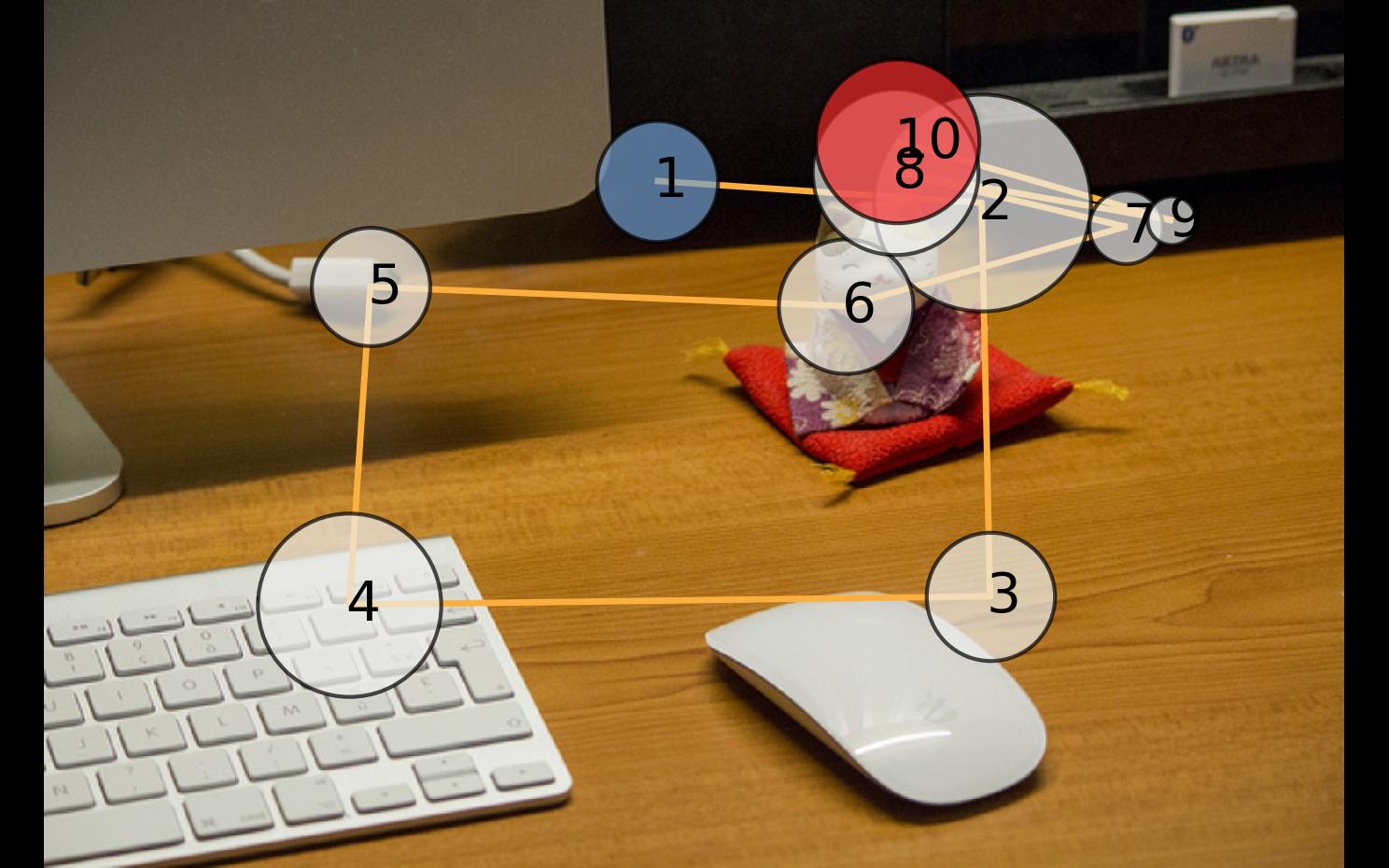} & \includegraphics[width=0.14\linewidth]{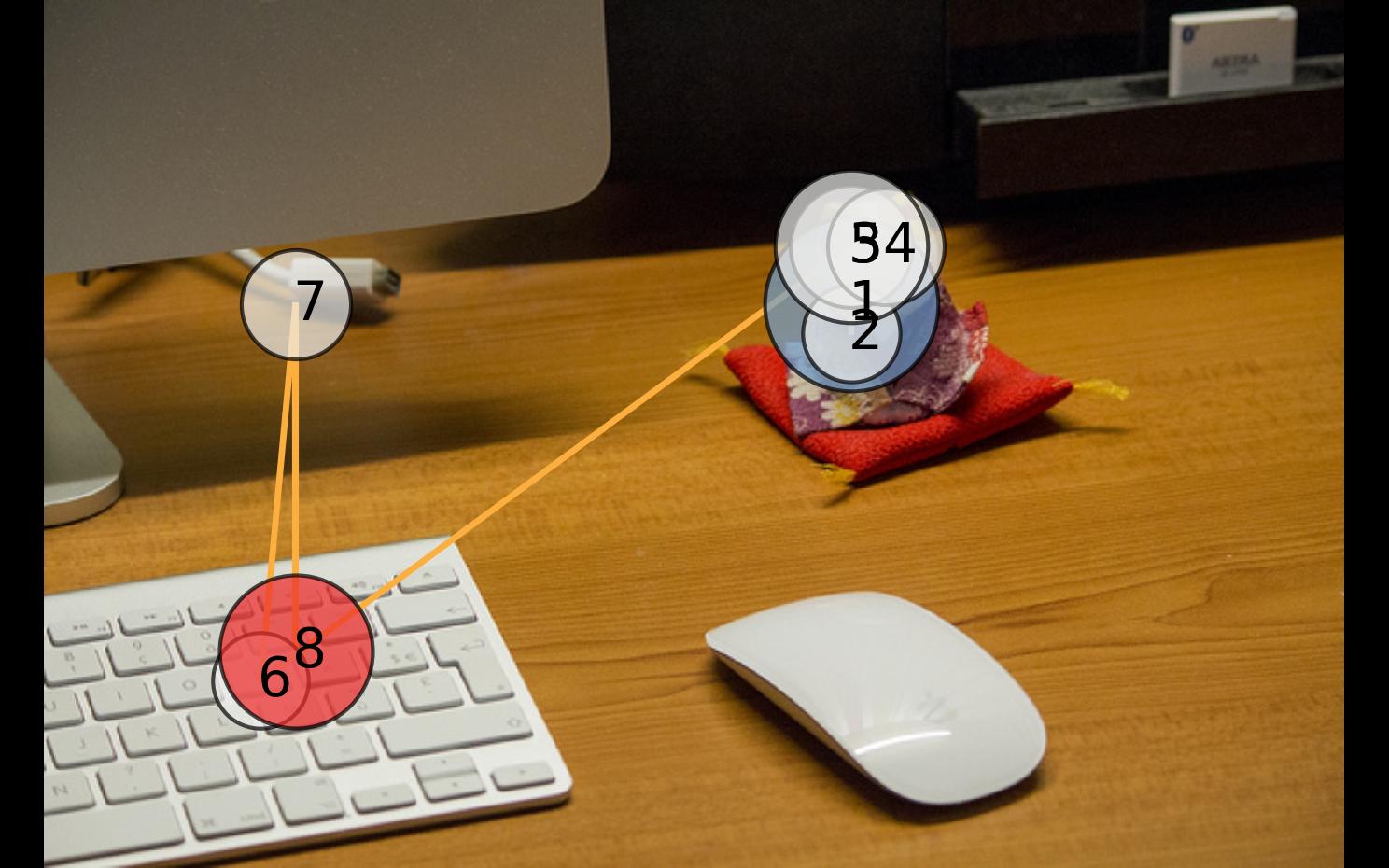} & \includegraphics[width=0.14\linewidth]{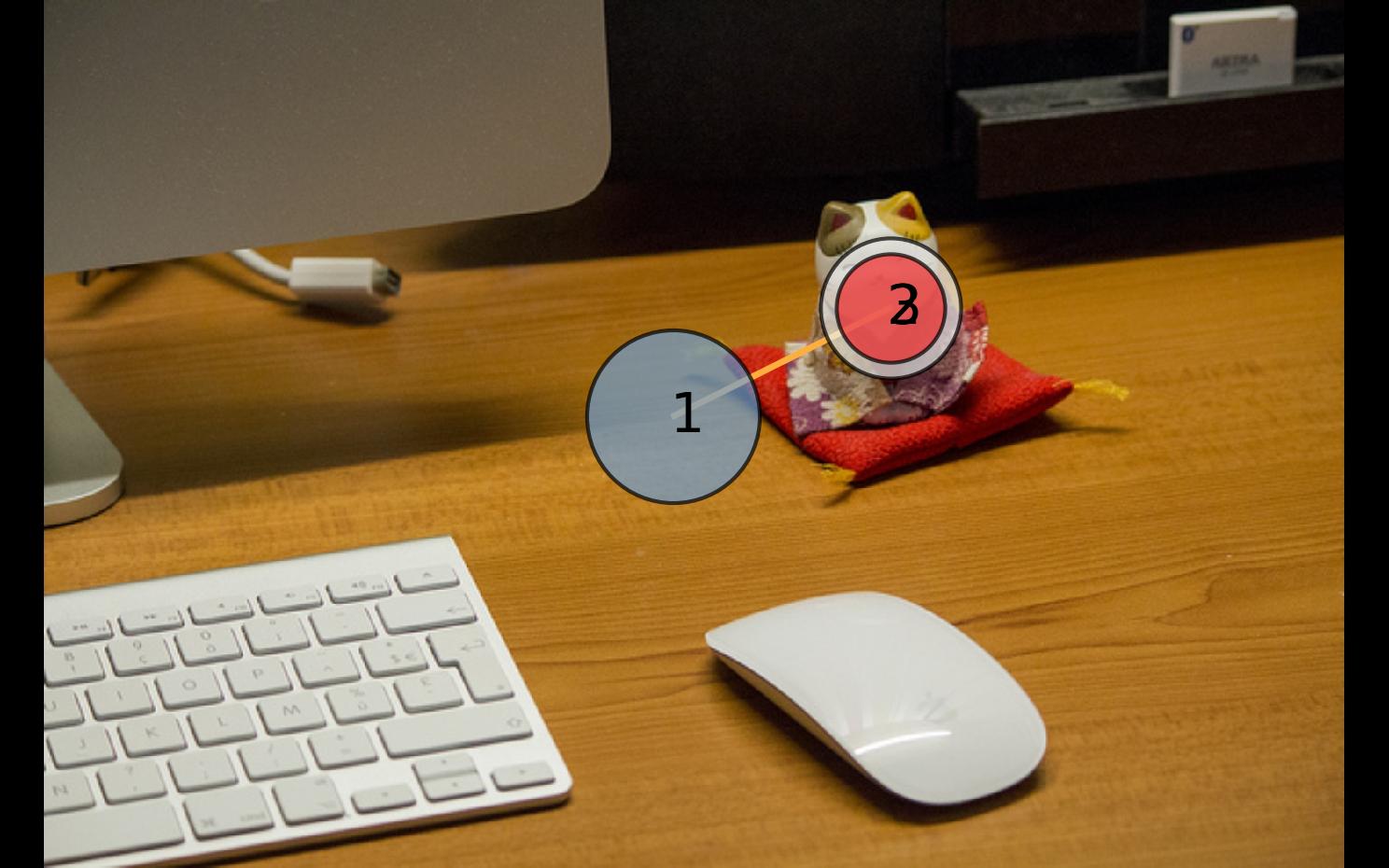} & \includegraphics[width=0.14\linewidth]{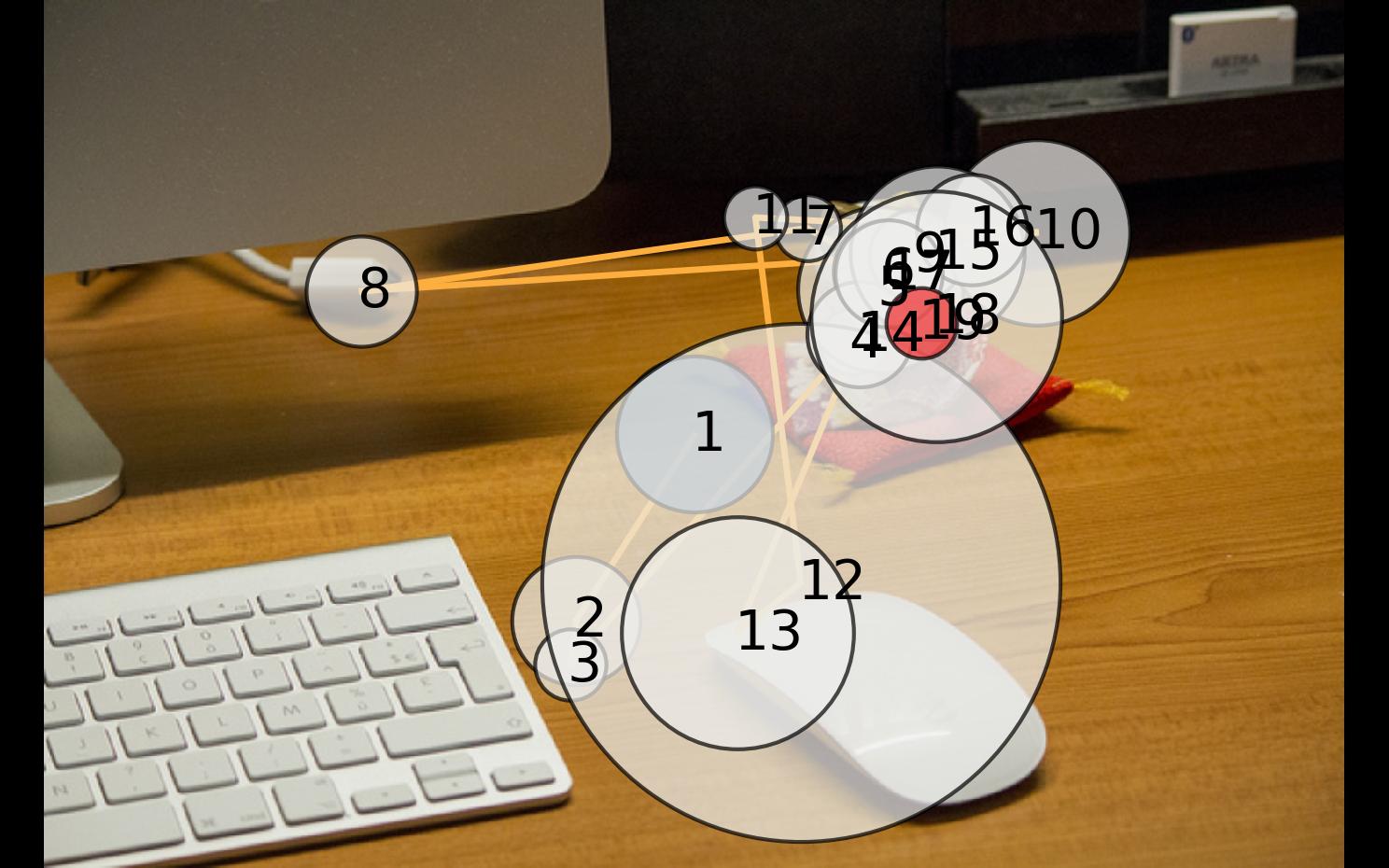} & \includegraphics[width=0.14\linewidth]{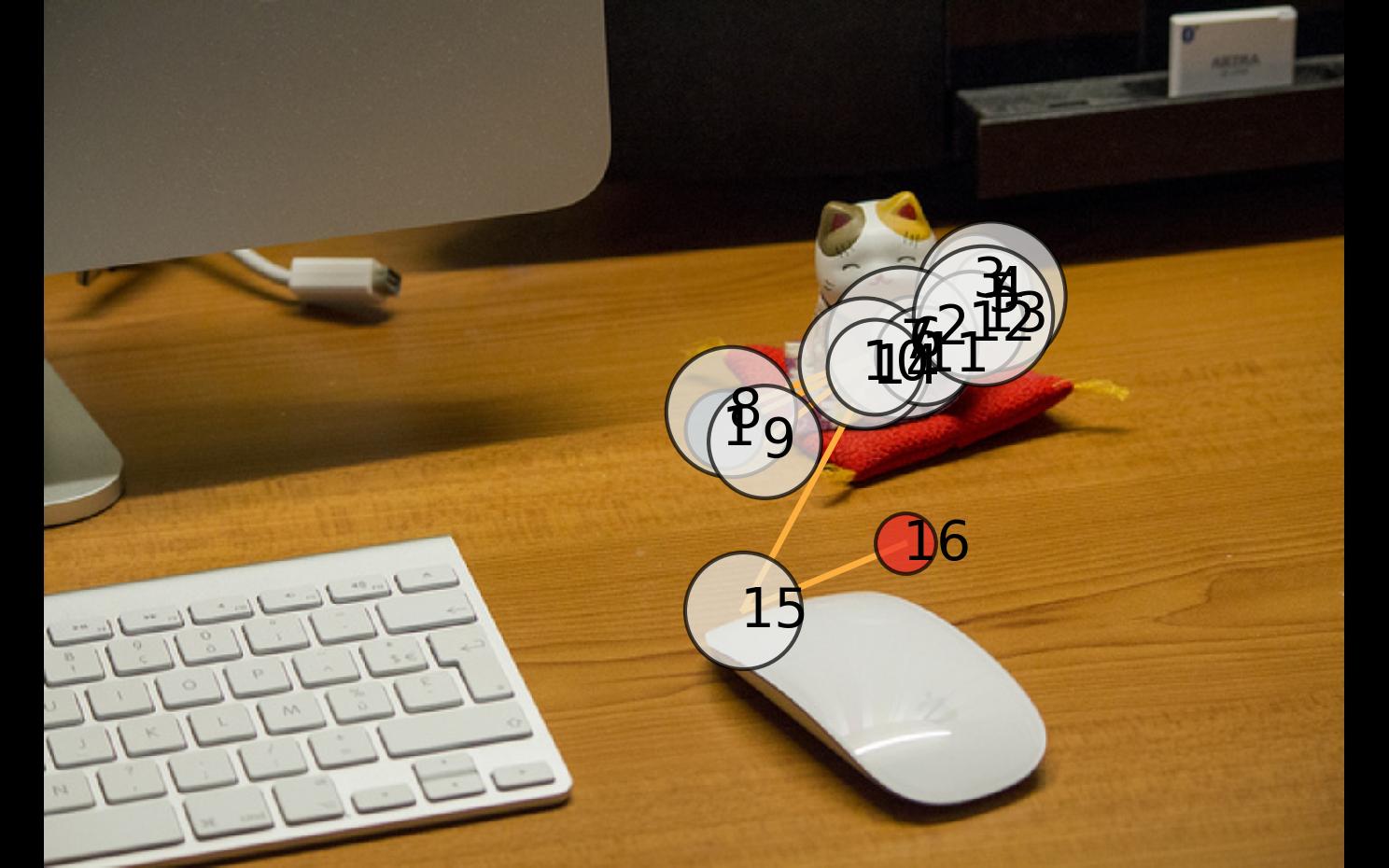} & \includegraphics[width=0.14\linewidth]{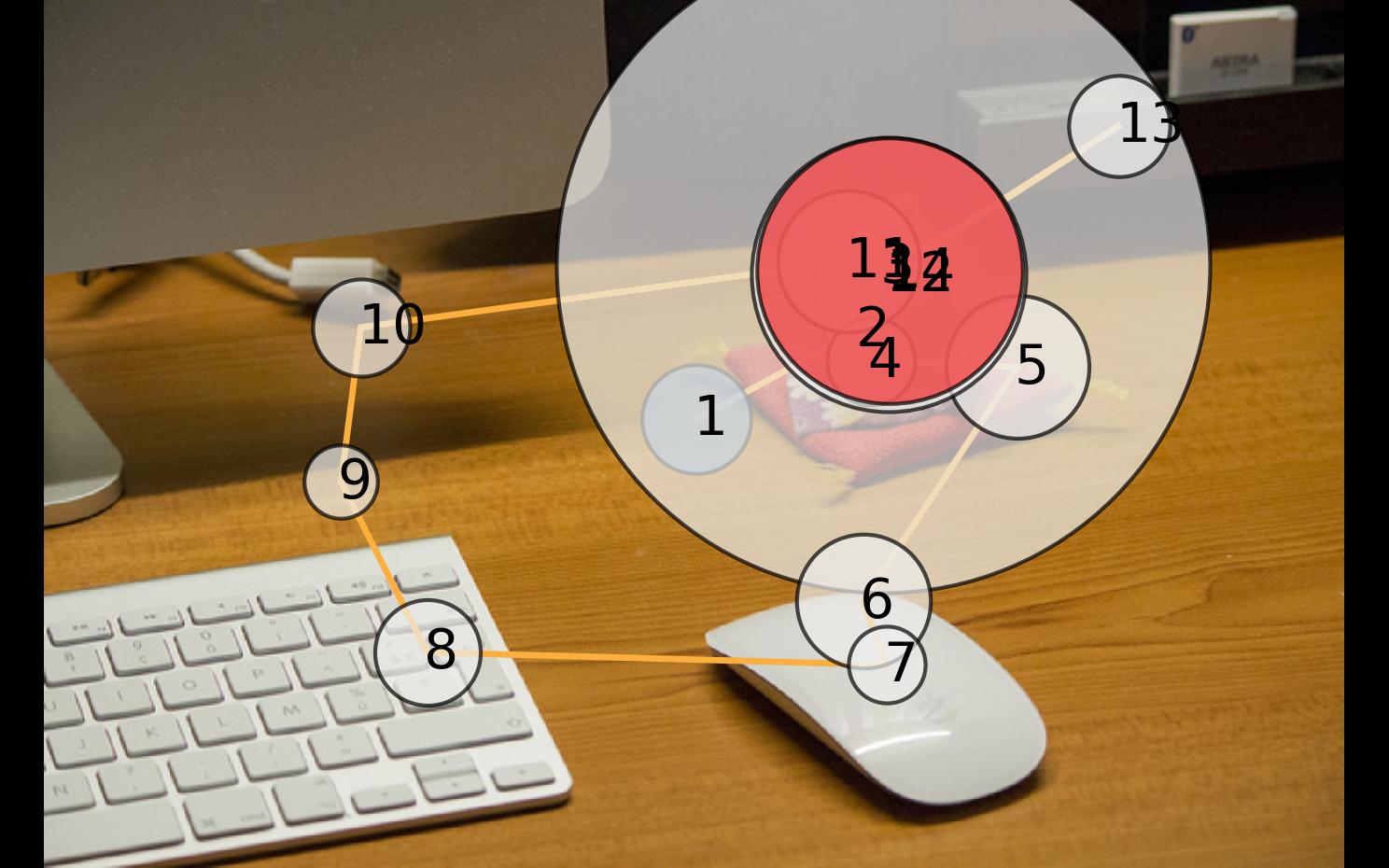} \\
    \end{tabular}
    }
    \vspace{-0.15cm}
    \caption{Qualitative comparison of the variability in simulated and human scanpaths on the COCOFreeView dataset. Each row corresponds to a different simulation or a different human observer.}
    \label{fig:qualitatives_variab_cocoFV}
    \vspace{-0.4cm}
\end{figure*}

\begin{figure*}[t]
    \footnotesize
    \setlength{\tabcolsep}{.1em}
    \resizebox{\linewidth}{!}{
    \begin{tabular}{cccccc}
        \scriptsize IOR-ROI-LSTM~\cite{chen2018scanpath}  & \scriptsize ChenLSTM~\cite{chen2021predicting} & \scriptsize GazeXplain~\cite{chen2024gazexplain} & \scriptsize TPP-Gaze~\cite{damelio2025tpp} & \scriptsize \ours (Ours) & \scriptsize Humans \\
        \includegraphics[width=0.14\linewidth]{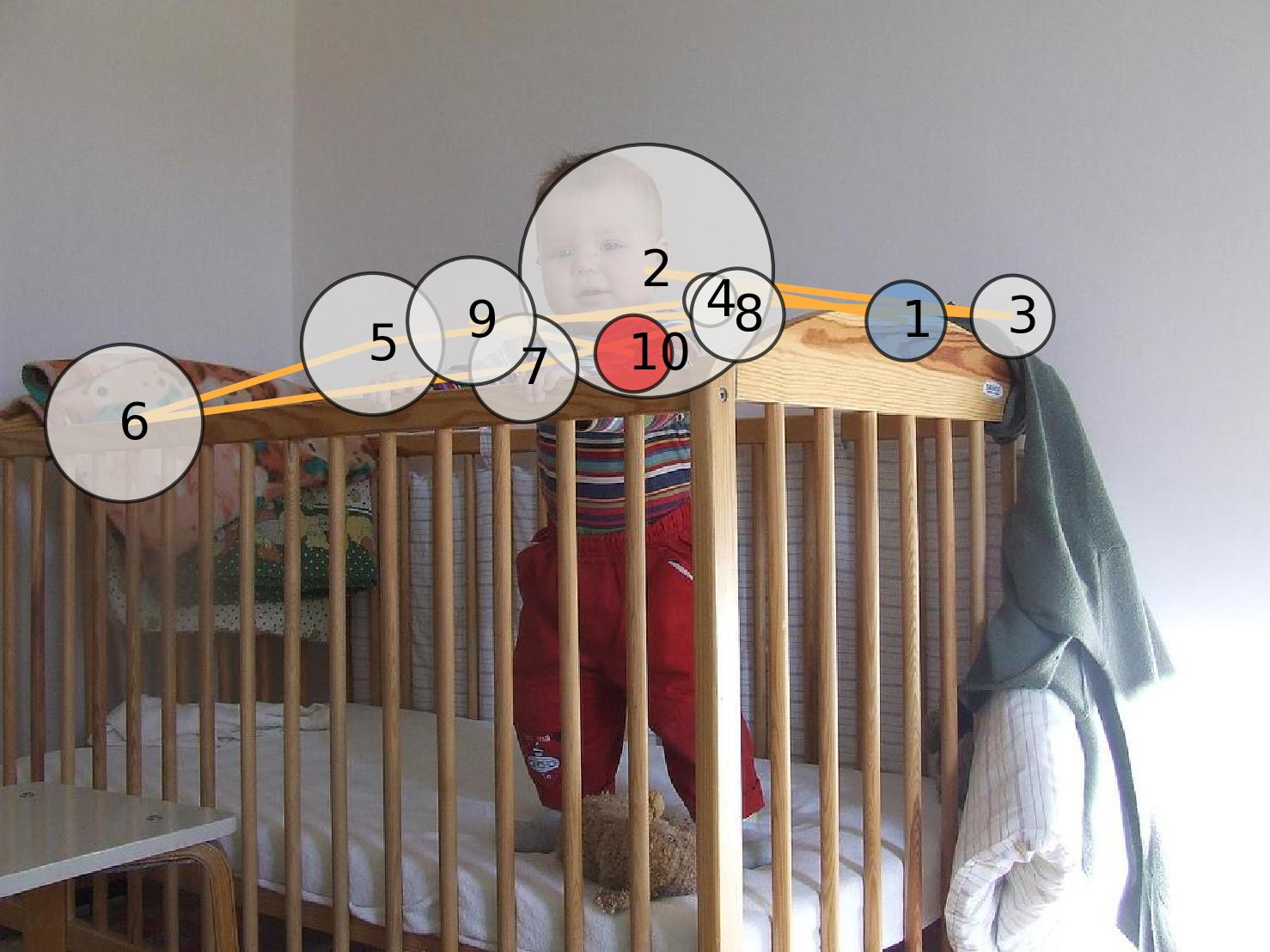} & \includegraphics[width=0.14\linewidth]{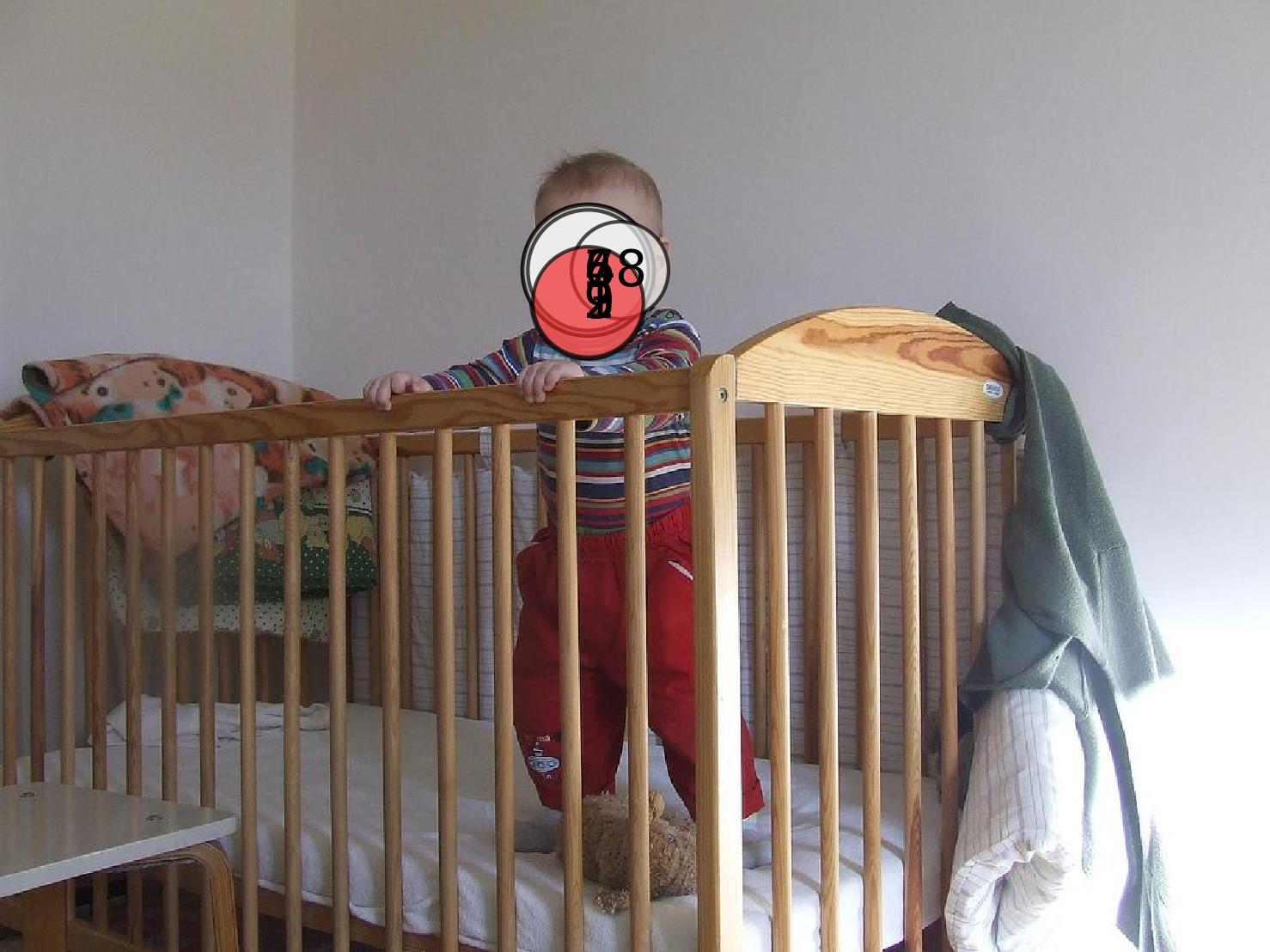} & \includegraphics[width=0.14\linewidth]{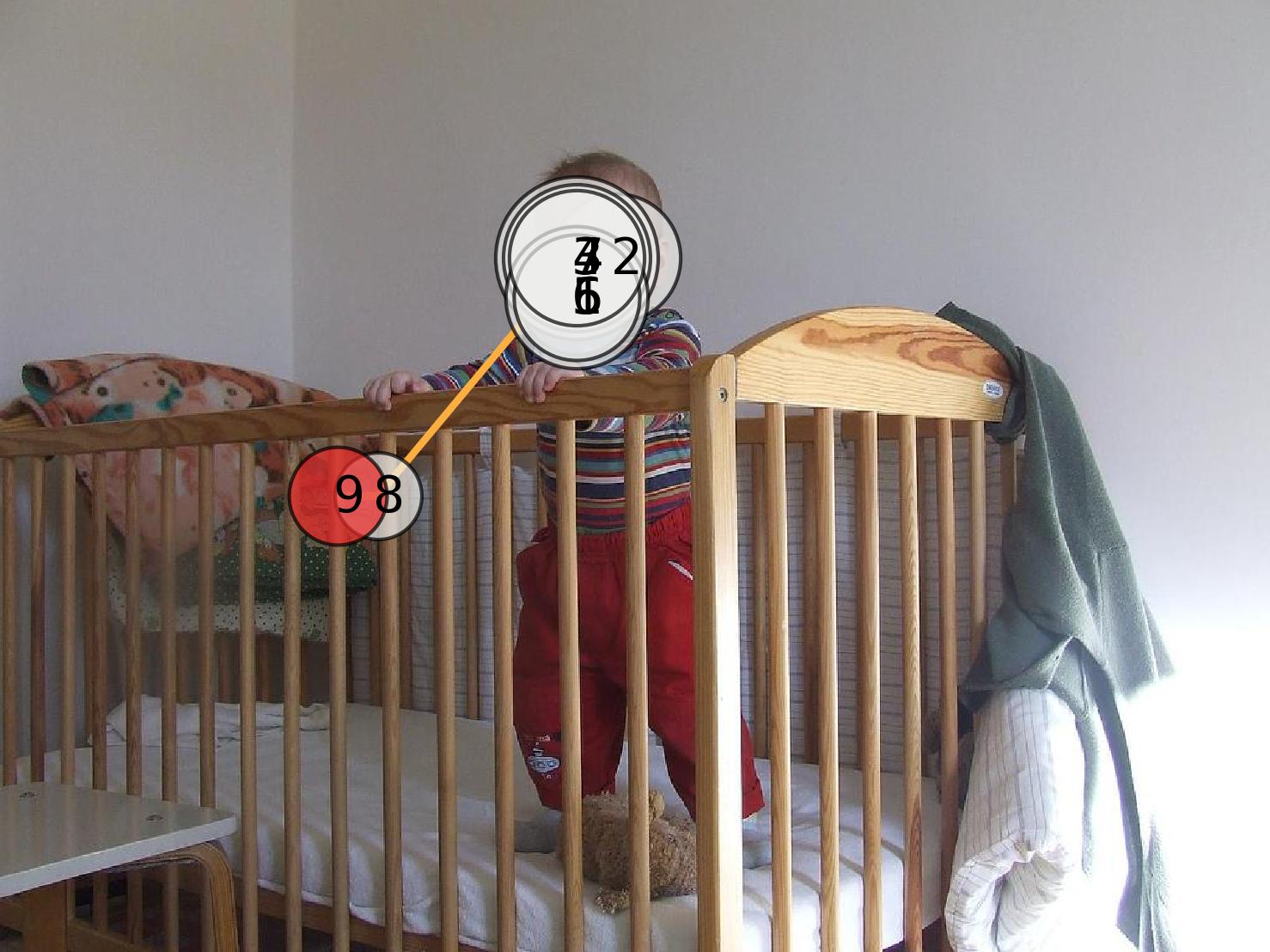} & \includegraphics[width=0.14\linewidth]{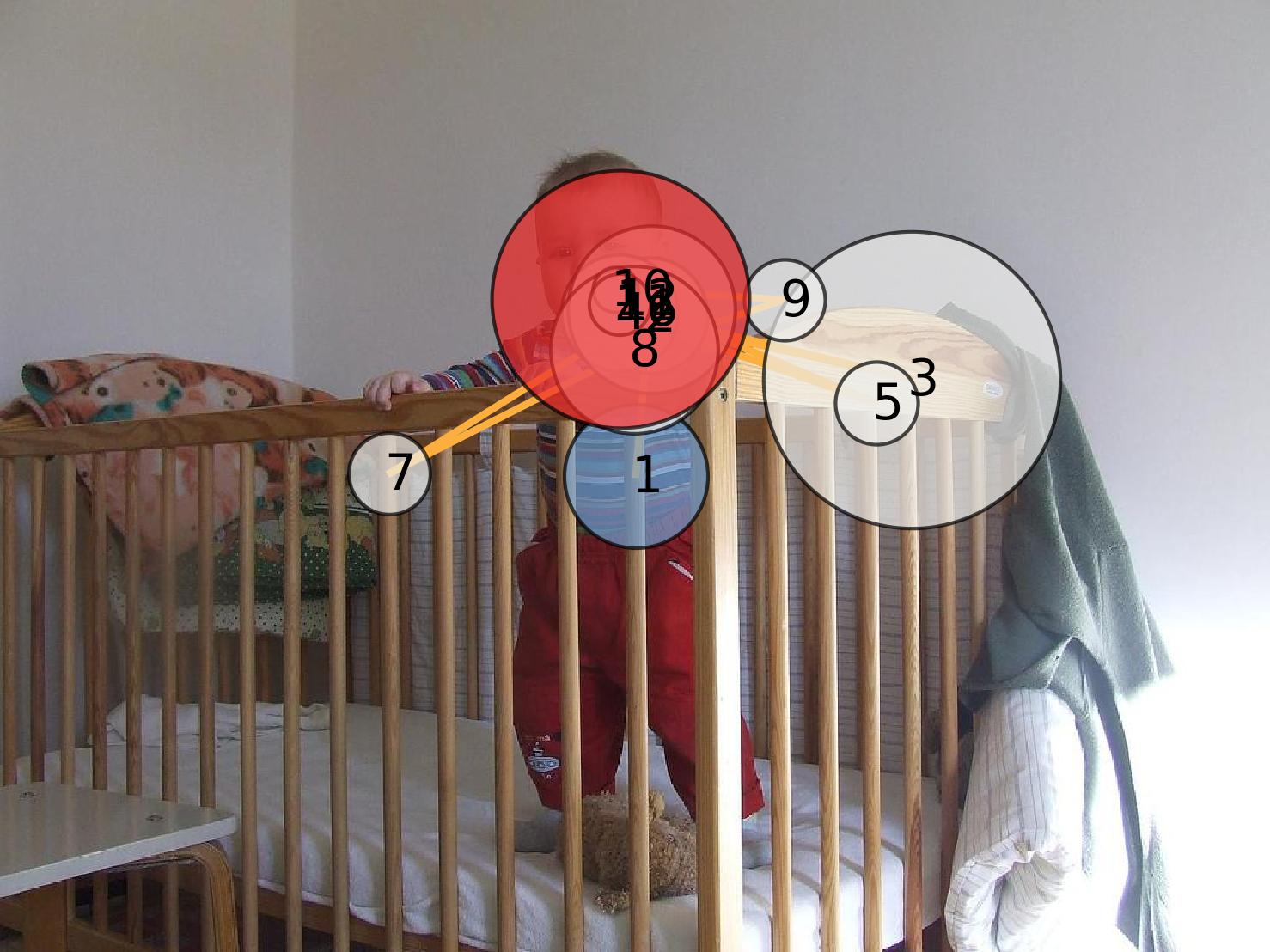} & \includegraphics[width=0.14\linewidth]{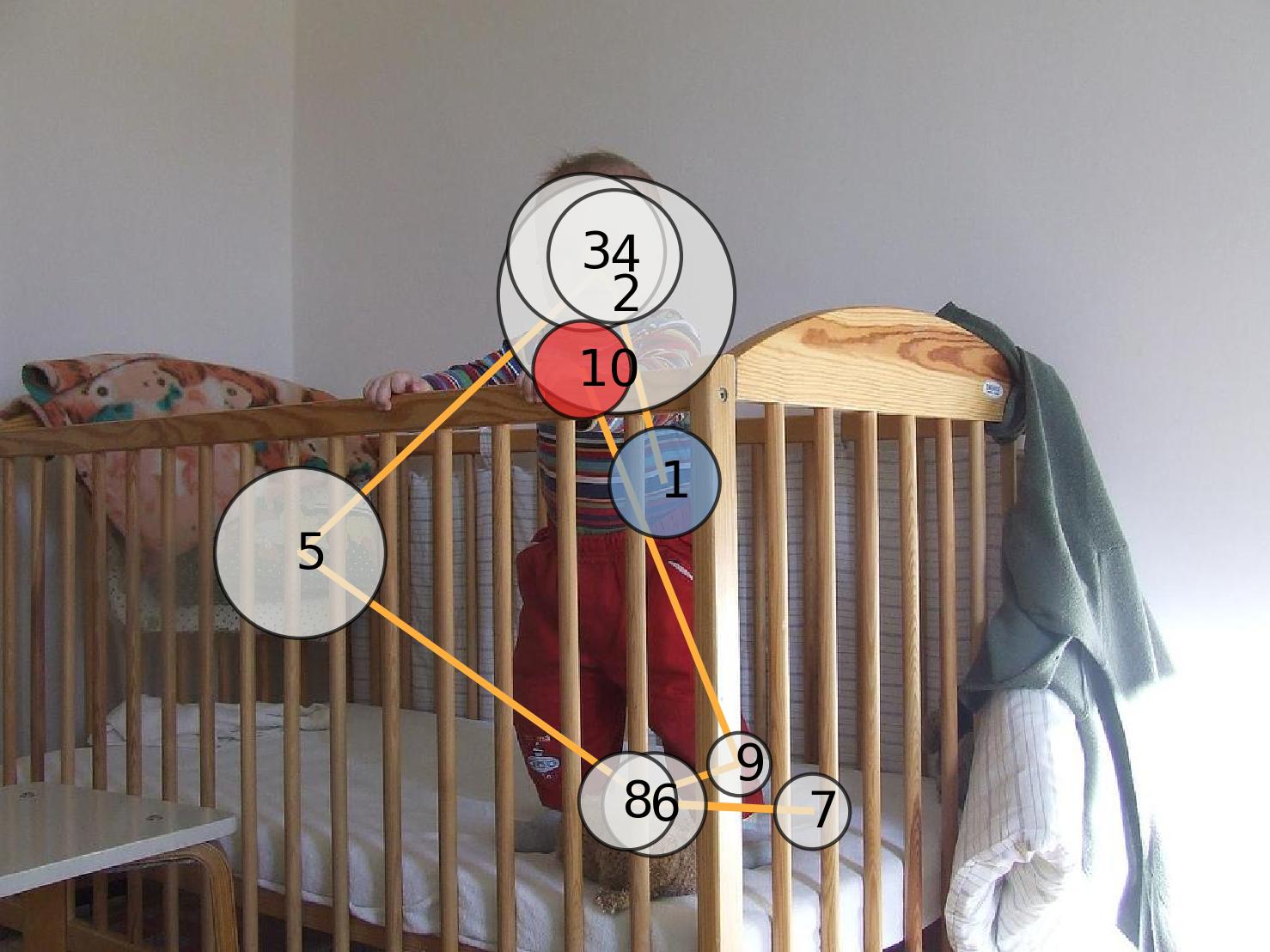} & \includegraphics[width=0.14\linewidth]{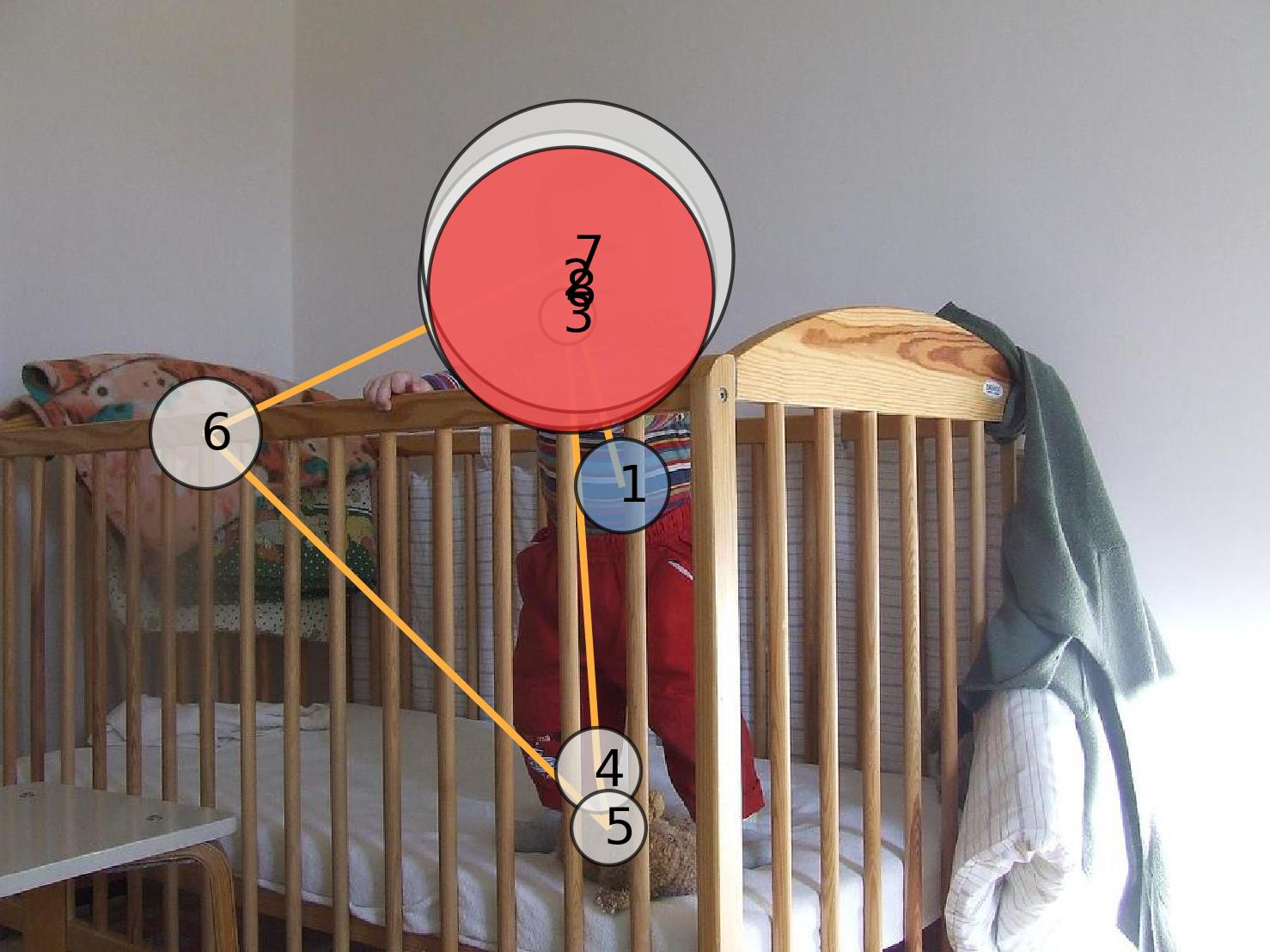} \\
         \includegraphics[width=0.14\linewidth]{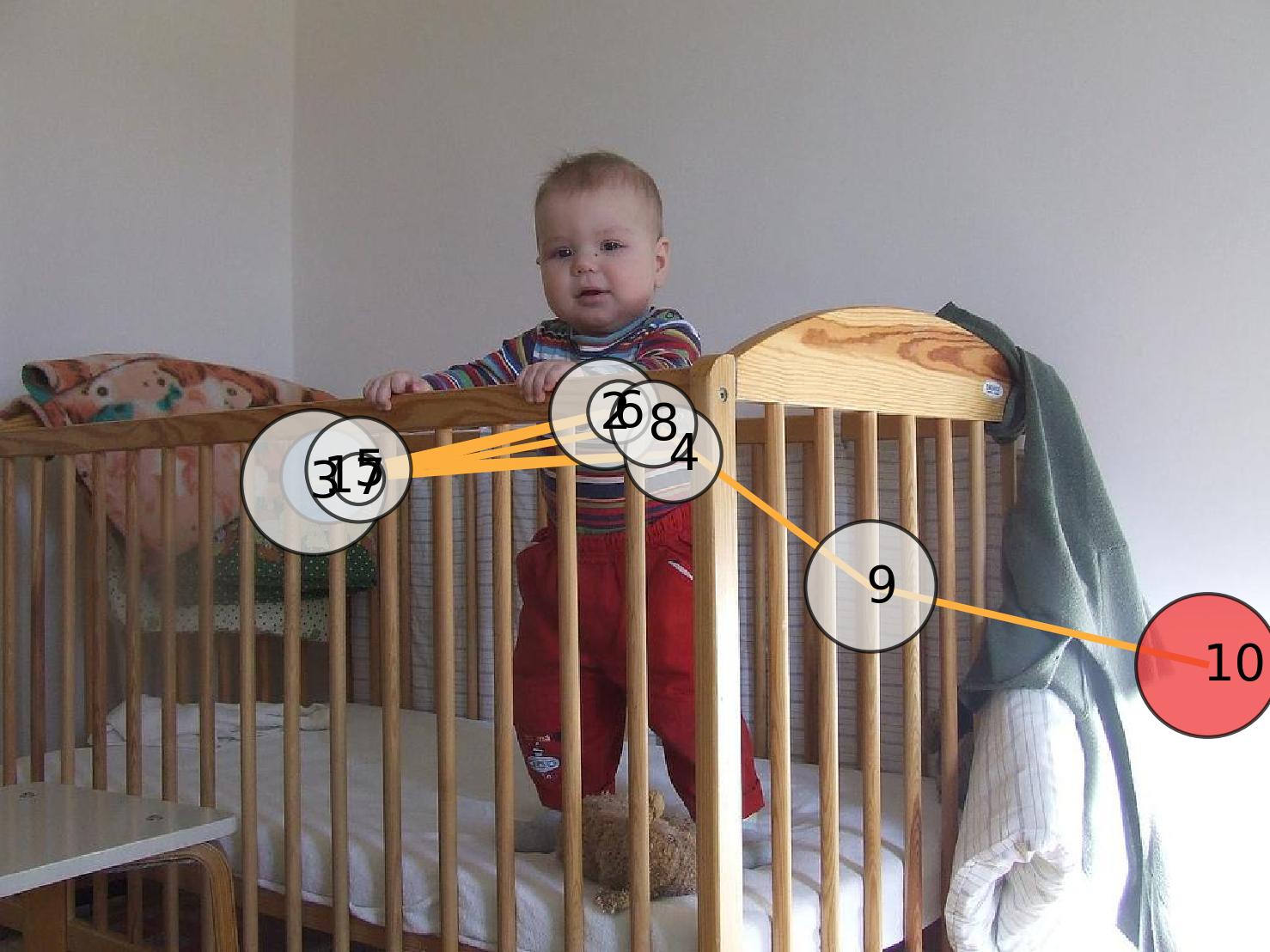} & \includegraphics[width=0.14\linewidth]{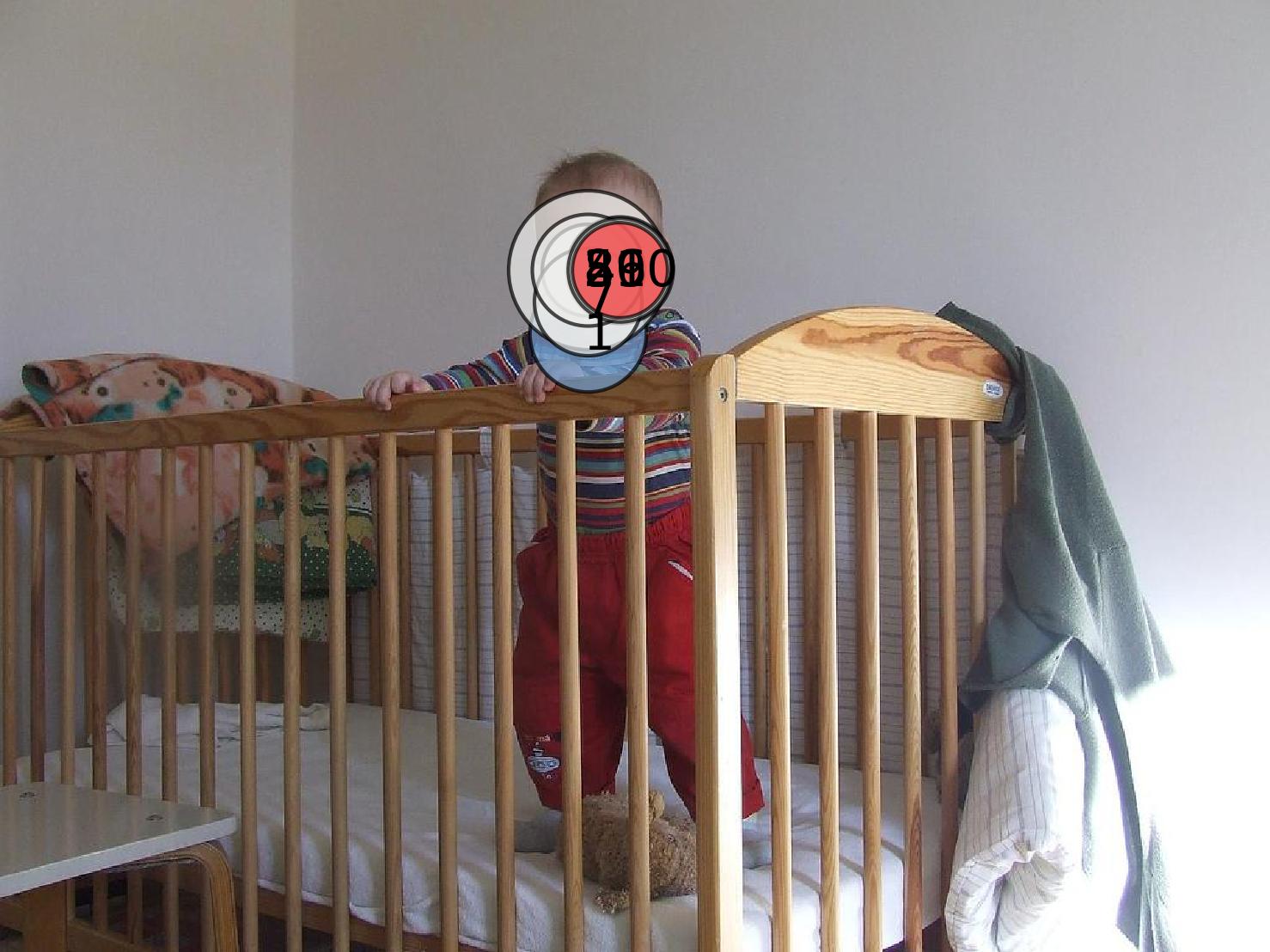} & \includegraphics[width=0.14\linewidth]{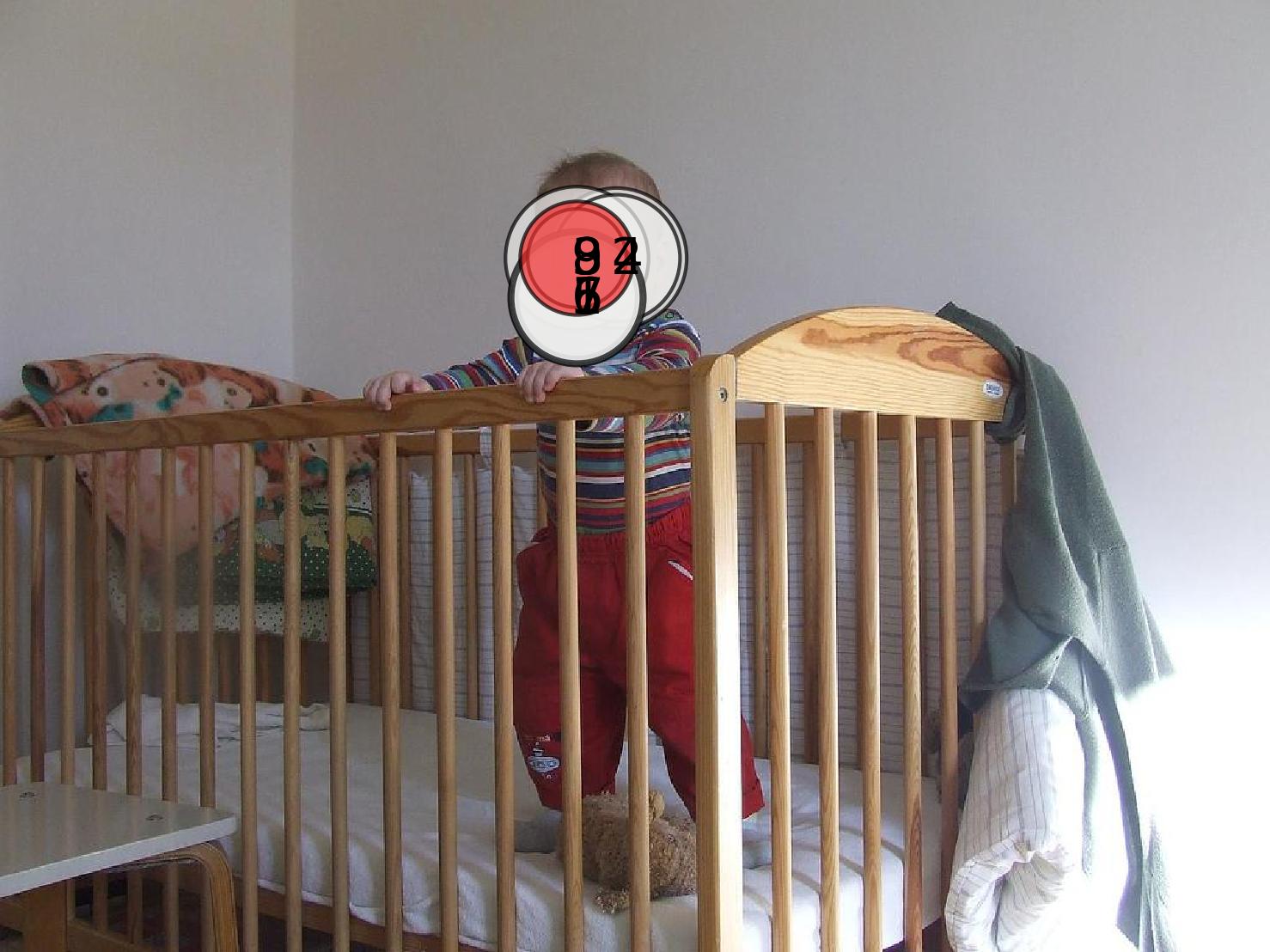} & \includegraphics[width=0.14\linewidth]{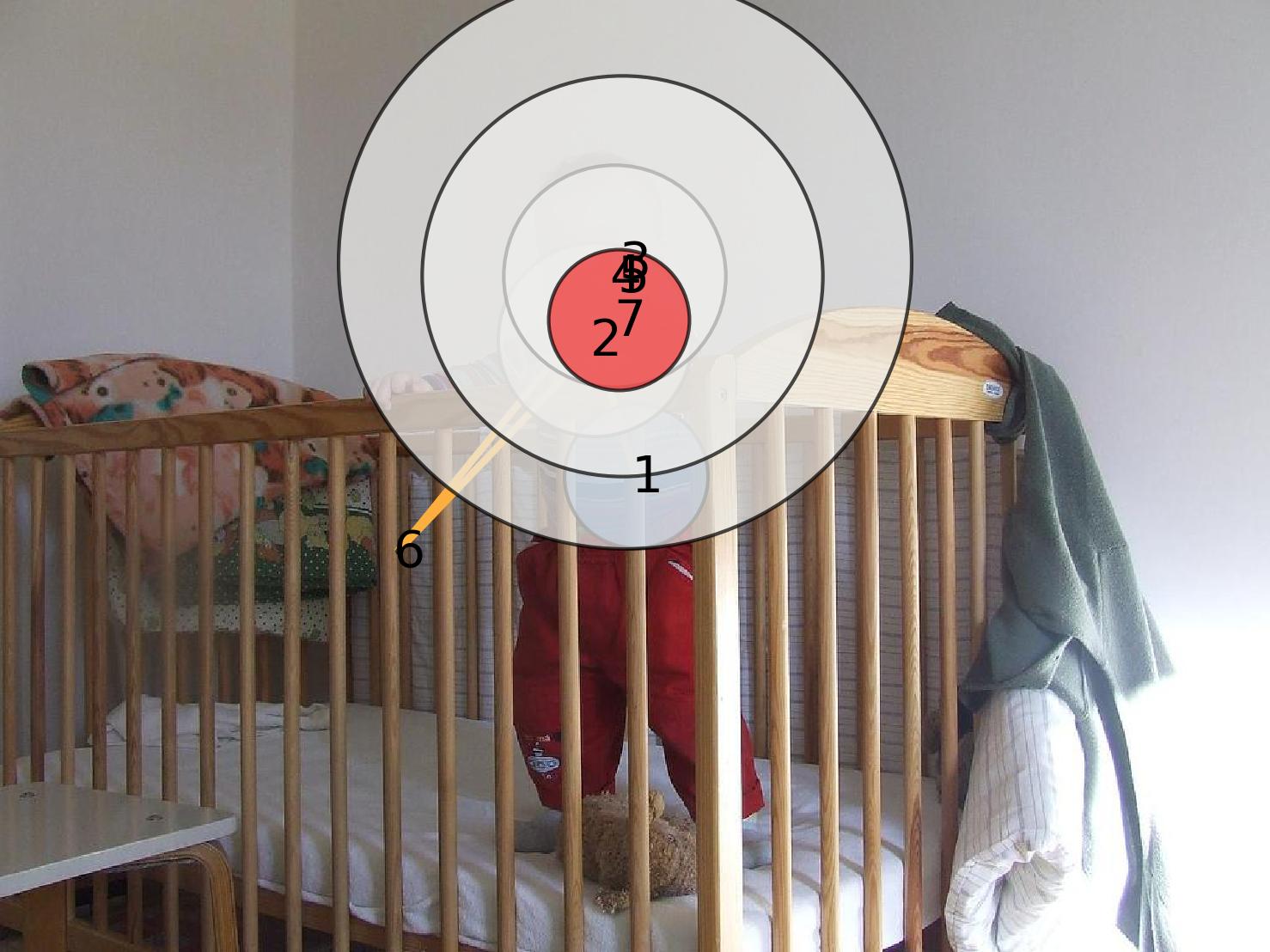} & \includegraphics[width=0.14\linewidth]{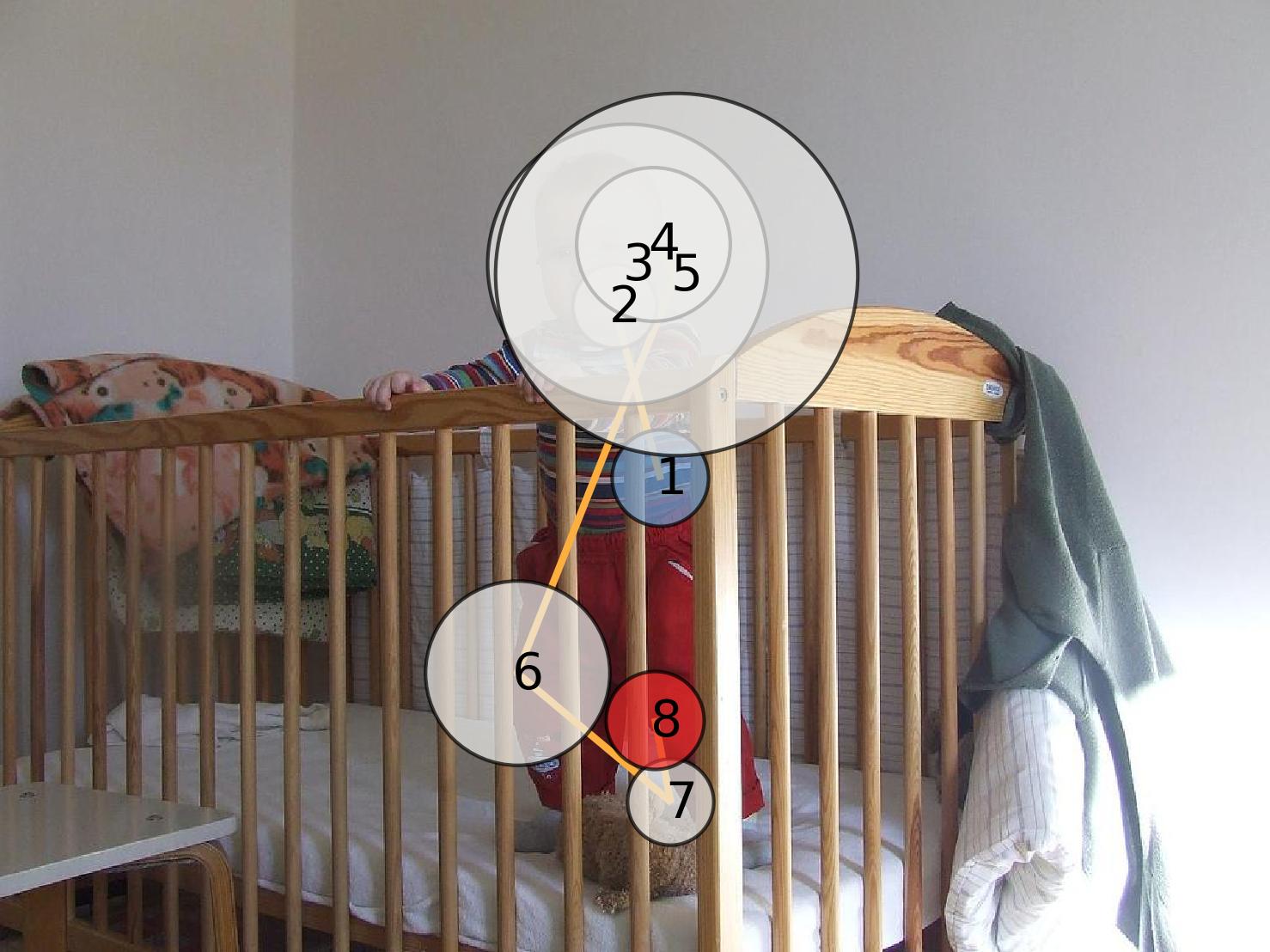} & \includegraphics[width=0.14\linewidth]{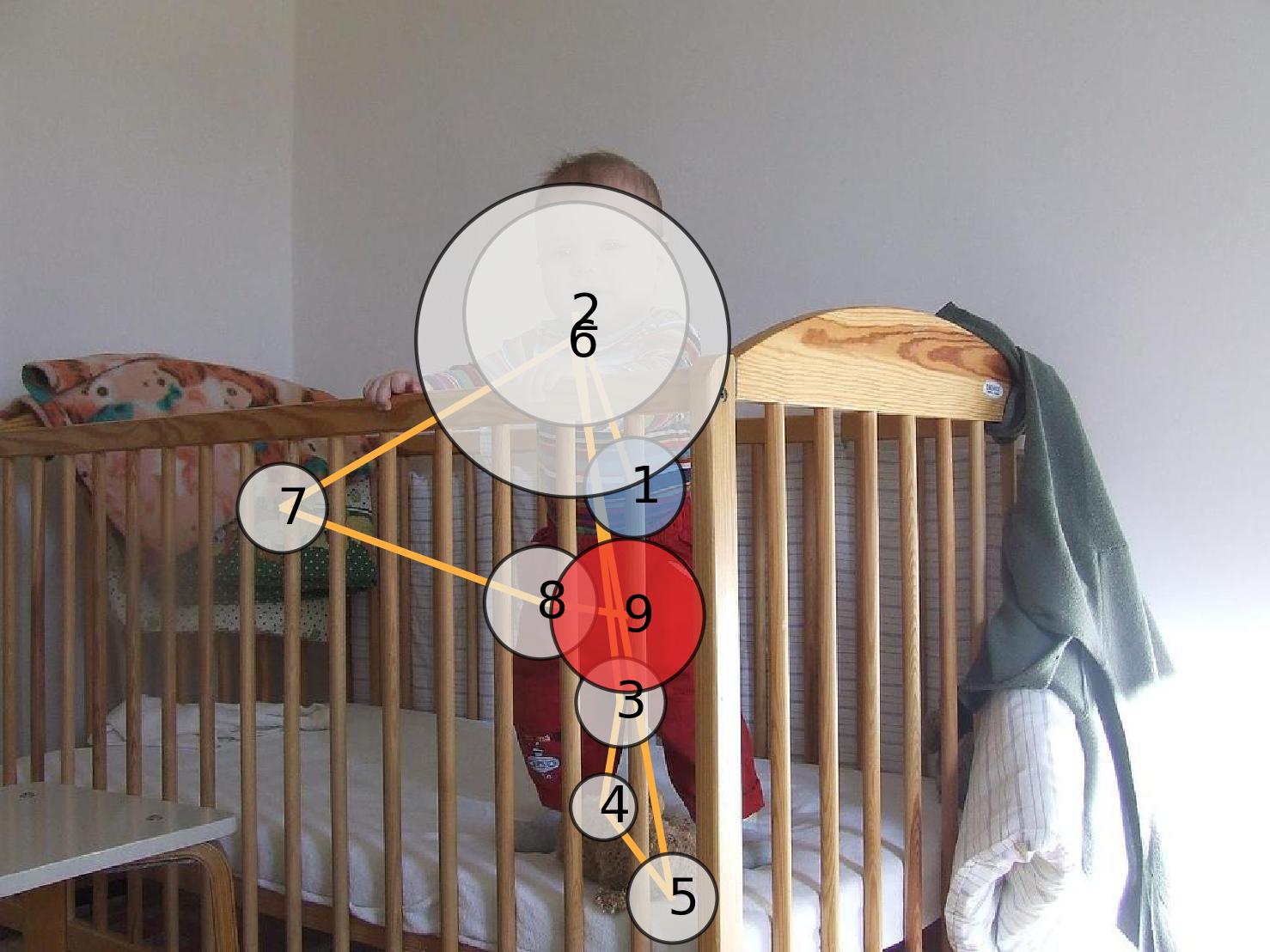} \\
         \includegraphics[width=0.14\linewidth]{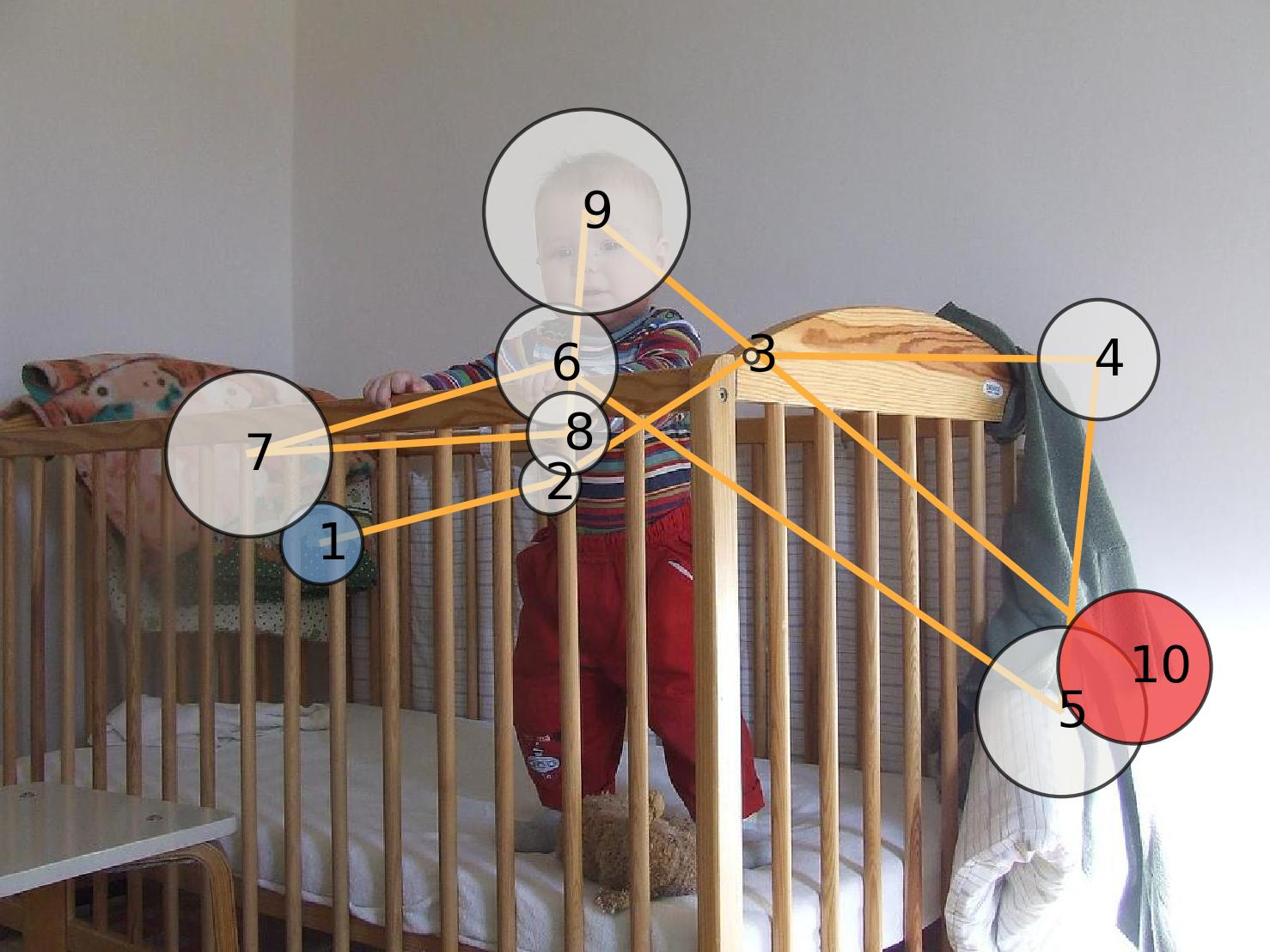} & \includegraphics[width=0.14\linewidth]{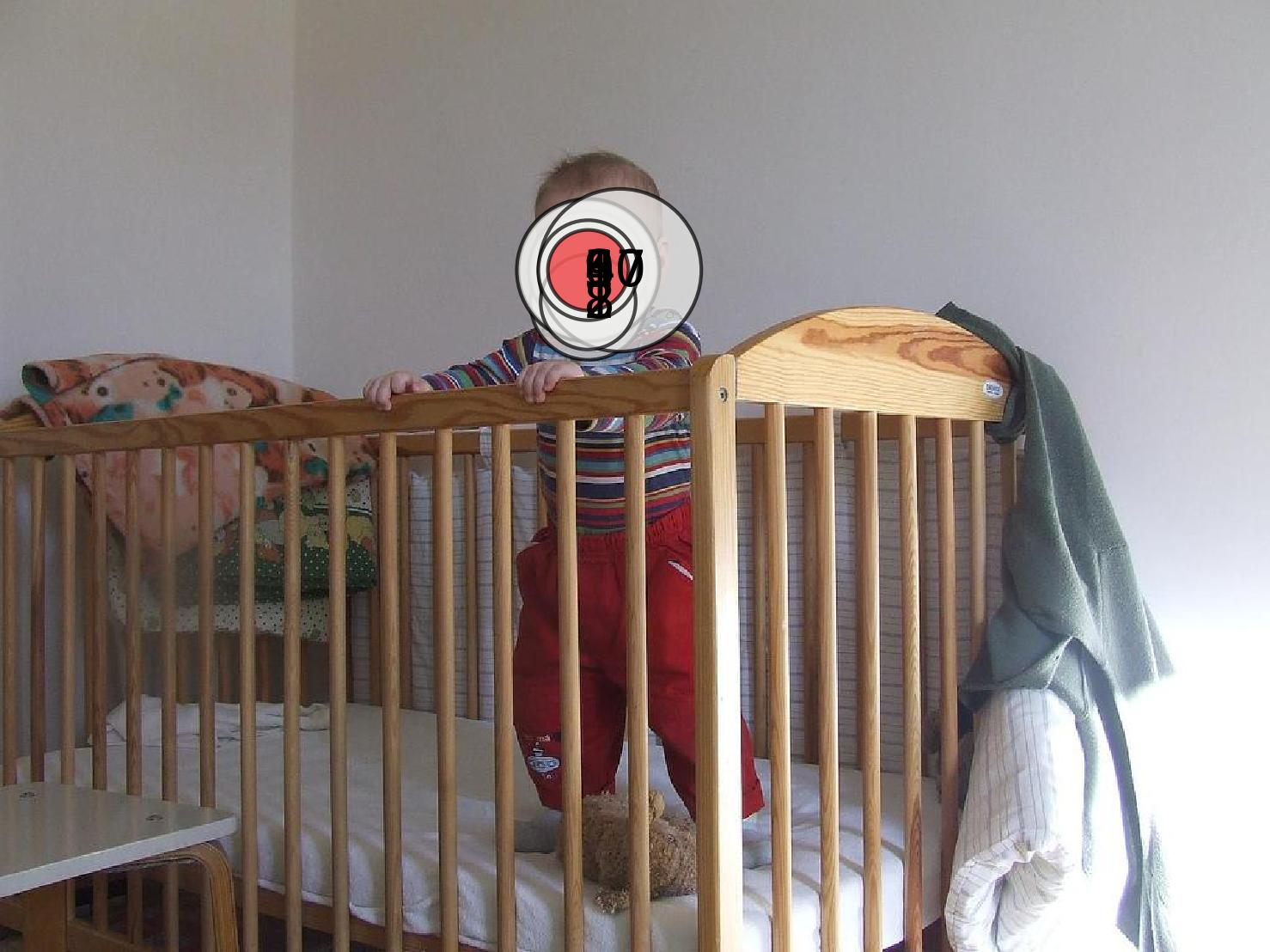} & \includegraphics[width=0.14\linewidth]{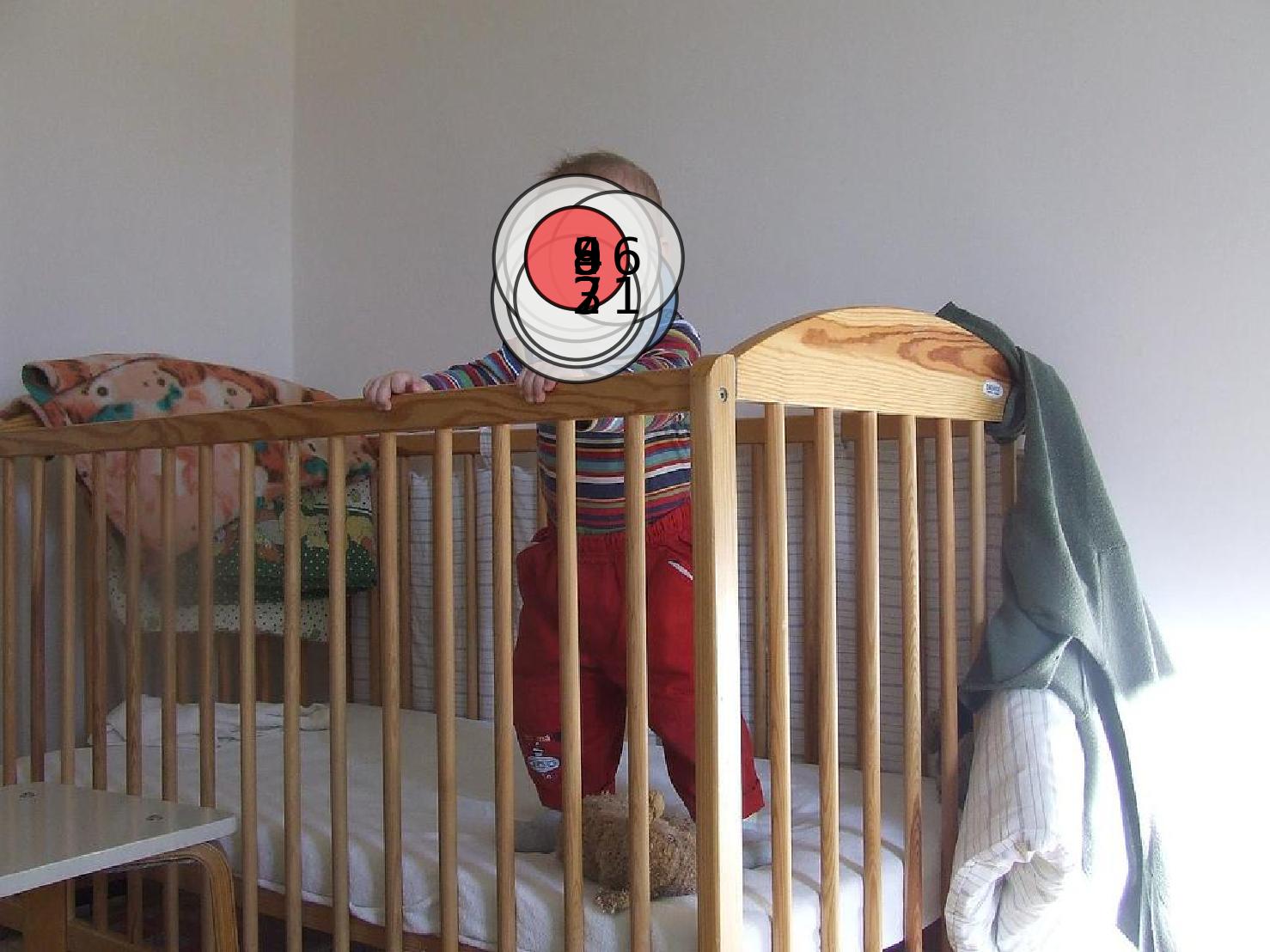} & \includegraphics[width=0.14\linewidth]{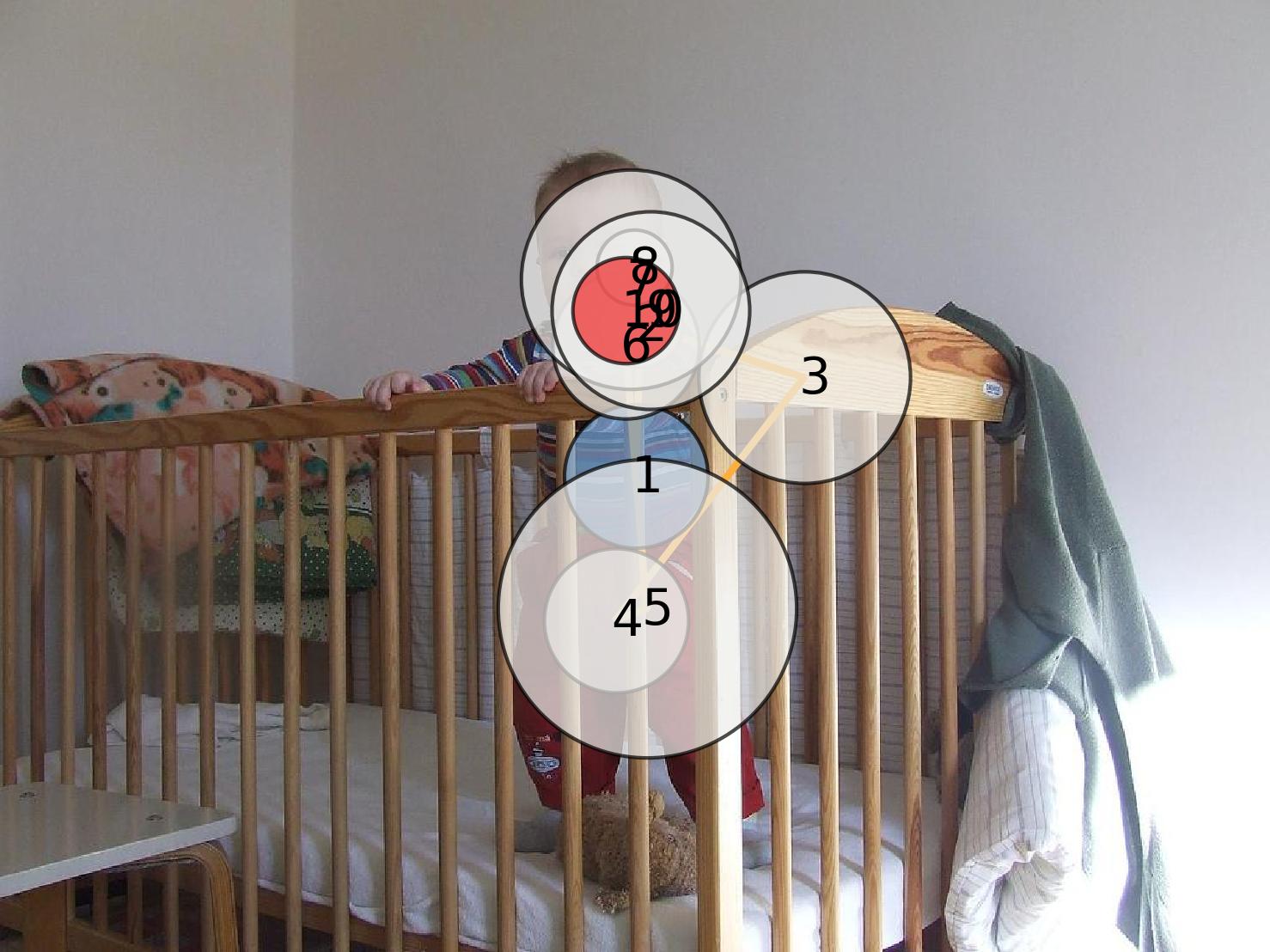} & \includegraphics[width=0.14\linewidth]{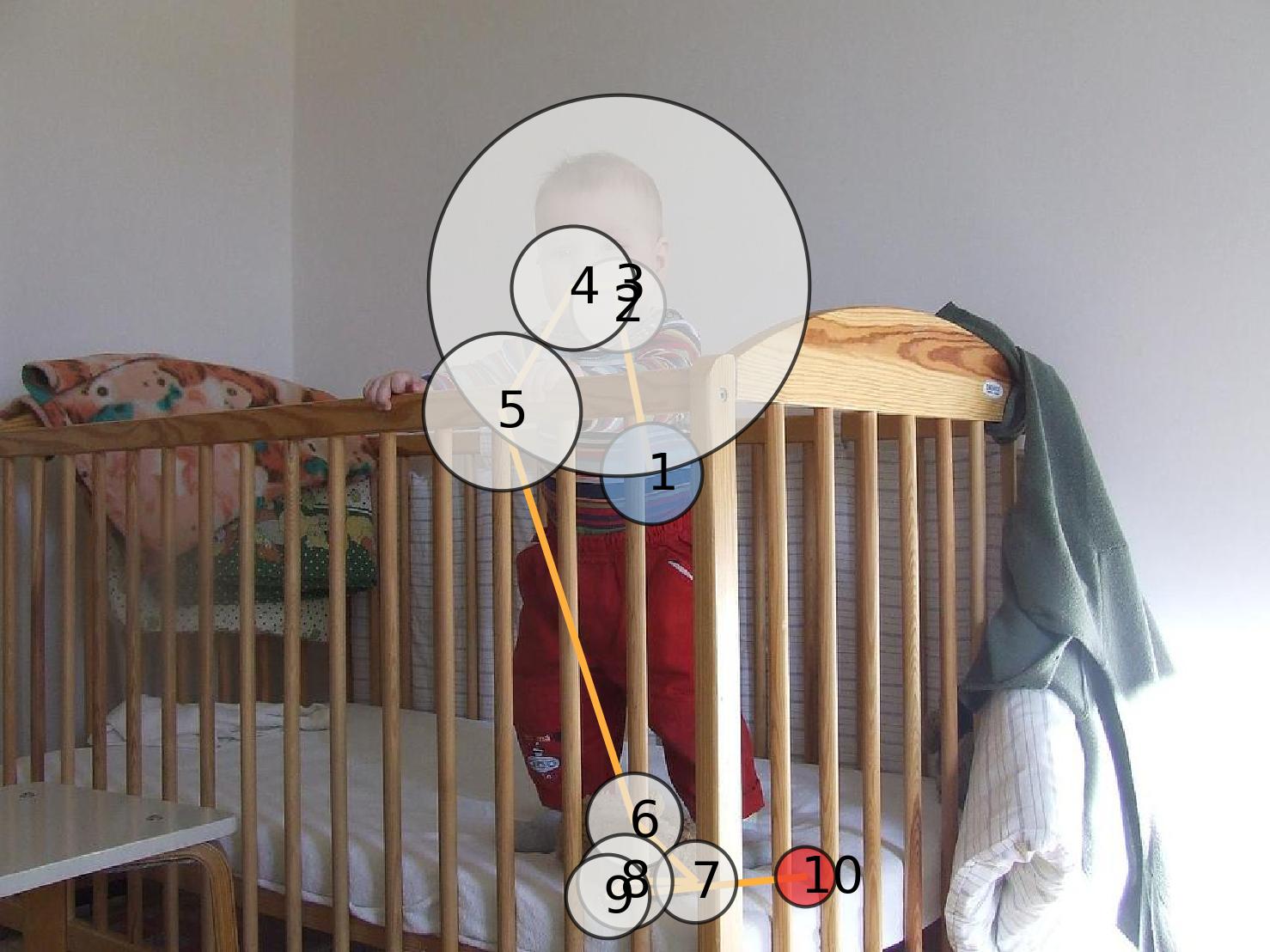} & \includegraphics[width=0.14\linewidth]{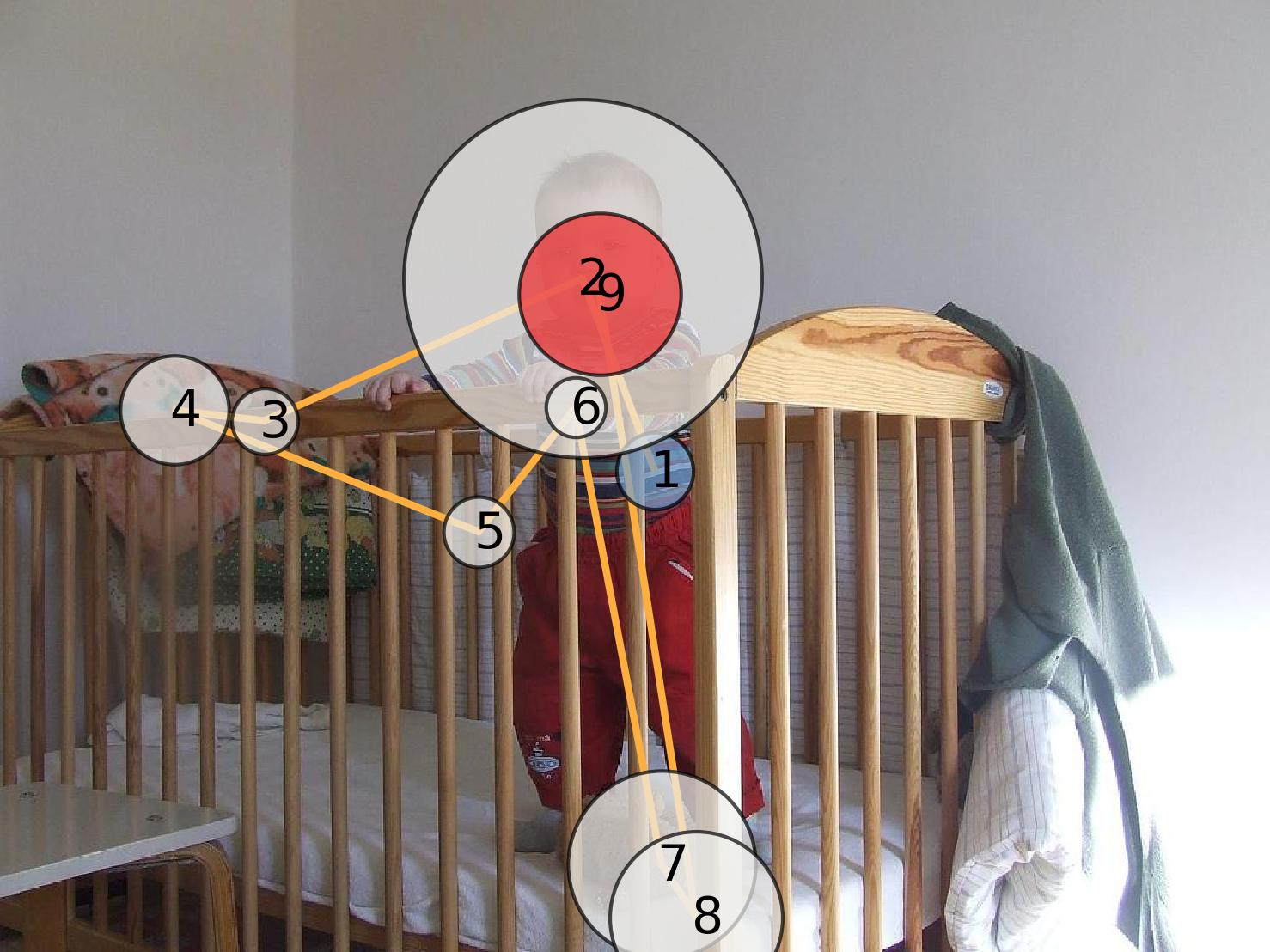} \\
         \includegraphics[width=0.14\linewidth]{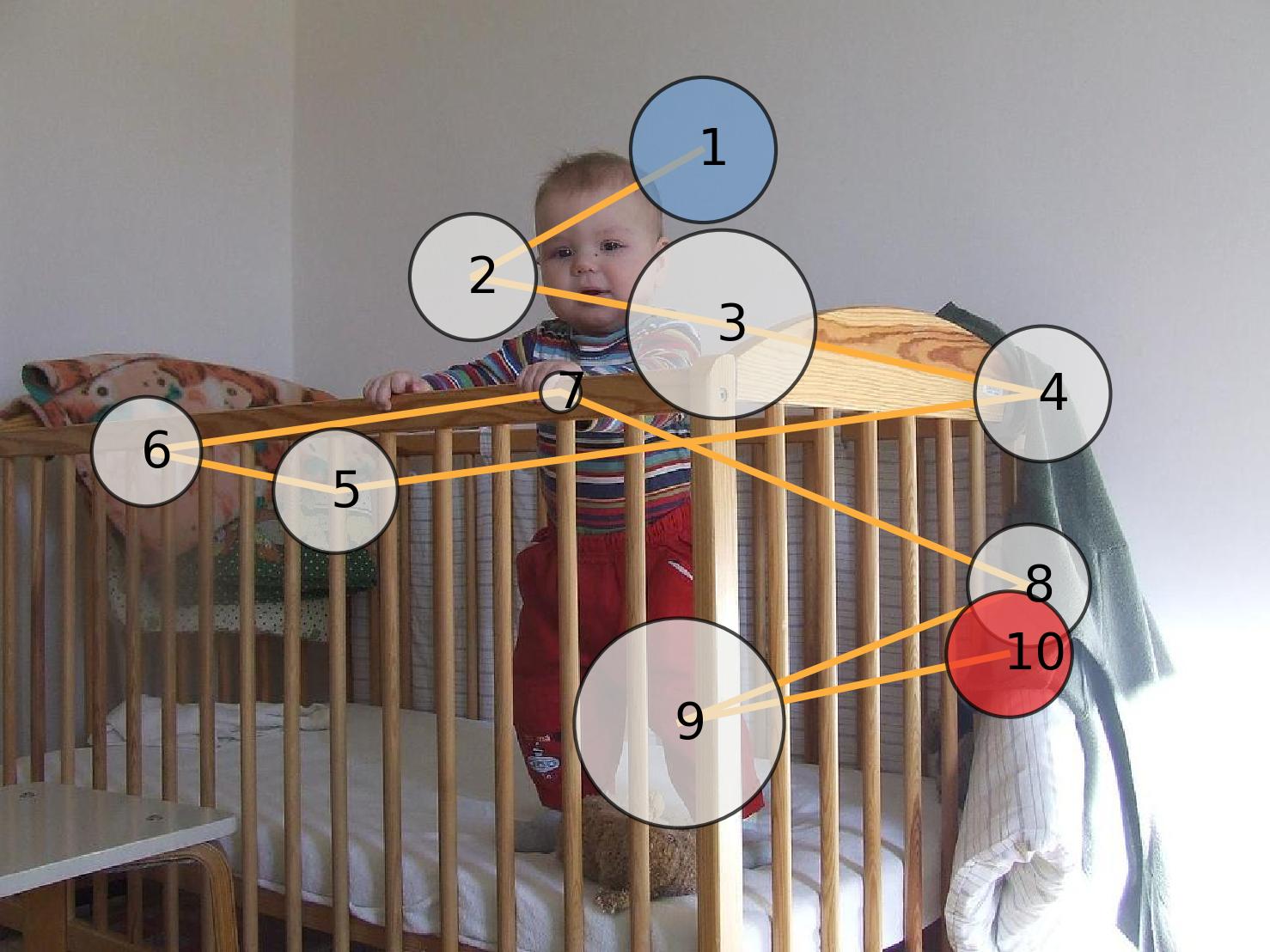} & \includegraphics[width=0.14\linewidth]{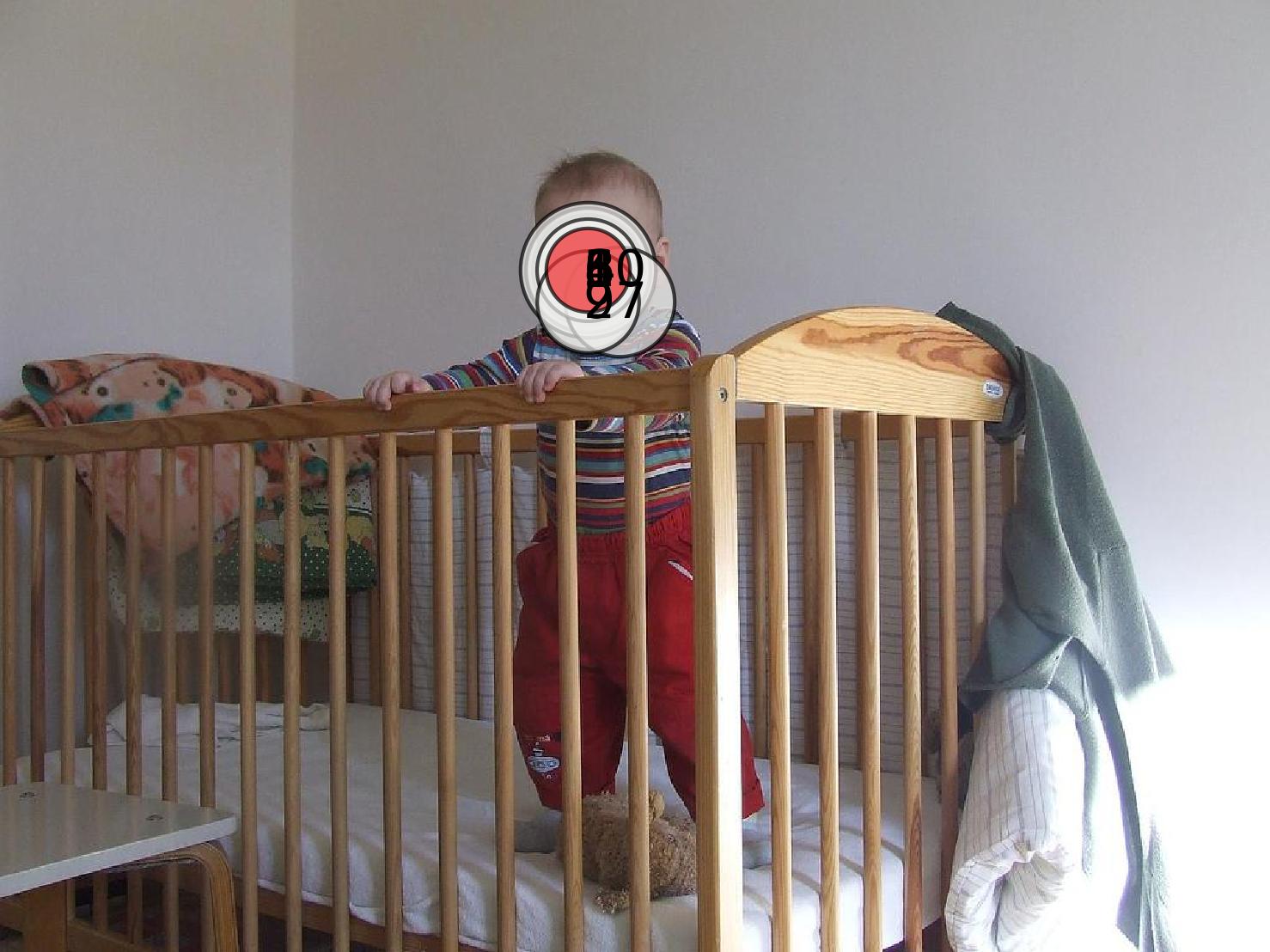} & \includegraphics[width=0.14\linewidth]{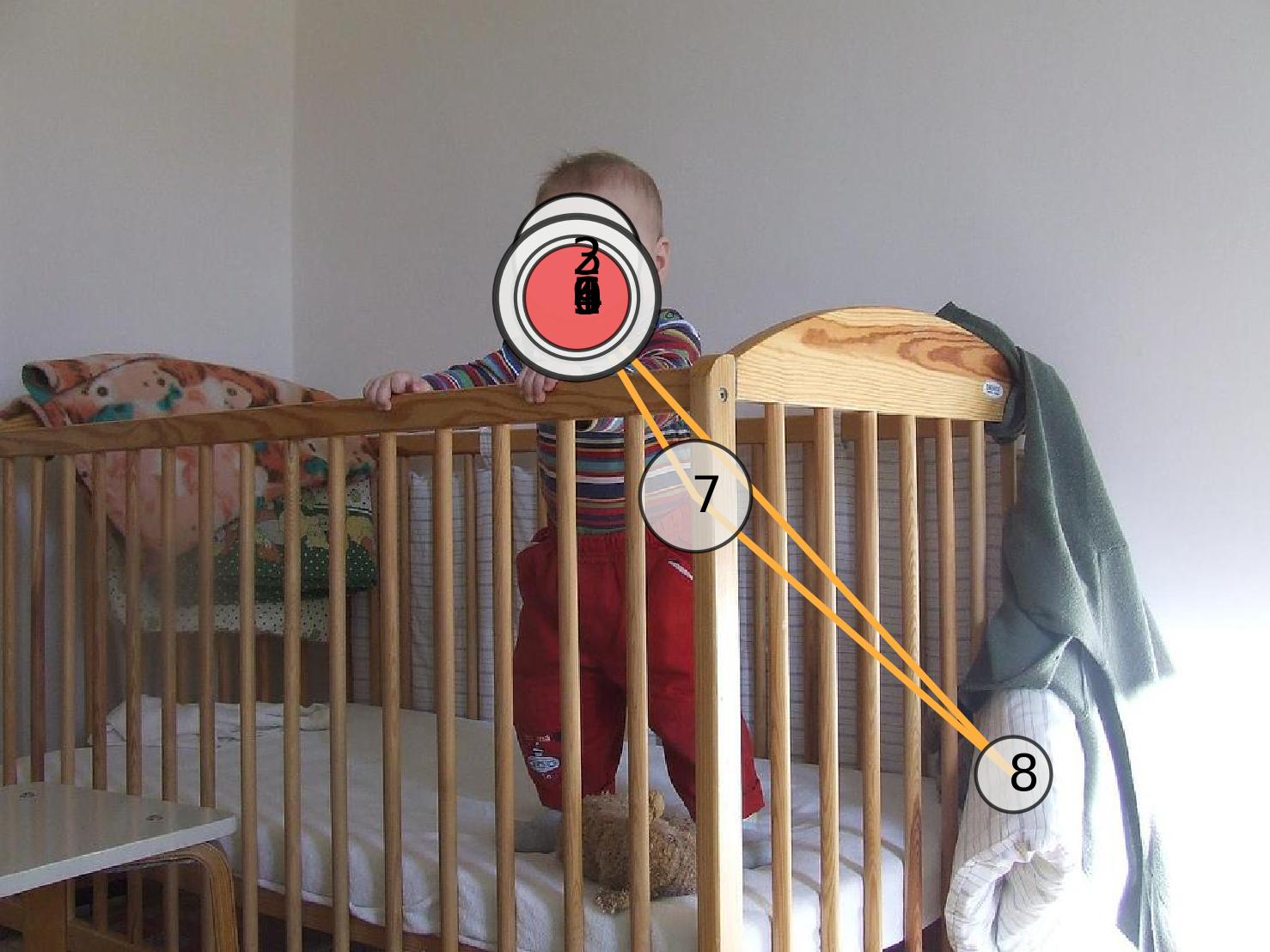} & \includegraphics[width=0.14\linewidth]{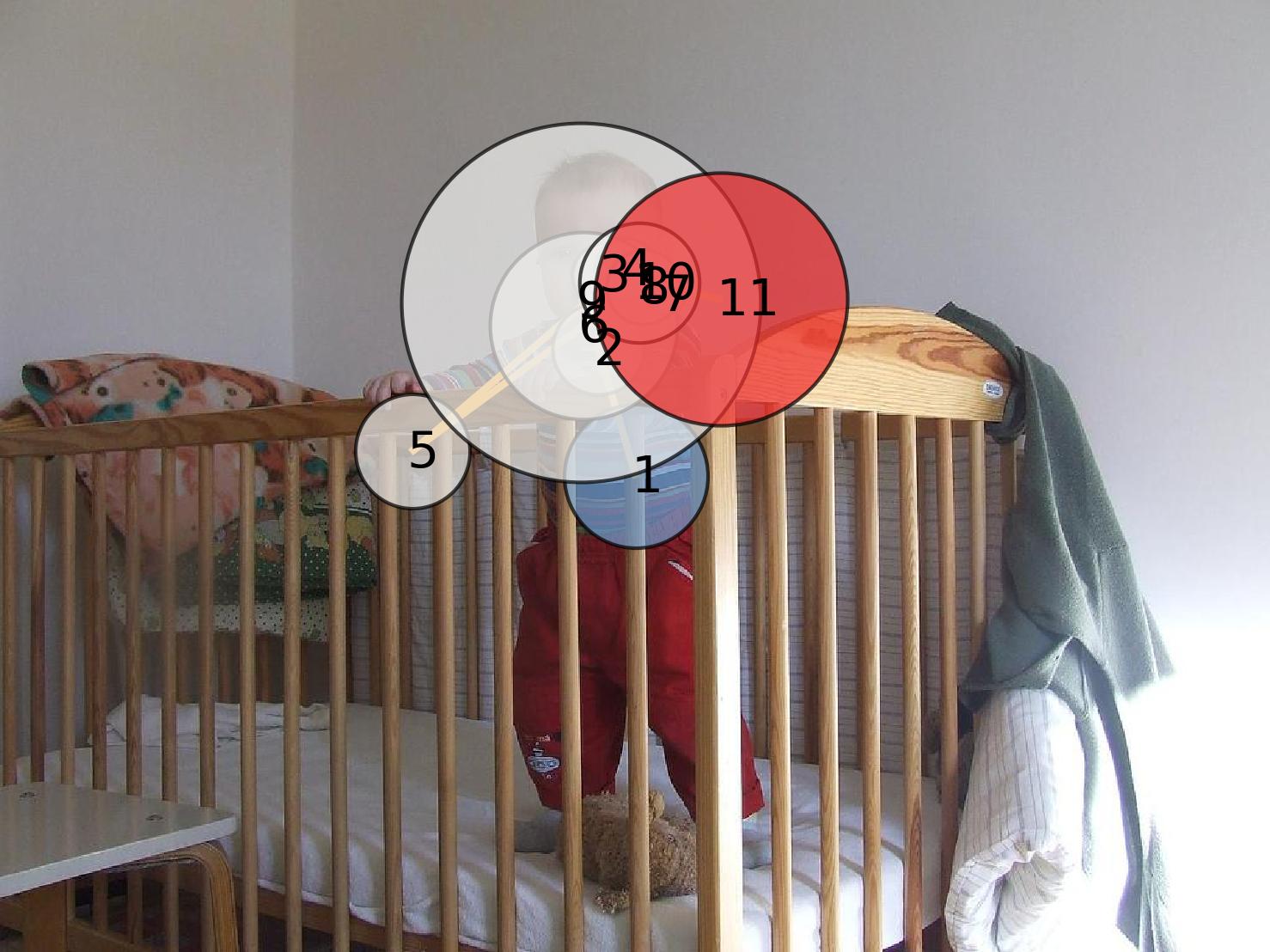} & \includegraphics[width=0.14\linewidth]{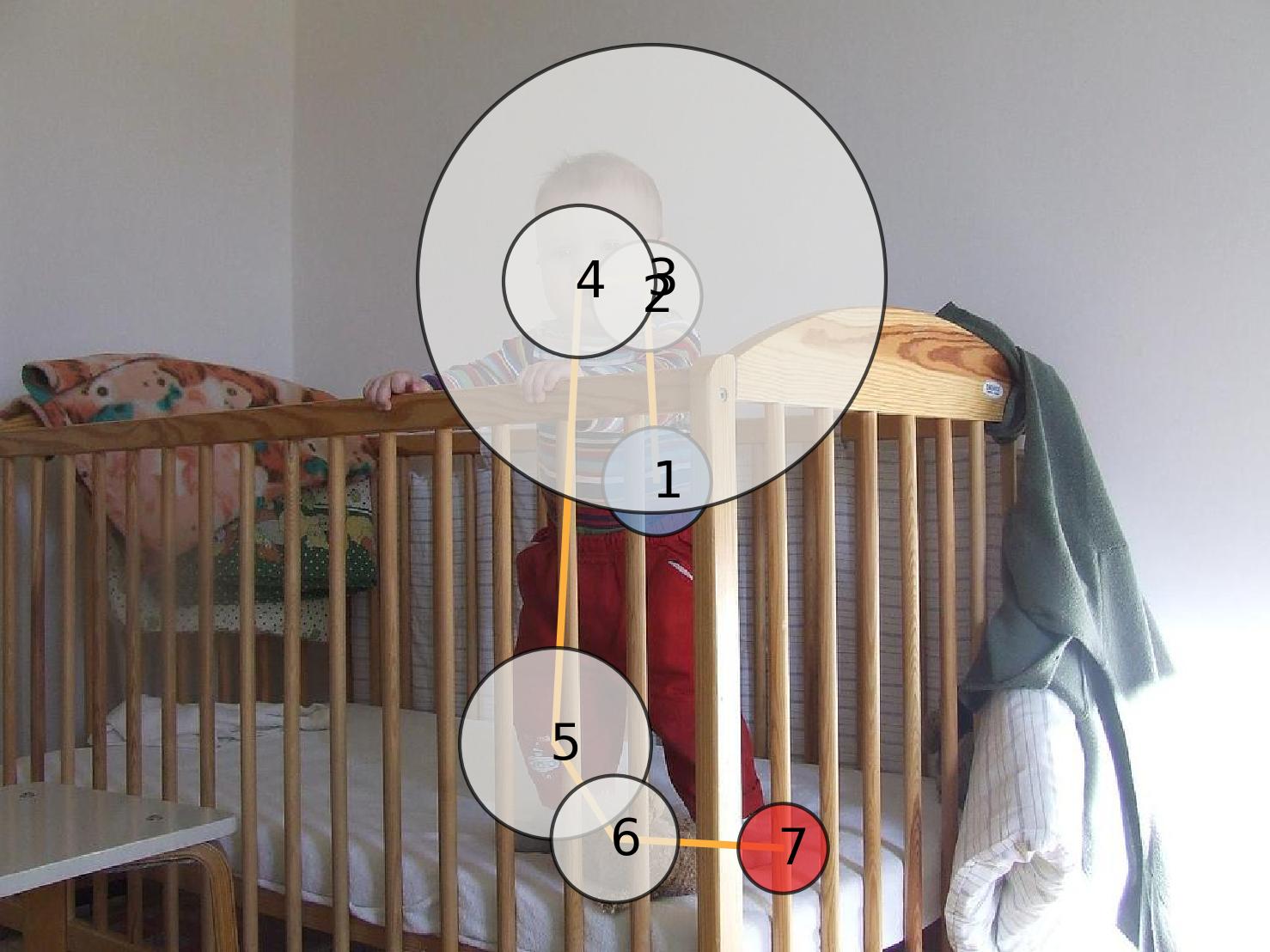} & \includegraphics[width=0.14\linewidth]{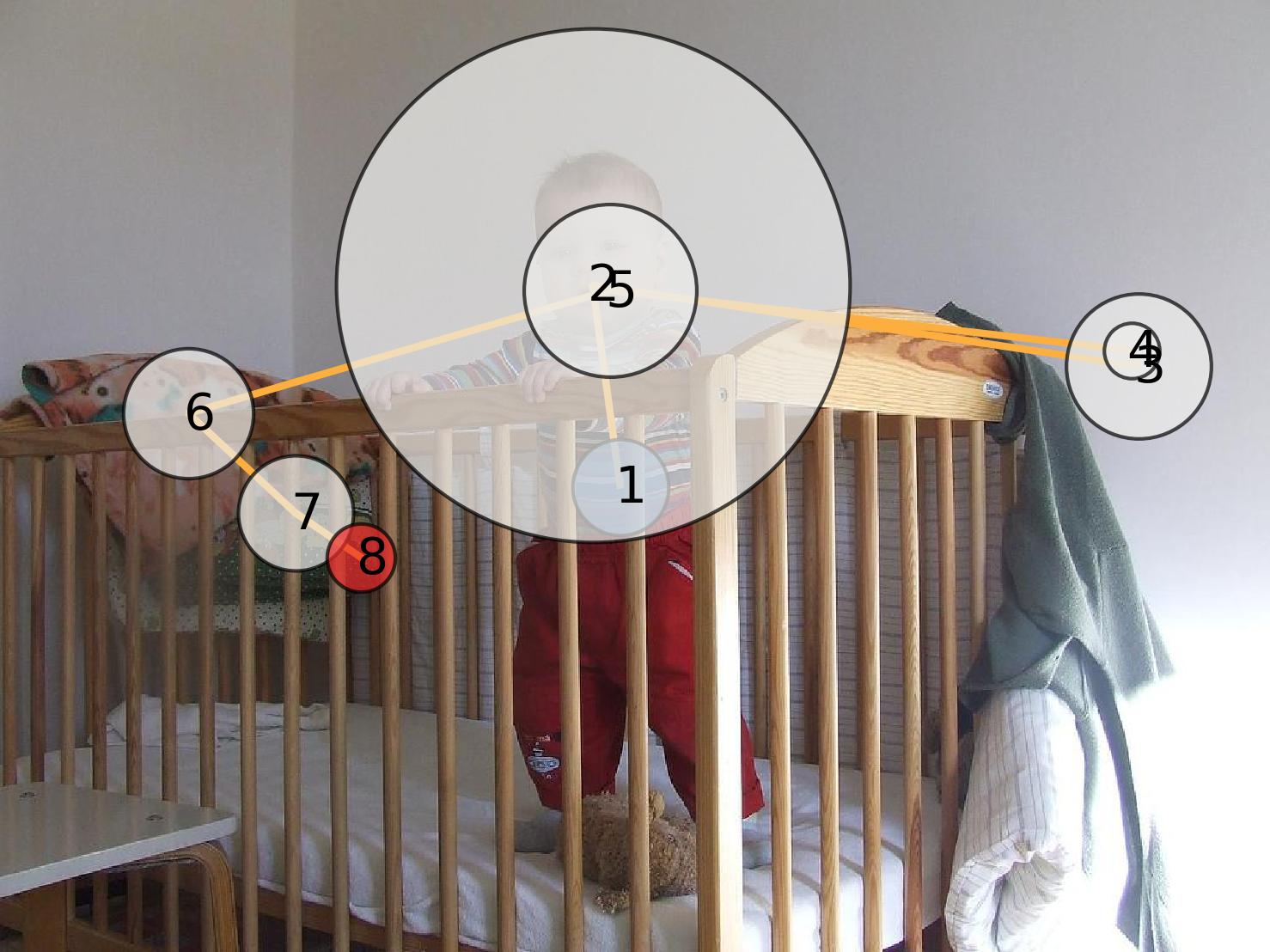} \\
        \includegraphics[width=0.14\linewidth]{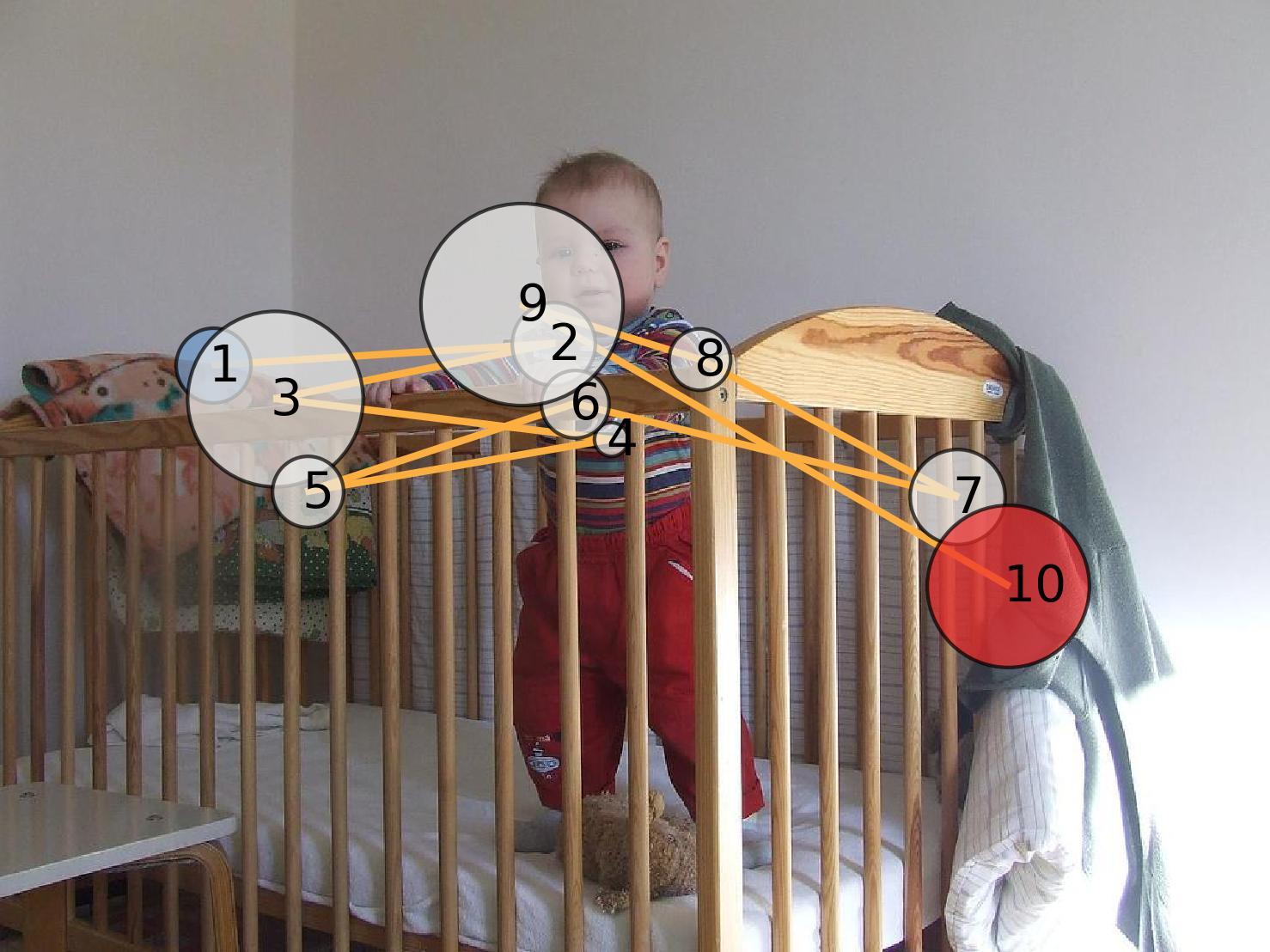} & \includegraphics[width=0.14\linewidth]{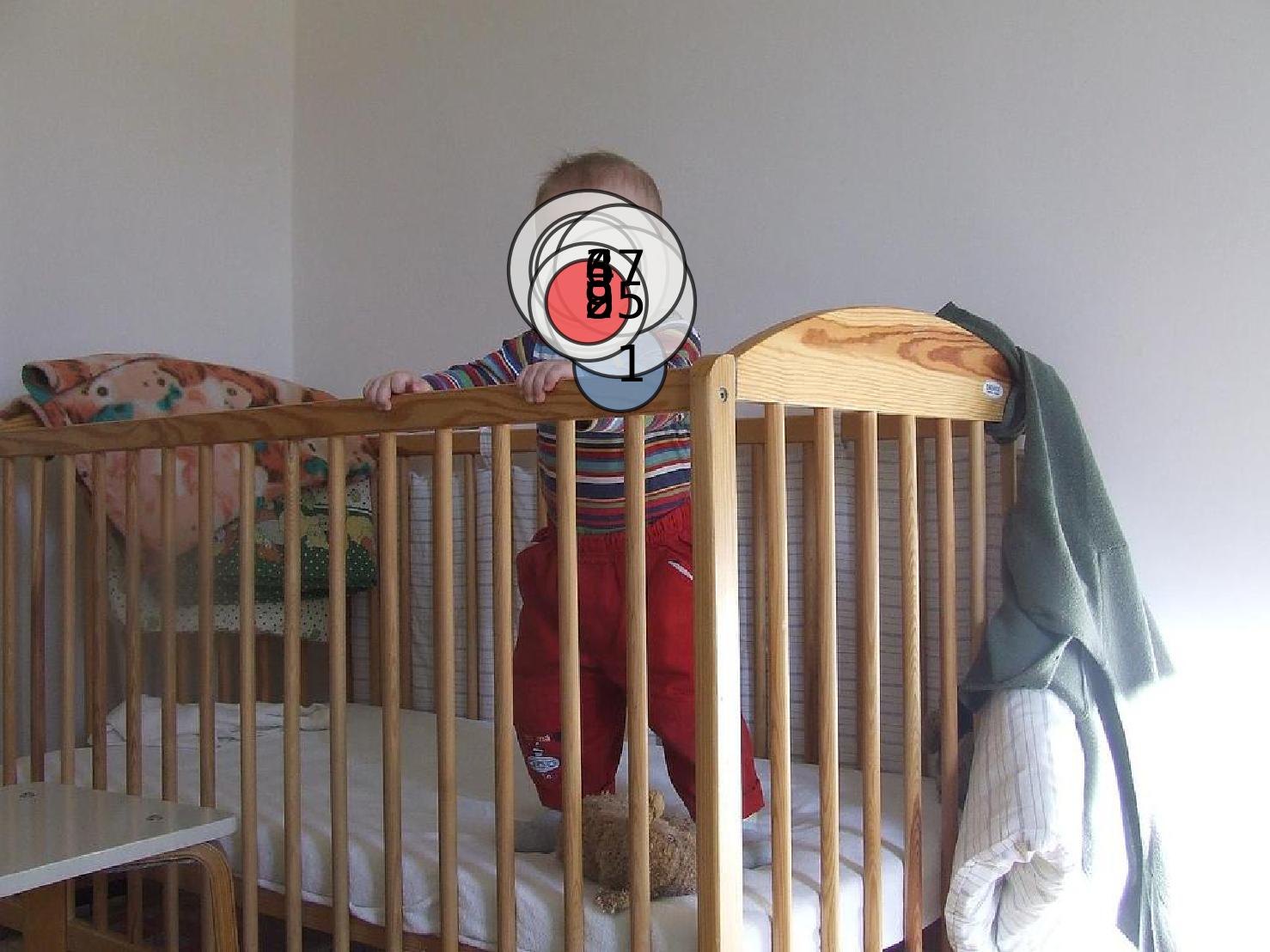} & \includegraphics[width=0.14\linewidth]{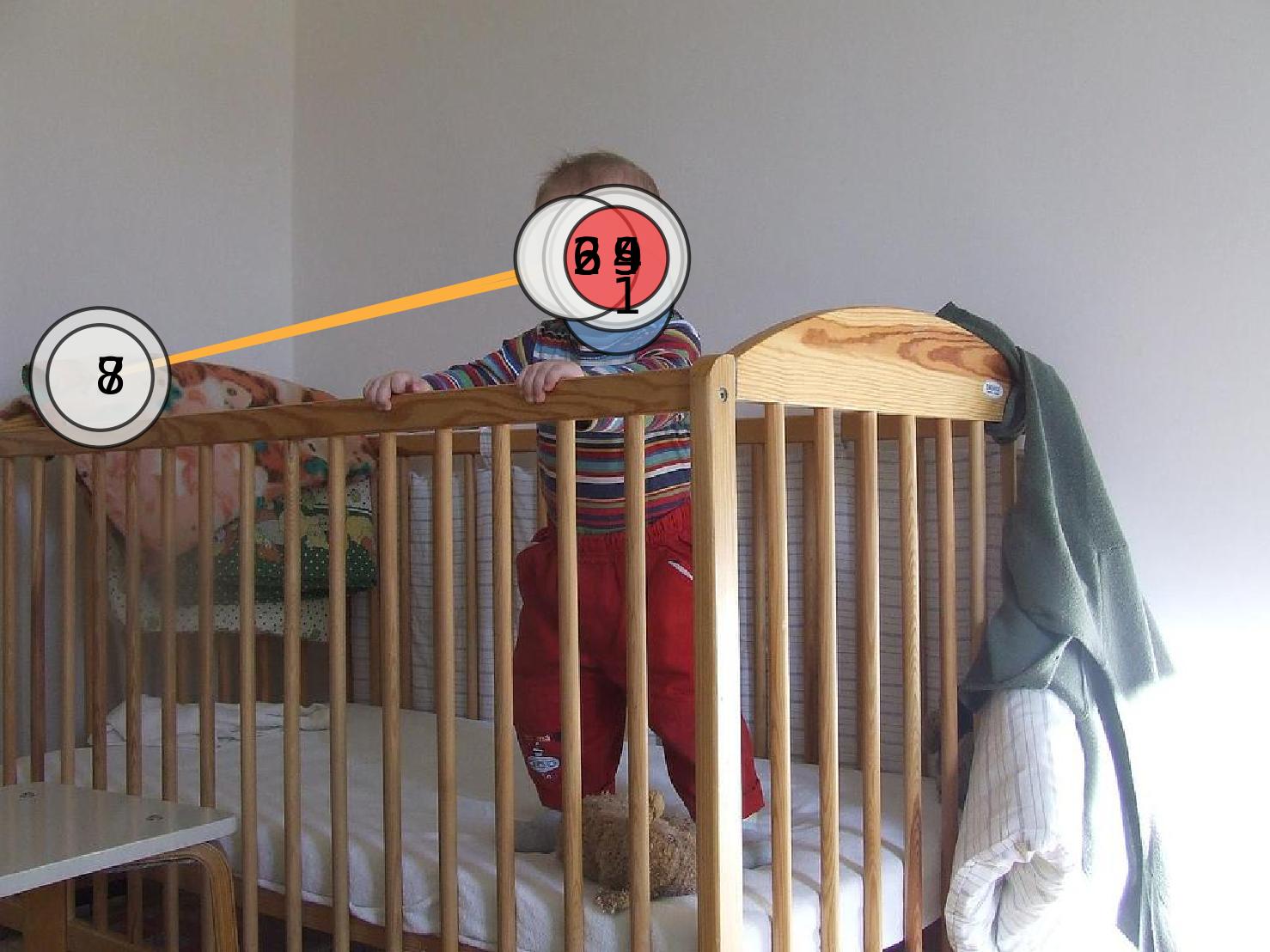} & \includegraphics[width=0.14\linewidth]{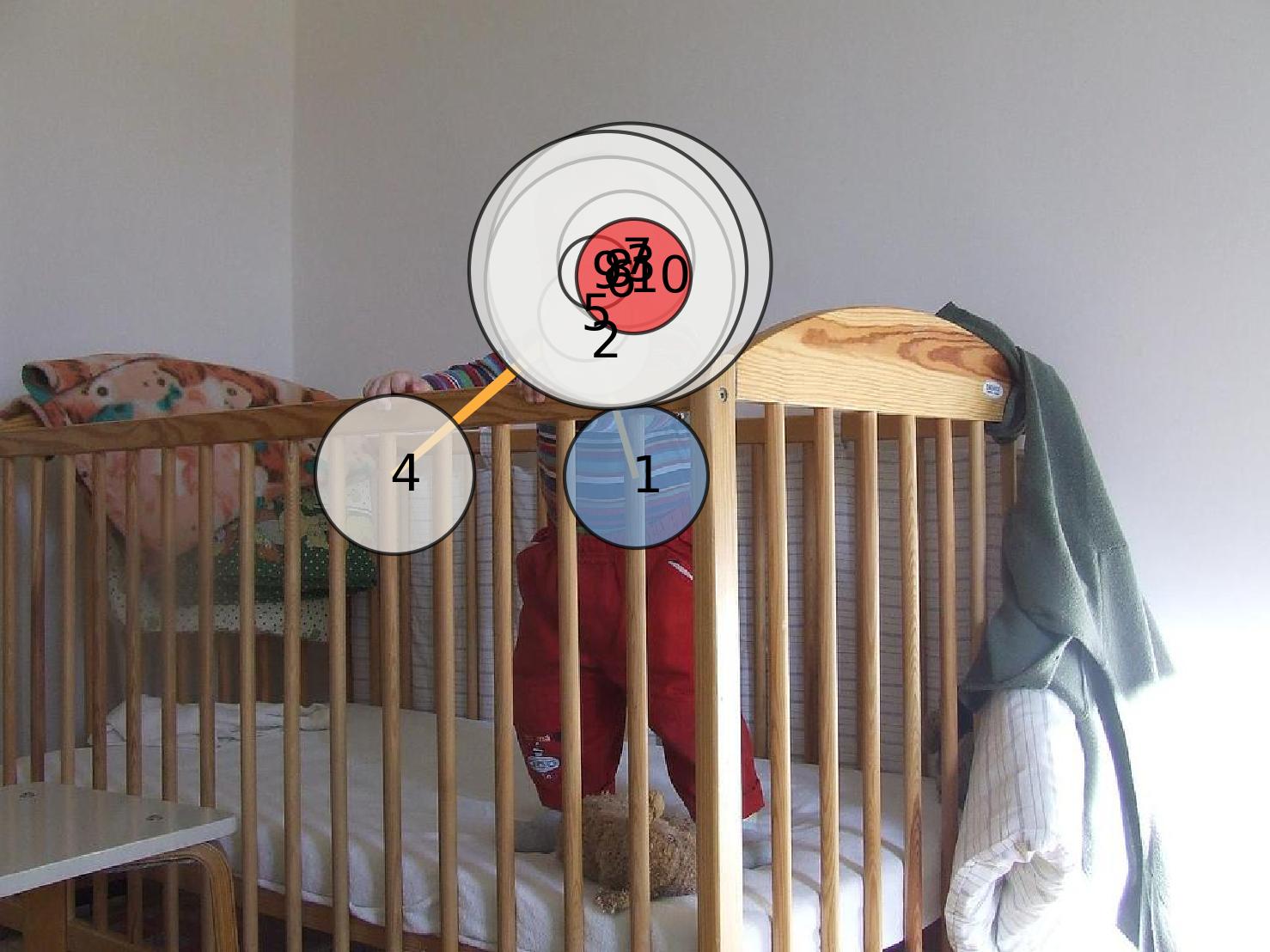} & \includegraphics[width=0.14\linewidth]{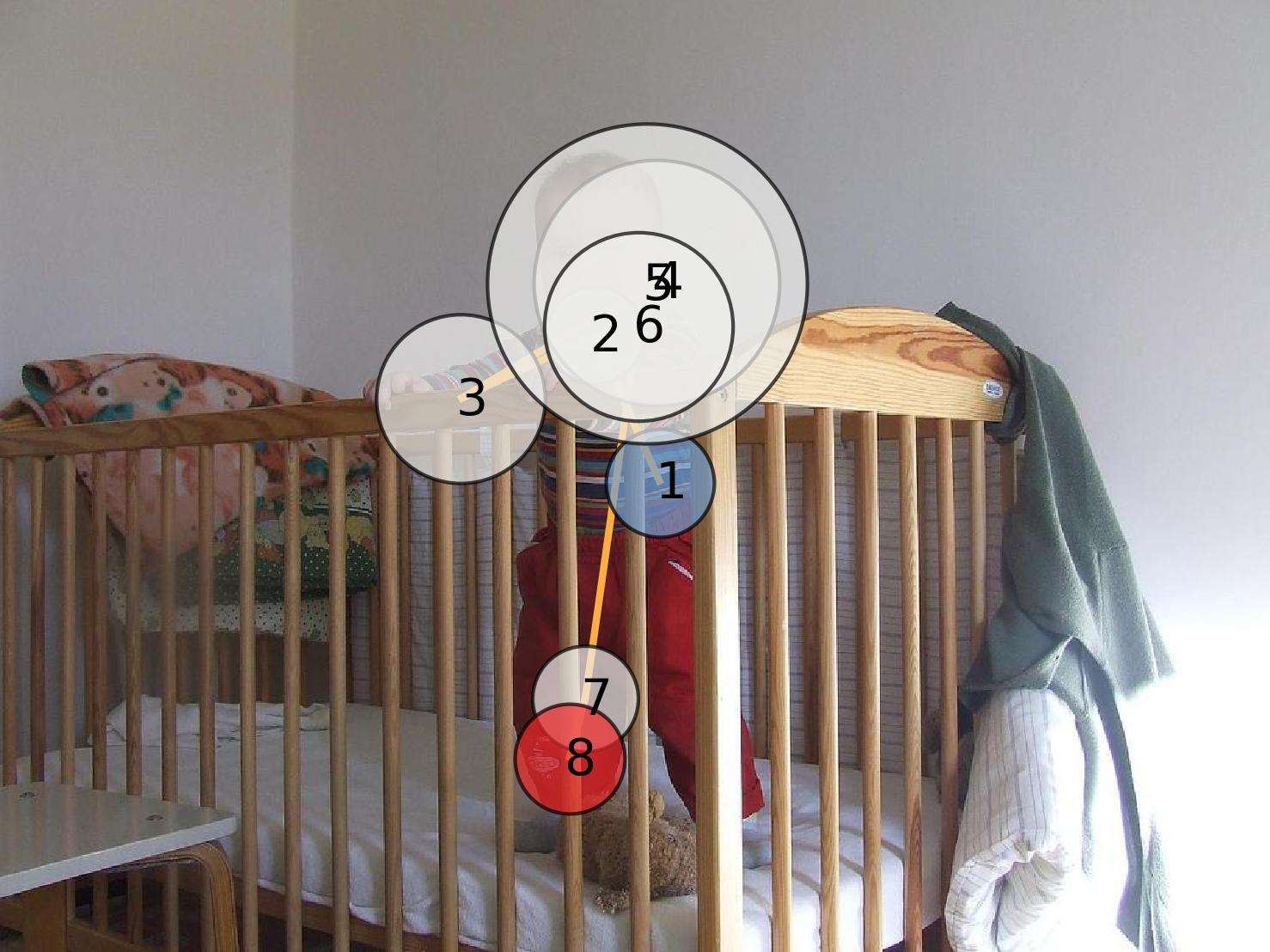} & \includegraphics[width=0.14\linewidth]{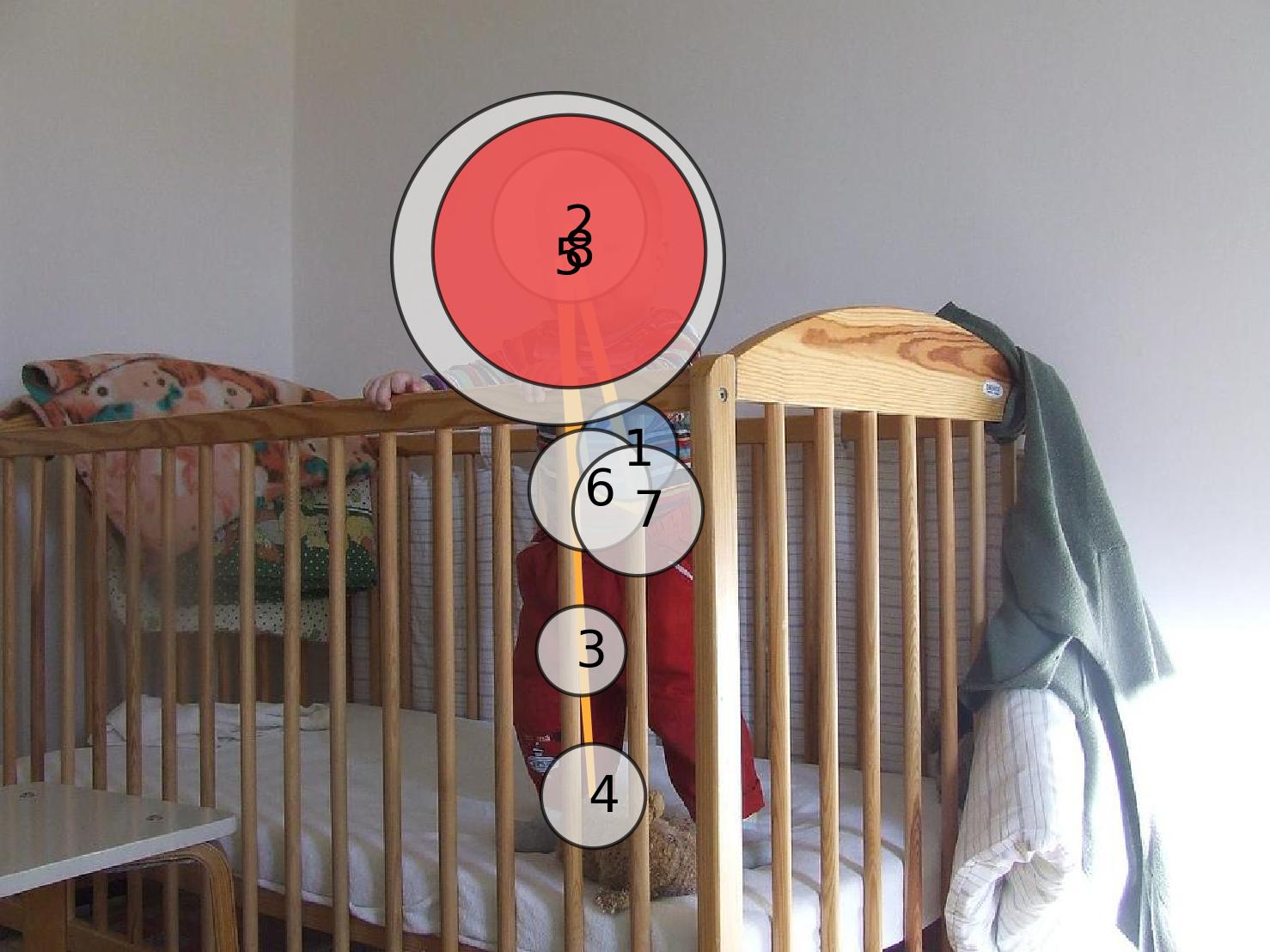} \\
    \end{tabular}
    }
    \vspace{-0.15cm}
    \caption{Qualitative comparison of the variability in simulated and human scanpaths on the MIT1003 dataset. Each row corresponds to a different simulation or a different human observer.}
    \label{fig:qualitatives_variab_mit1003}
    \vspace{-0.4cm}
\end{figure*}

\begin{figure*}[t]
    \footnotesize
    \setlength{\tabcolsep}{.1em}
    \resizebox{\linewidth}{!}{
    \begin{tabular}{cccccc}
        \scriptsize ChenLSTM~\cite{chen2021predicting} & \scriptsize Gazeformer~\cite{mondal2023gazeformer} & \scriptsize GazeXplain~\cite{chen2024gazexplain} & \scriptsize TPP-Gaze~\cite{damelio2025tpp} & \scriptsize \ours (Ours) & \scriptsize Humans \\
        \includegraphics[width=0.15\linewidth]{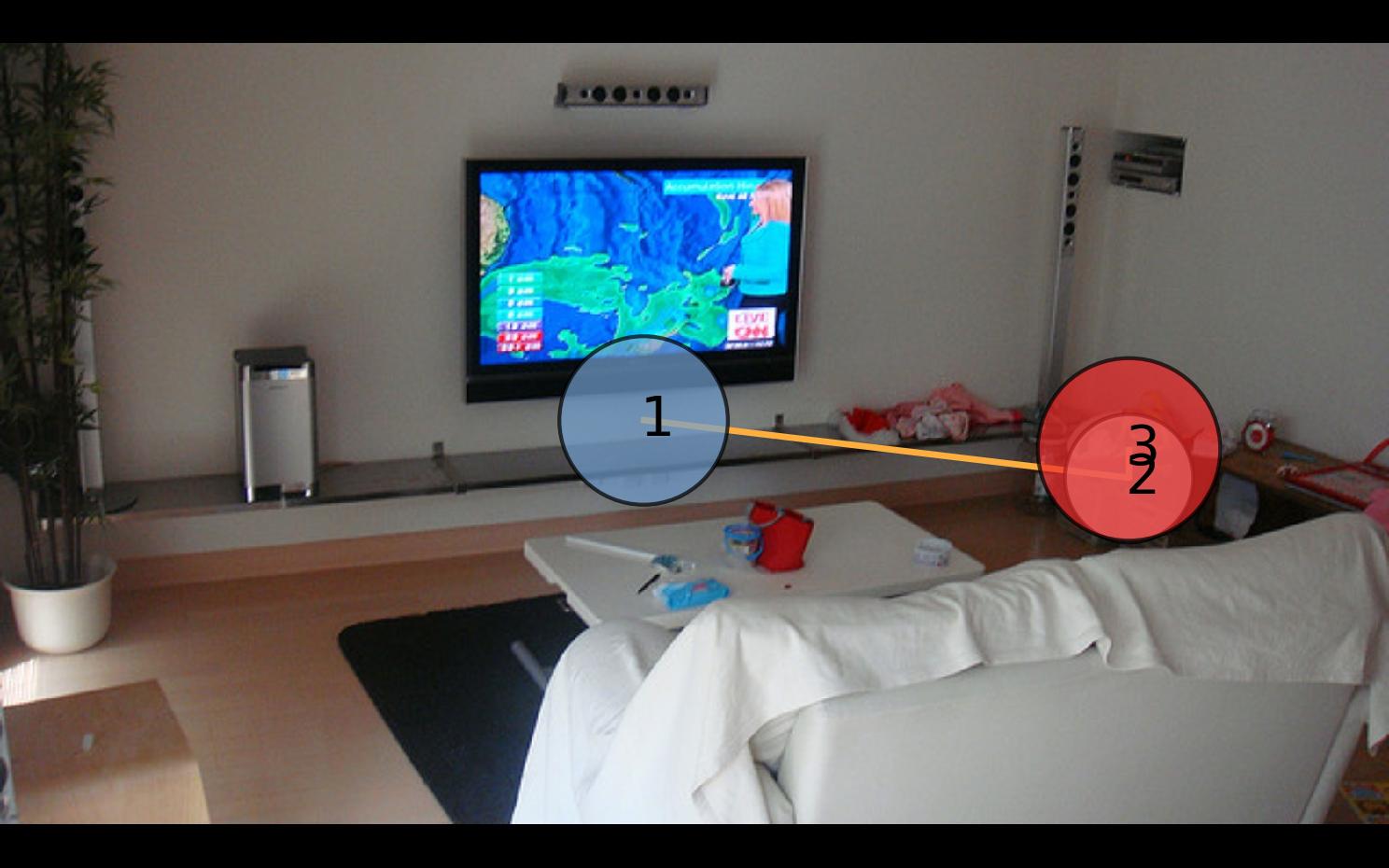} & \includegraphics[width=0.15\linewidth]{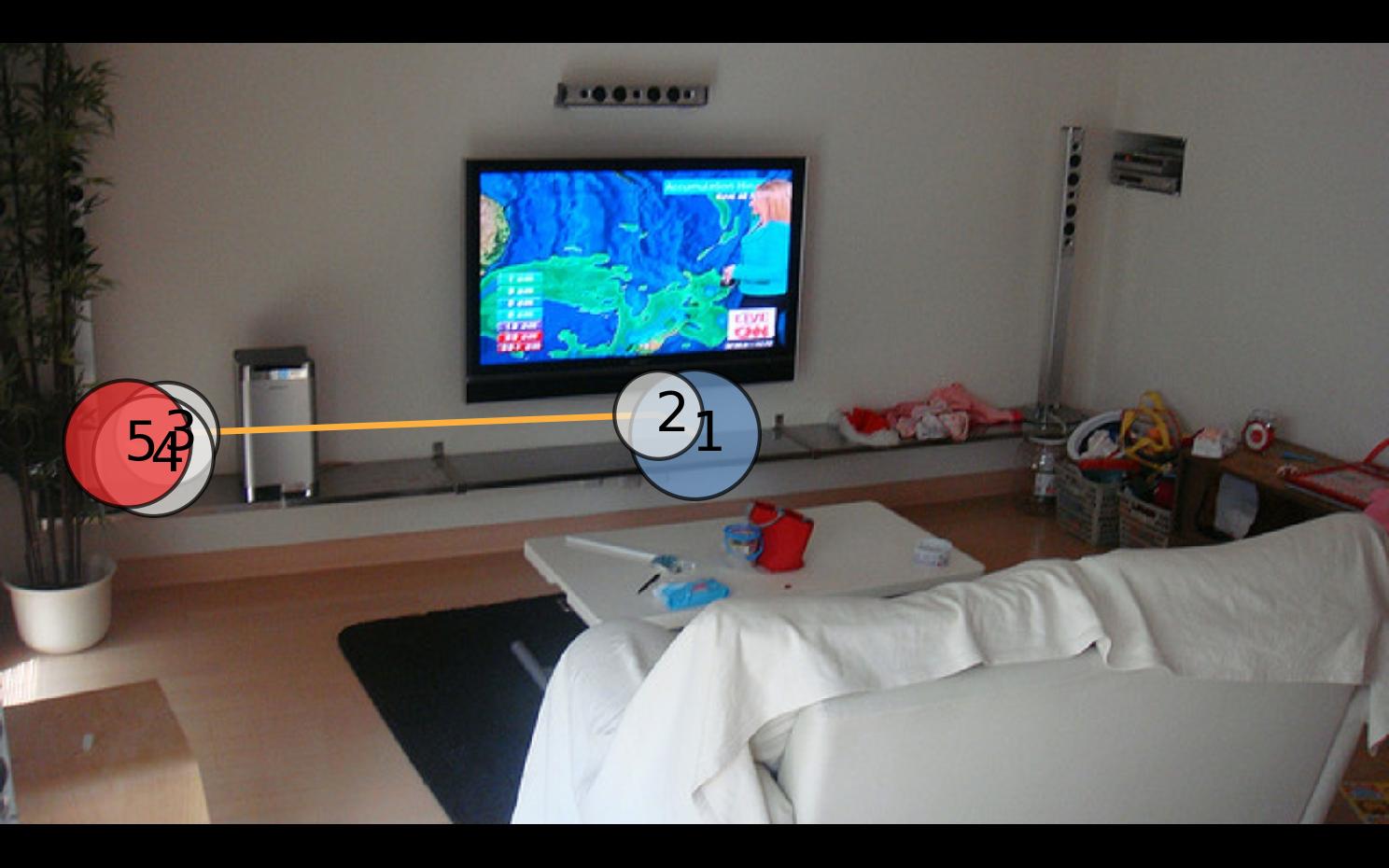} & \includegraphics[width=0.15\linewidth]{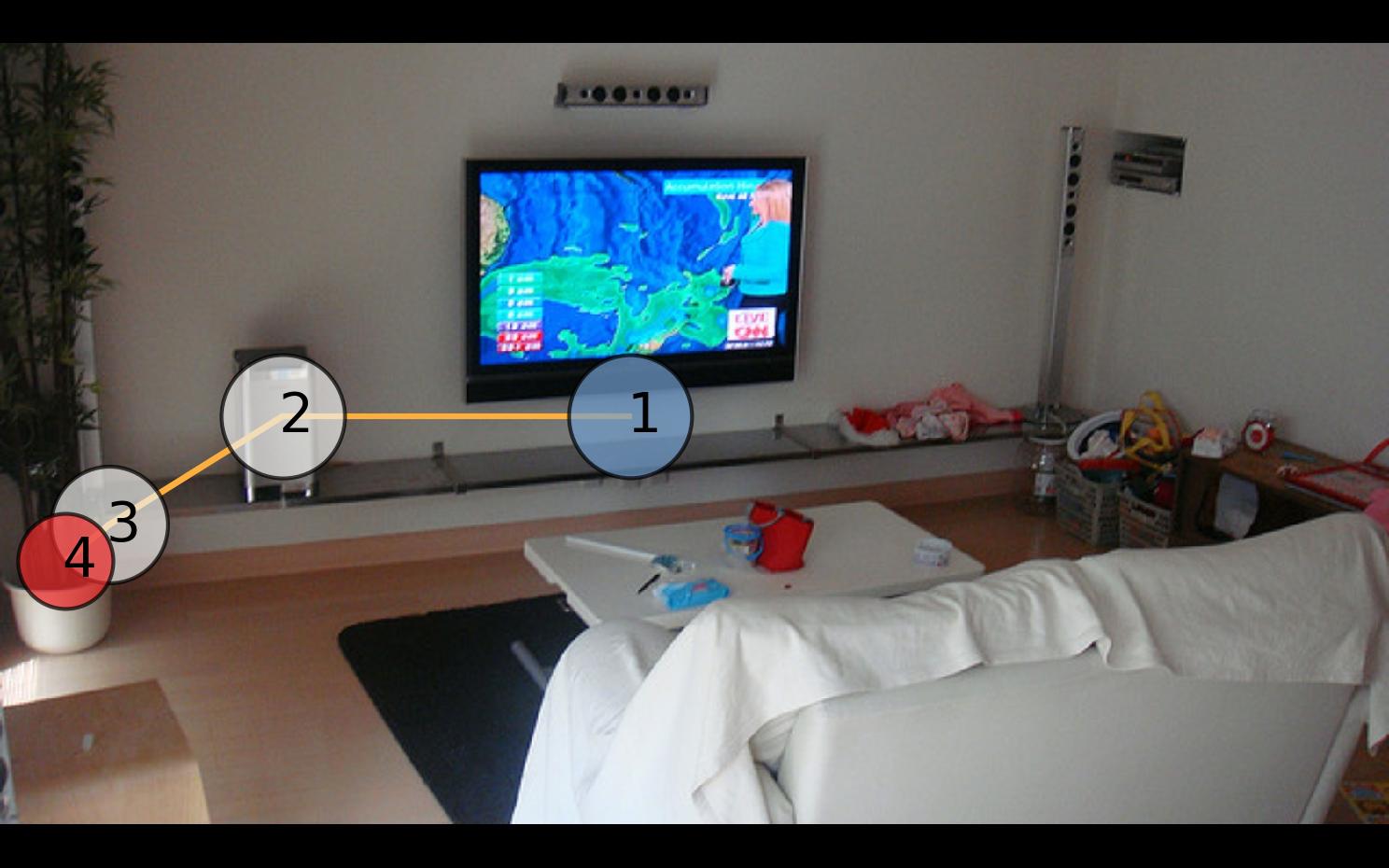} & \includegraphics[width=0.15\linewidth]{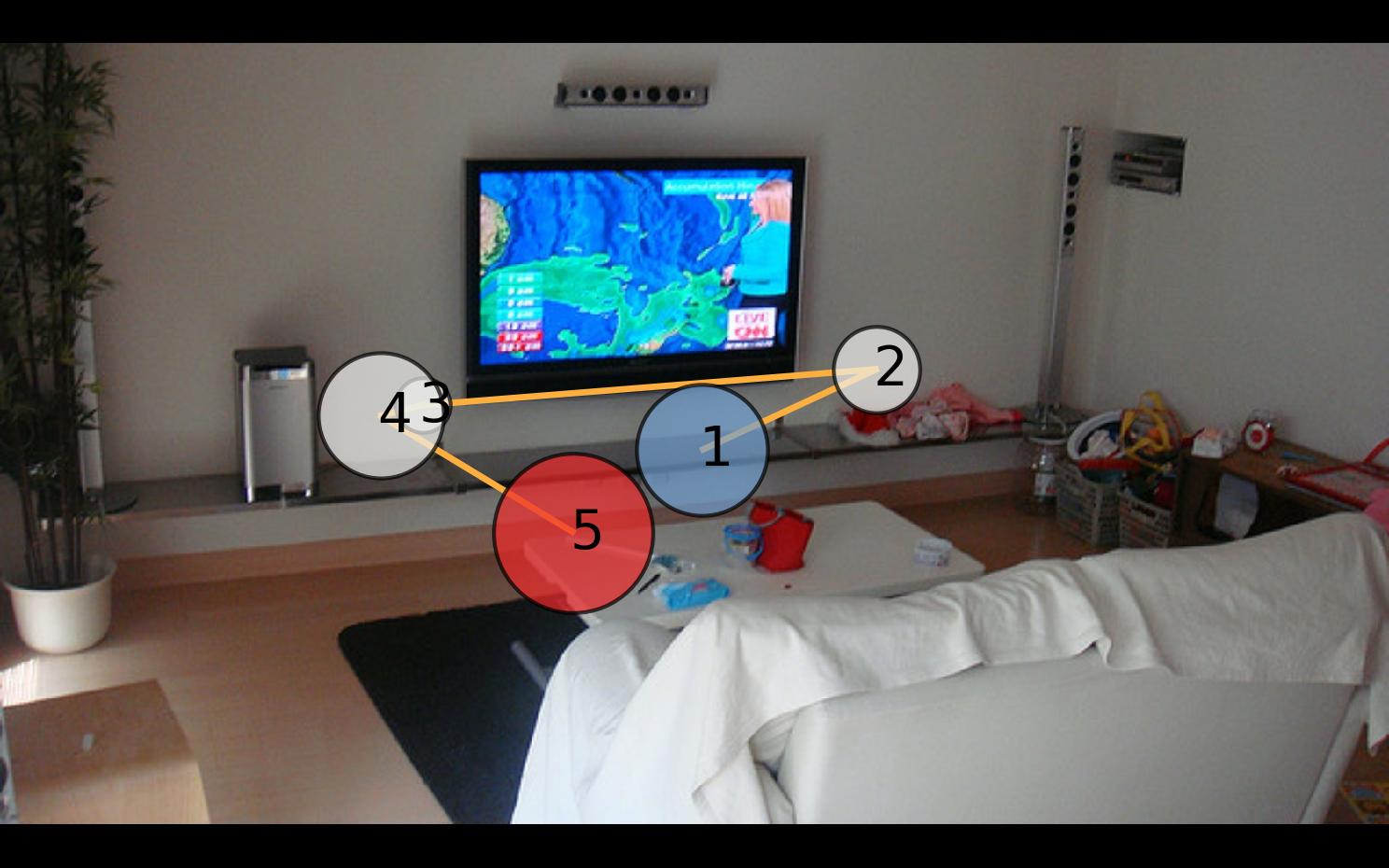} & \includegraphics[width=0.15\linewidth]{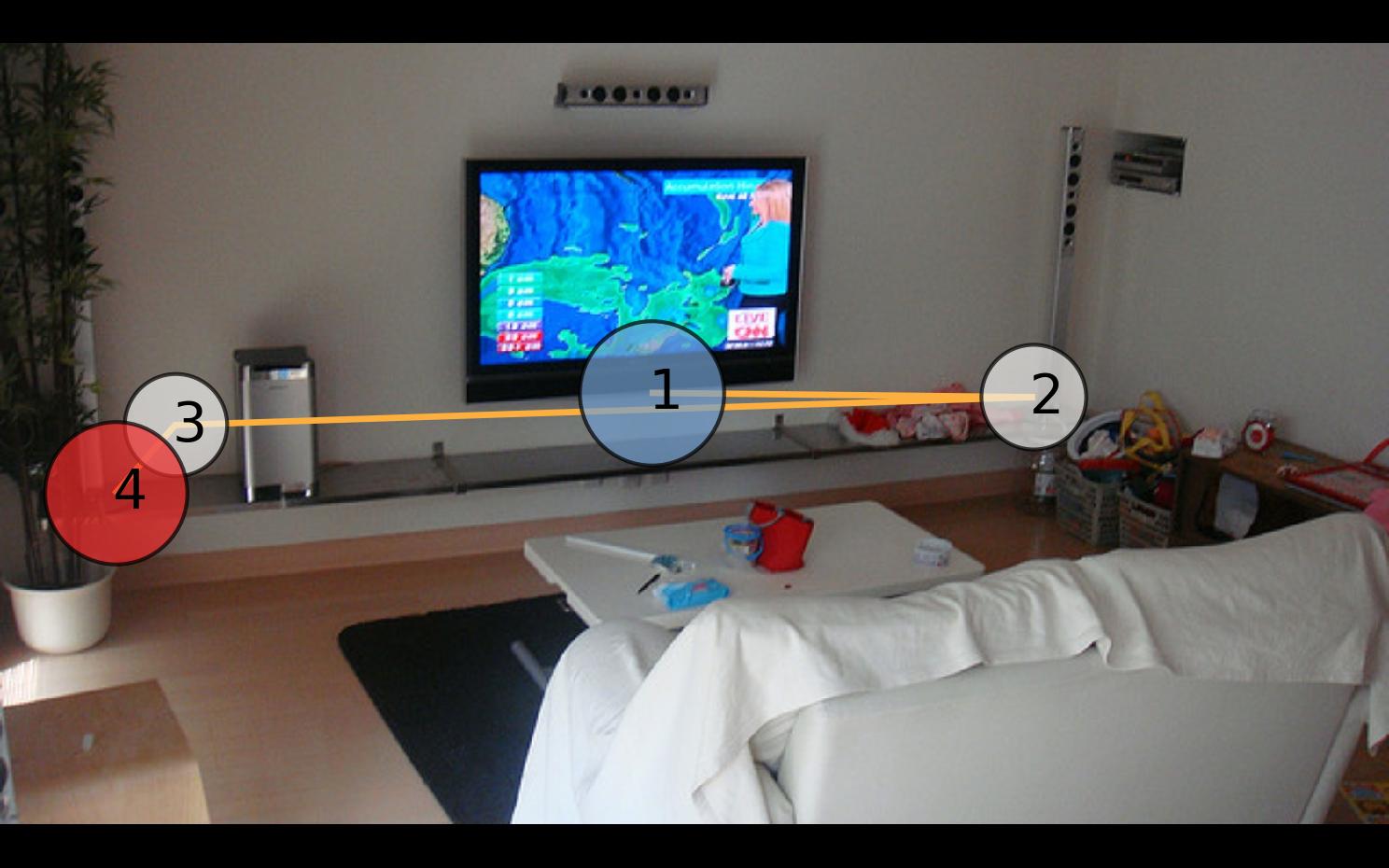} & \includegraphics[width=0.15\linewidth]{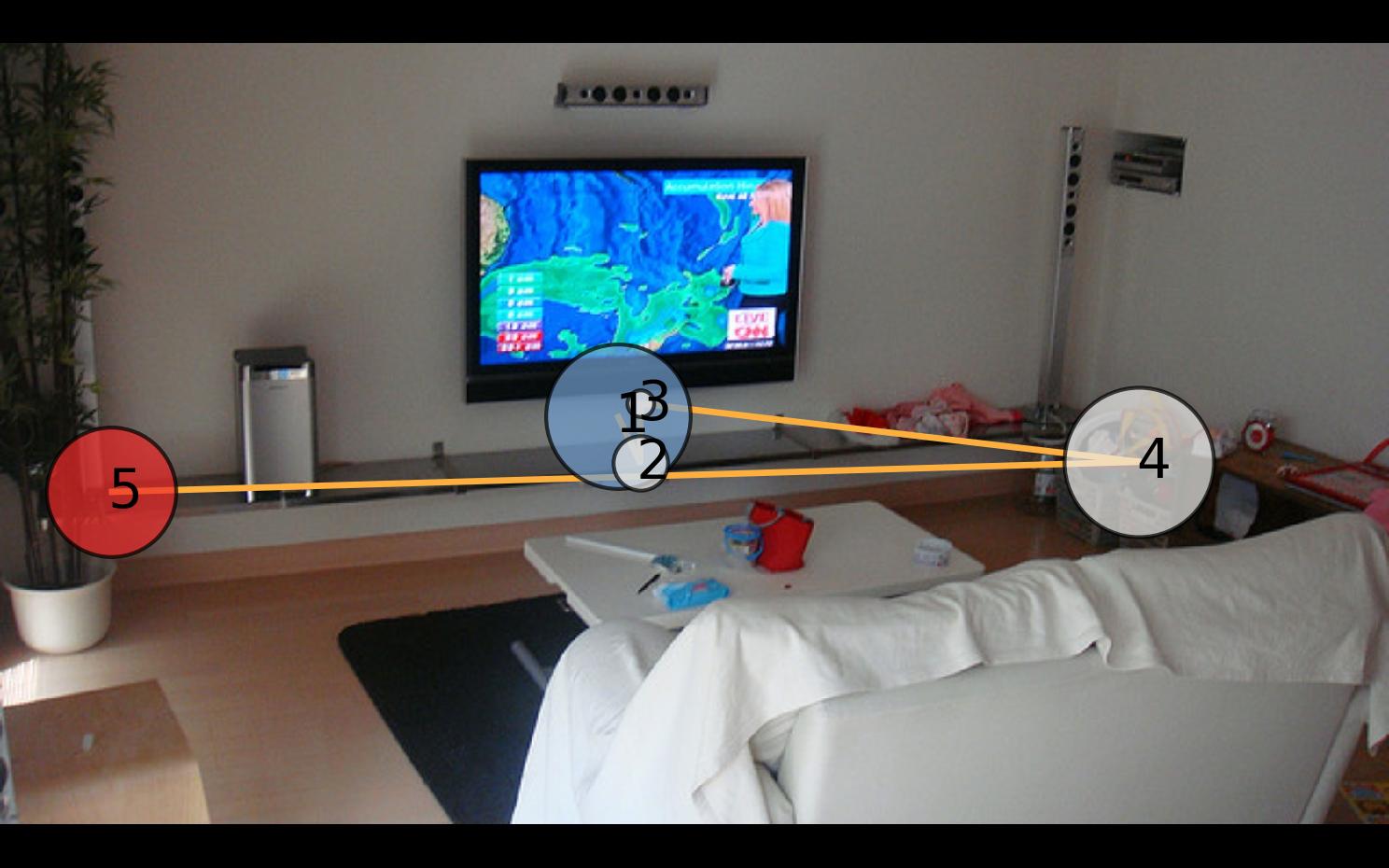} \\
        \includegraphics[width=0.15\linewidth]{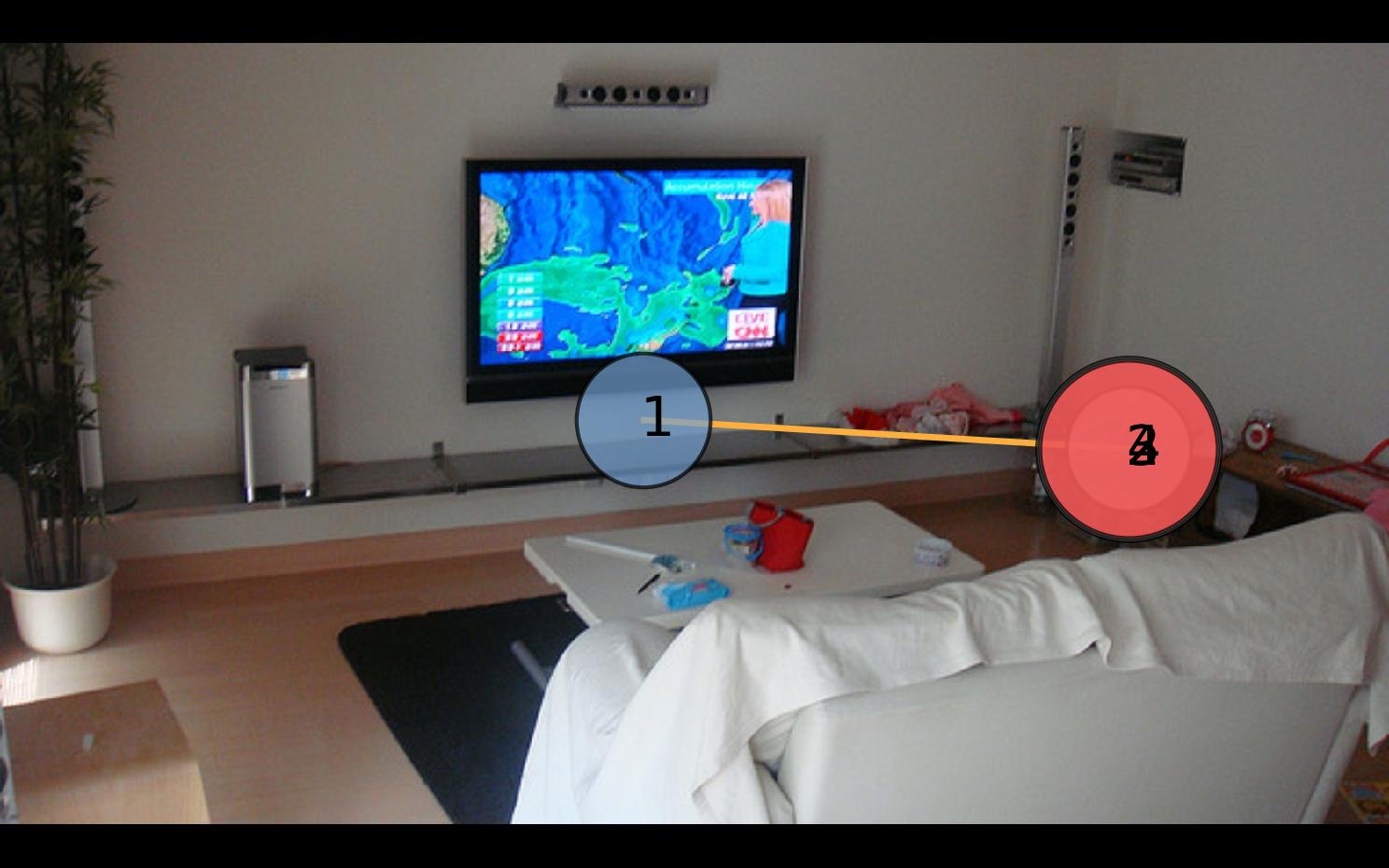} & \includegraphics[width=0.15\linewidth]{images/variabilitycocotp/gazeformer1.jpg} & \includegraphics[width=0.15\linewidth]{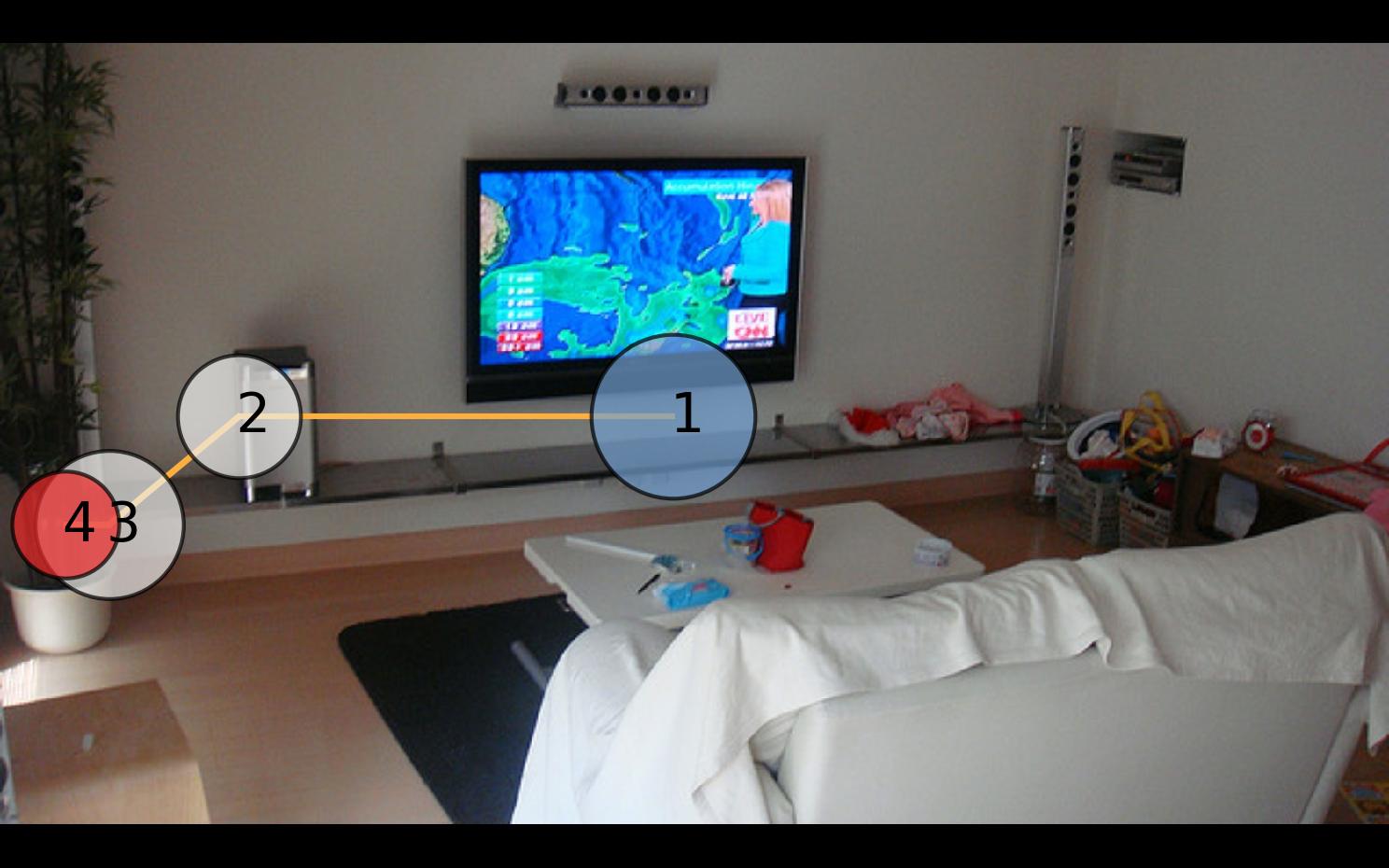} & \includegraphics[width=0.15\linewidth]{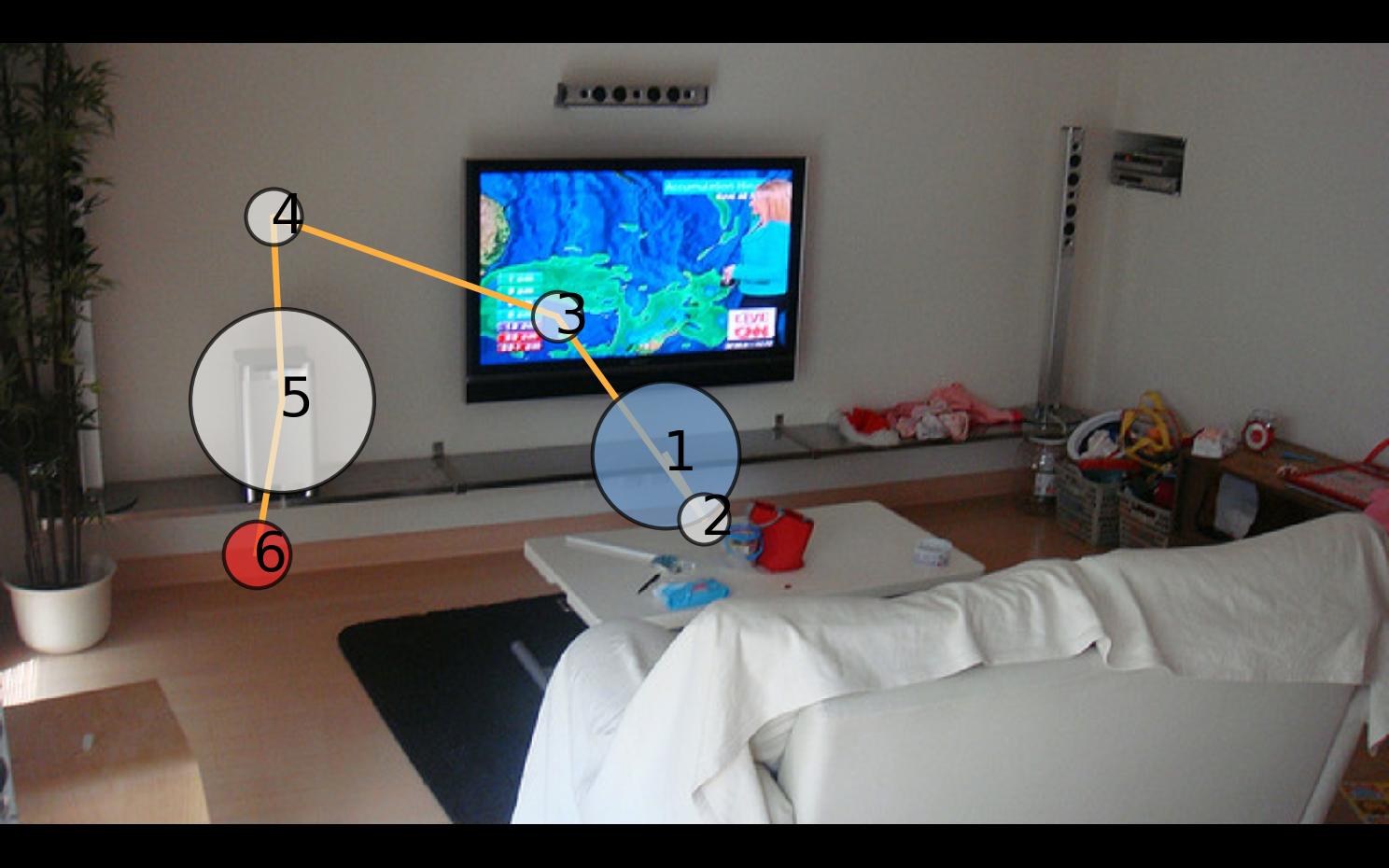} & \includegraphics[width=0.15\linewidth]{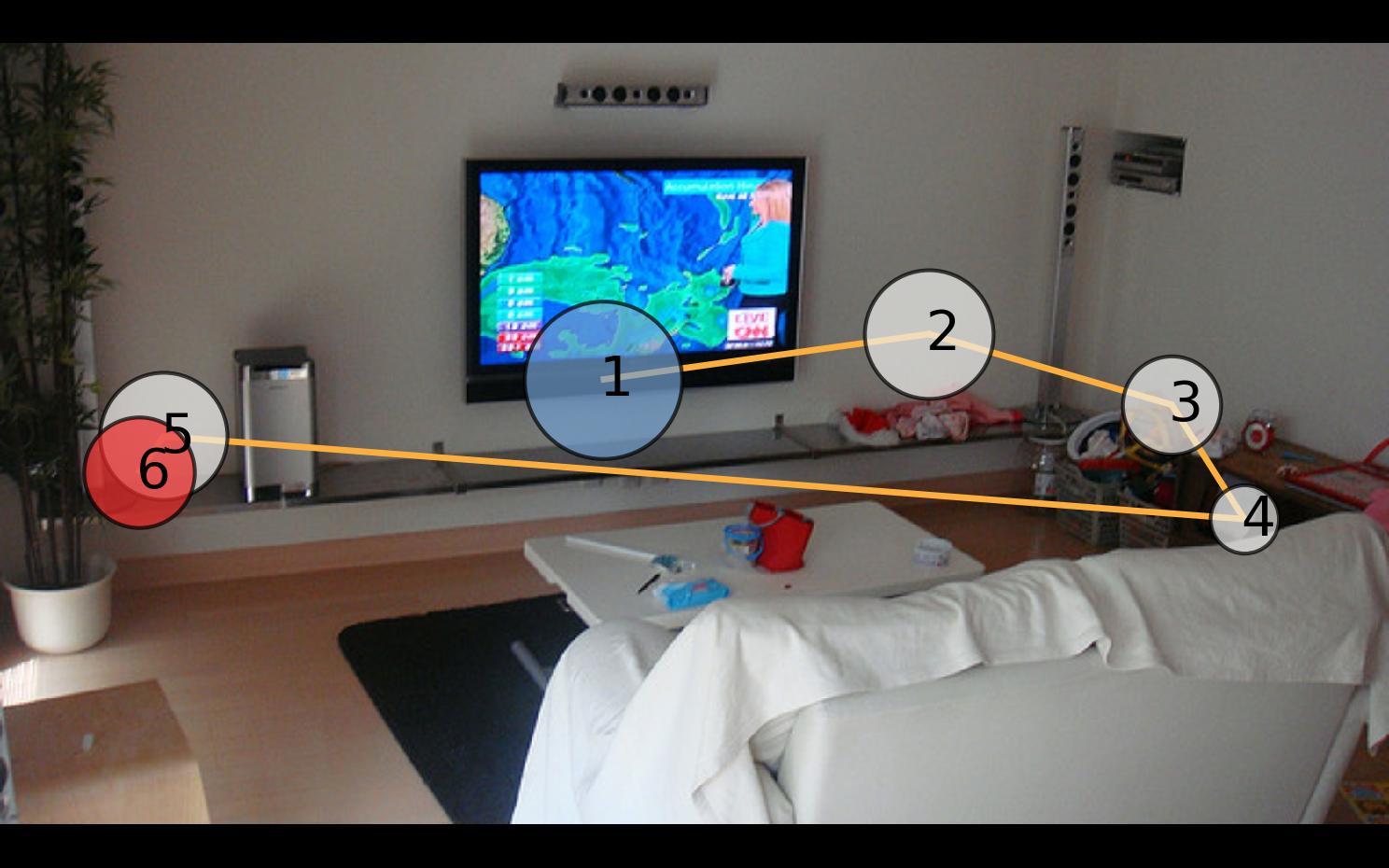} & \includegraphics[width=0.15\linewidth]{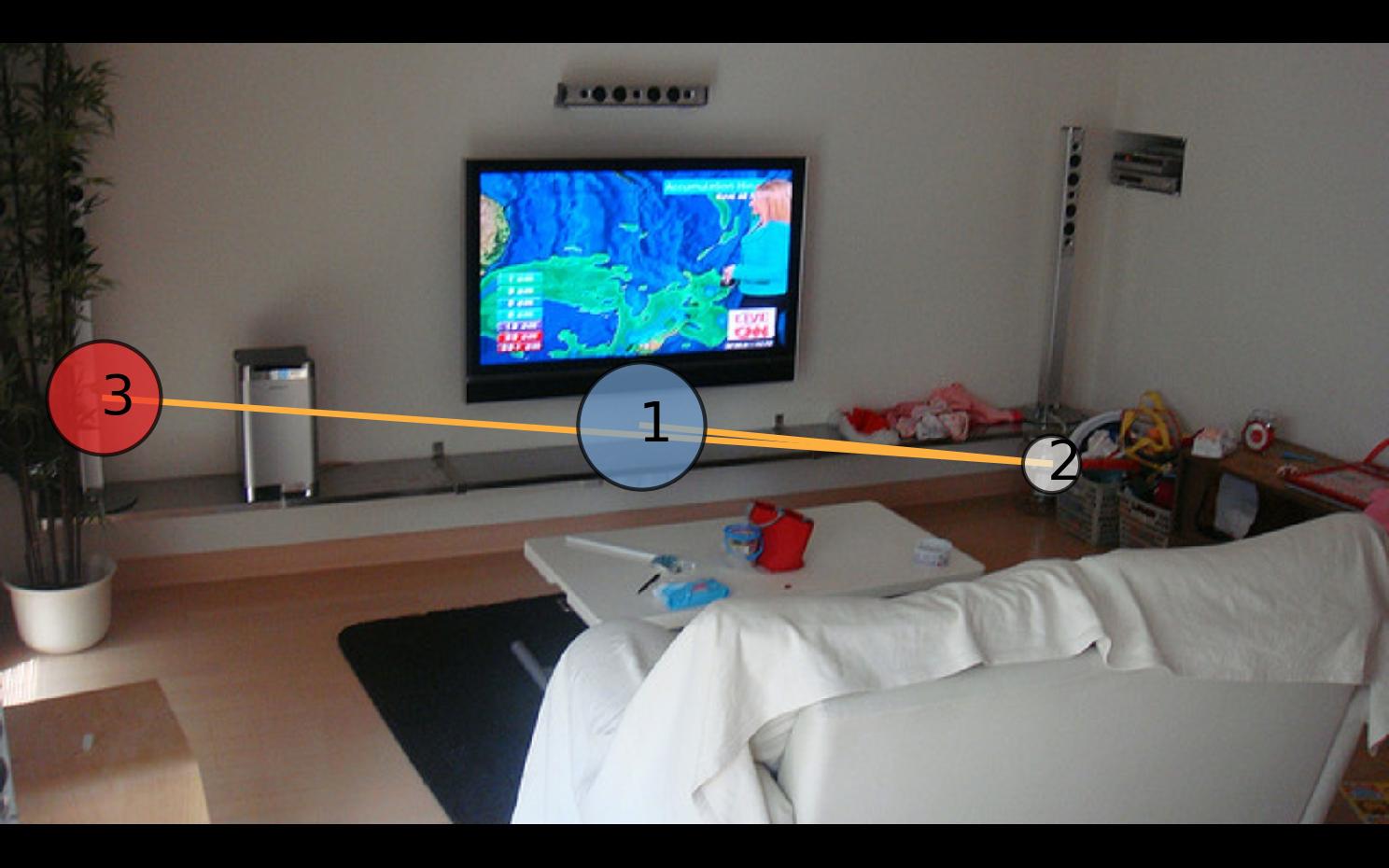} \\
        \includegraphics[width=0.15\linewidth]{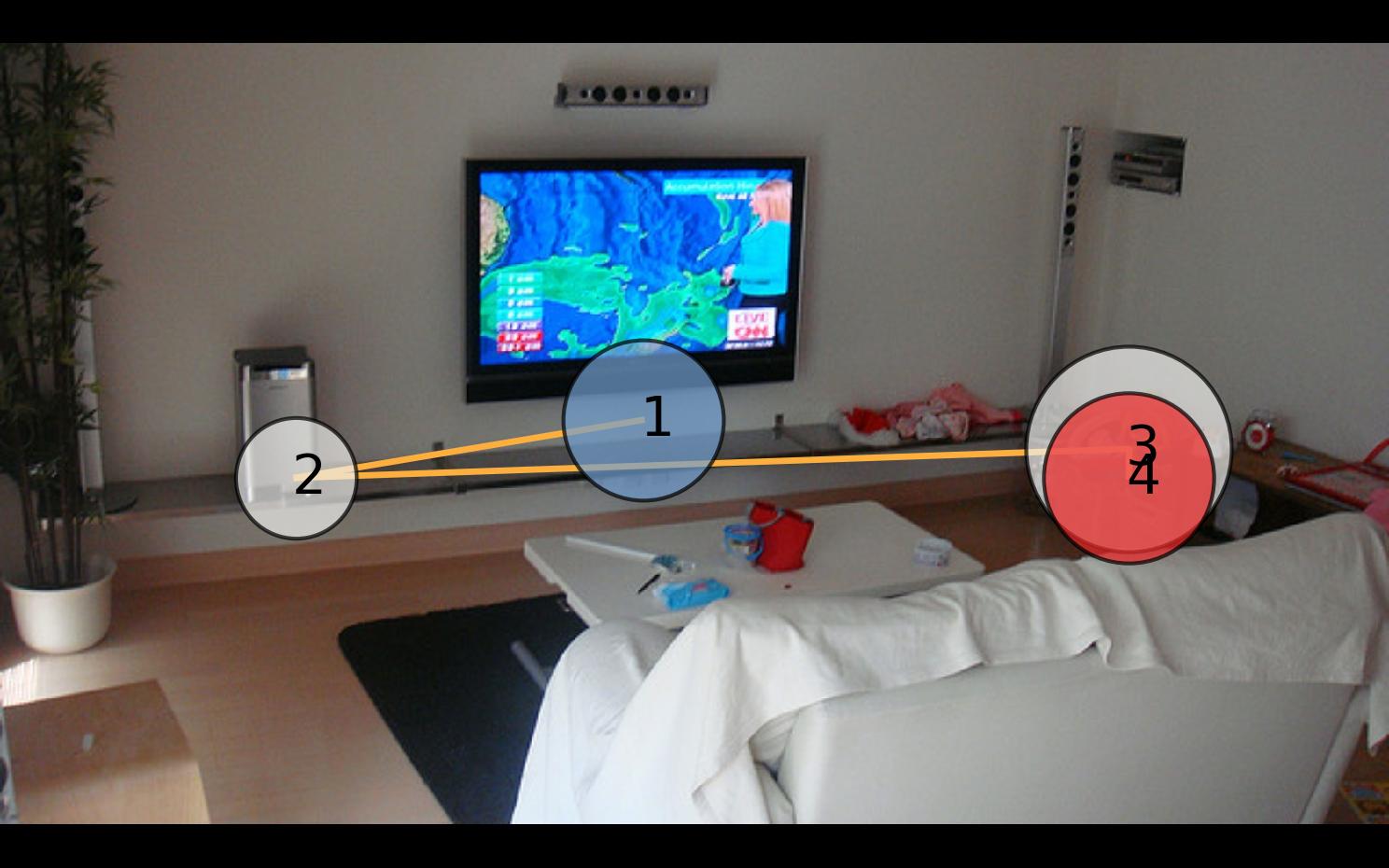} & \includegraphics[width=0.15\linewidth]{images/variabilitycocotp/gazeformer1.jpg} & \includegraphics[width=0.15\linewidth]{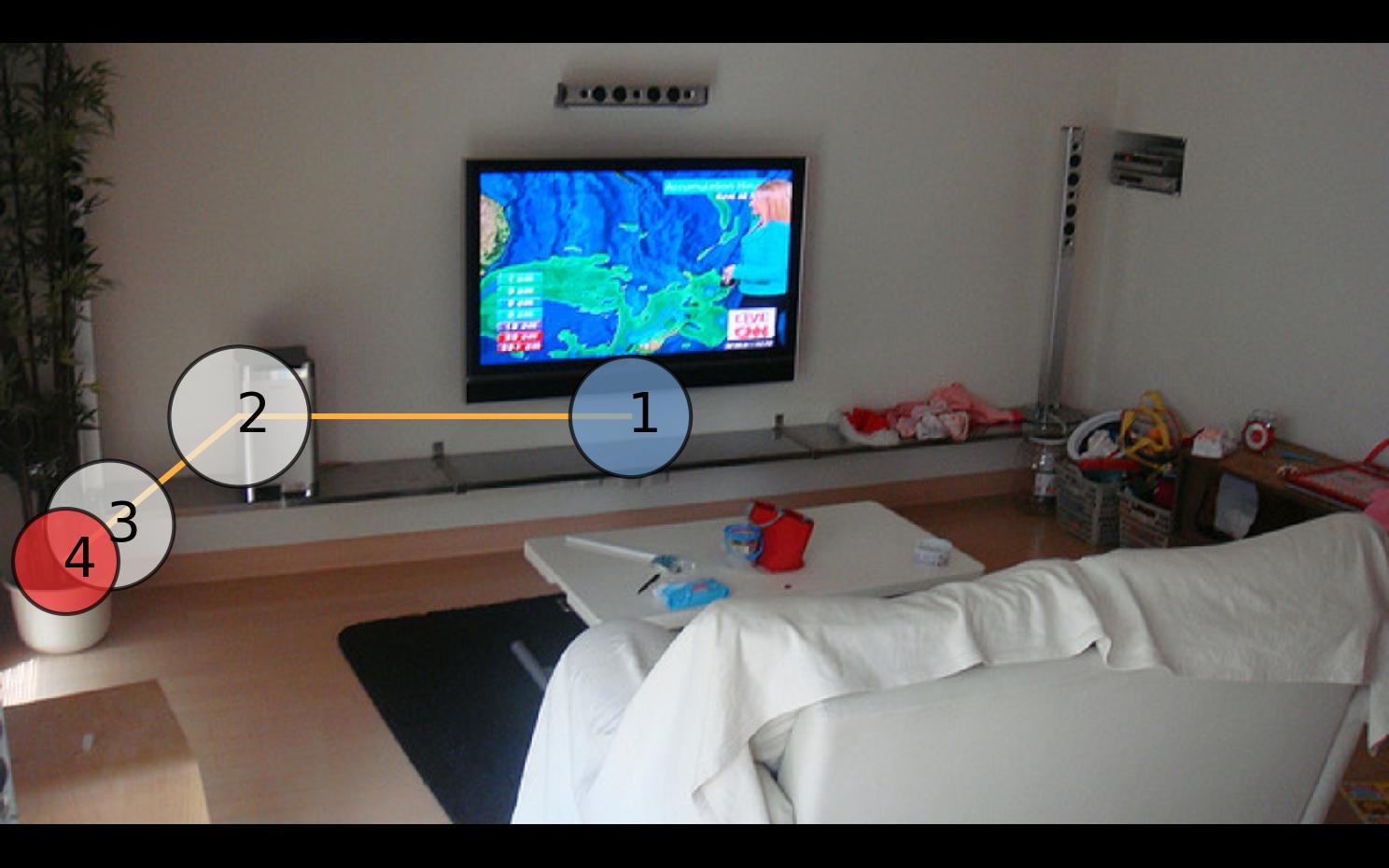} & \includegraphics[width=0.15\linewidth]{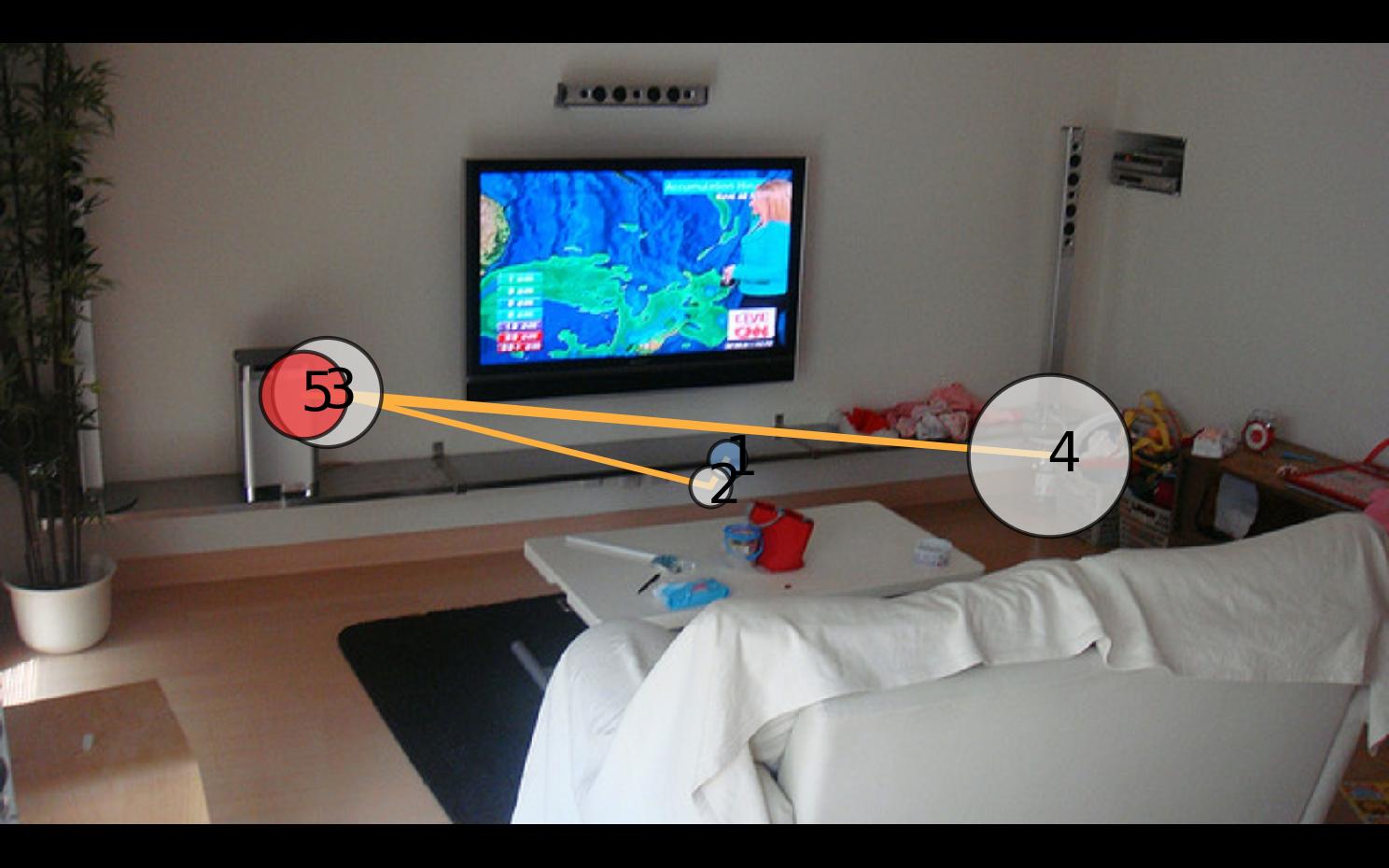} & \includegraphics[width=0.15\linewidth]{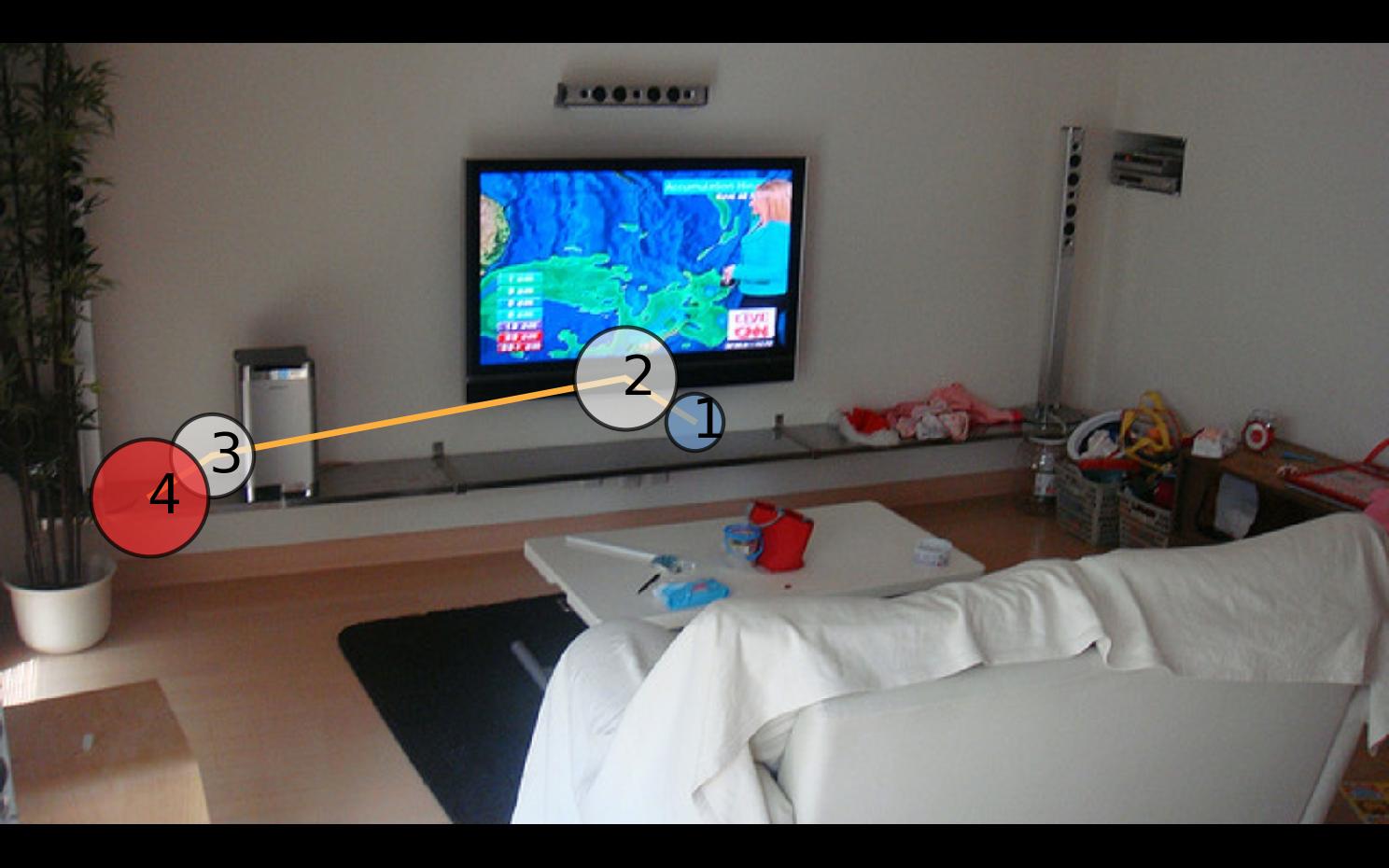} & \includegraphics[width=0.15\linewidth]{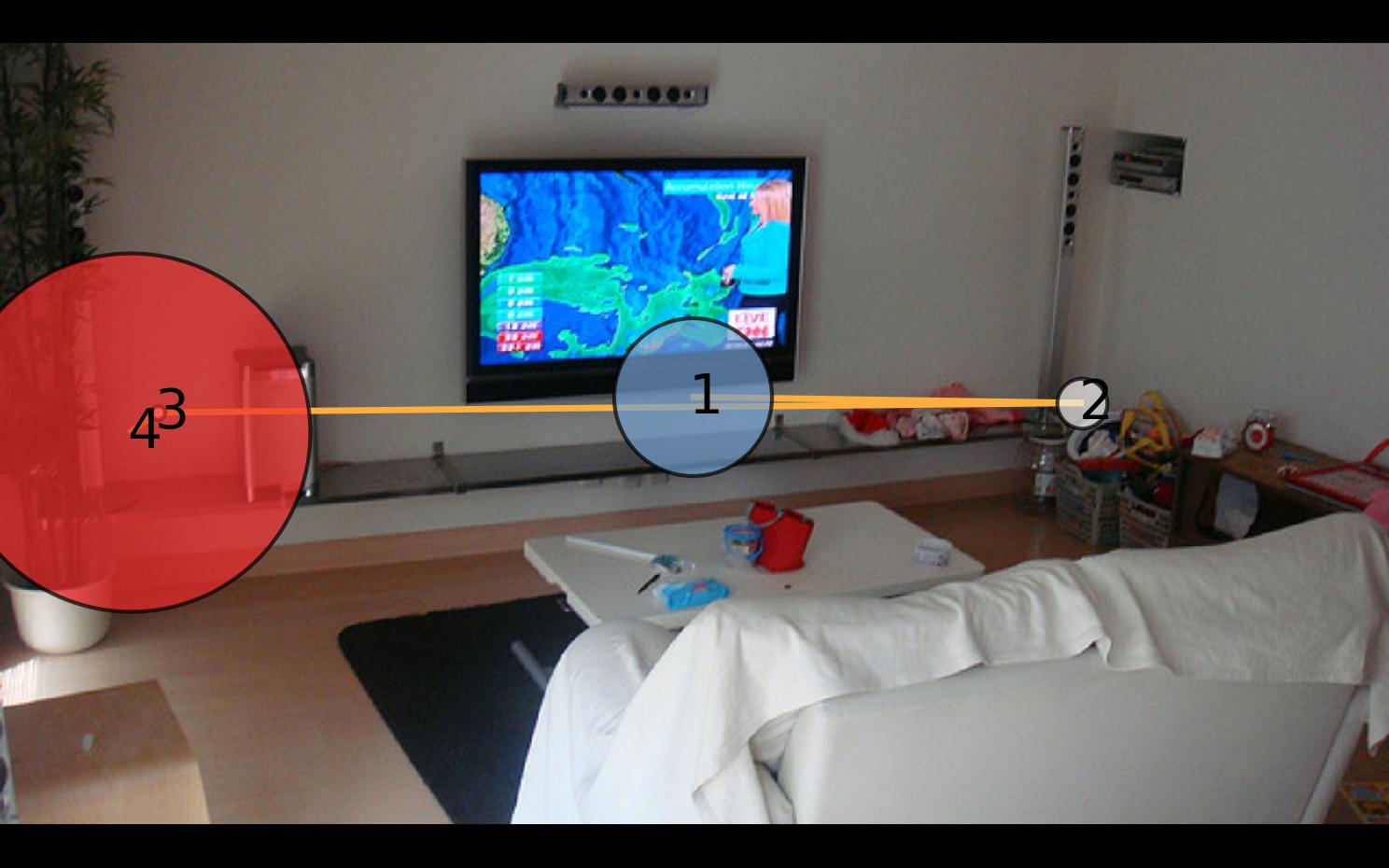} \\
        \includegraphics[width=0.15\linewidth]{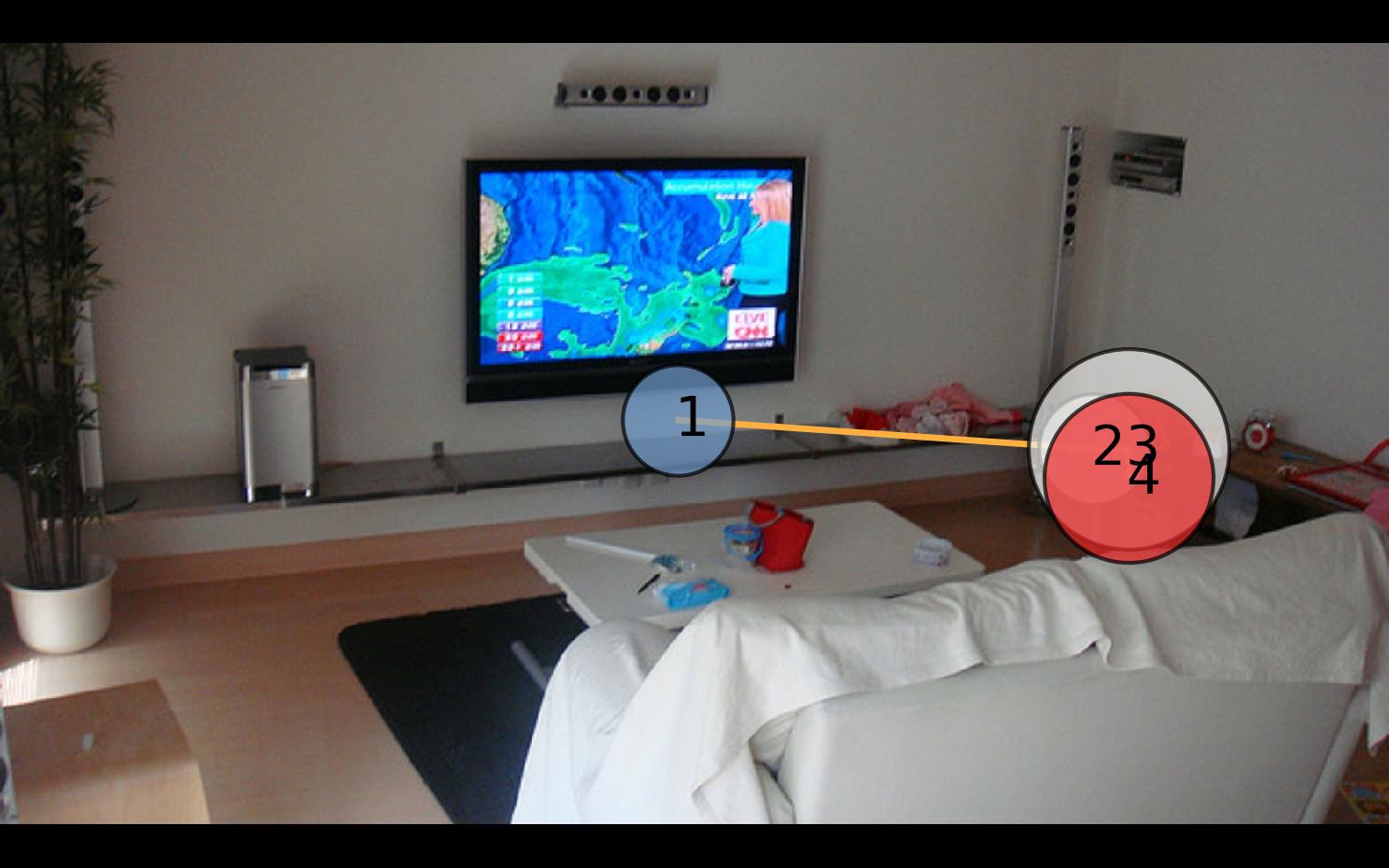} & \includegraphics[width=0.15\linewidth]{images/variabilitycocotp/gazeformer1.jpg} & \includegraphics[width=0.15\linewidth]{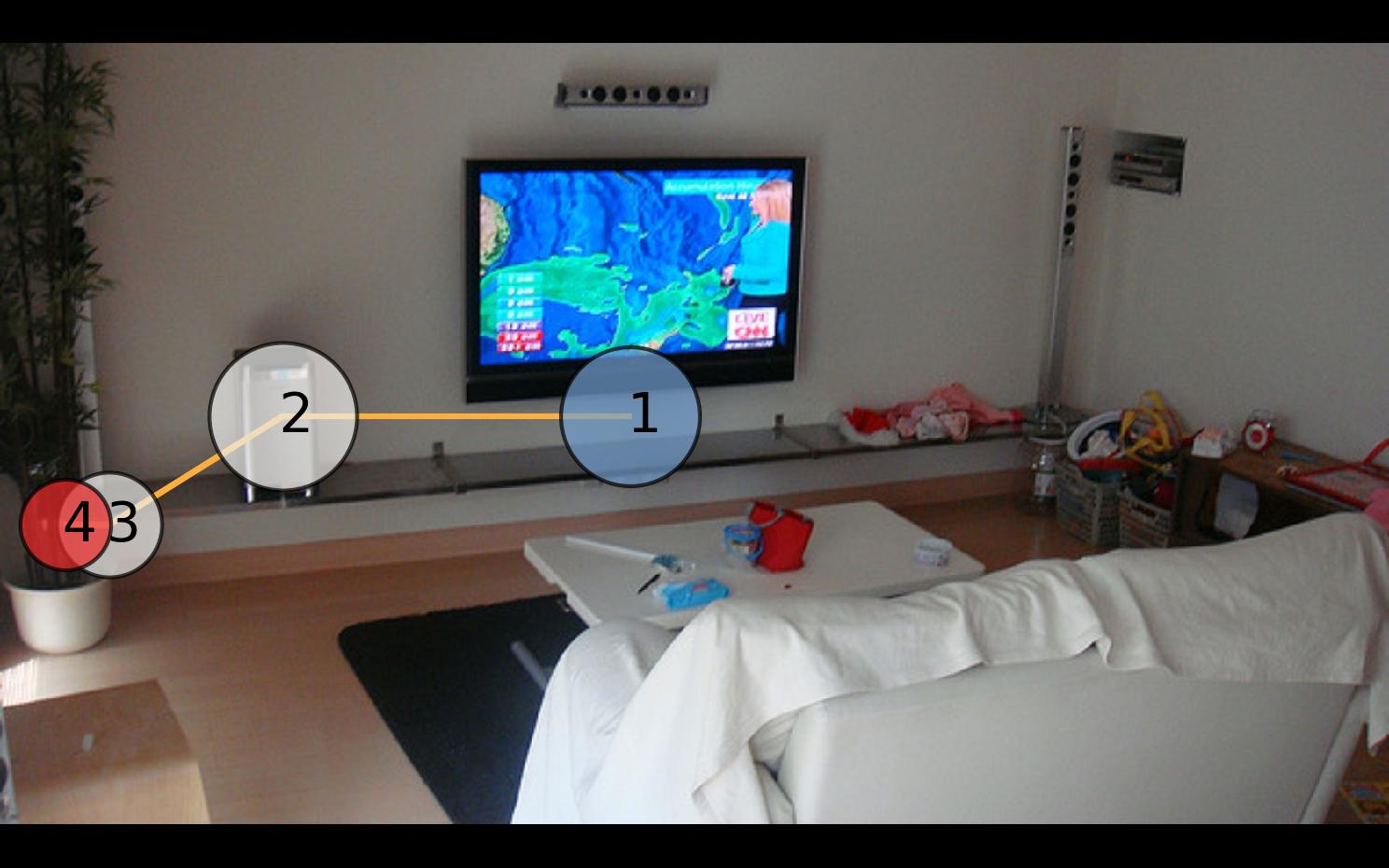} & \includegraphics[width=0.15\linewidth]{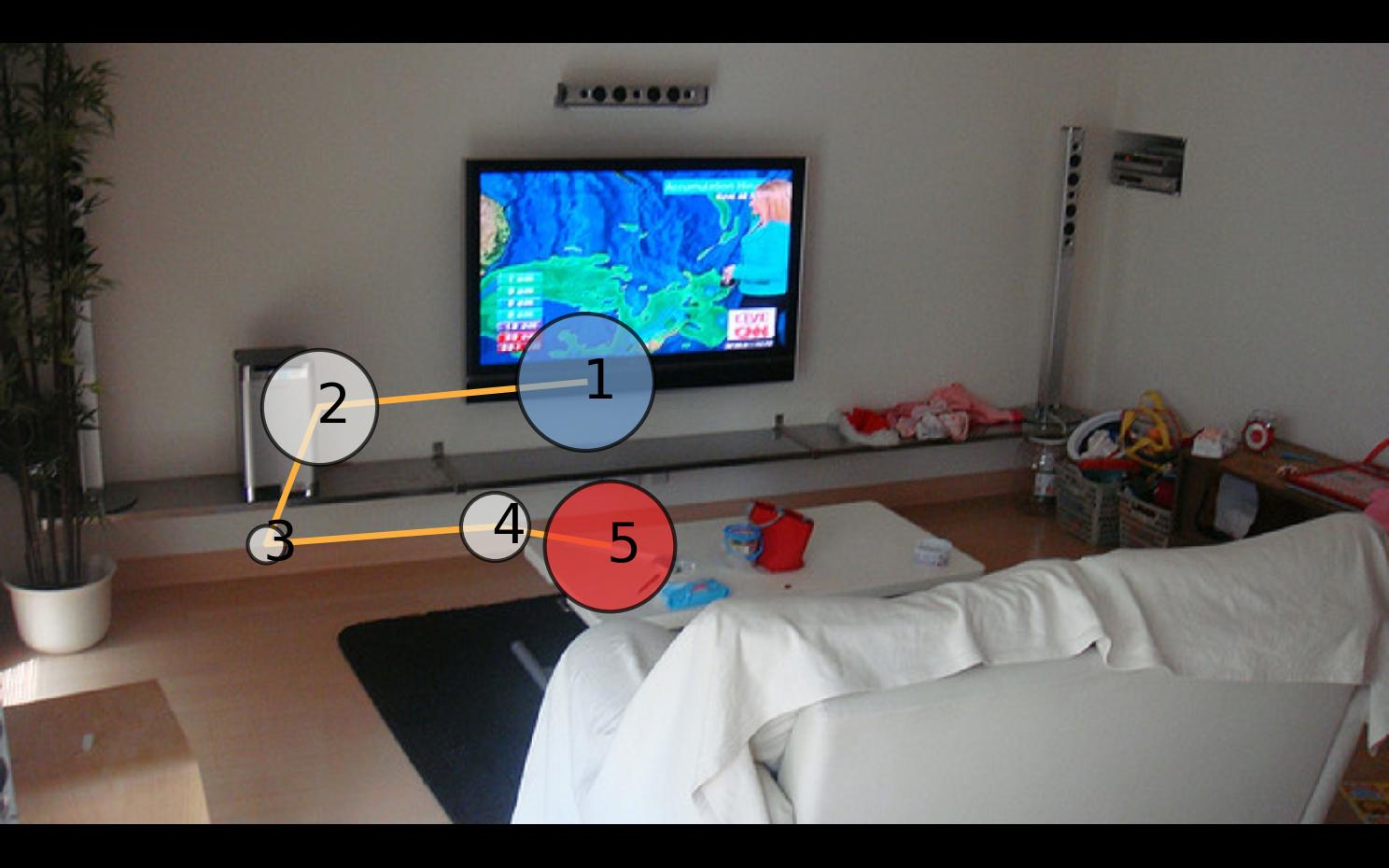} & \includegraphics[width=0.15\linewidth]{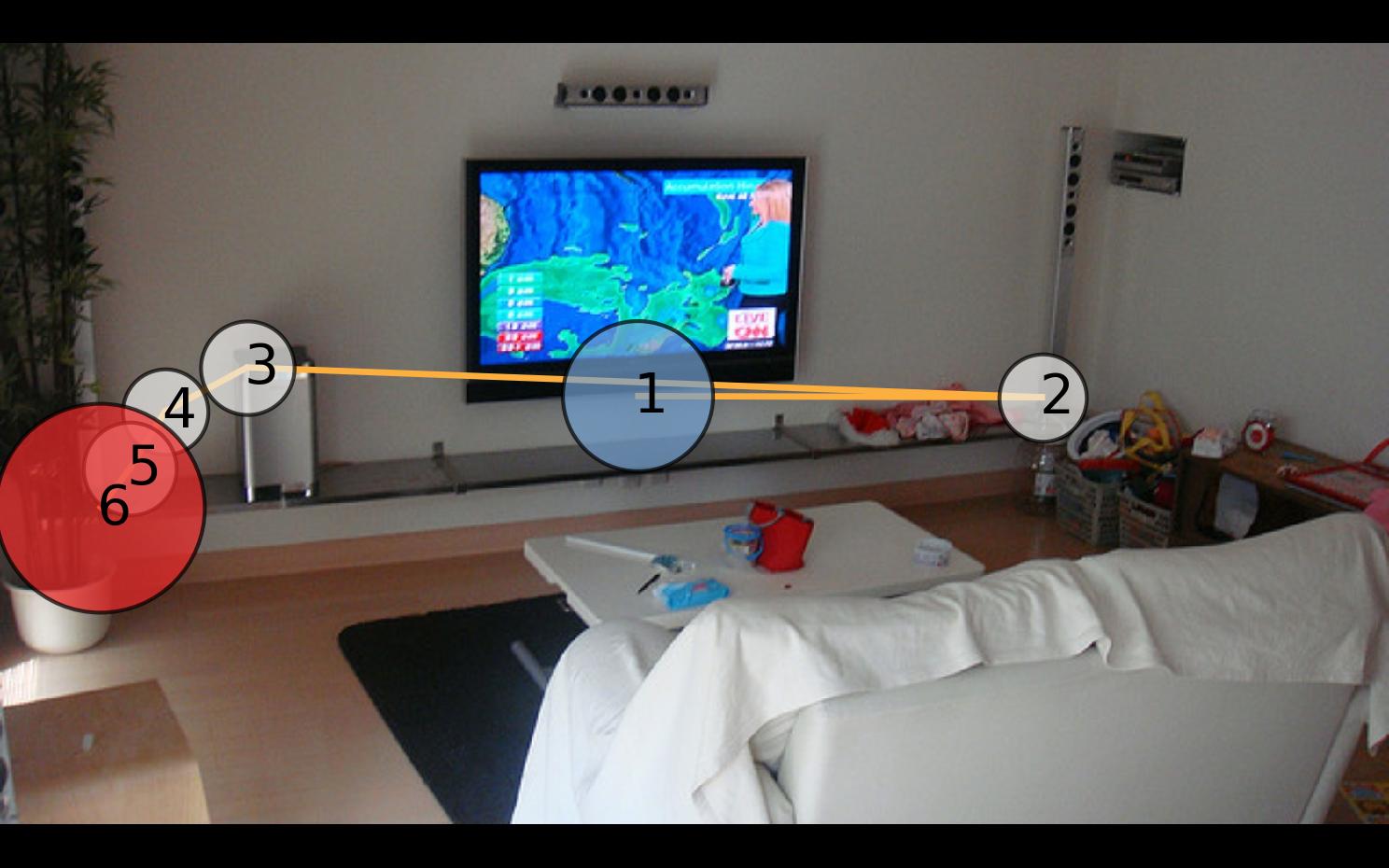} & \includegraphics[width=0.15\linewidth]{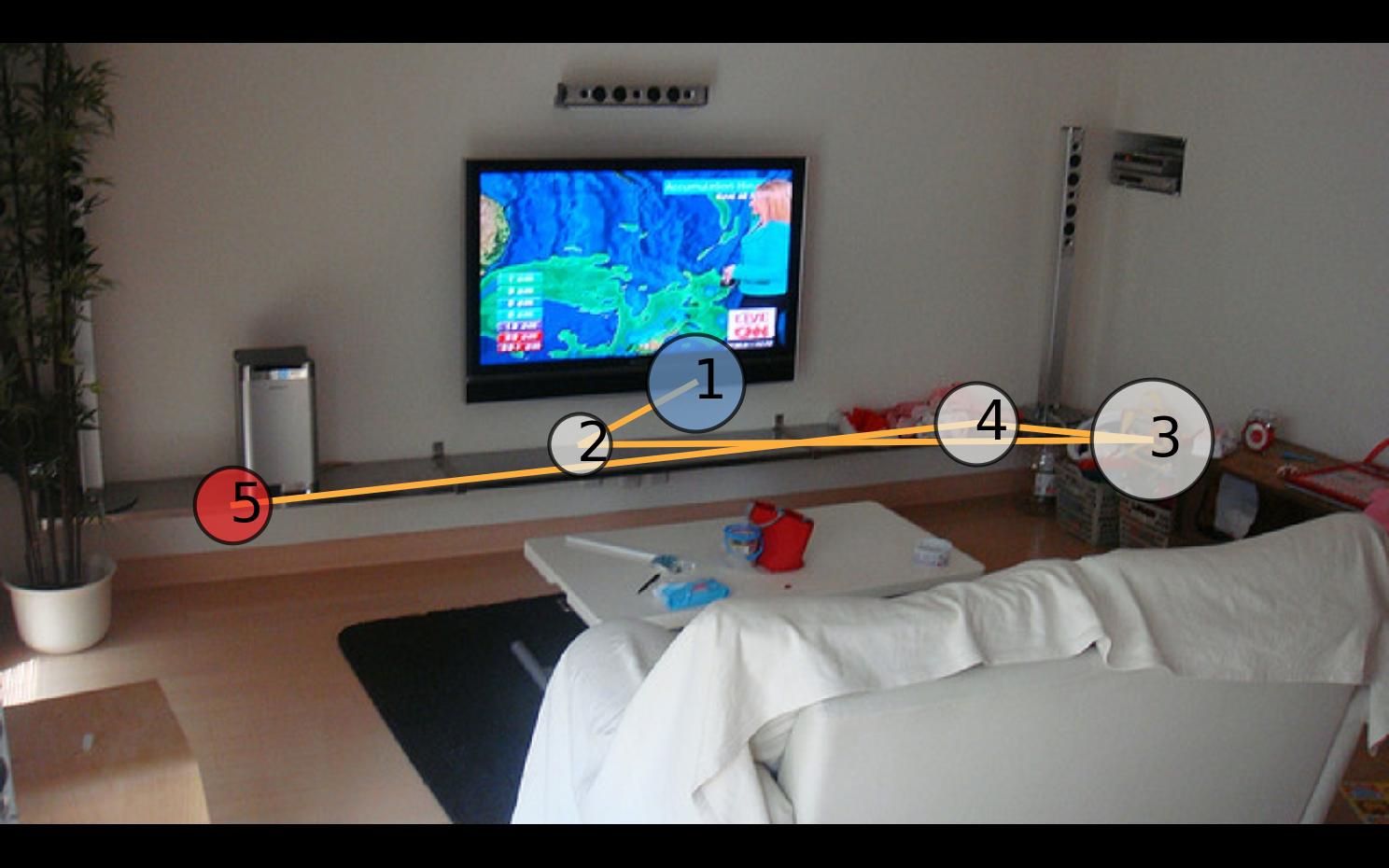} \\
        \includegraphics[width=0.15\linewidth]{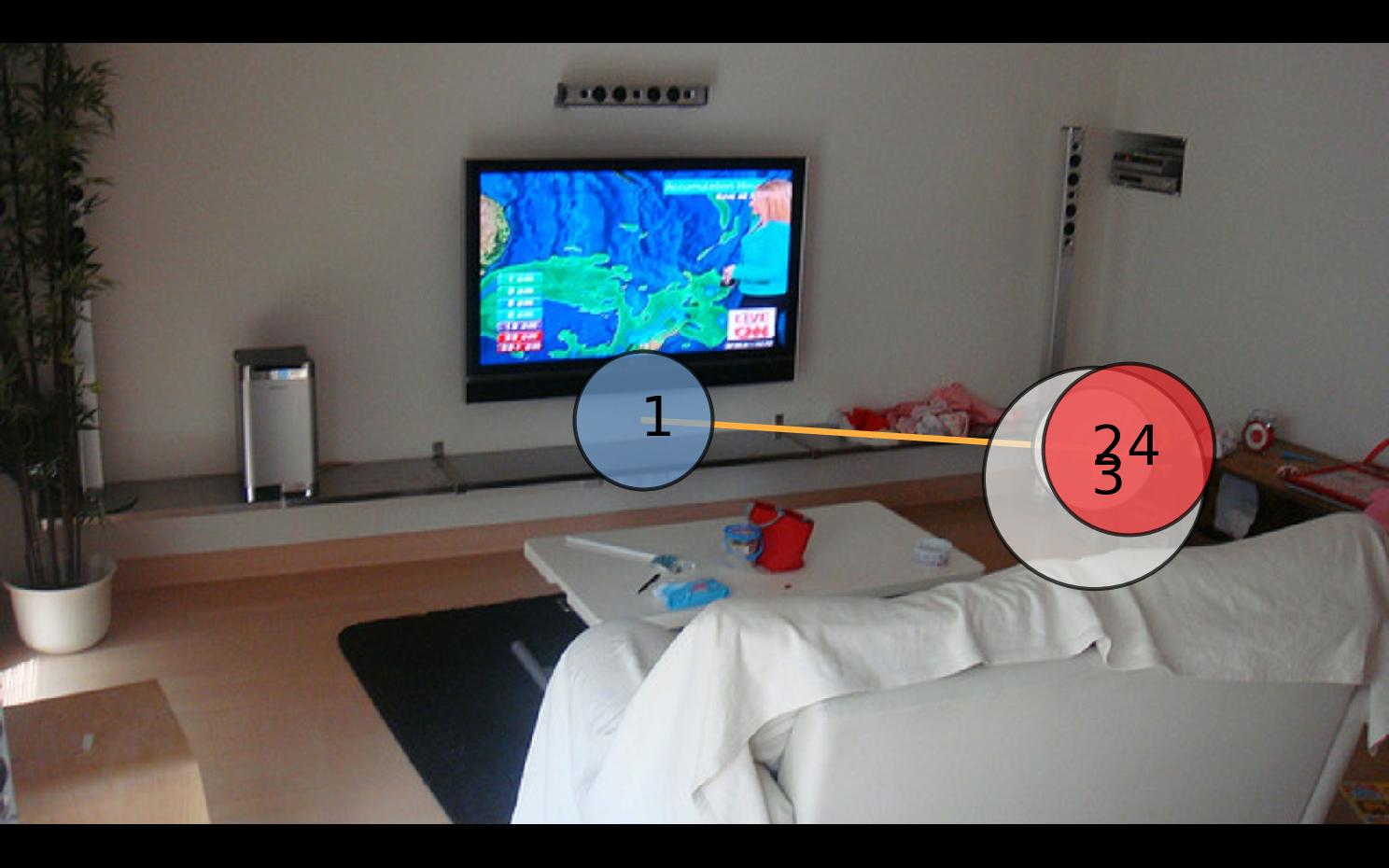} & \includegraphics[width=0.15\linewidth]{images/variabilitycocotp/gazeformer1.jpg} & \includegraphics[width=0.15\linewidth]{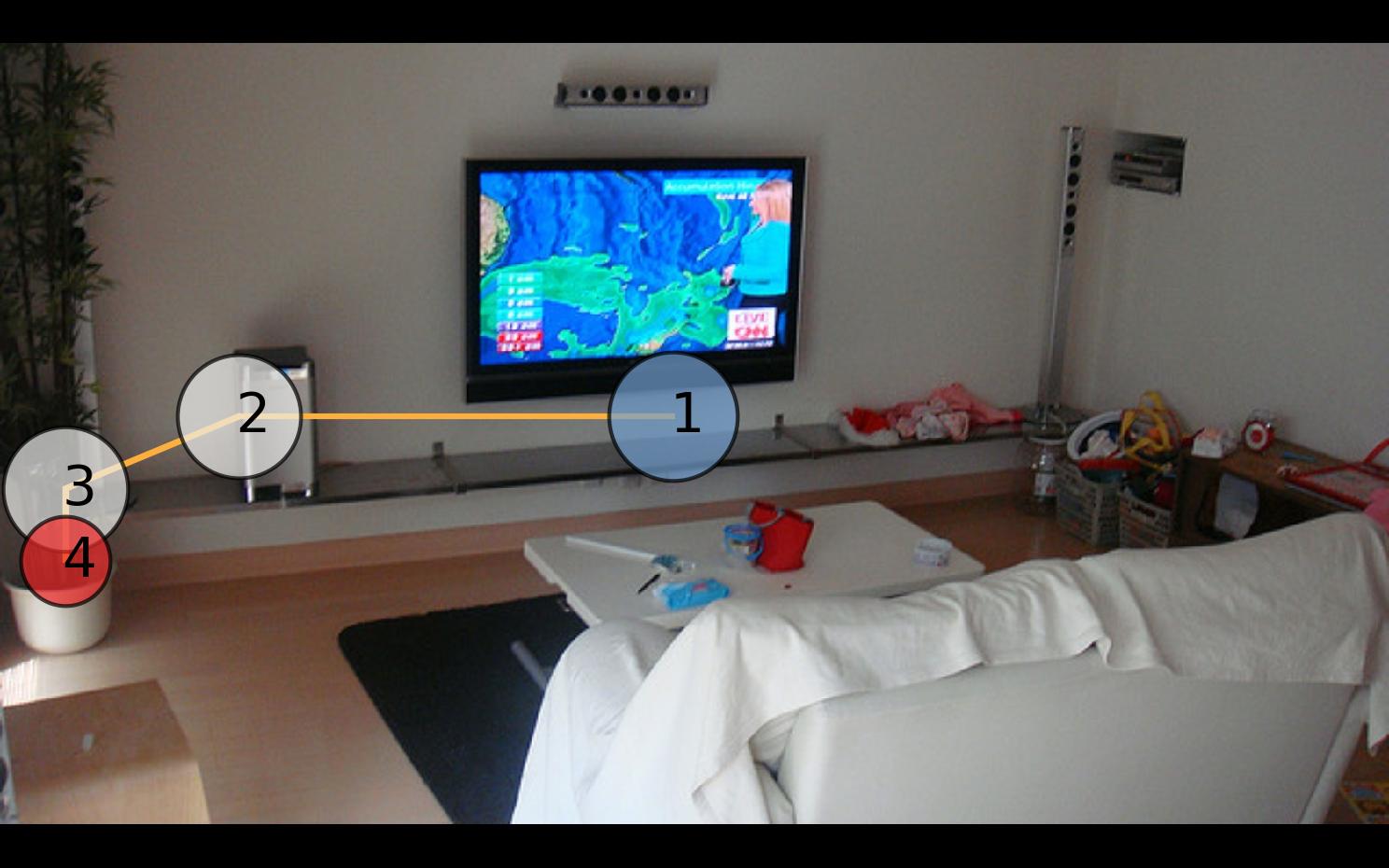} & \includegraphics[width=0.15\linewidth]{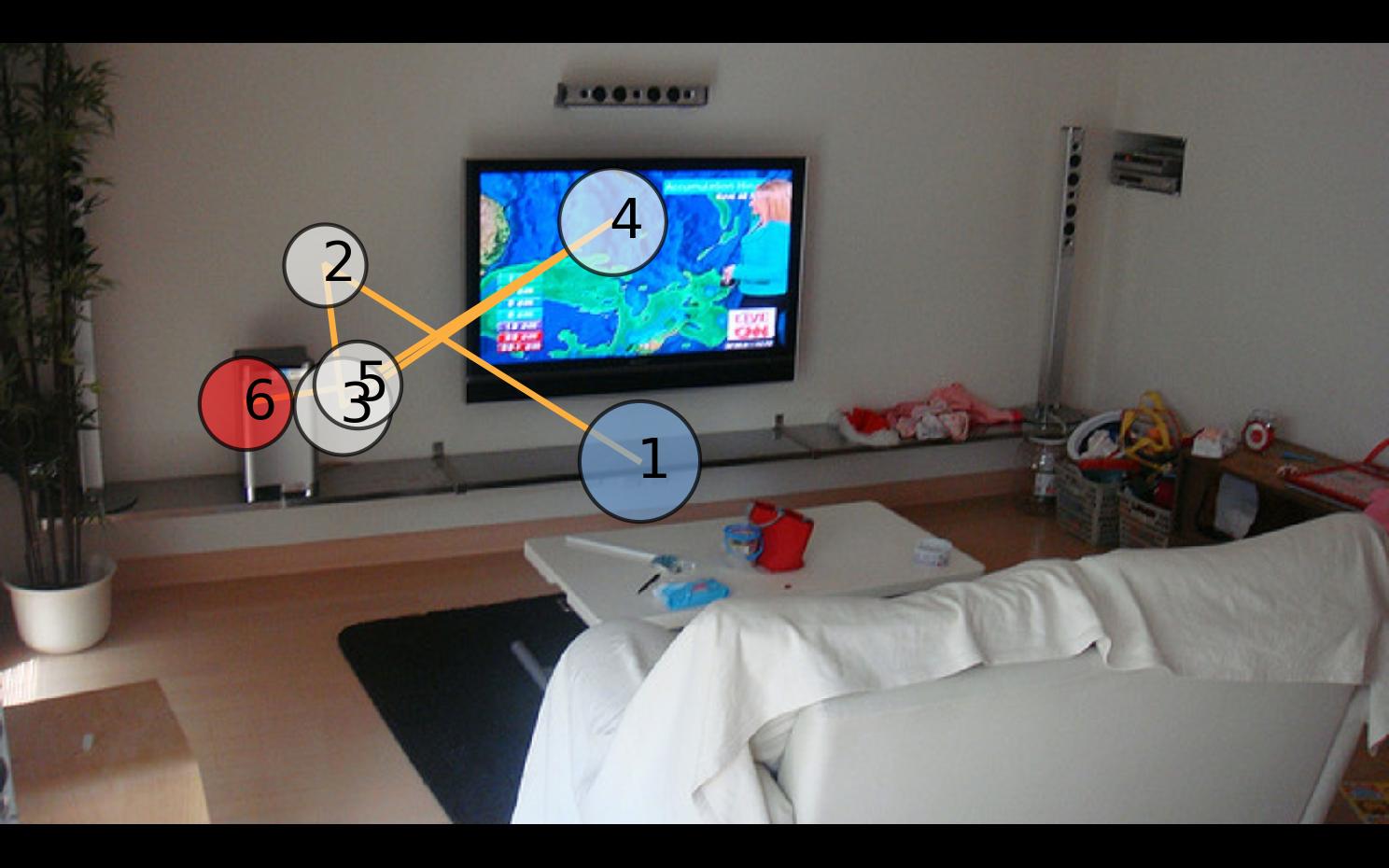} & \includegraphics[width=0.15\linewidth]{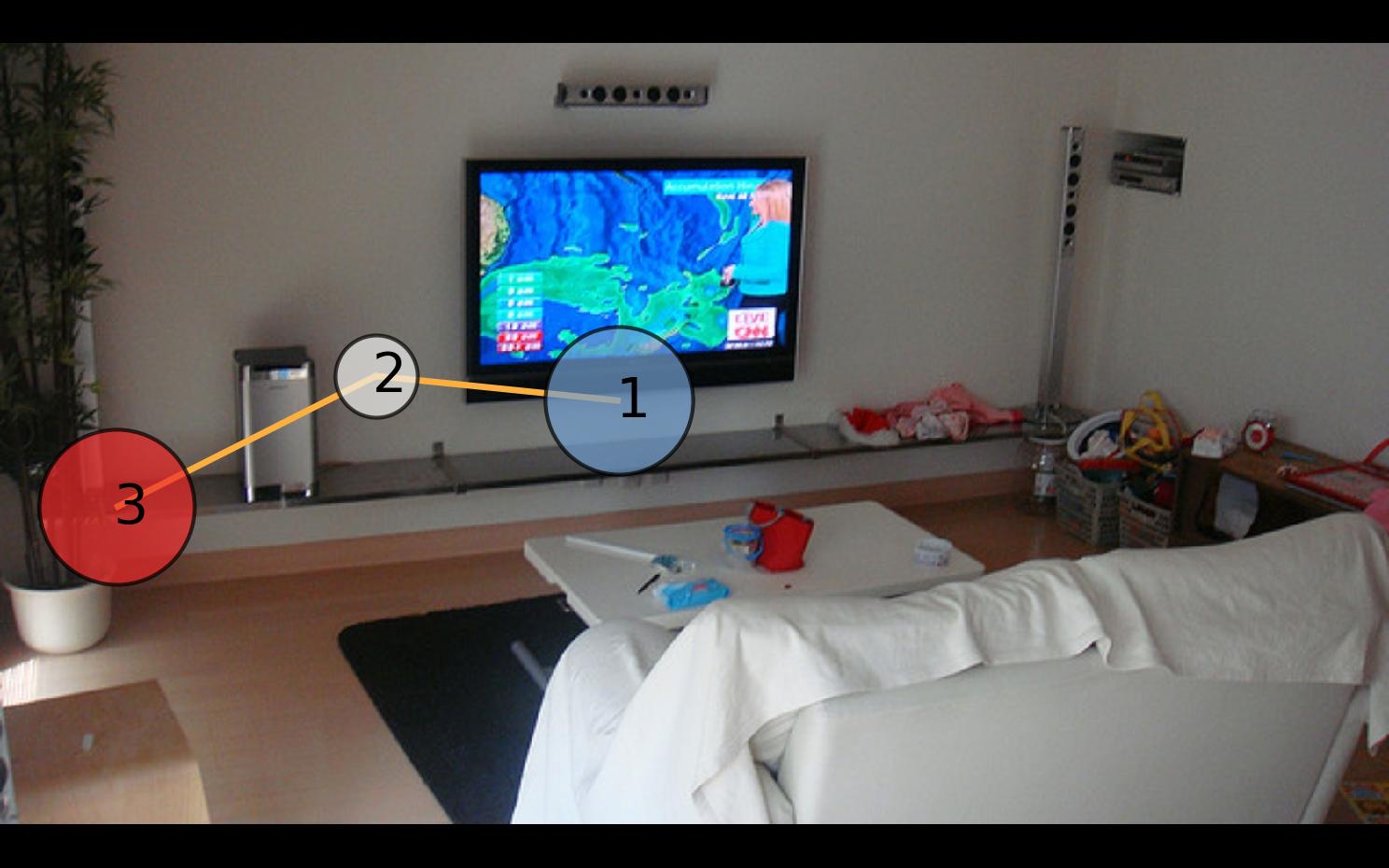} & \includegraphics[width=0.15\linewidth]{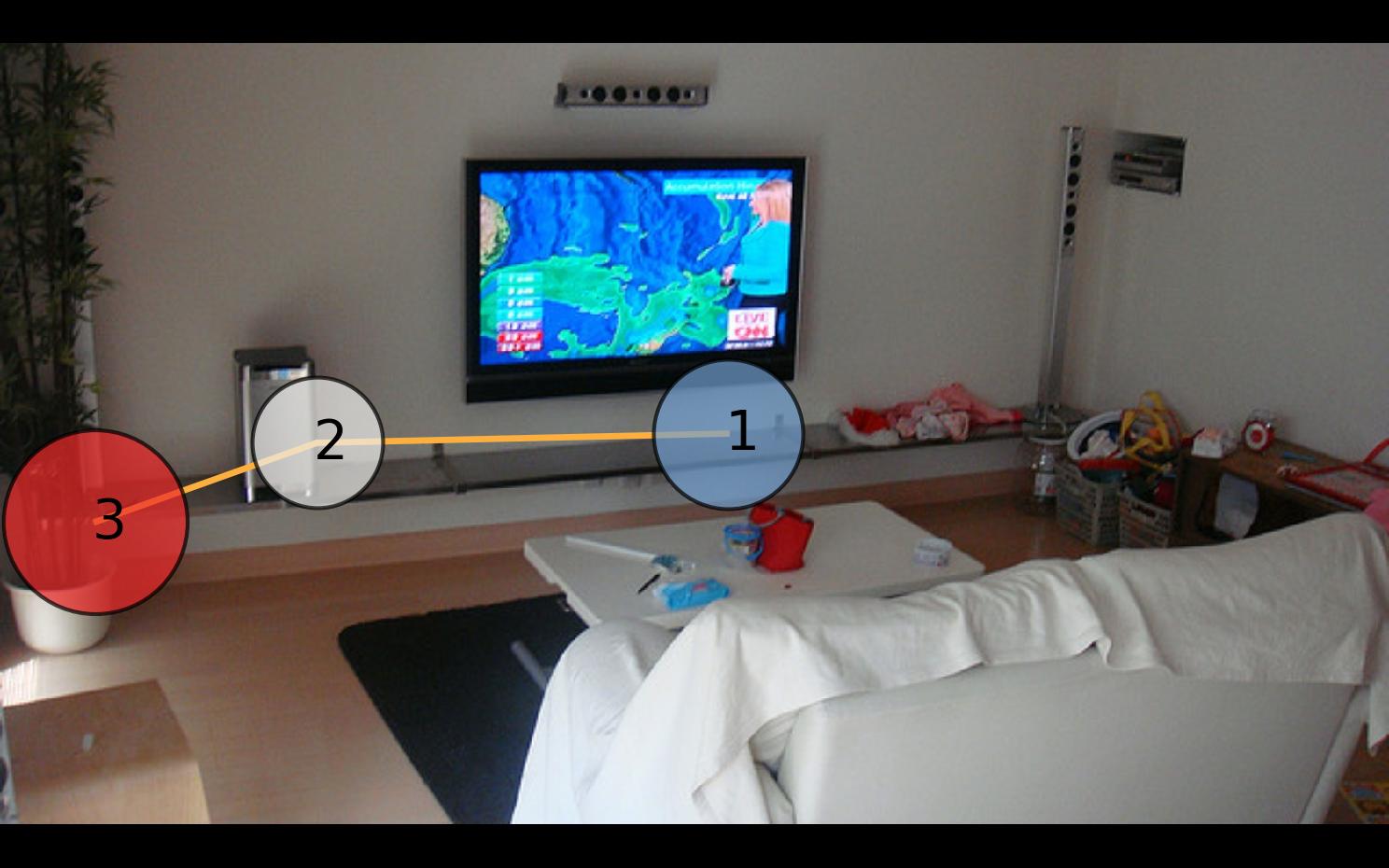} \\
    \end{tabular}
    }
    \vspace{-0.15cm}
    \caption{Qualitative comparison of the variability of simulated and human scanpaths on the COCO-Search18 (TP) dataset for the viewing task: \texttt{potted plant}. Each row corresponds to a different simulation or a different human observer.}
    \label{fig:qualitatives_variab_cocoT}
    \vspace{-0.4cm}
\end{figure*}

\begin{figure*}[t]
    \footnotesize
    \setlength{\tabcolsep}{.1em}
    \resizebox{\linewidth}{!}{
    \begin{tabular}{cccccc}
        \scriptsize ChenLSTM~\cite{chen2021predicting} & \scriptsize Gazeformer~\cite{mondal2023gazeformer} & \scriptsize GazeXplain~\cite{chen2024gazexplain} & \scriptsize TPP-Gaze~\cite{damelio2025tpp} & \scriptsize \ours (Ours) & \scriptsize Humans \\
        \includegraphics[width=0.15\linewidth]{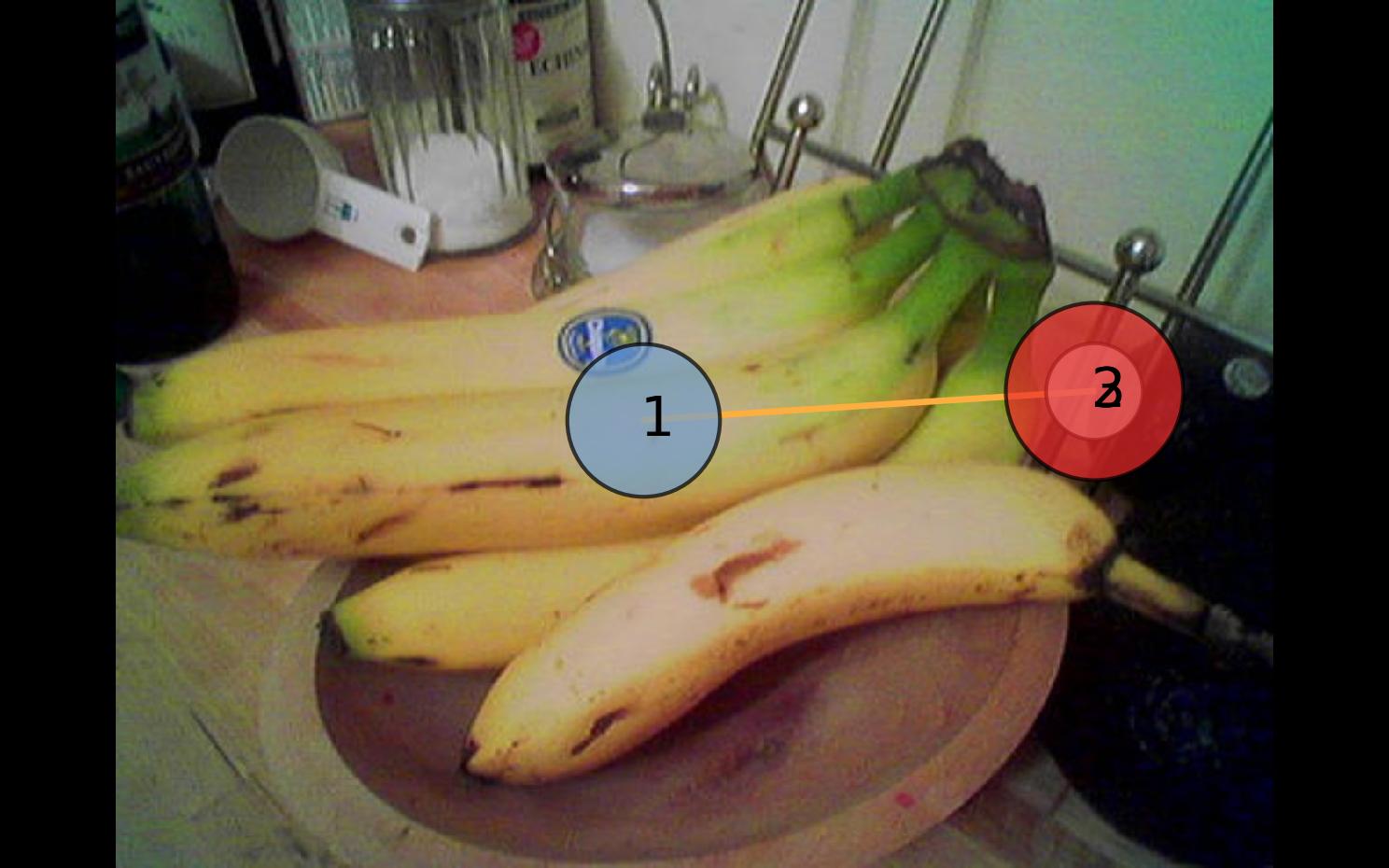} & \includegraphics[width=0.15\linewidth]{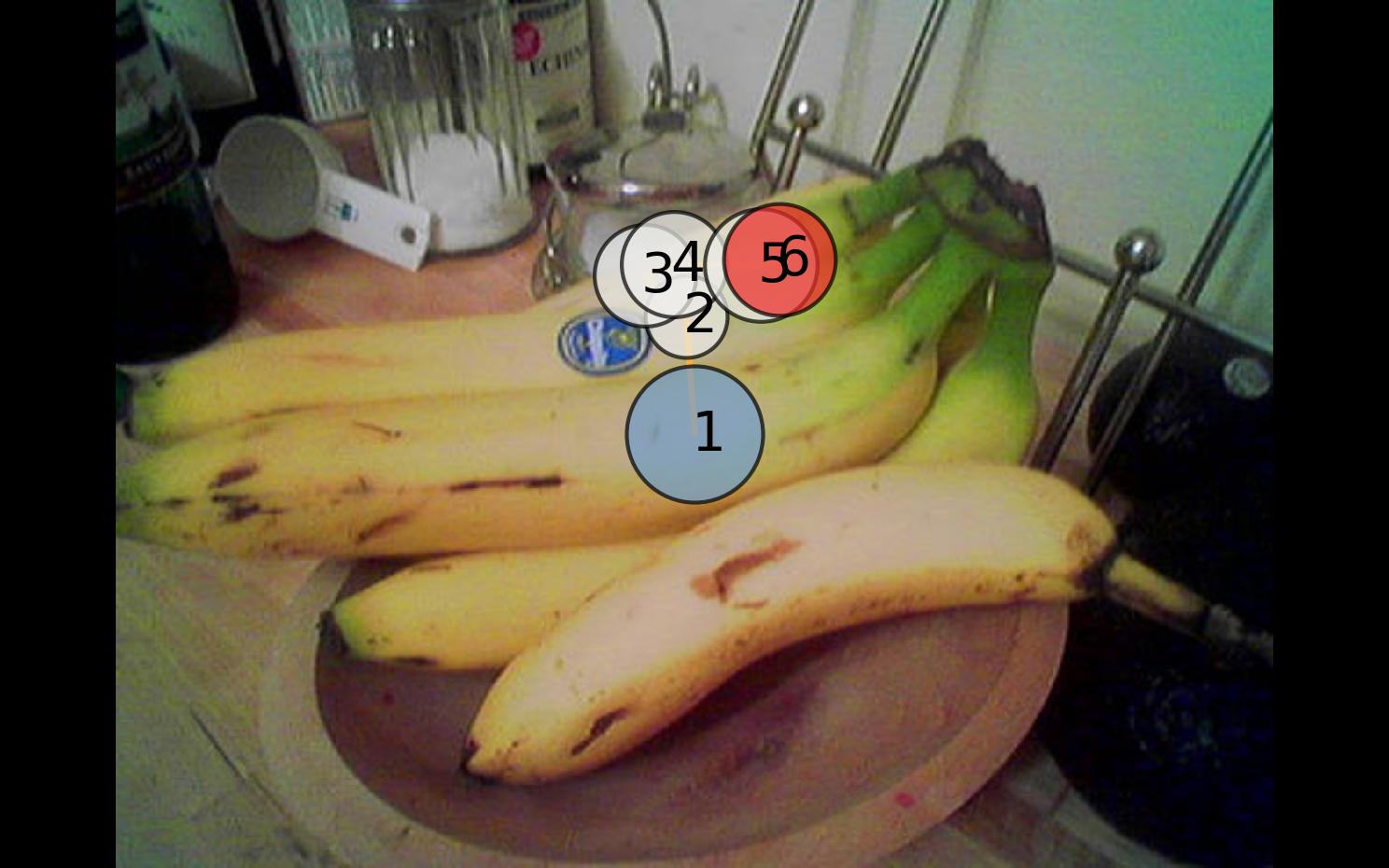} & \includegraphics[width=0.15\linewidth]{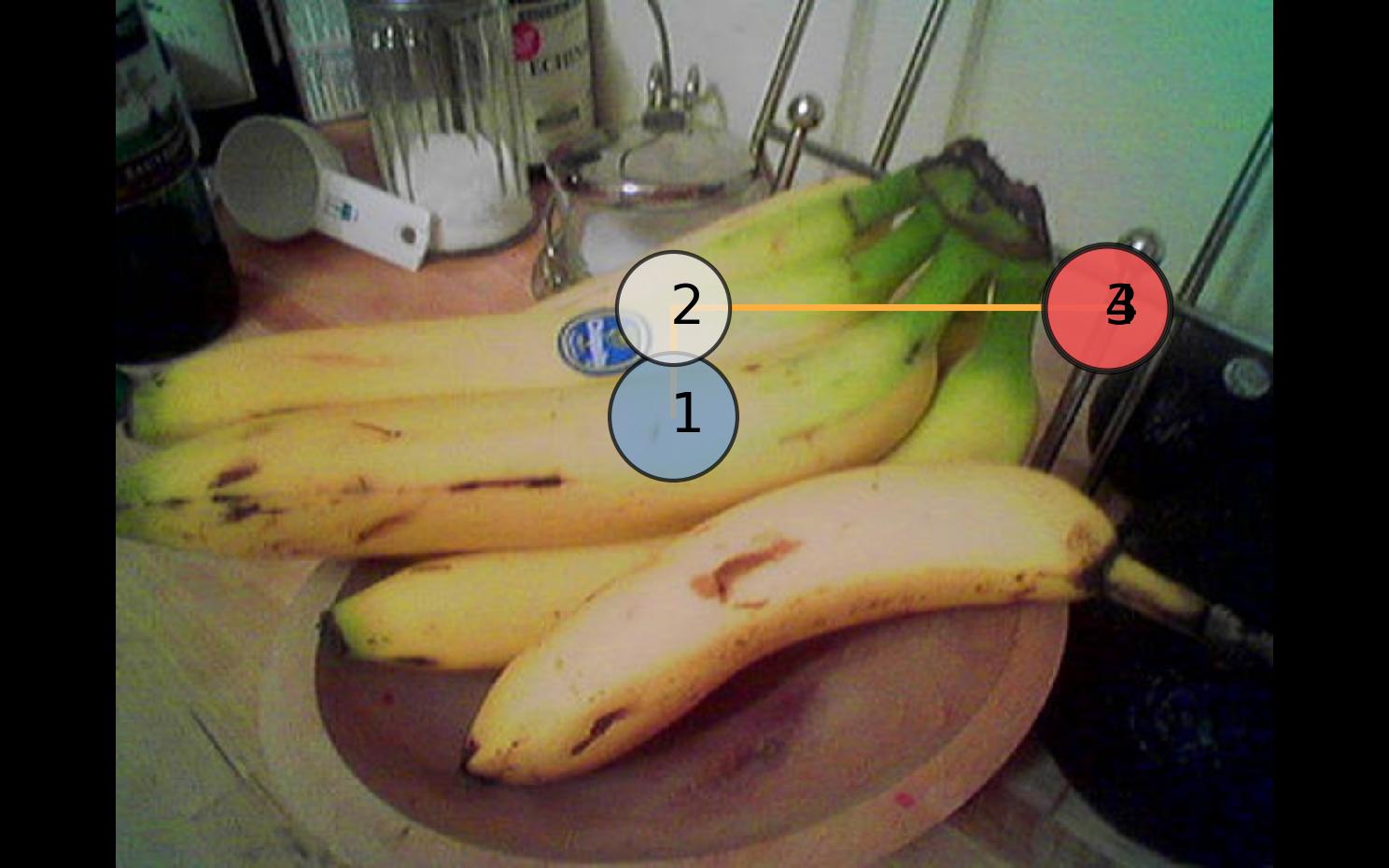} & \includegraphics[width=0.15\linewidth]{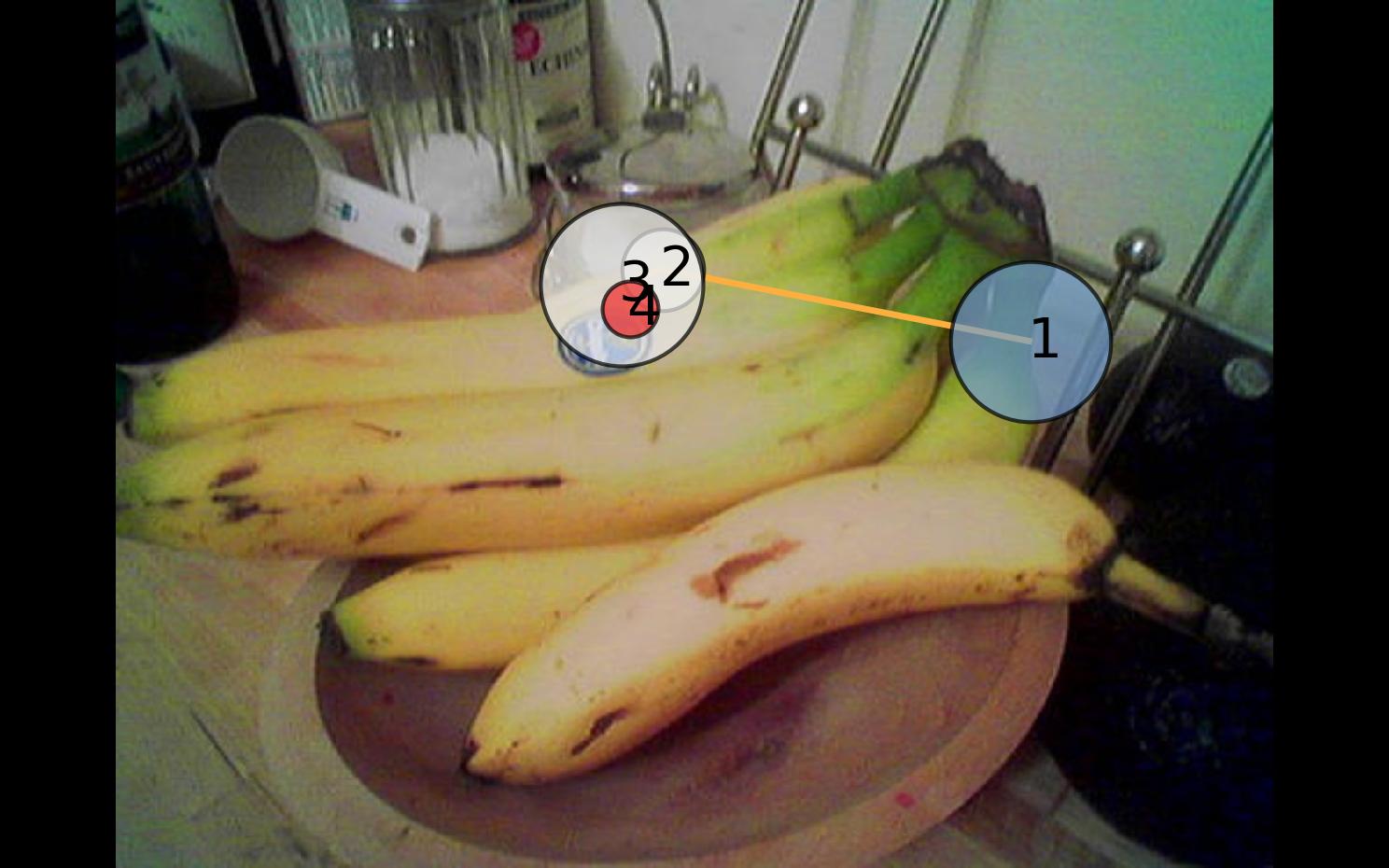} & \includegraphics[width=0.15\linewidth]{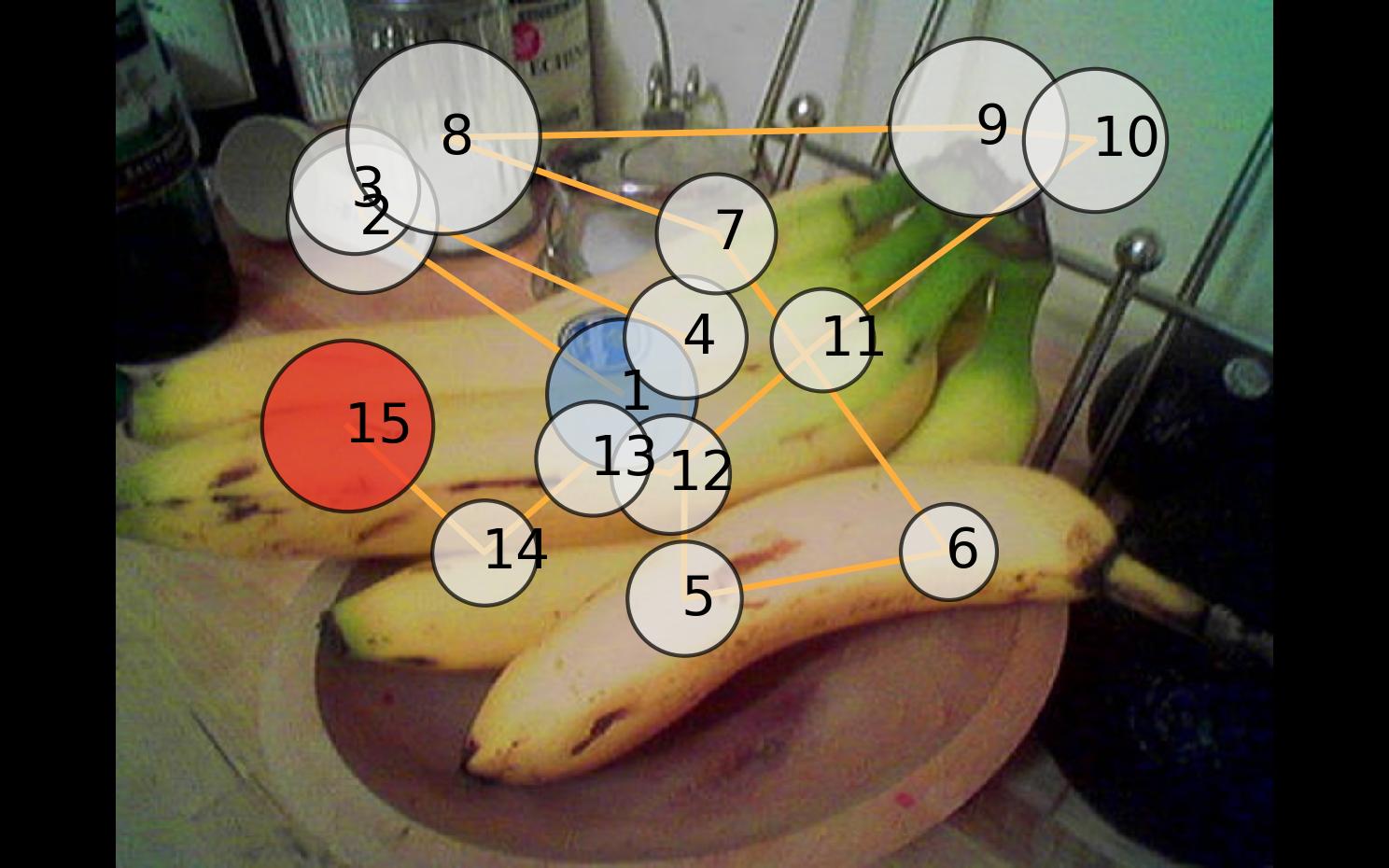} & \includegraphics[width=0.15\linewidth]{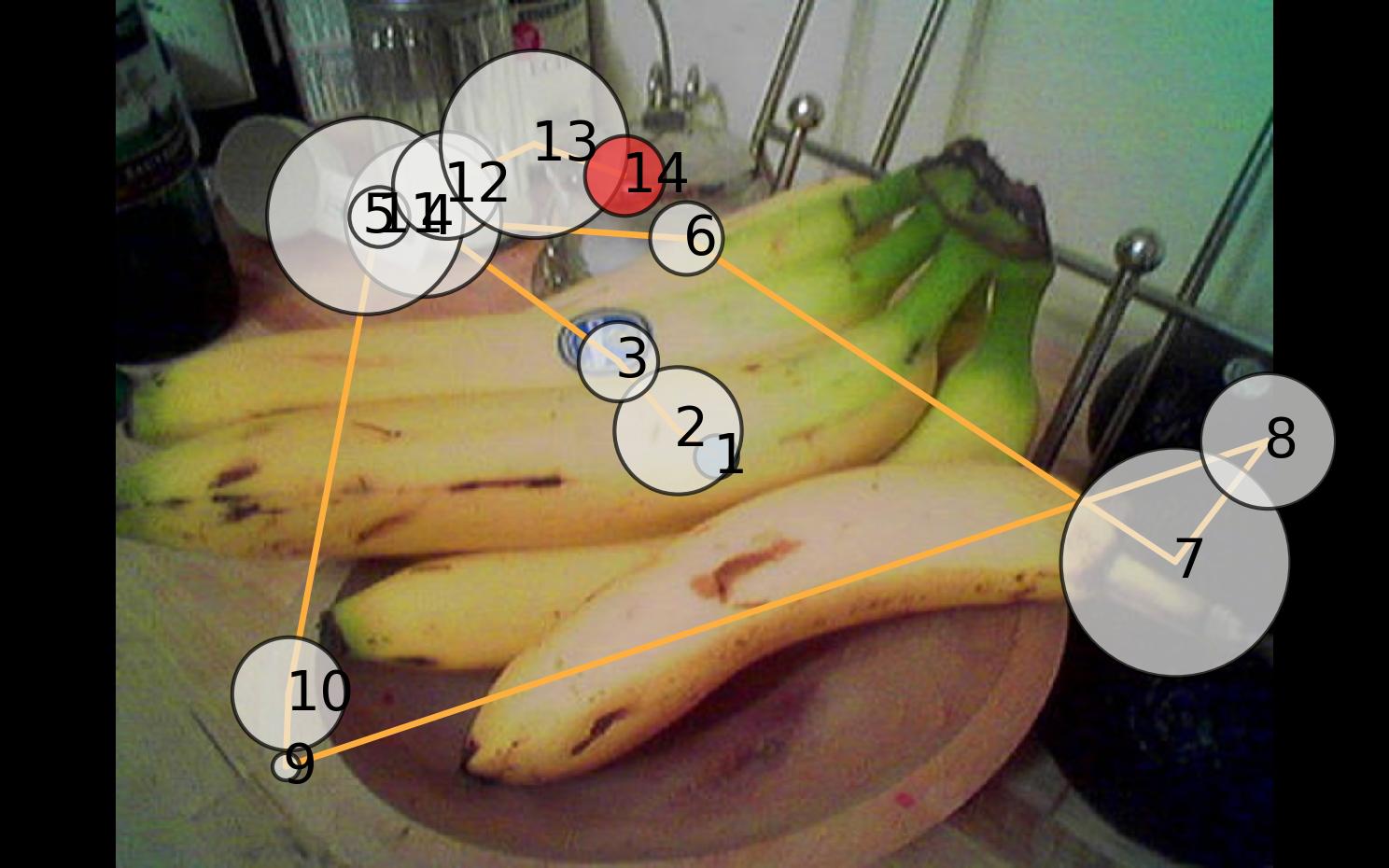} \\
        \includegraphics[width=0.15\linewidth]{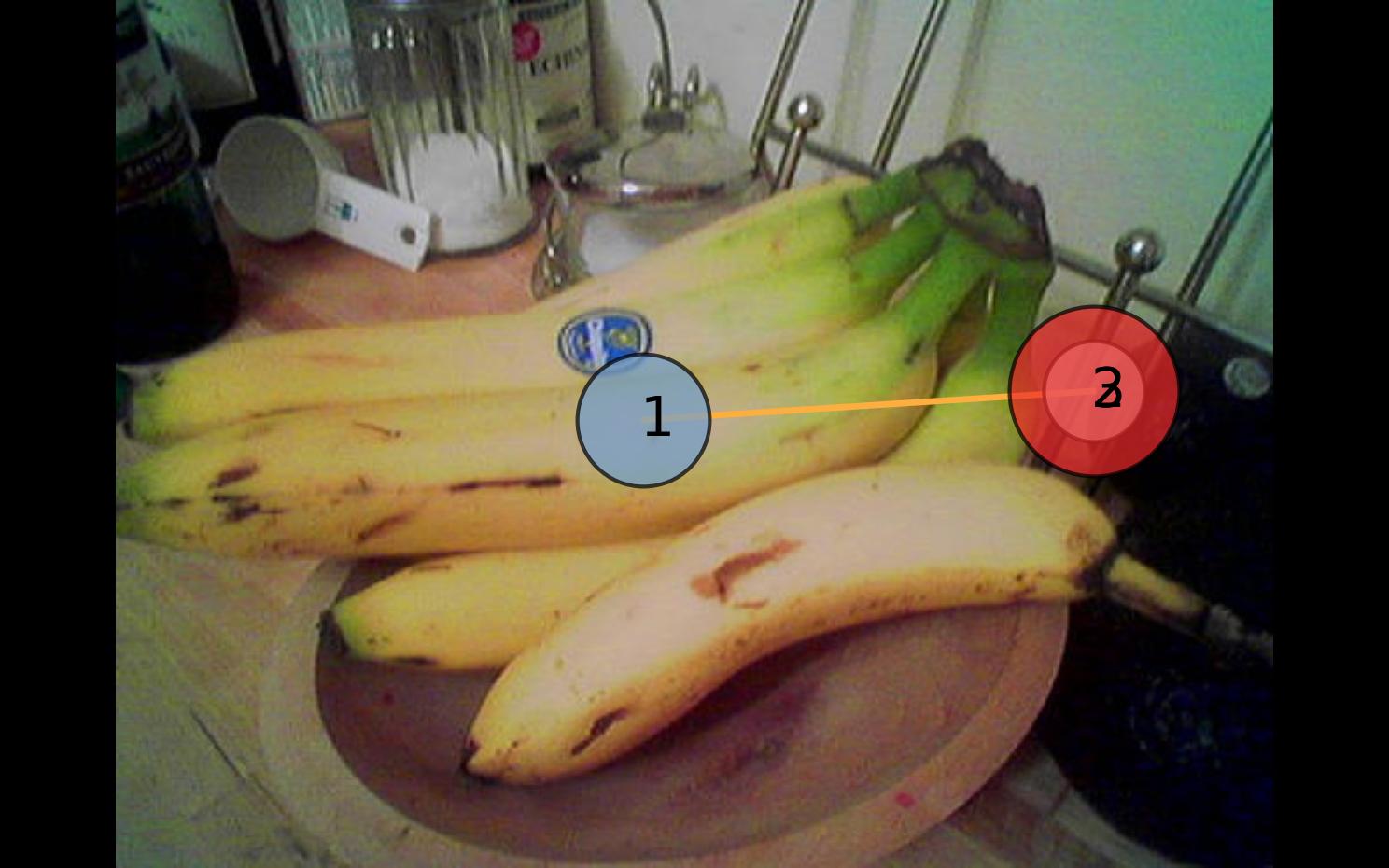} & \includegraphics[width=0.15\linewidth]{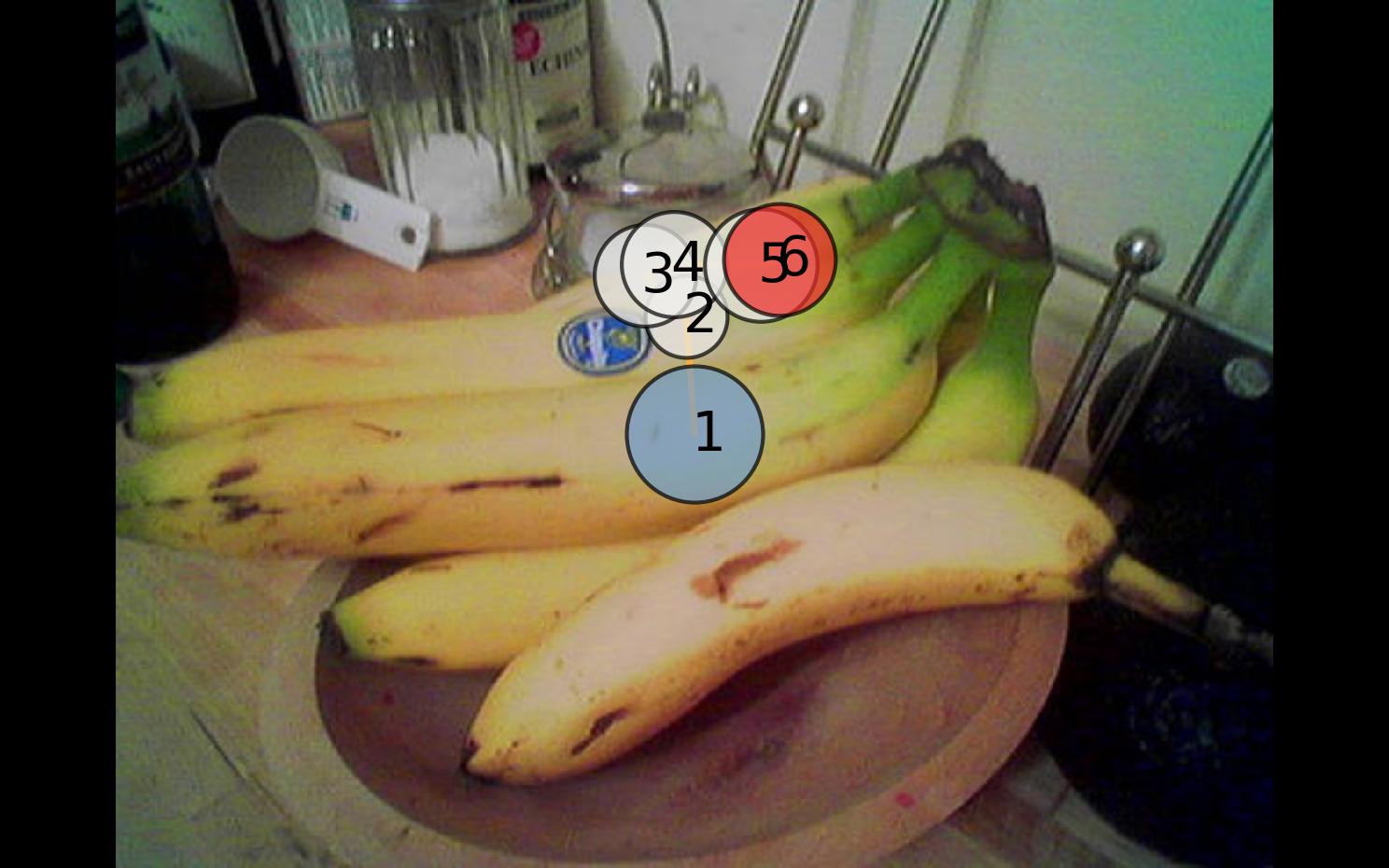} & \includegraphics[width=0.15\linewidth]{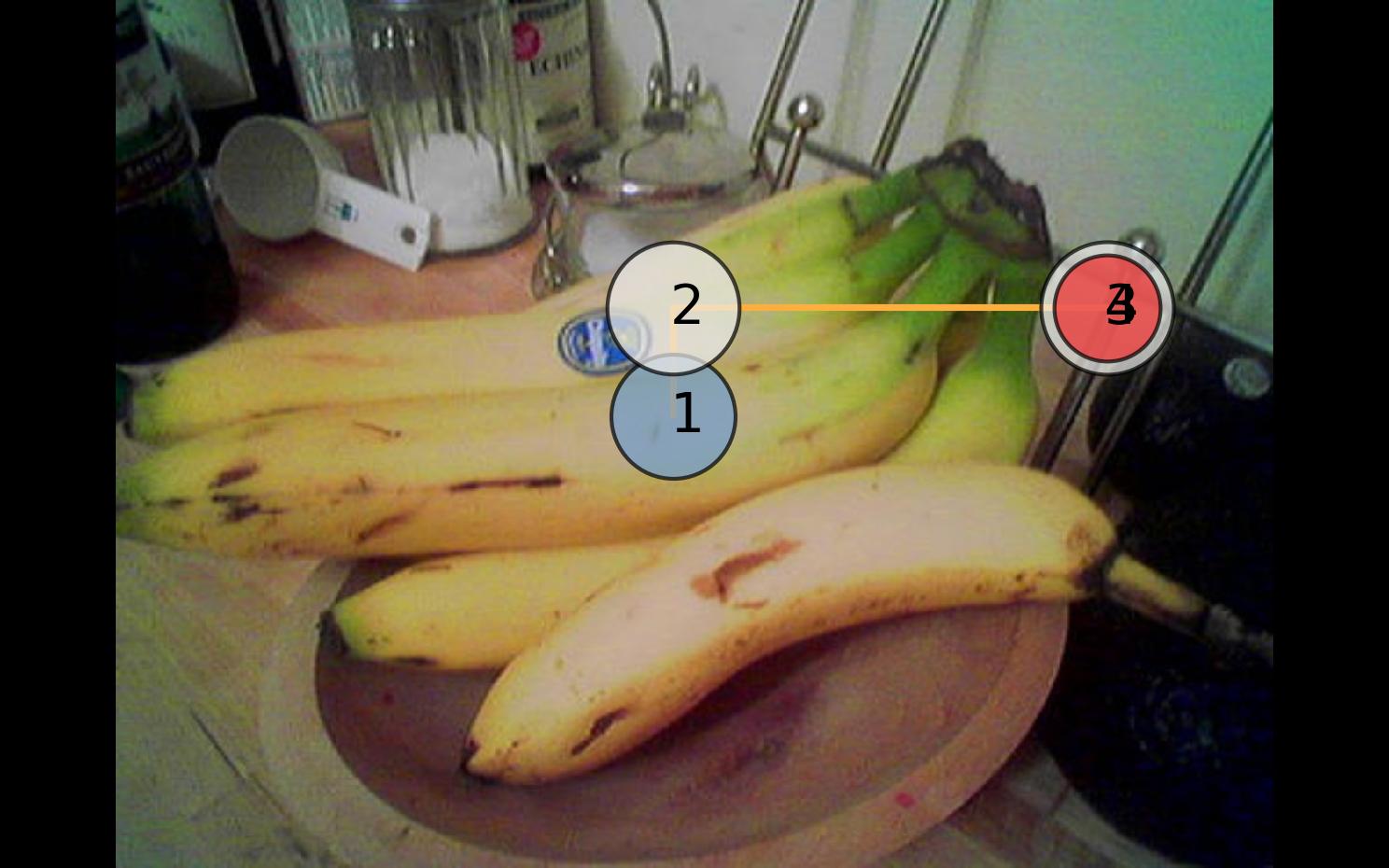} & \includegraphics[width=0.15\linewidth]{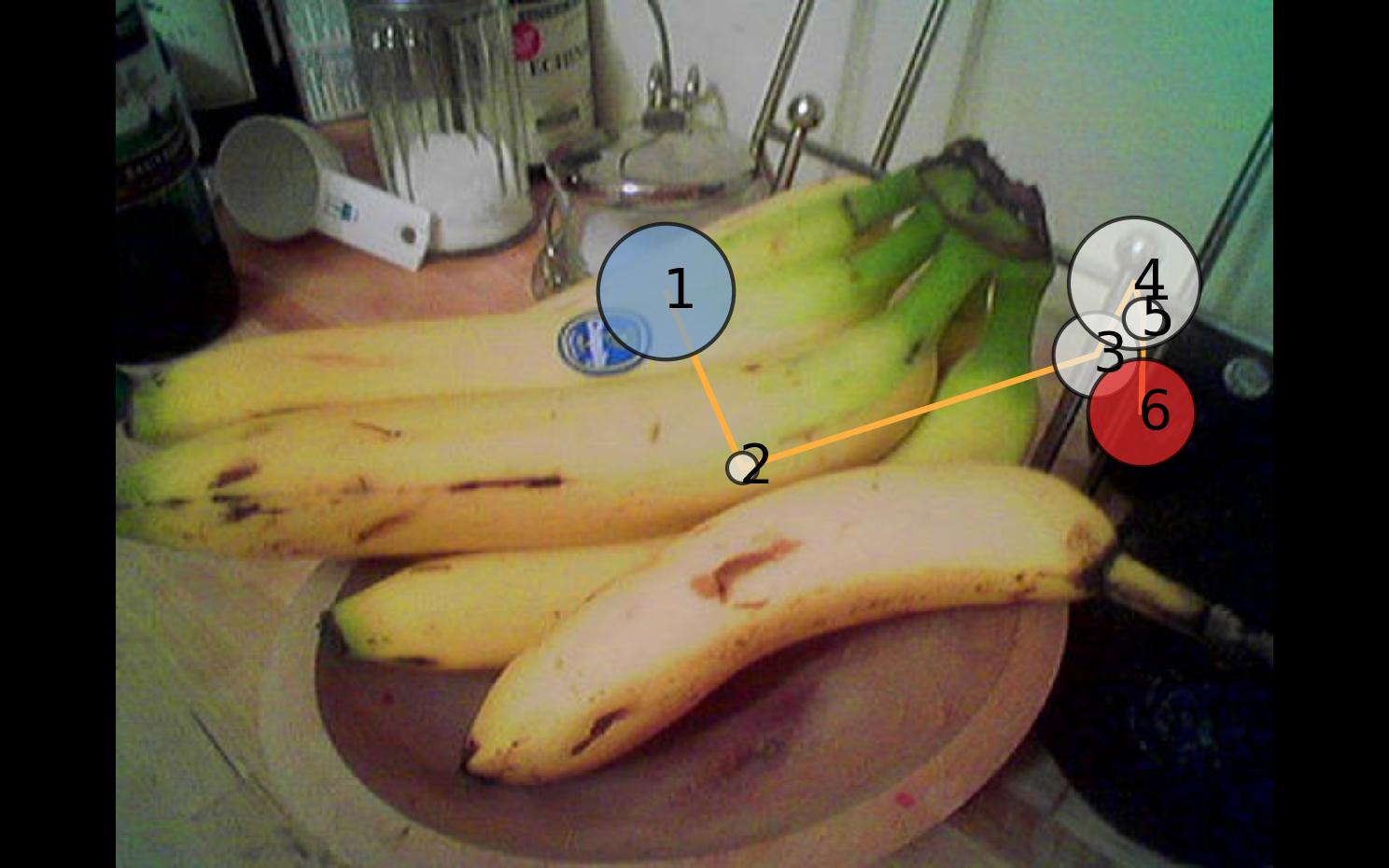} & \includegraphics[width=0.15\linewidth]{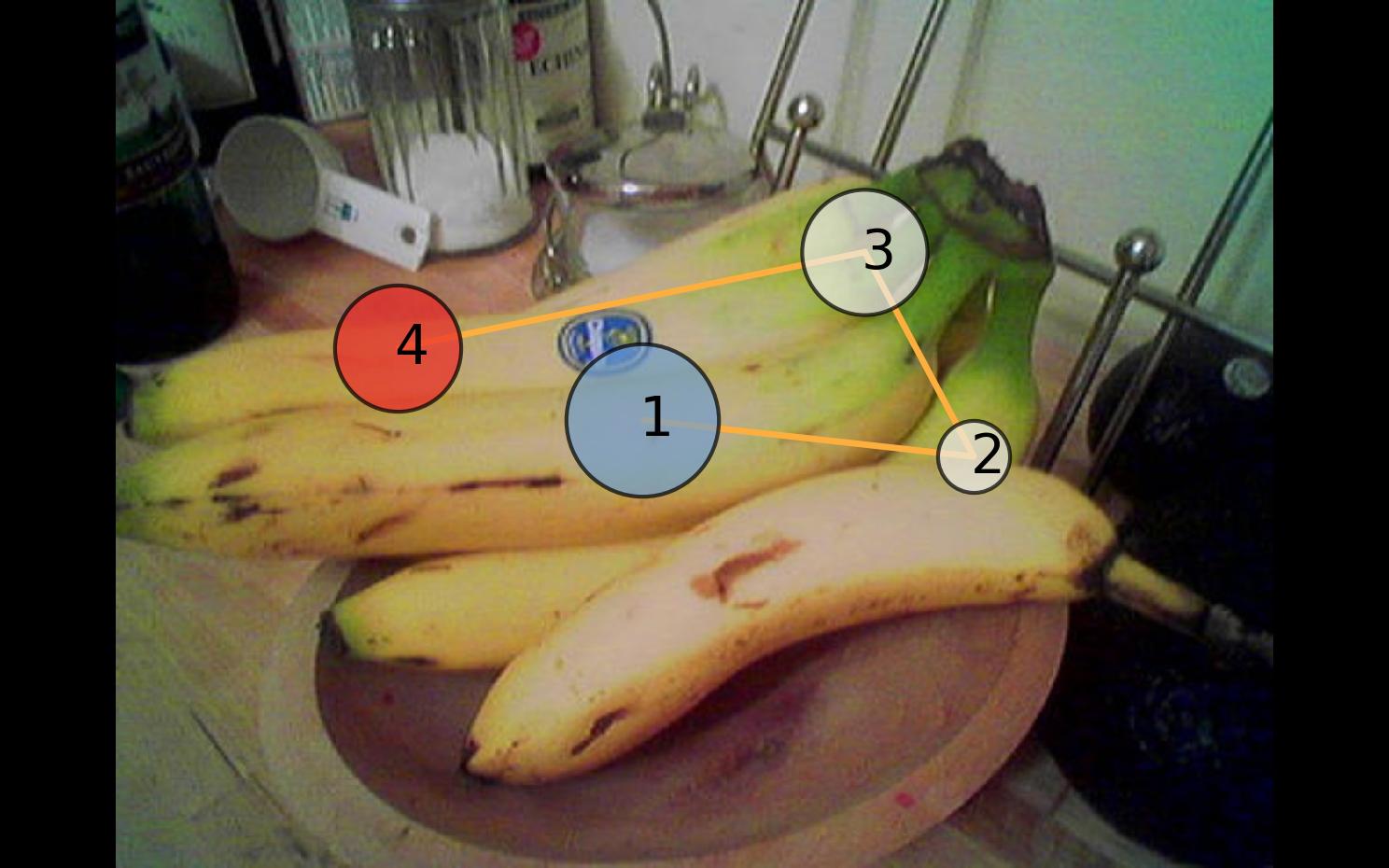} & \includegraphics[width=0.15\linewidth]{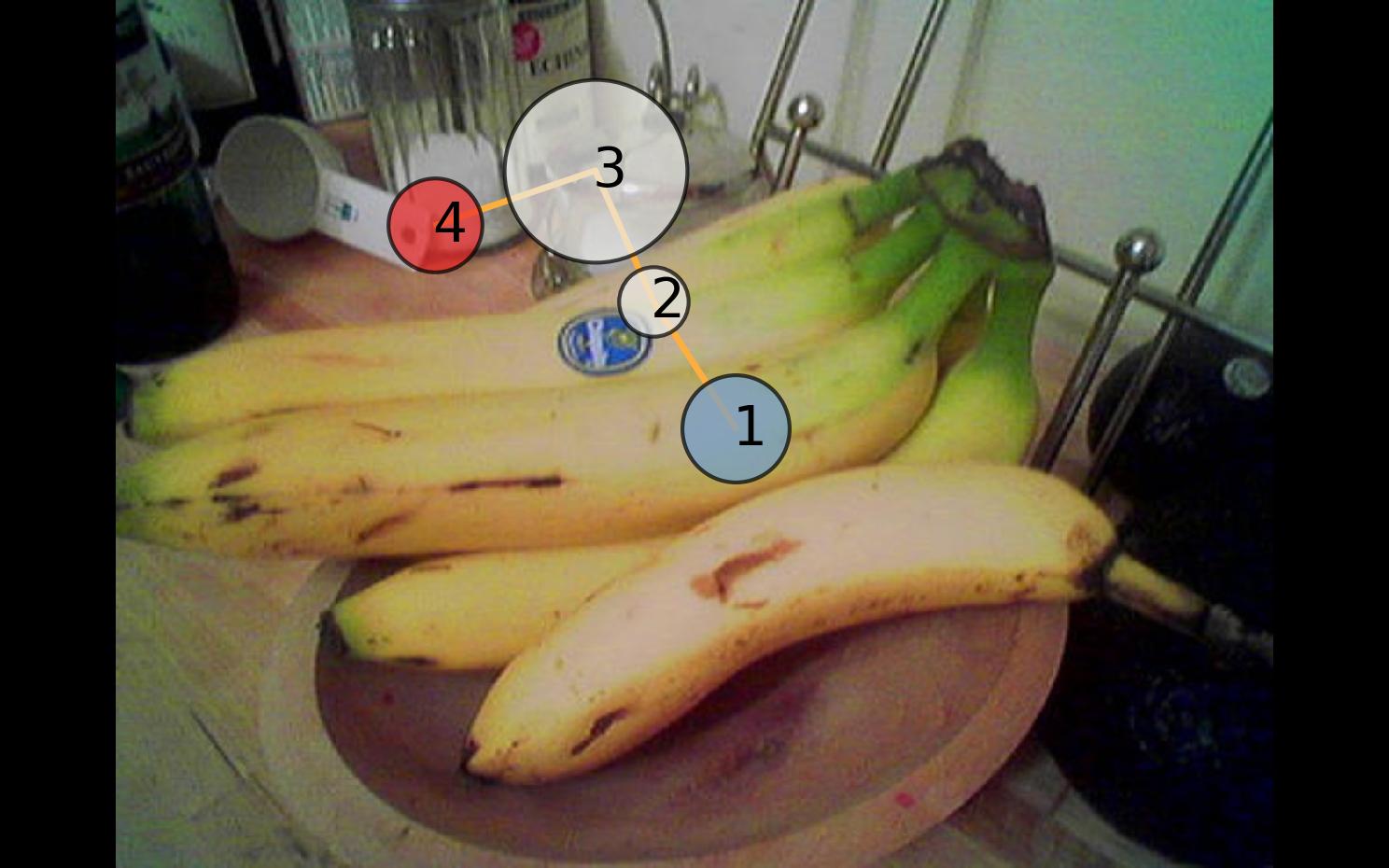} \\
        \includegraphics[width=0.15\linewidth]{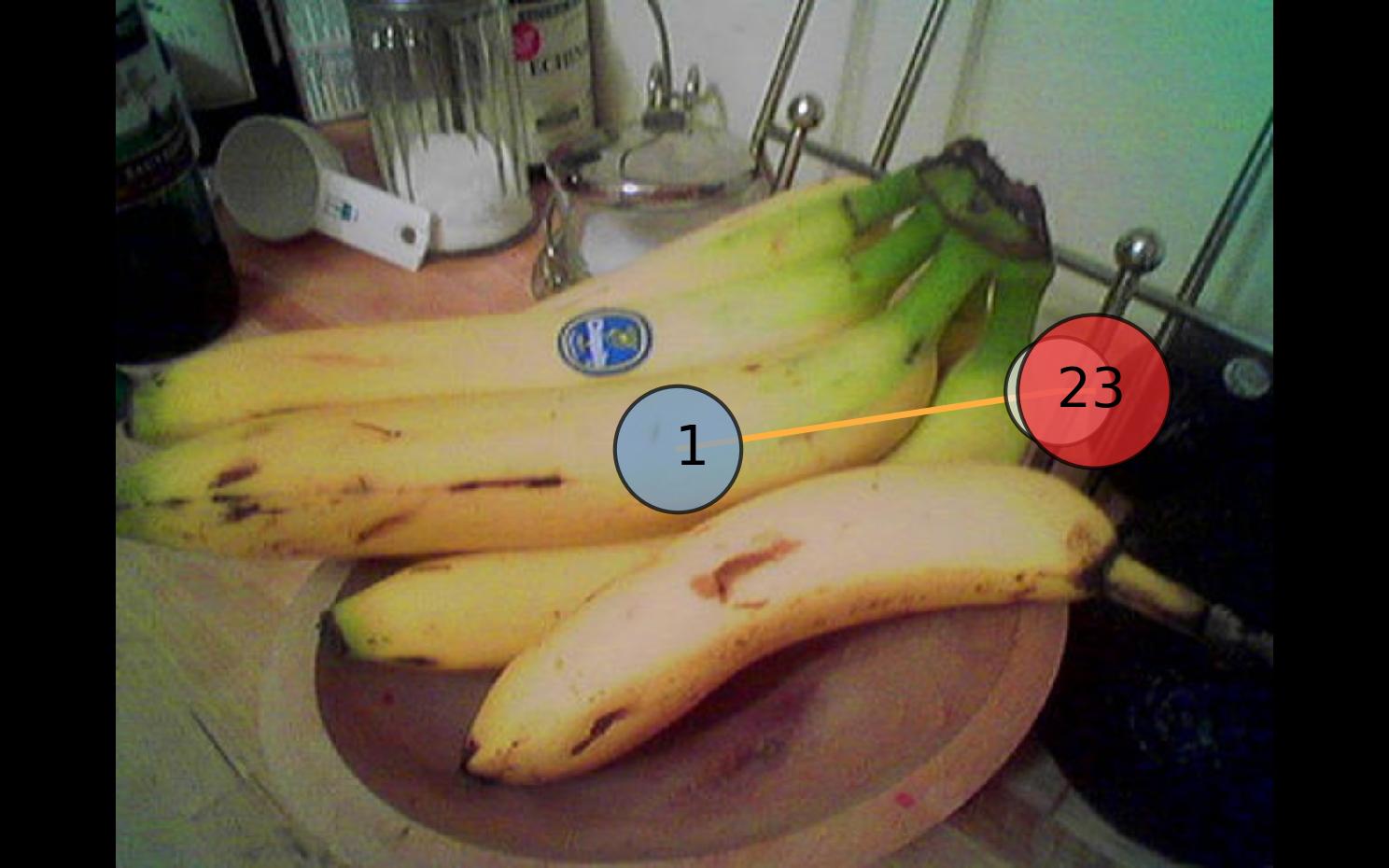} & \includegraphics[width=0.15\linewidth]{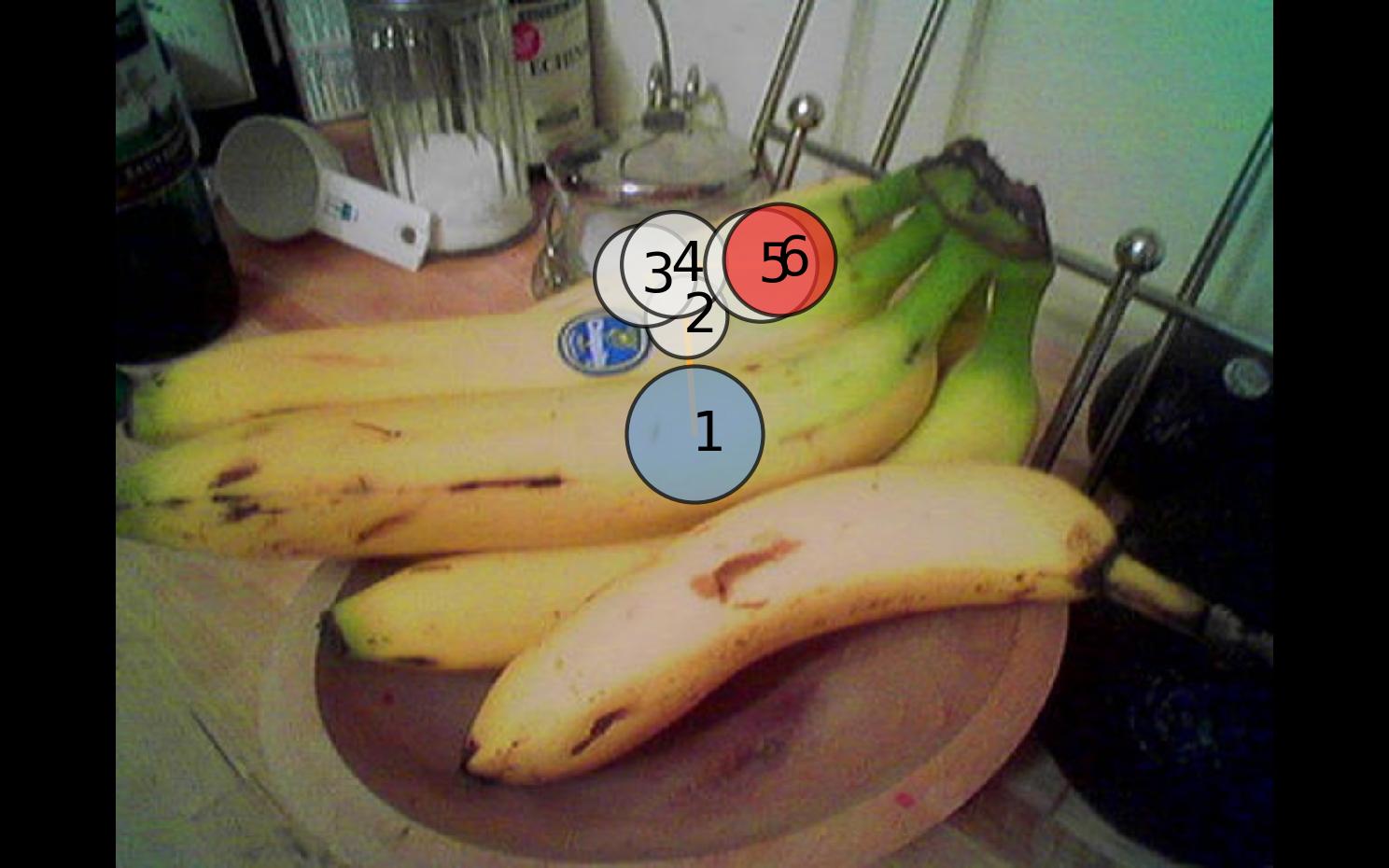} & \includegraphics[width=0.15\linewidth]{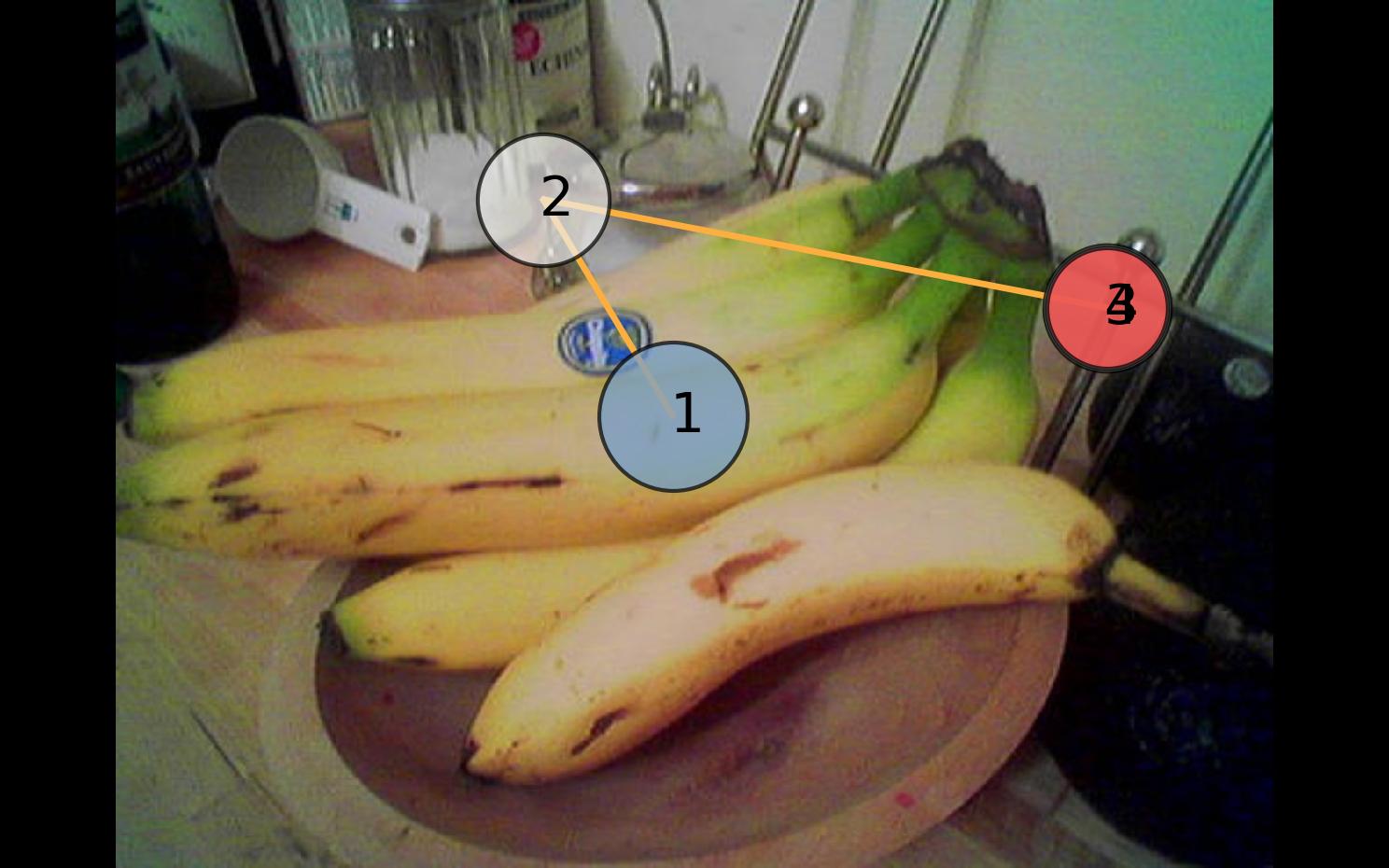} & \includegraphics[width=0.15\linewidth]{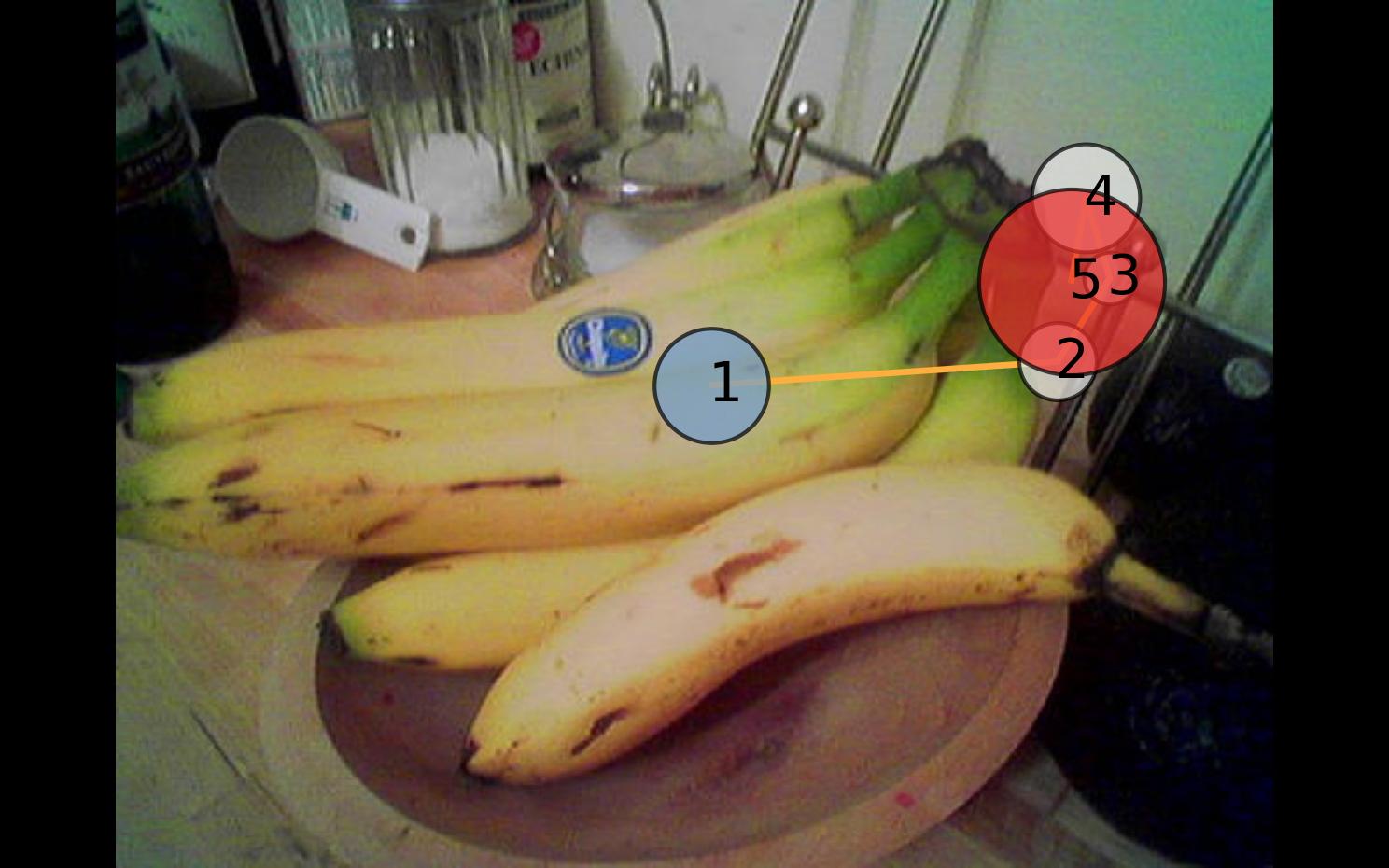} & \includegraphics[width=0.15\linewidth]{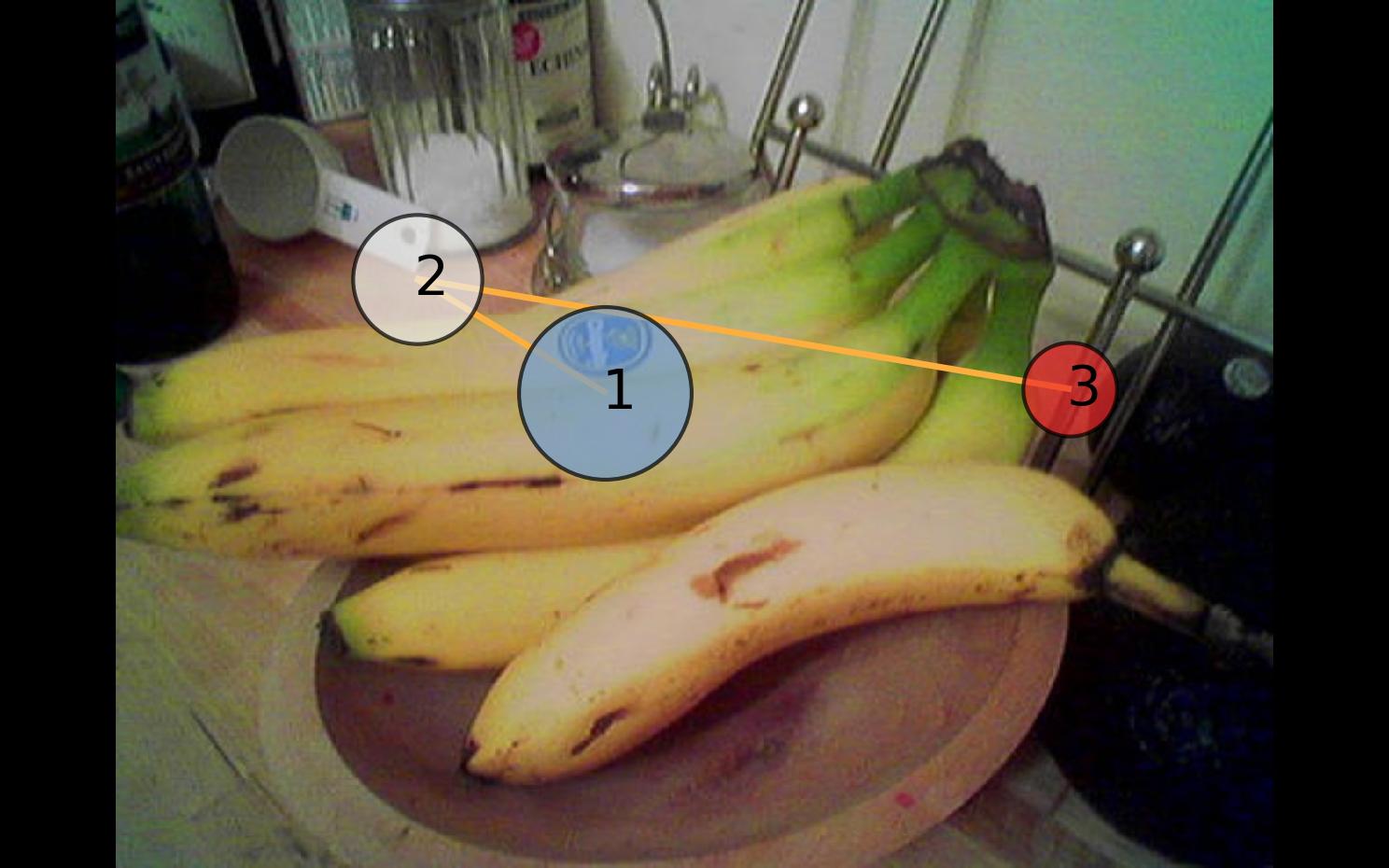} & \includegraphics[width=0.15\linewidth]{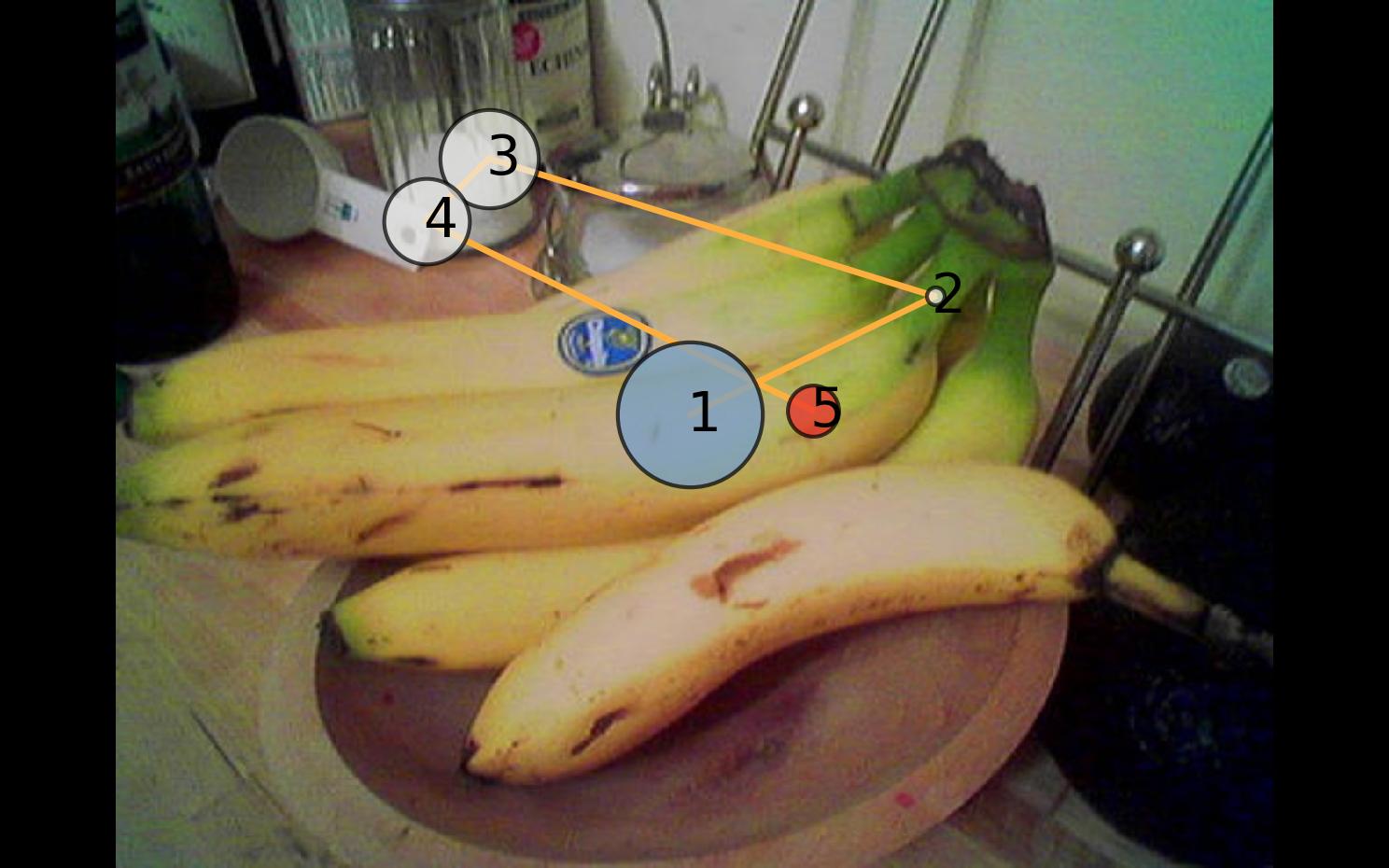} \\
        \includegraphics[width=0.15\linewidth]{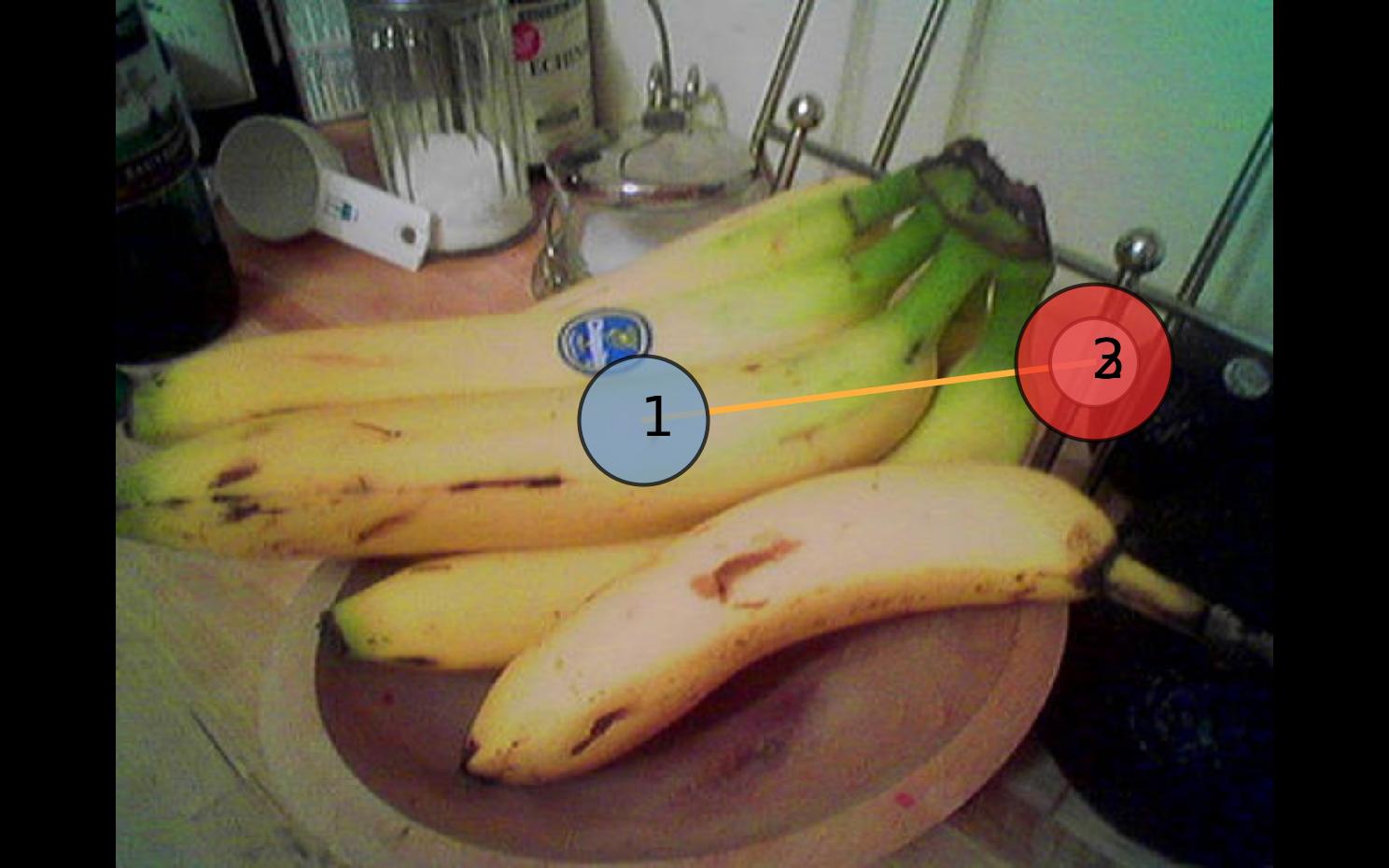} & \includegraphics[width=0.15\linewidth]{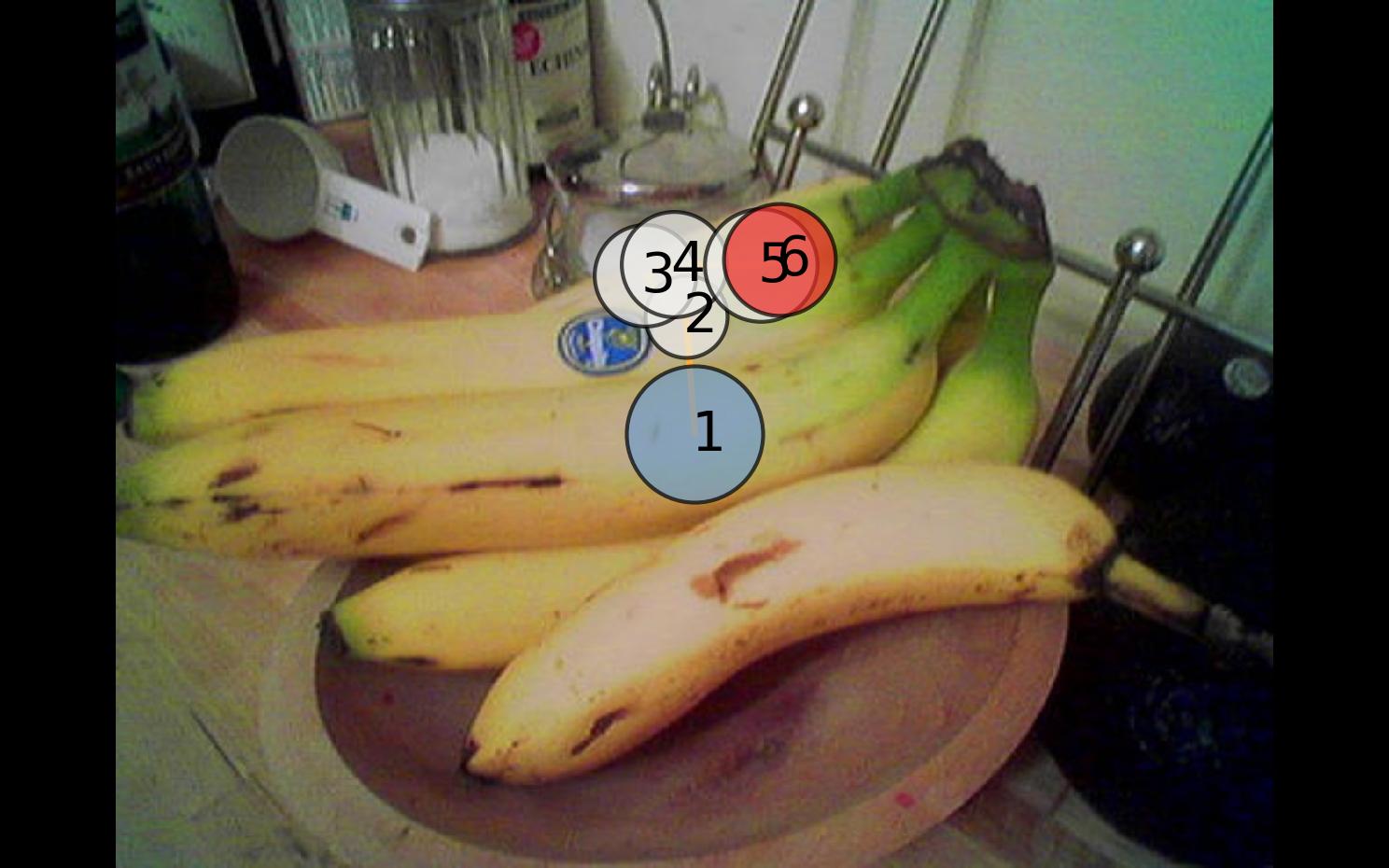} & \includegraphics[width=0.15\linewidth]{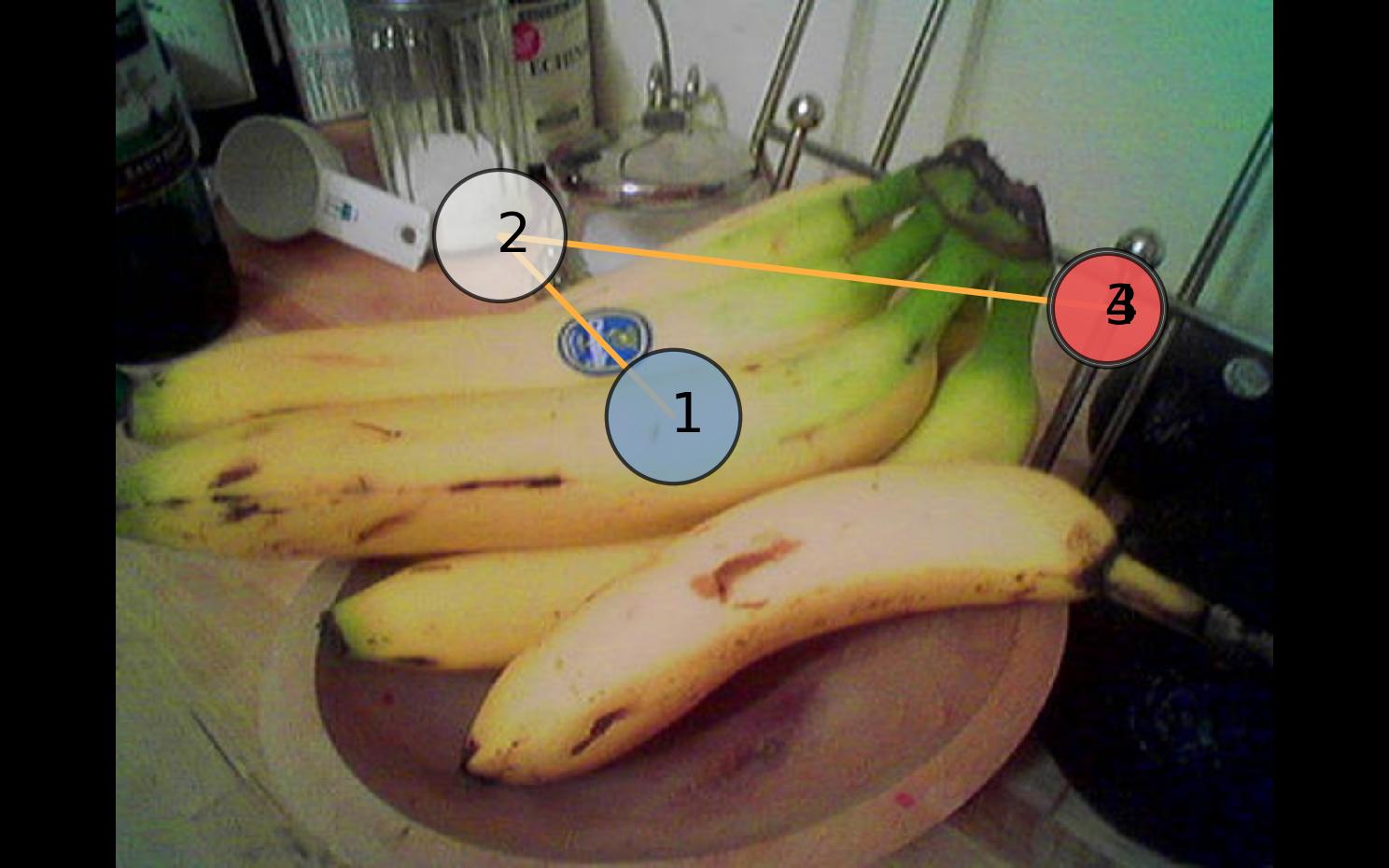} & \includegraphics[width=0.15\linewidth]{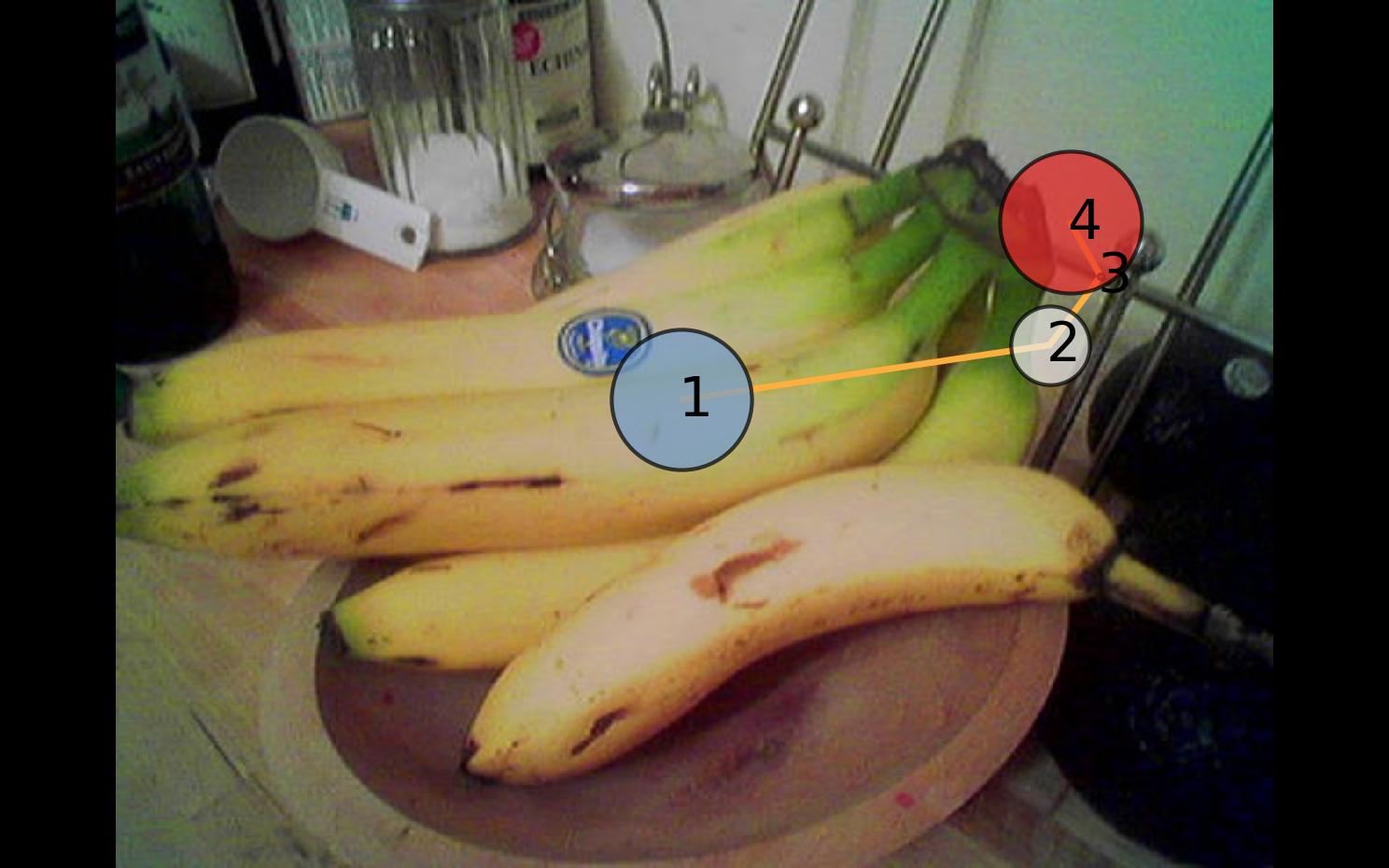} & \includegraphics[width=0.15\linewidth]{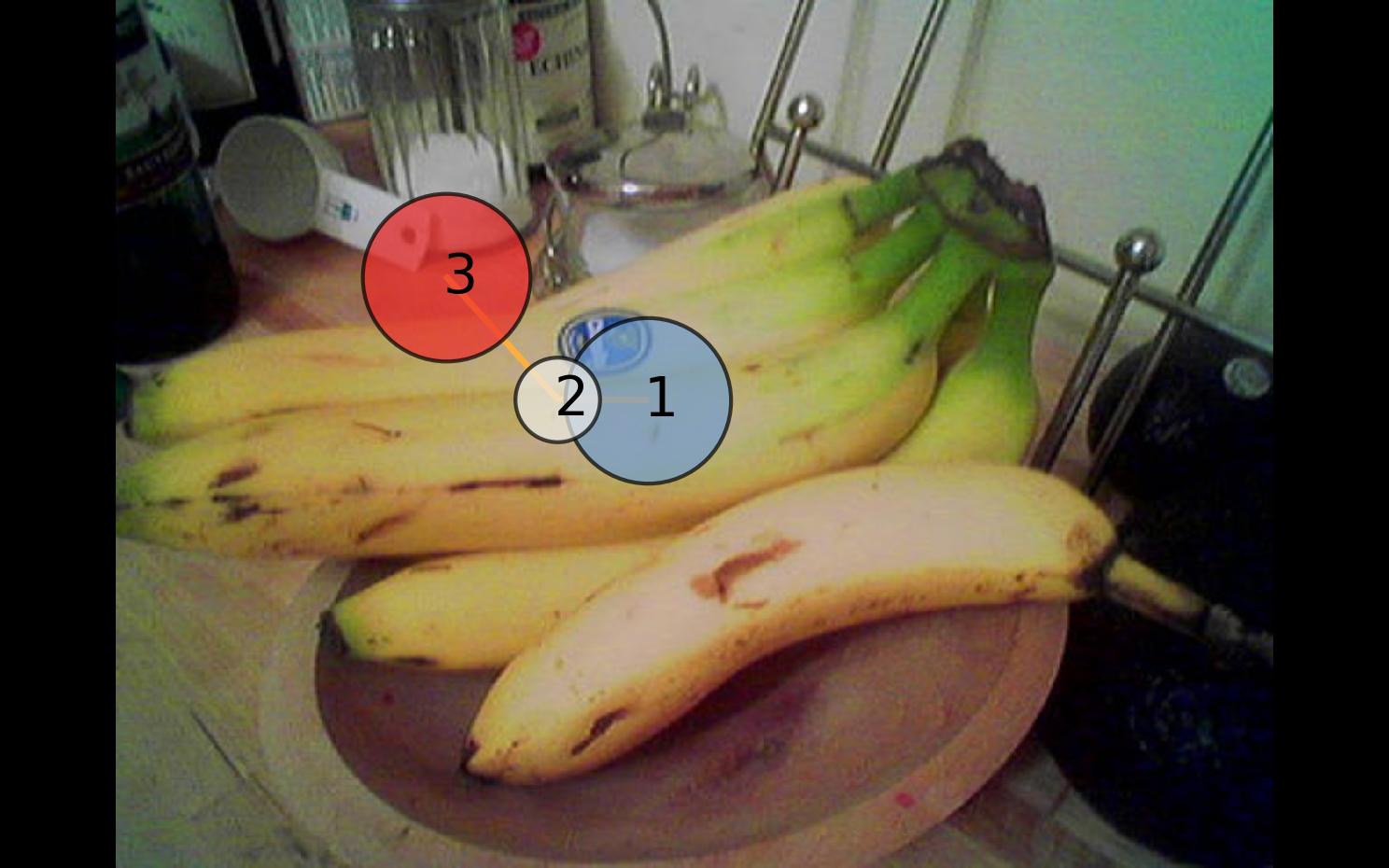} & \includegraphics[width=0.15\linewidth]{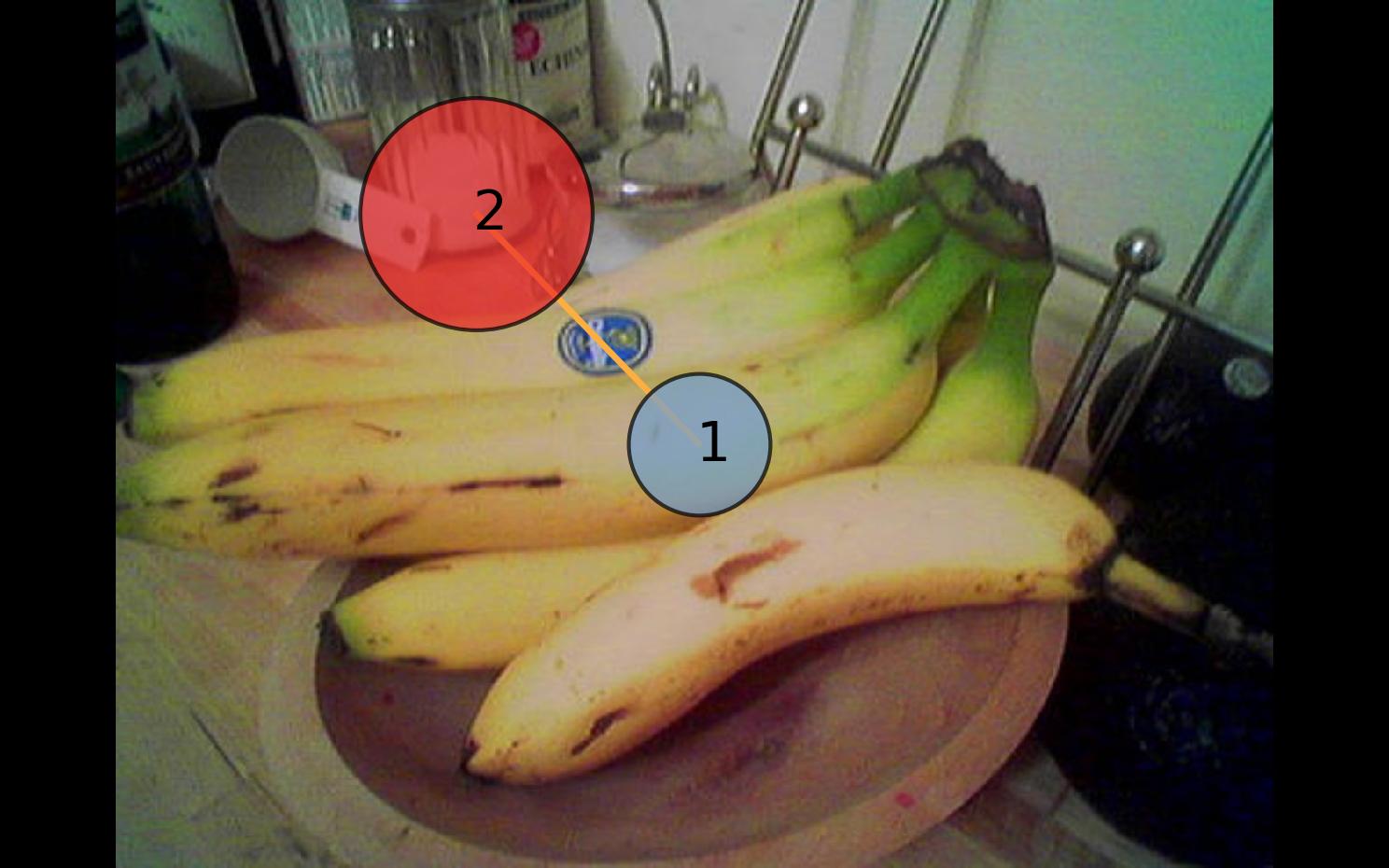} \\
        \includegraphics[width=0.15\linewidth]{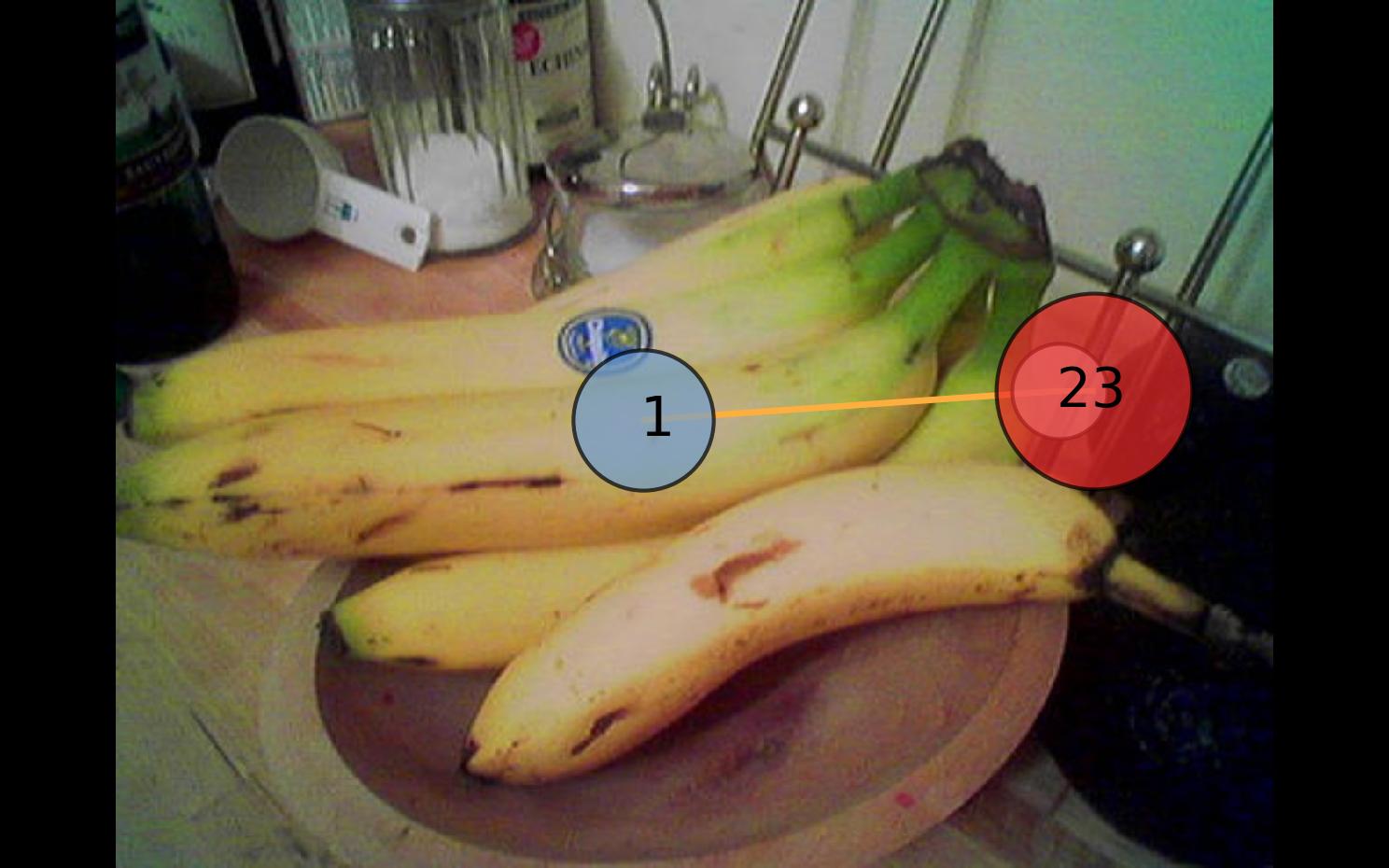} & \includegraphics[width=0.15\linewidth]{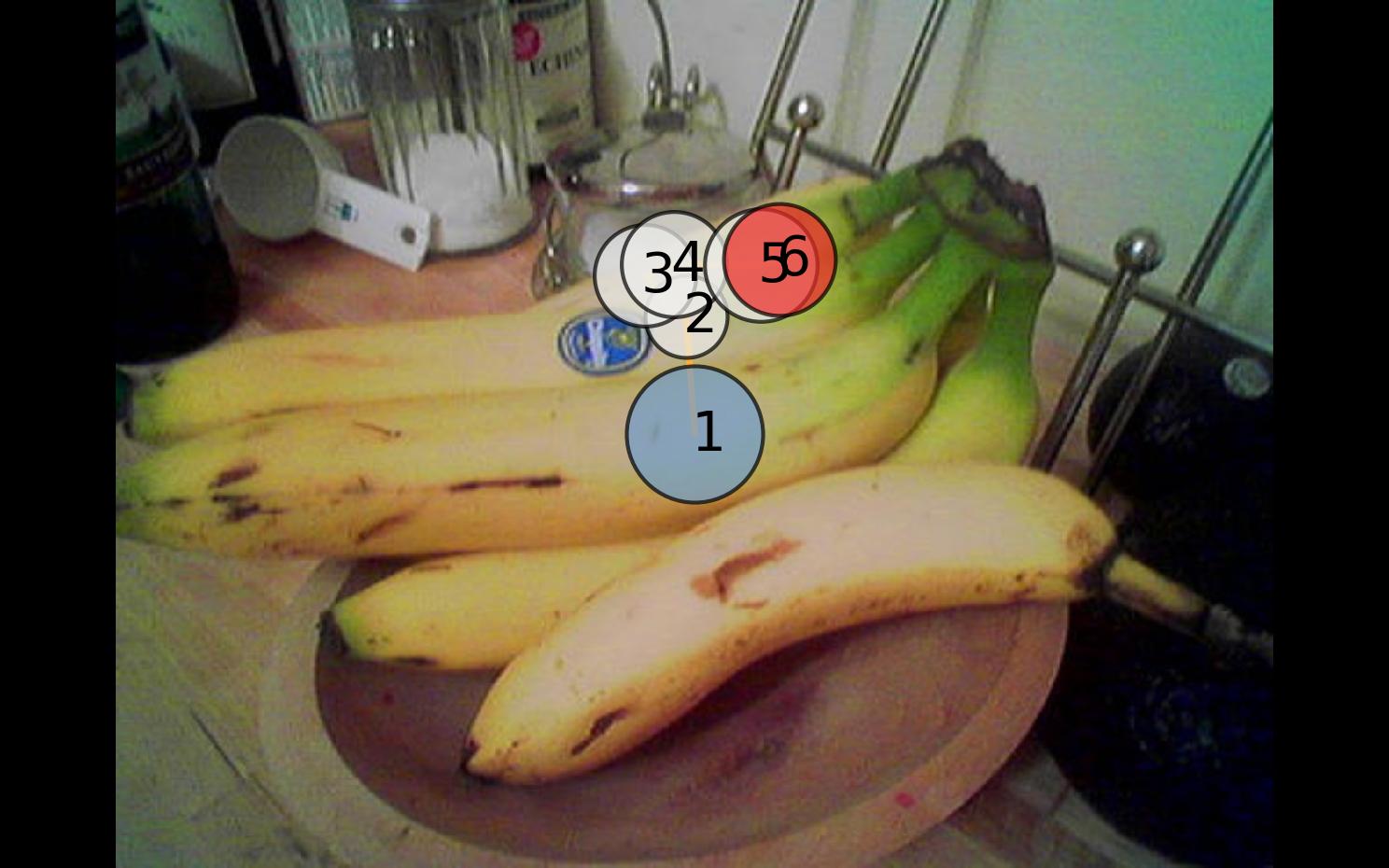} & \includegraphics[width=0.15\linewidth]{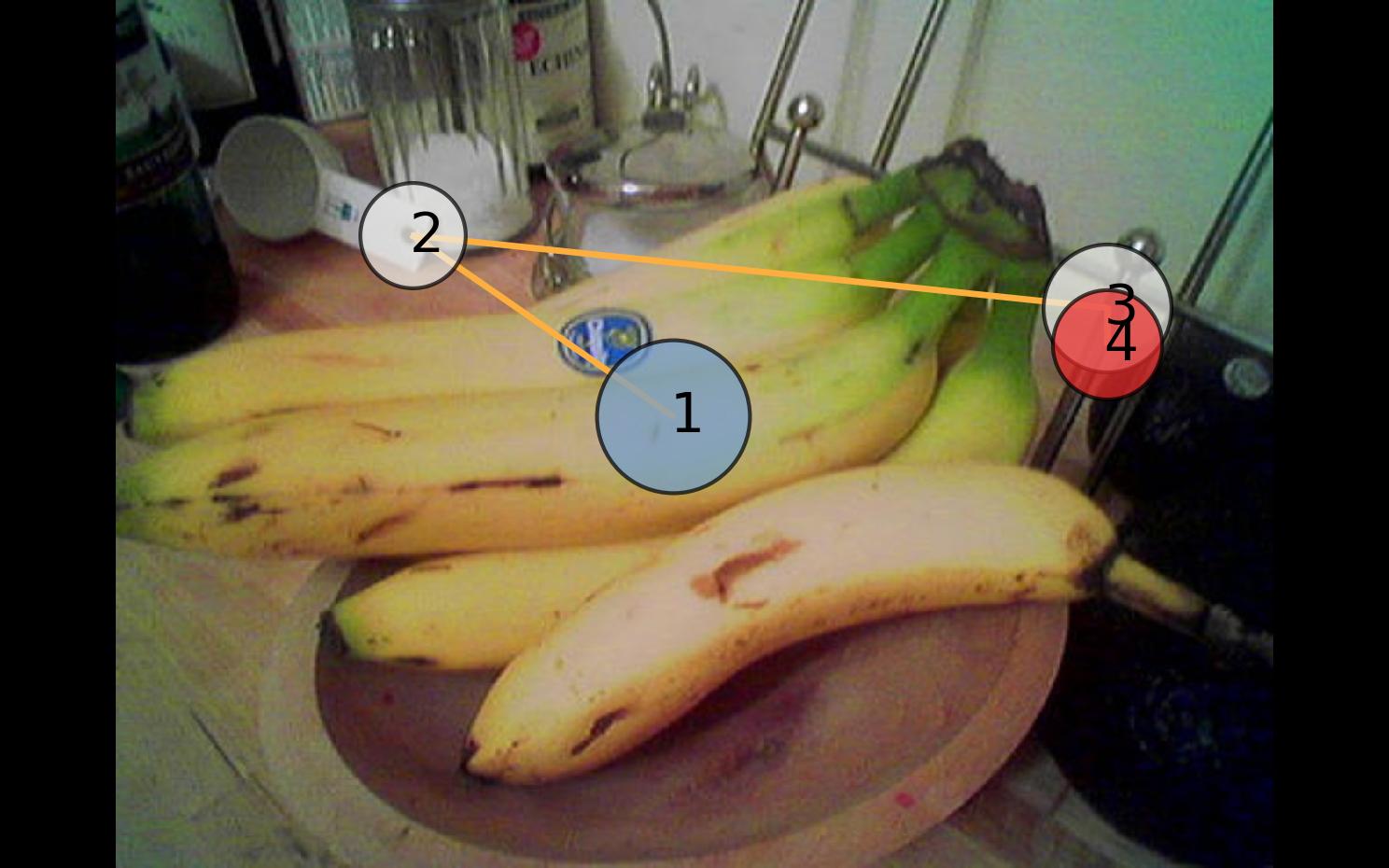} & \includegraphics[width=0.15\linewidth]{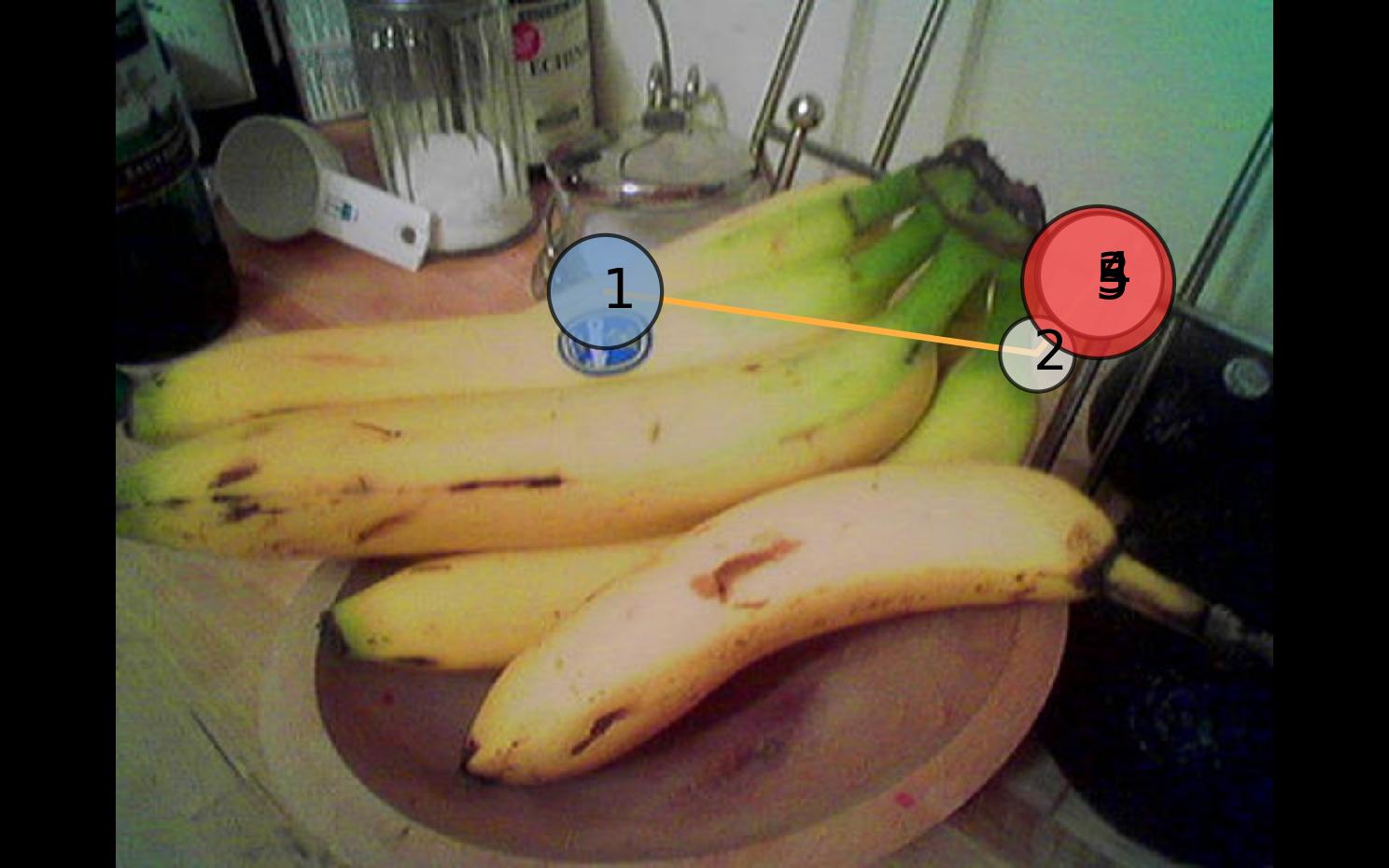} & \includegraphics[width=0.15\linewidth]{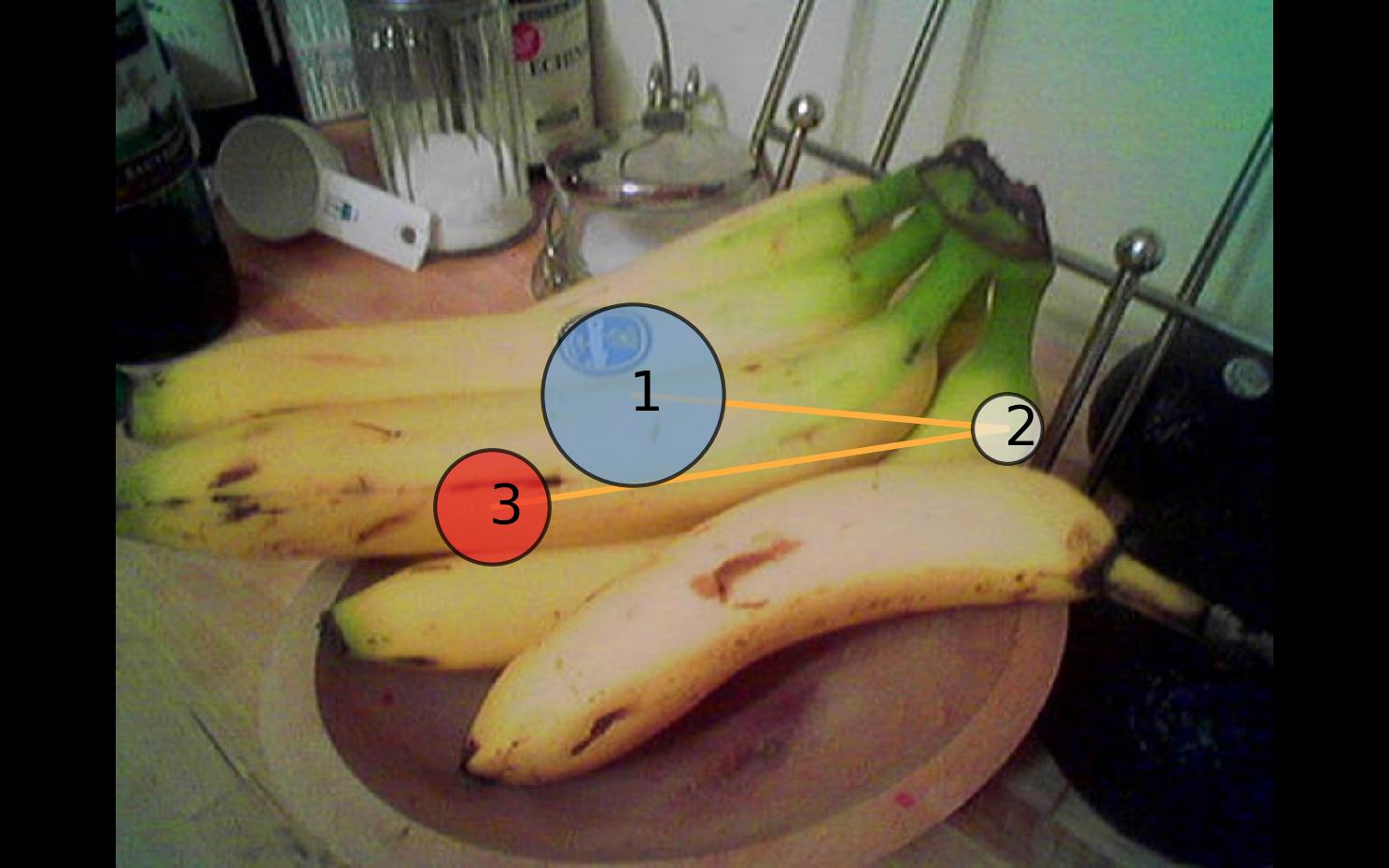} & \includegraphics[width=0.15\linewidth]{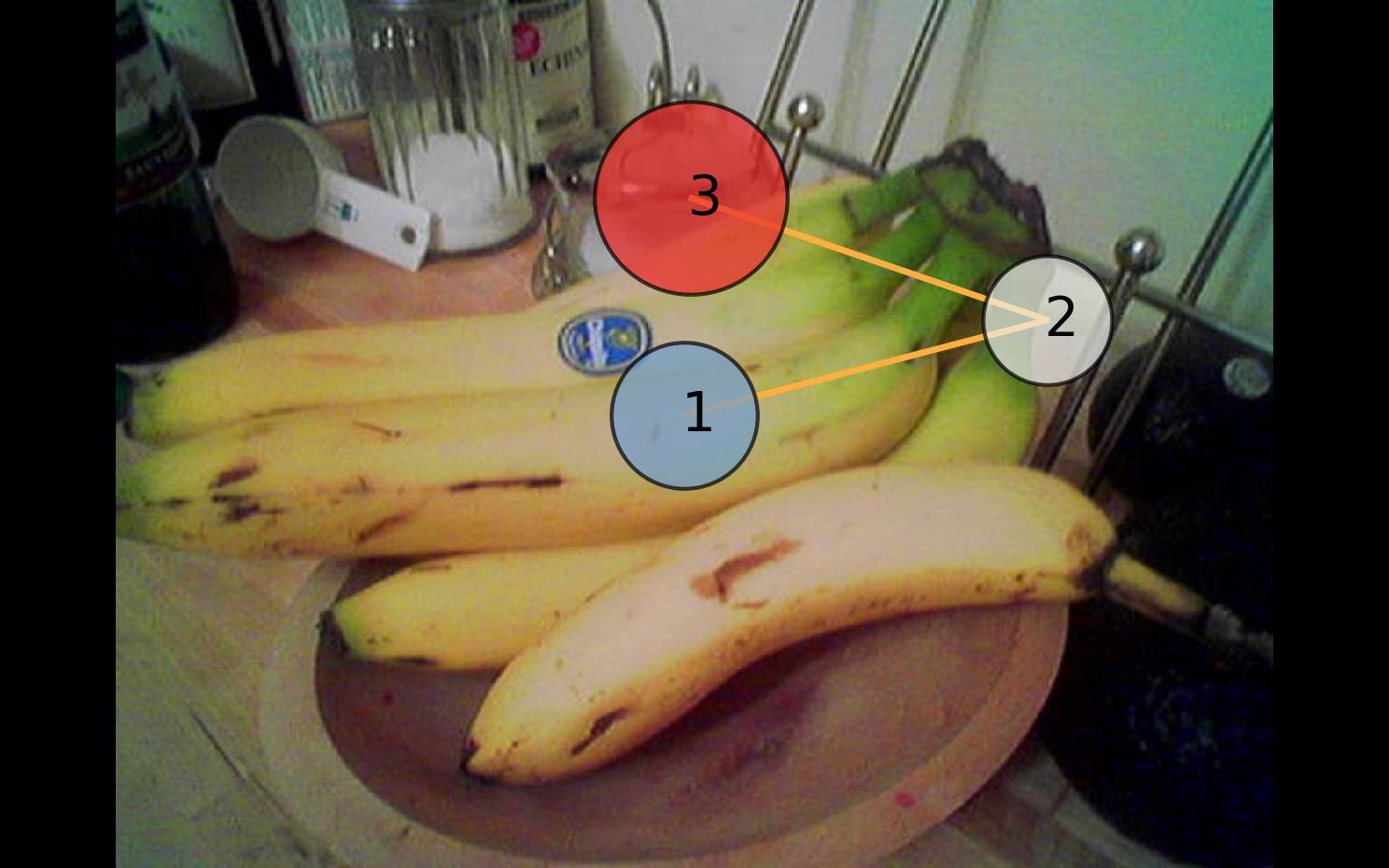} \\
    \end{tabular}
    }
    \vspace{-0.15cm}
    \caption{Qualitative comparison of the variability of simulated and human scanpaths on the COCO-Search18 (TA) dataset for the viewing task: \texttt{fork}. Each row corresponds to a different simulation or a different human observer.}
    \label{fig:qualitatives_variab_cocoTA}
    \vspace{-0.4cm}
\end{figure*}

The qualitative results further support the findings of our scanpath variability analysis, highlighting the ability of \scandiff to generate diverse yet human-like scanpaths across different viewing tasks. As shown in the comparison evaluation from \cref{fig:qualitatives_variab_cocoFV} to \cref{fig:qualitatives_variab_cocoTA} existing methods often produce scanpaths that are either overly deterministic -- failing to capture the natural variability of human gaze behavior -- or overly stochastic, resulting in implausible trajectories.
This is particularly true for Gazeformer that, as shown in Fig.~\ref{fig:qualitatives_variab_cocoT} and Fig.~\ref{fig:qualitatives_variab_cocoTA}, produces scanpaths that are identical to each other over the simulations, essentially generating the same fixation pattern repeatedly regardless of the inherent variability present in human visual attention processes. While Gazeformer achieves reasonable performance on task-oriented datasets as shown in the quantitative results, its deterministic nature fundamentally limits its ability to model the stochastic aspects of human gaze behavior that our approach successfully captures.

\begin{figure*}[t]
    \footnotesize
    \setlength{\tabcolsep}{.1em}
    \resizebox{\linewidth}{!}{
    \begin{tabular}{cc ccc}
         & & \tiny ChenLSTM~\cite{chen2021predicting} & \tiny Gazeformer~\cite{mondal2023gazeformer} & \tiny GazeXplain~\cite{chen2024gazexplain} \\
         \rotatebox{90}{\parbox[t]{0.12\linewidth}{\hspace*{\fill}\tiny \textbf{Target:} \texttt{bottle}\hspace*{\fill}}} 
         & & \includegraphics[height=0.12\linewidth]{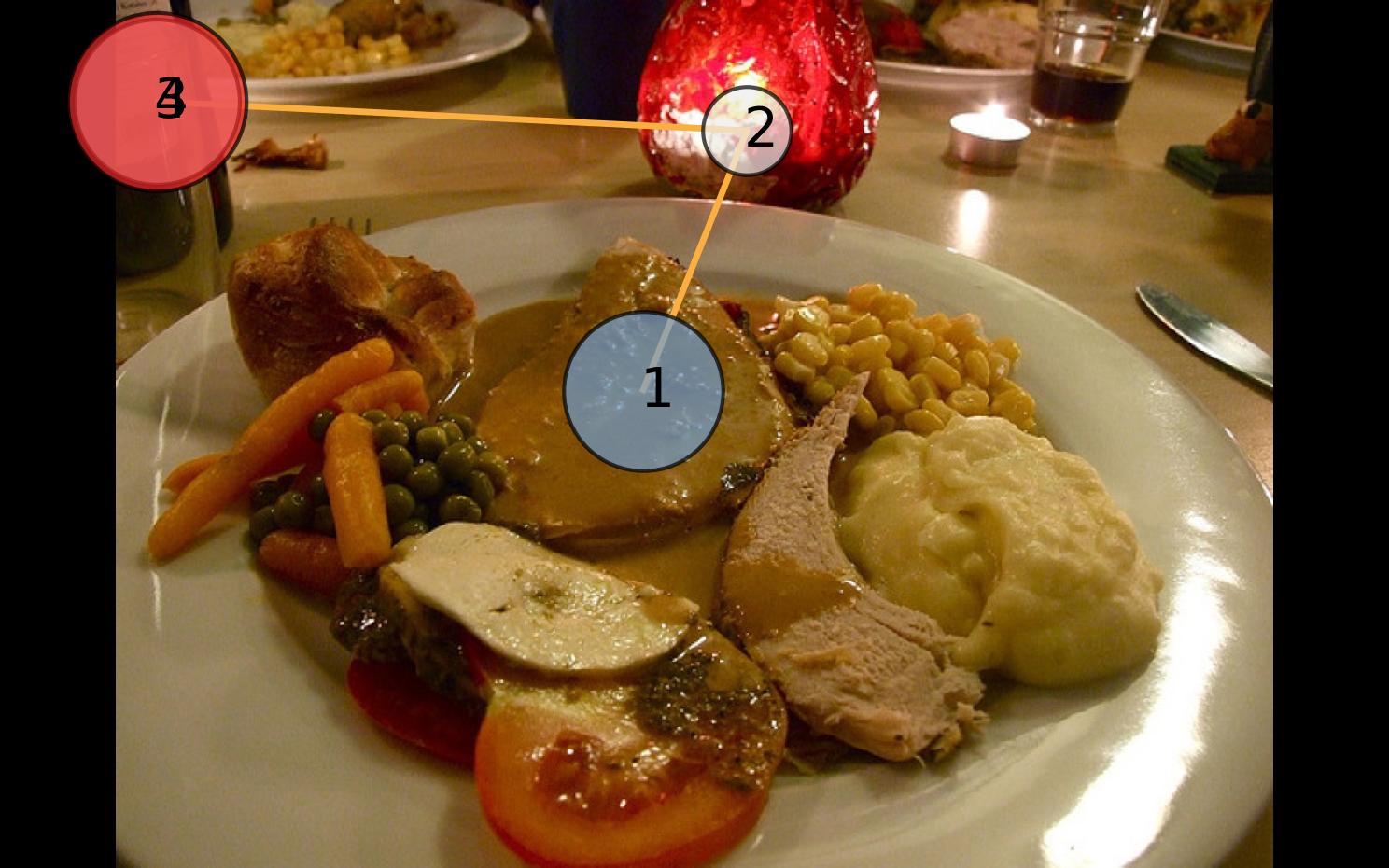} & 
         \includegraphics[height=0.12\linewidth]{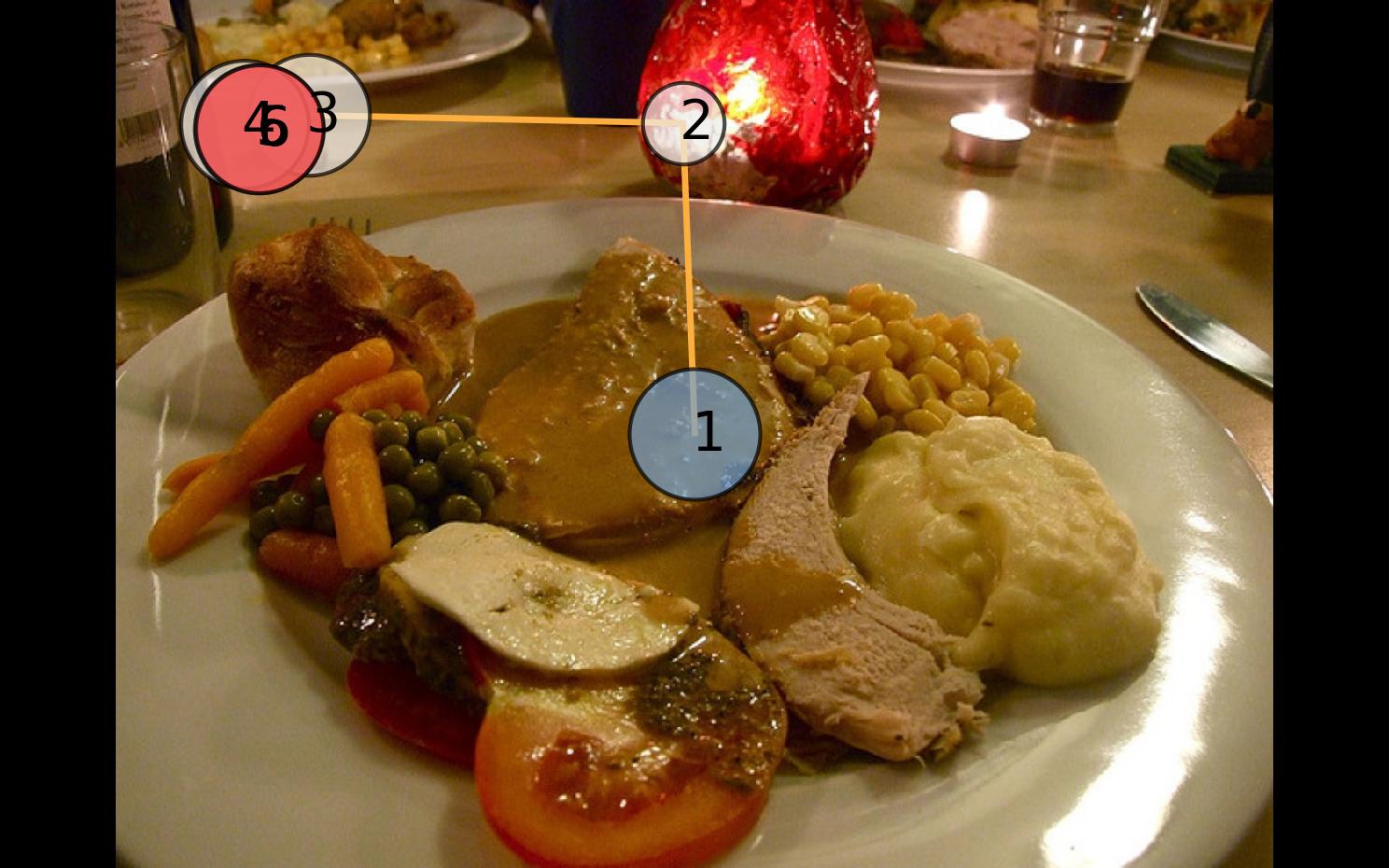} & 
         \includegraphics[height=0.12\linewidth]{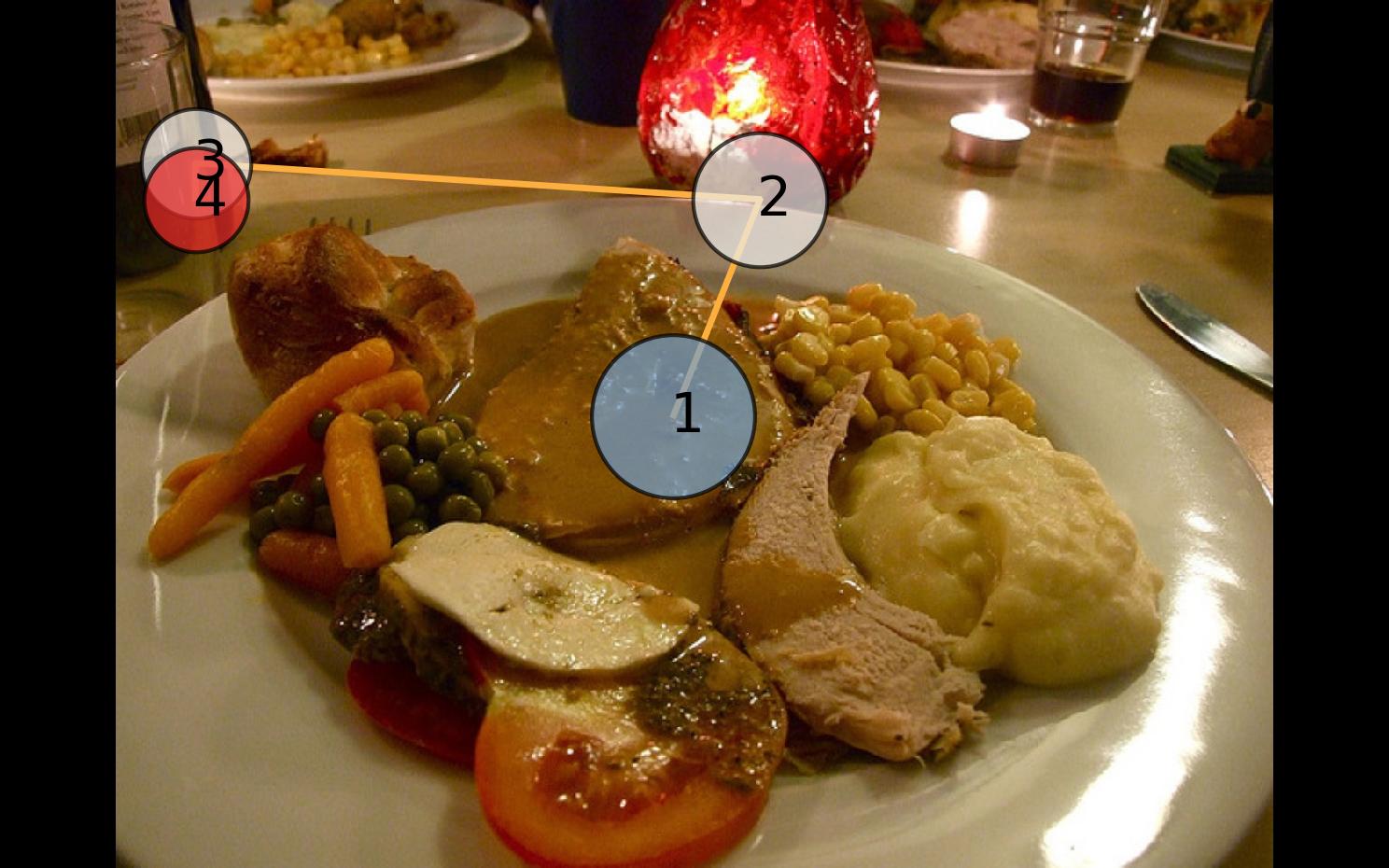} \\         
         & & \tiny TPP-Gaze~\cite{damelio2025tpp} & \tiny \textbf{\ours (Ours)} & \tiny Humans \\
         & & \includegraphics[height=0.12\linewidth]{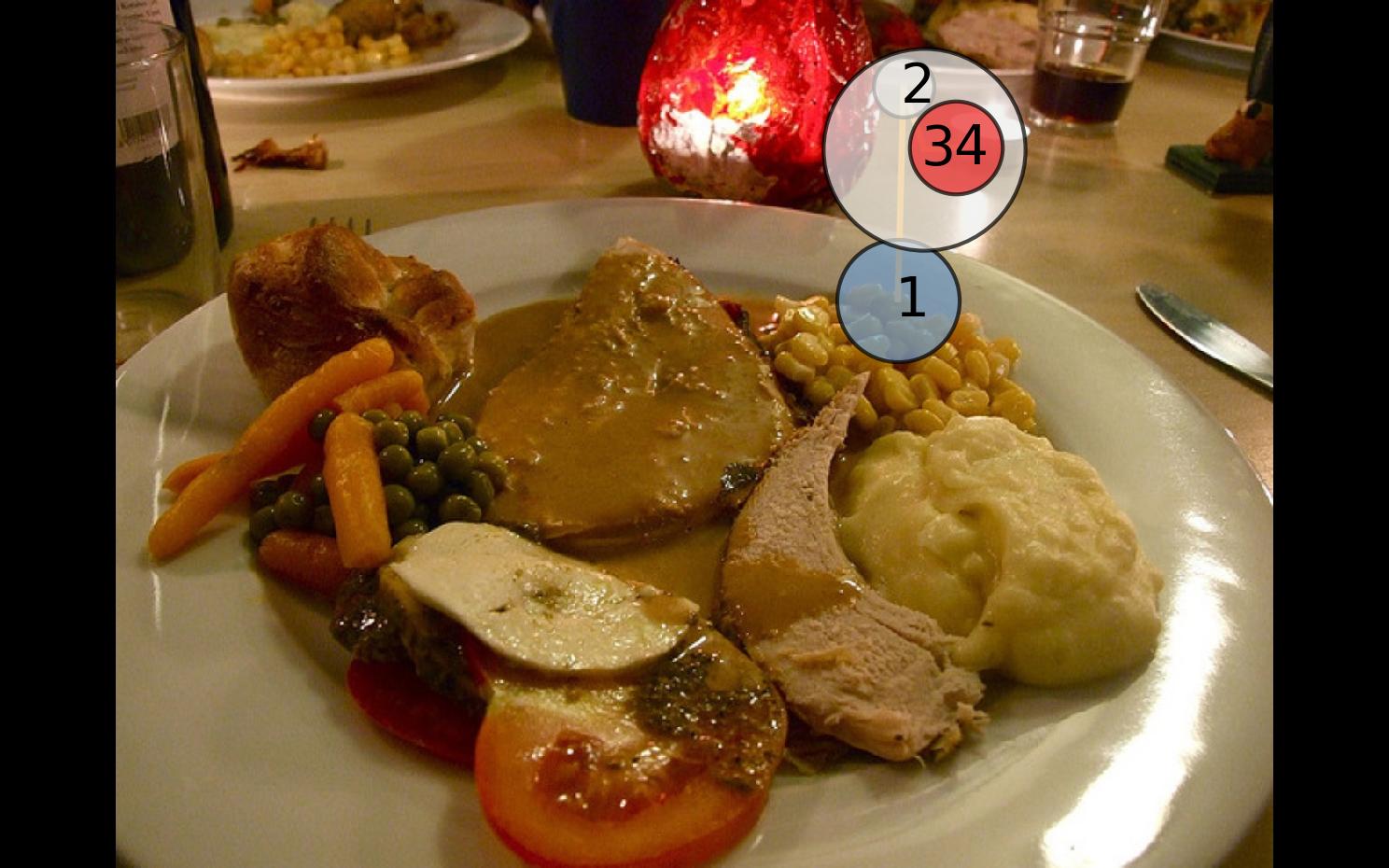} & 
         \includegraphics[height=0.12\linewidth]{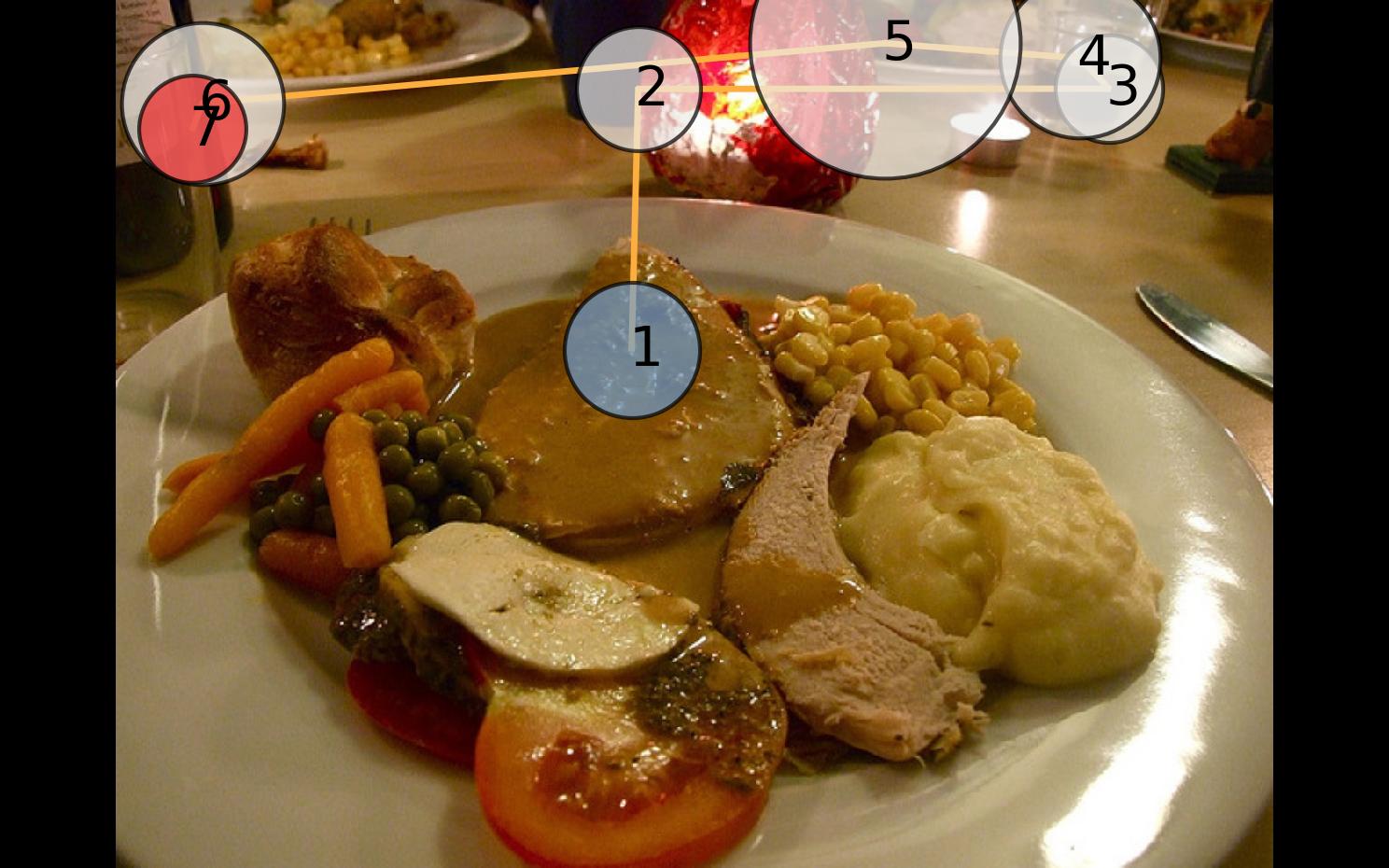} & 
         \includegraphics[height=0.12\linewidth]{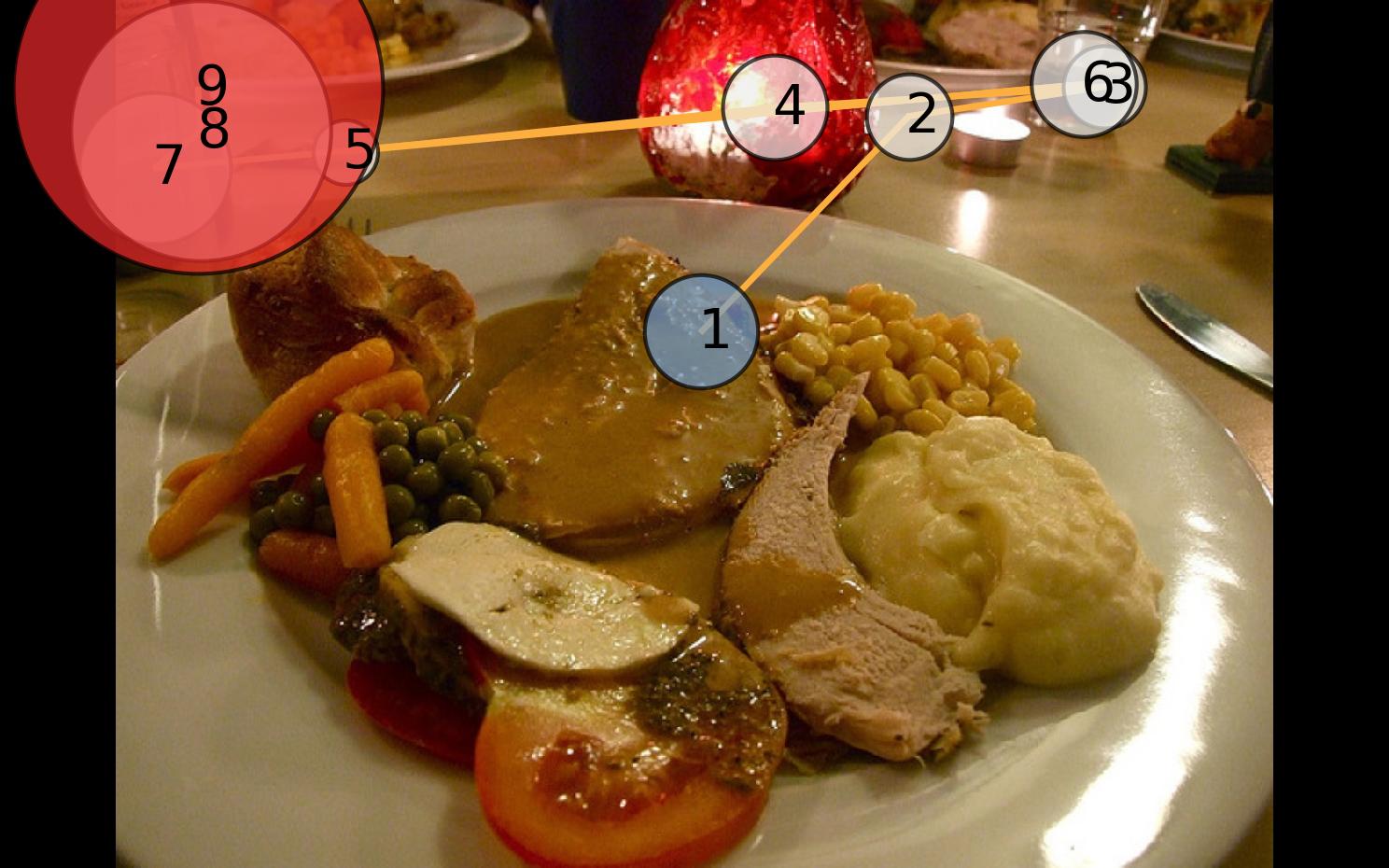} \\

         \addlinespace[0.6cm]
         
         & & \tiny ChenLSTM~\cite{chen2021predicting} & \tiny Gazeformer~\cite{mondal2023gazeformer} & \tiny GazeXplain~\cite{chen2024gazexplain} \\
         \rotatebox{90}{\parbox[t]{0.12\linewidth}{\hspace*{\fill}\tiny \textbf{Target:} \texttt{keyboard}\hspace*{\fill}}} 
         & & \includegraphics[height=0.12\linewidth]{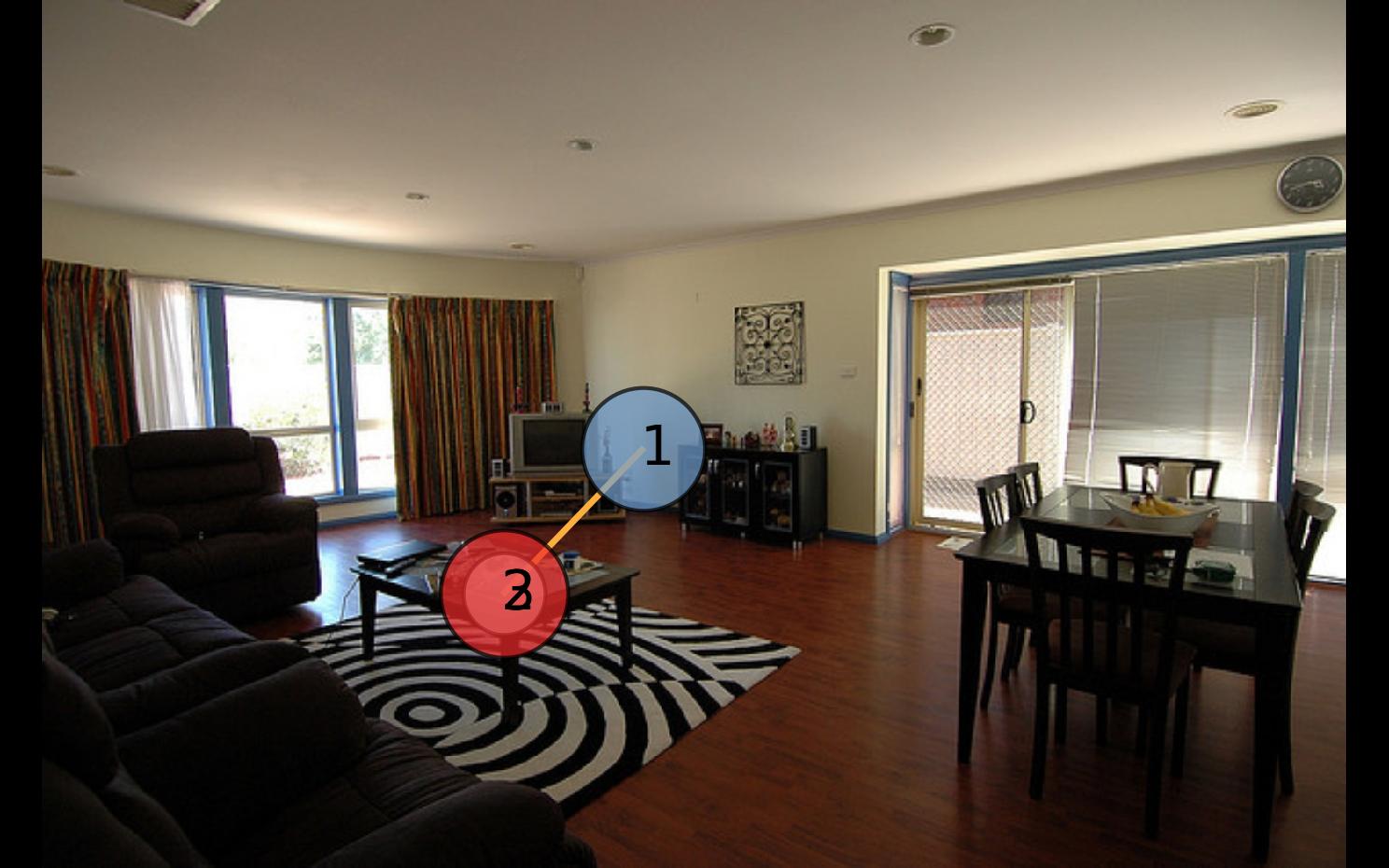} & 
         \includegraphics[height=0.12\linewidth]{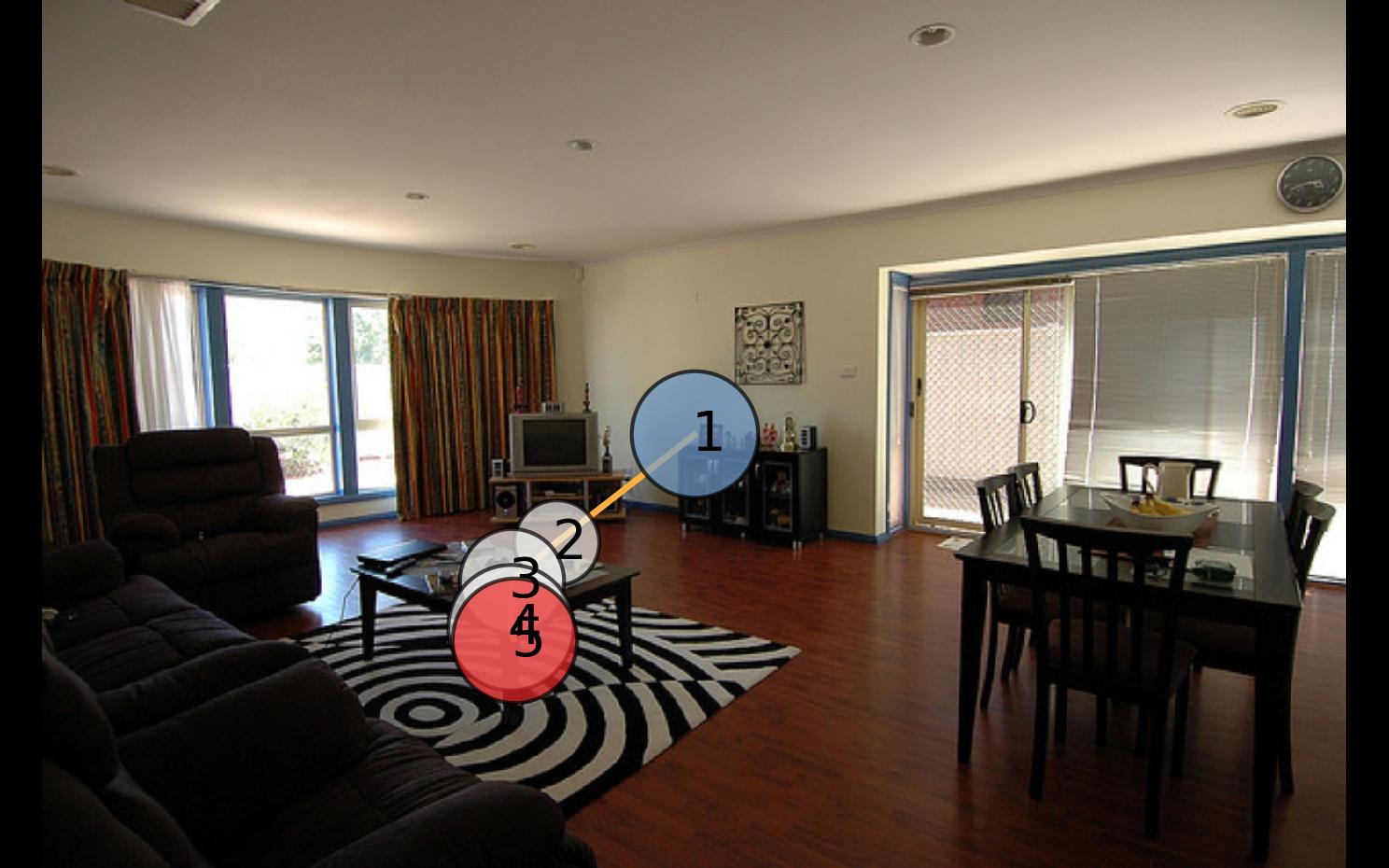} & 
         \includegraphics[height=0.12\linewidth]{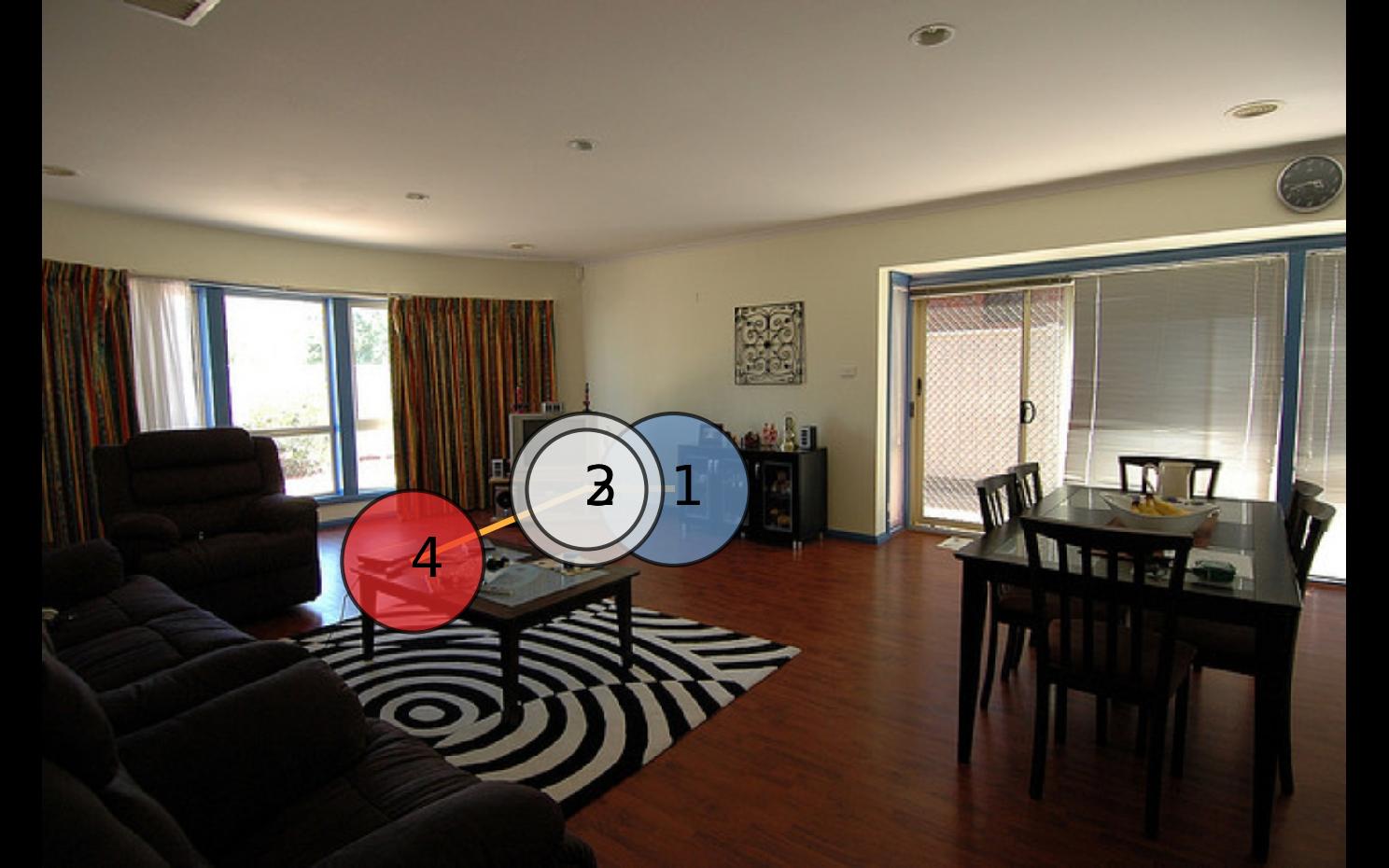} \\         
         & & \tiny TPP-Gaze~\cite{damelio2025tpp} & \tiny \textbf{\ours (Ours)} & \tiny Humans \\
         & & \includegraphics[height=0.12\linewidth]{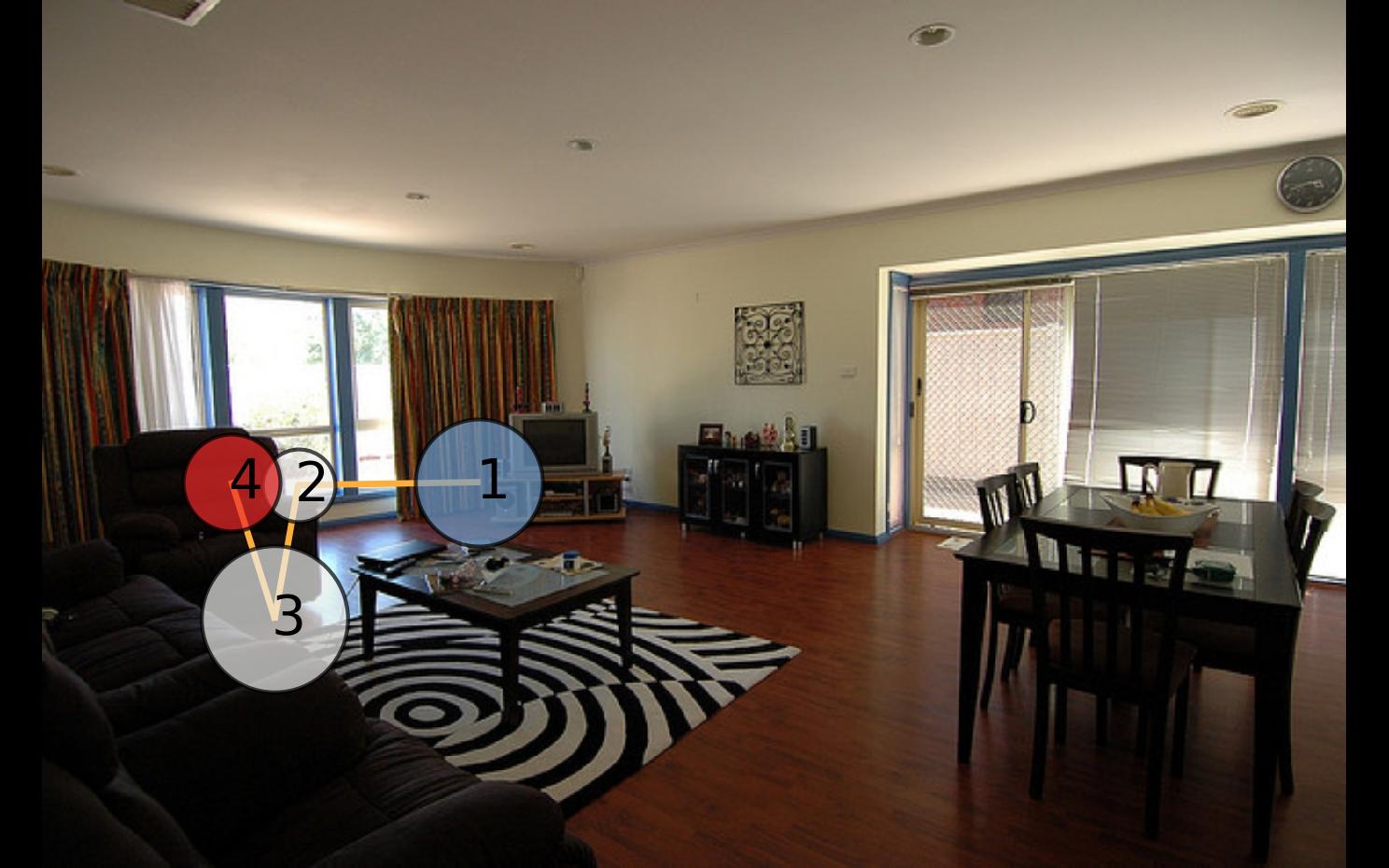} & 
         \includegraphics[height=0.12\linewidth]{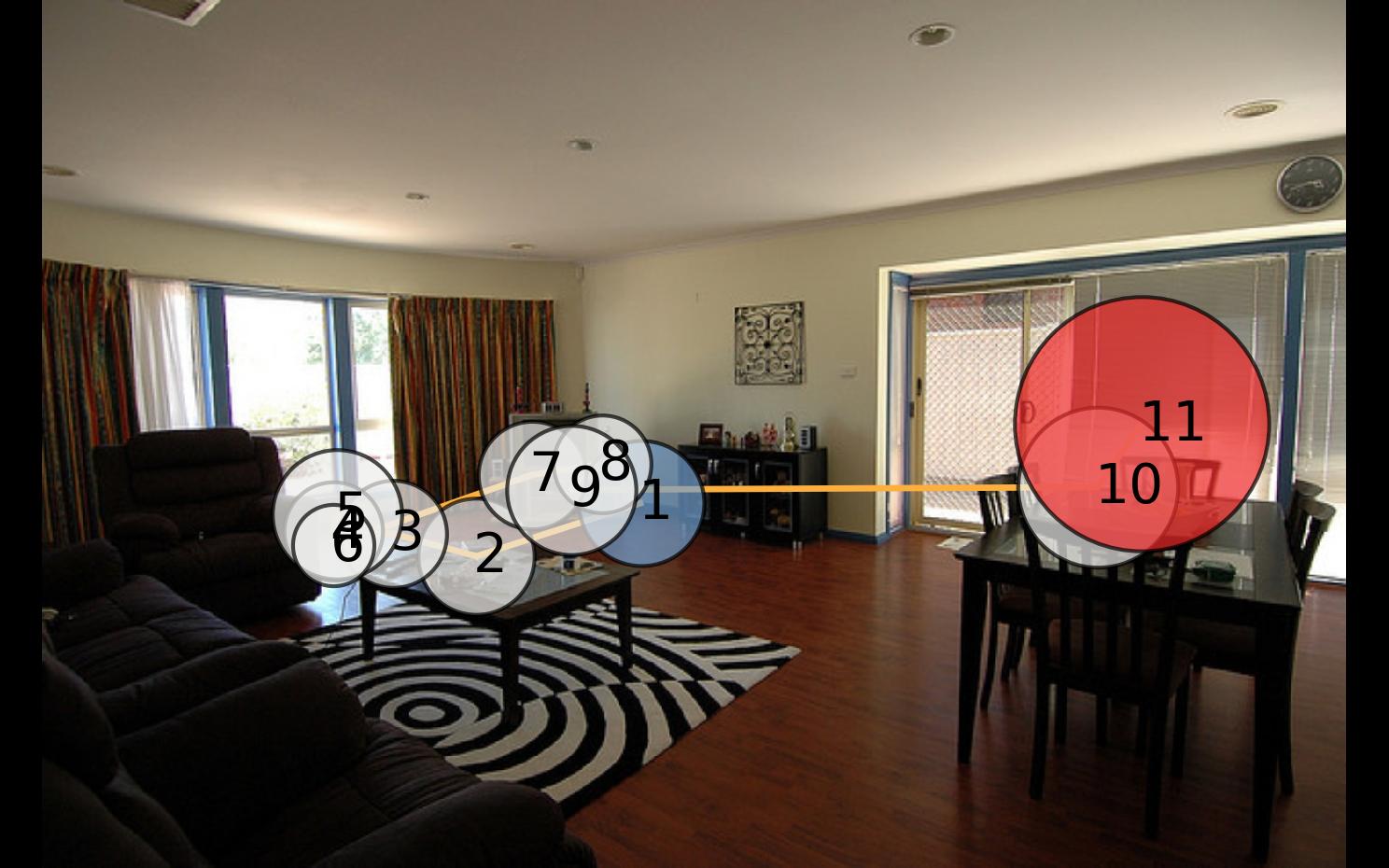} & 
         \includegraphics[height=0.12\linewidth]{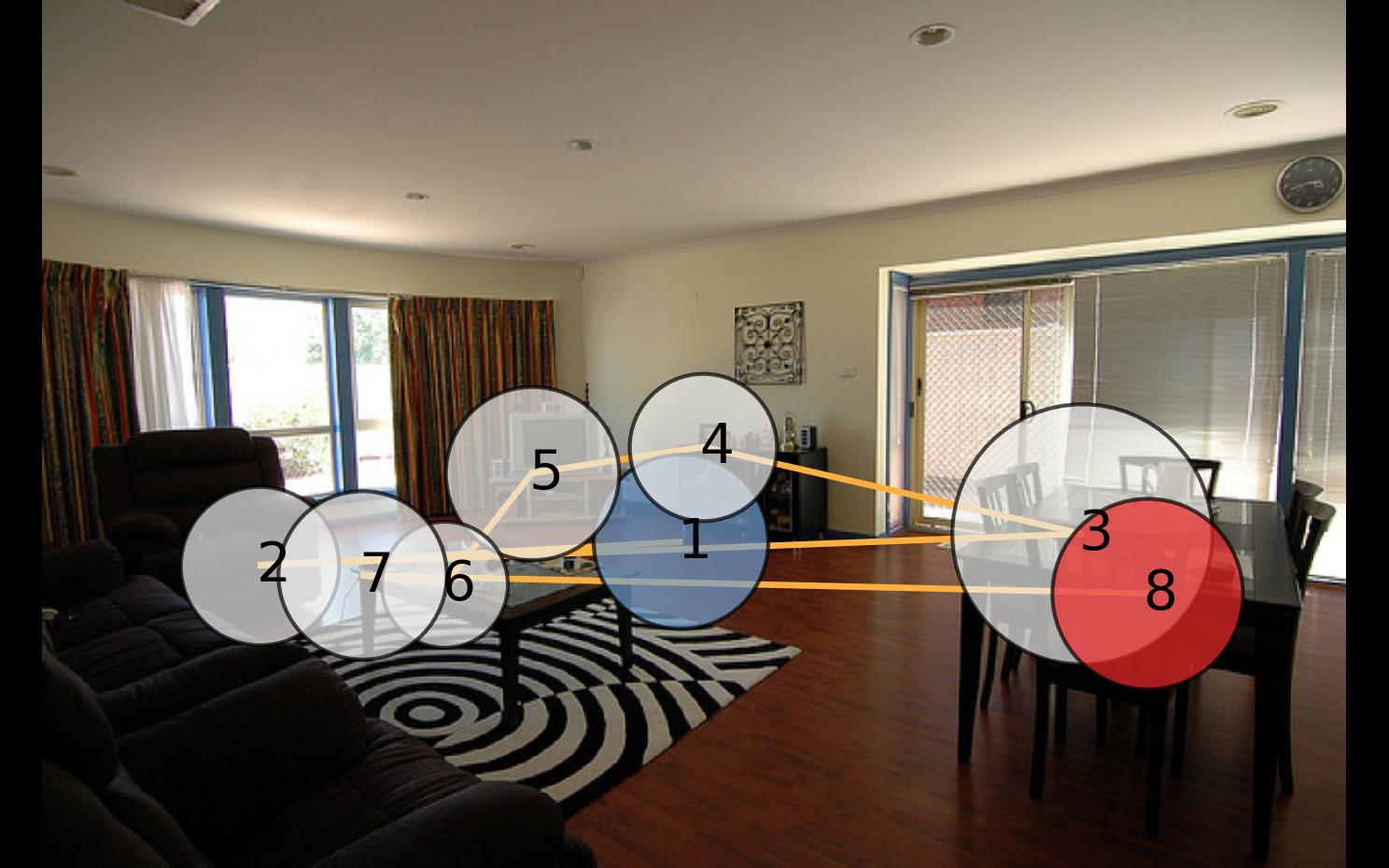} \\
         
    \end{tabular}
    }
    \vspace{-0.15cm}
    \caption{Qualitative comparison of simulated and human scanpaths on the COCO-Search18 (TA) dataset for the visual search task.}
    \label{fig:qualitatives_COCOSearch_TA}
    \vspace{-0.4cm}
\end{figure*}

\end{document}